%% file: main.tex
\documentclass{book}

\usepackage{listings}
\usepackage{times}
\usepackage{latexsym}

\usepackage[T1]{fontenc}
\usepackage[utf8]{inputenc}

\usepackage{microtype}
\usepackage{inconsolata}
\usepackage{natbib}

\usepackage{tikz}
\usetikzlibrary{positioning, shapes.geometric, arrows}

\usepackage{algorithm}
\usepackage{algpseudocode}
\usepackage[export]{adjustbox}
\usepackage[mathscr]{euscript}
\usepackage{verbatim}
\usepackage{hyperref}
\usepackage{graphicx}
\usepackage{dsfont}
\usepackage{bm}
\usepackage{bbm}
\usepackage{multirow}
\usepackage{multicol}

\usepackage{amsthm}
\usepackage{amsmath}
\usepackage{amssymb}
\usepackage{amsfonts}

\usepackage{mathtools}

\usepackage{makecell}
\usepackage{caption}
\usepackage{subcaption}
\usepackage{stmaryrd}
\DeclareMathOperator*{\argmax}{arg\,max}
\DeclareMathOperator*{\argmin}{arg\,min}

\newcommand{\pred}{\bm{f}}
\newcommand{\vg}{\bm{G}}
\newcommand{\vx}{\bm{X}}
\newcommand{\vy}{\bm{Y}}
\newcommand{\vu}{\bm{U}}
\newcommand{\vw}{\bm{W}}

\newcommand{\va}{\bm{A}}
\newcommand{\vm}{\bm{M}}
\newcommand{\Var}{\mathbb{V}}

\theoremstyle{definition}
\newtheorem{definition}{Definition}[section]

\usepackage{pdfpages}

\begin{document}
\includepdf{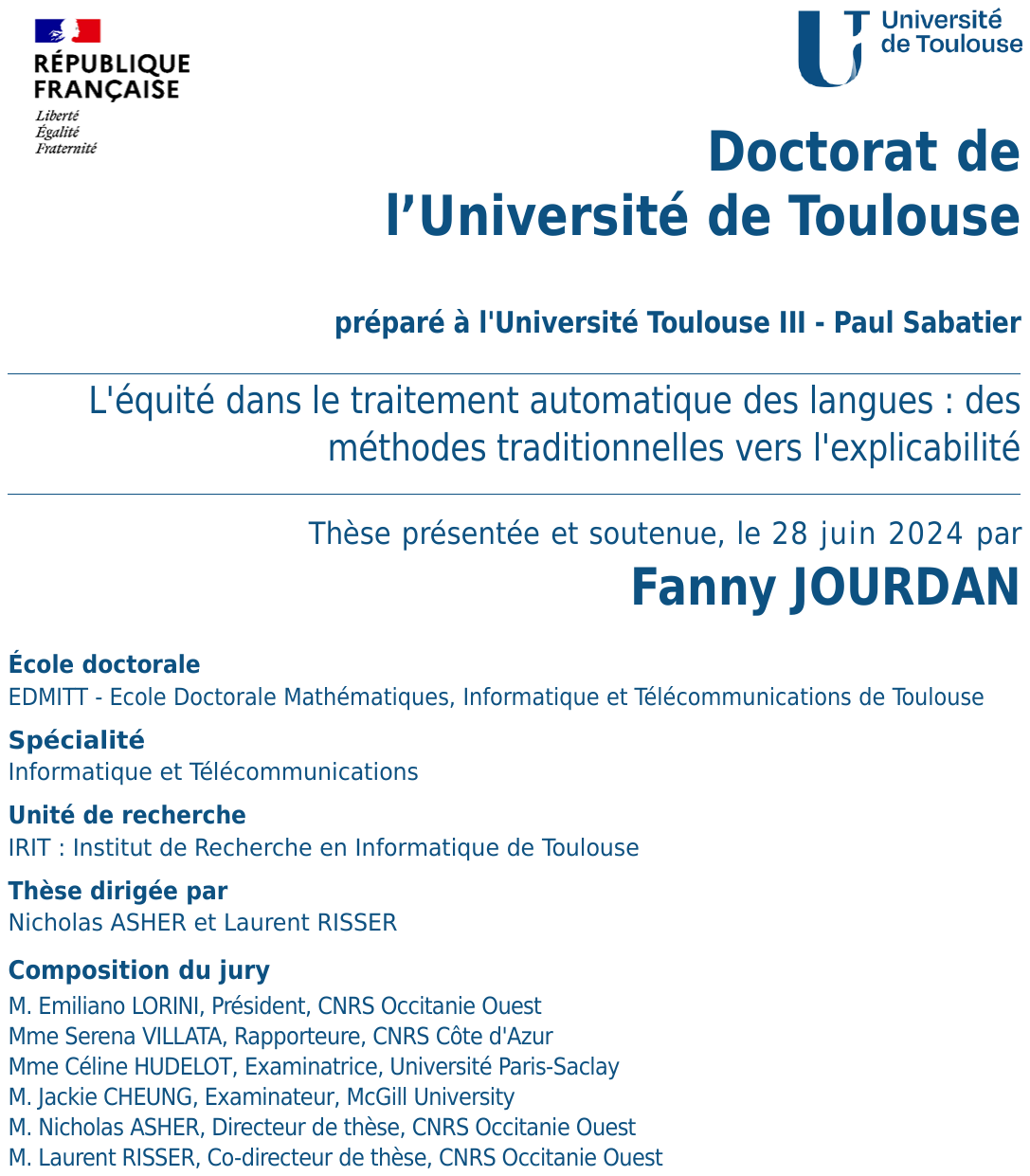}



\input{remerciements}
\tableofcontents
\input{Abstract_fr}
\input{Abstract}
\input{Introduction_fr}

\input{Introduction}
\input{Preliminaries}
\input{State_of_the_art}

\input{Fairness_reg} 
\input{Fairness_metrics} 
\input{Explainability}
\input{XAI_Fairness}
\input{Conclusion}
\input{Conclusion_fr}

\bibliography{custom}

\bibliographystyle{natbib}

\appendix
\input{AppendixRoBERTatraining}
\input{AppendixA}
\includepdf{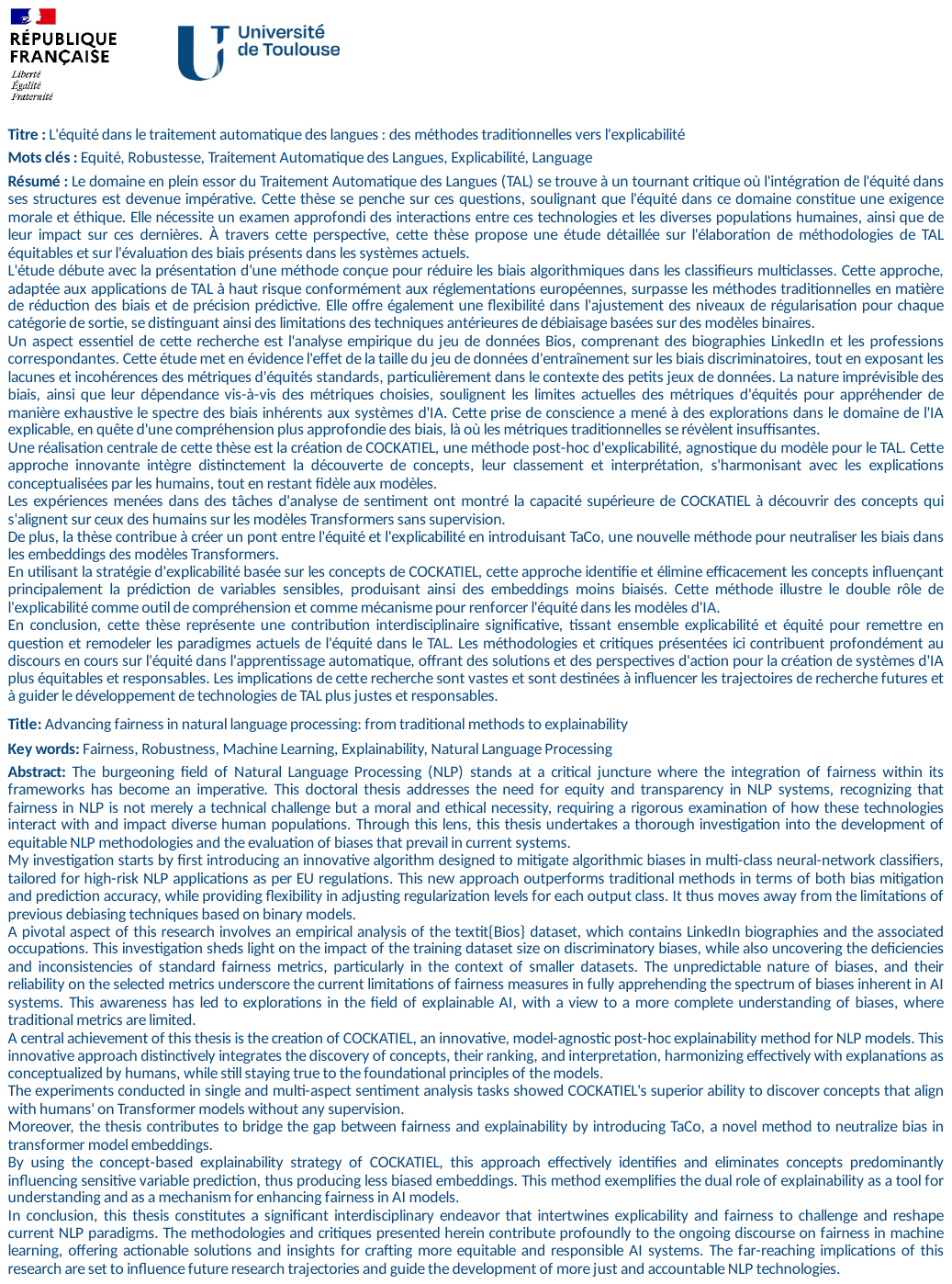}
\end{document}

%% file: remerciements.tex
\chapter*{Remerciements}

Tout d'abord, je tiens à exprimer ma profonde gratitude envers mes directeurs de thèse, Nicholas Asher et Laurent Risser, pour leur accompagnement et leurs précieux conseils tout au long de ces trois années passionnantes. Merci à vous deux d'avoir été la direction de thèse dont j'avais besoin pour réussir dans cette voie. Vous avez su être présents et m'apporter un soutien constant quand il le fallait, tout en me laissant la liberté nécessaire pour m'épanouir et devenir une chercheuse autonome. Votre rigueur scientifique et votre grande pédagogie resteront pour moi un exemple à suivre dans la suite de ma carrière.

Je souhaite également adresser un grand merci à mes rapporteuses, Michèle Sebag et Serena Villata, pour le temps consacré à l'examen approfondi de mon manuscrit et pour leurs conseils pertinents, grâce auxquels j'ai pu présenter cette version finale. Je suis honorée que deux chercheuses que j'admire beaucoup aient pris le temps de lire en détail mon travail et de me faire des retours si constructifs. Mes remerciements s'étendent aux autres membres du jury, Céline Hudelot, Jackie Cheung, et particulièrement à Emiliano Lorini pour avoir accepté de présider mon jury.

Grâce à la nature pluridisciplinaire de ma thèse portée par ANITI, j'ai eu l'opportunité de collaborer et d'échanger avec de nombreuses personnes. Je remercie donc ceux qui ont rendu cette thèse possible grâce à leurs rôles clefs au sein d'ANITI : Corinne Joffre, Nicholas Asher (en tant que directeur scientifique d'ANITI cette fois), Serge Gratton, Nicolas Viallet et Romaric Redon. Un grand merci aussi à Ana Gonzales et Grégory Flandin pour leur accueil chaleureux dès ma seconde année de thèse au sein de l'équipe DEEL.

Merci à tous mes collègues et amis de l'Institut de Mathématique de Toulouse (IMT), Alain, Alberto, Alexandre, Anthony, Armand, Arnaud, Benjamin (x2), Candice, Clément, Corentin, Elio, Etienne, Fuhsuan, Javier, Joachim, Laeticia, Louis, Lucas, Mahmoud, Mathis, Maxime, Michèle, Mitja, Nicolas (x2), Paola, Perla, Sophia et Virgile, ainsi que ceux du B612, Corentin, David (x2), Etienne, Estèle, Frederic, Joseba, Justin, Luca, Lucas, Louise, Matthieu, Mouhcine, Nassima, Thibaut, Thomas (x2), Vuong et Yannick, pour avoir partagé cette aventure avec moi. Un remerciement tout particulier à mes cobureaux, Chifaa, Clément, El Medhi, Léo, pour tous ces moments précieux passés ensemble.

Je souhaite remercier spécialement Adil, Agustin, Alon, Antonin, Guilhem, Louis, Mehdi, Nicolas, Paul et Thomas pour tout le travail réalisé avec chacun d'entre eux. Merci infiniment pour tout le temps passé à travailler avec moi, pour votre aide précieuse et vos conseils. Sans vous, je n'aurais pas écrit la thèse que vous lisez aujourd'hui. Un remerciement particulier aussi pour Chloé, Franck, Marjorie, Mélanie, Nawal, Philippe, Ryma, Sébastien, Victor et Yann pour leurs conseils sur le plan académique mais surtout sur mon avenir post-thèse.

Enfin, je souhaite adresser des remerciements plus personnels à ma famille pour leur soutien et leur amour tout au long de ma vie et sans qui je ne serais pas là aujourd'hui. Merci d'avoir cru en moi avant tout le monde et de m'avoir toujours encouragée à suivre ma voie même quand ce n'était pas la plus évidente.  Je remercie également mes colocataires Marco et Victor, avec qui j'ai partagé ces trois années de thèse à Toulouse, ainsi que mes amies Loën et Fanny, qui ont été présentes même dans les moments difficiles et m'ont toujours soutenue. Vous êtes précieux.

%% file: Abstract_fr.tex
\chapter*{Abstract en français}

Le domaine en plein essor du Traitement Automatique des Langues (TAL) se trouve à un tournant critique où l'intégration de l'équité dans ses structures est devenue impérative. Cette thèse de doctorat se penche sur ces questions, soulignant que l'équité dans ce domaine ne constitue pas uniquement un défi technique, mais également une exigence morale et éthique. Elle nécessite un examen approfondi des interactions entre ces technologies et les diverses populations humaines, ainsi que de leur impact sur ces dernières. À travers cette perspective, cette thèse propose une étude détaillée sur l'élaboration de méthodologies de TAL équitables et sur l'évaluation des biais présents dans les systèmes actuels.

L'étude débute avec la présentation d'un algorithme novateur, conçu pour réduire les biais algorithmiques dans les classifieurs neuronaux multiclasses. Cette approche, adaptée aux applications de TAL à haut risque conformément aux réglementations européennes, surpasse les méthodes traditionnelles en matière de réduction des biais et de précision prédictive. Elle offre également une flexibilité dans l'ajustement des niveaux de régularisation pour chaque catégorie de sortie, se distinguant ainsi des limitations des techniques antérieures de débiaisage basées sur des modèles binaires.

Un aspect essentiel de cette recherche est l'analyse empirique du jeu de données \textit{Bios}, comprenant des biographies LinkedIn et les professions correspondantes. Cette étude met en évidence l'effet de la taille du jeu de données d'entraînement sur les biais discriminatoires, tout en exposant les lacunes et incohérences des métriques d'équités standards, particulièrement dans le contexte des petits jeux de données. La nature imprévisible des biais, ainsi que leur dépendance vis-à-vis des métriques choisies, soulignent les limites actuelles des métriques d'équités pour appréhender de manière exhaustive le spectre des biais inhérents aux systèmes d'Intelligence Artificielle (IA). Cette prise de conscience a mené à des explorations dans le domaine de l'IA explicable, en quête d'une compréhension plus approfondie des biais, là où les métriques traditionnelles se révèlent insuffisantes.

Une réalisation centrale de cette thèse est la création de COCKATIEL, une méthode post-hoc d'explicabilité agnostique du modèle pour les modèles de TAL. Cette approche innovante intègre distinctement la découverte de concepts, leur classement et interprétation, s'harmonisant efficacement avec les explications conceptualisées par les humains, tout en restant fidèle aux principes fondamentaux des modèles.
Les expériences menées dans des tâches d'analyse de sentiment ont montré la capacité supérieure de COCKATIEL à découvrir des concepts qui s'alignent sur ceux des humains dans les modèles Transformers sans aucune supervision.

De plus, la thèse contribue à créer un pont entre l'équité et l'explicabilité en introduisant TaCo, une nouvelle méthode pour neutraliser les biais dans les embeddings des modèles Transformers.
En utilisant la stratégie d'explicabilité basée sur les concepts de COCKATIEL, cette approche identifie et élimine efficacement les concepts influençant principalement la prédiction de variables sensibles, produisant ainsi des embeddings moins biaisés. Cette méthode illustre le double rôle de l'explicabilité comme outil de compréhension et comme mécanisme pour renforcer l'équité dans les modèles d'IA.

En conclusion, cette thèse représente une contribution interdisciplinaire significative, alliant explicabilité et équité pour remettre en question et remodeler les paradigmes actuels de l'équité dans le TAL. Les méthodologies et critiques présentées ici contribuent profondément au discours en cours sur l'équité dans l'apprentissage automatique, offrant des solutions et des perspectives d'action pour la création de systèmes d'IA plus équitables et responsables. Les implications de cette recherche sont vastes et sont destinées à influencer les trajectoires de recherche futures et à guider le développement de technologies de TAL plus justes et responsables.

%% file: Abstract.tex
\chapter*{Abstract}

The burgeoning field of Natural Language Processing (NLP) stands at a critical juncture where the integration of fairness within its frameworks has become an imperative. This doctoral thesis addresses the need for equity and transparency in NLP systems, recognizing that fairness in NLP is not merely a technical challenge but a moral and ethical necessity, requiring a rigorous examination of how these technologies interact with and impact diverse human populations. Through this lens, this thesis undertakes a thorough investigation into the development of equitable NLP methodologies and the evaluation of biases that prevail in current systems.

My investigation starts by first introducing an innovative algorithm designed to mitigate algorithmic biases in multi-class neural-network classifiers, tailored for high-risk NLP applications as per EU regulations. This new approach outperforms traditional methods in terms of both bias mitigation and prediction accuracy, while providing flexibility in adjusting regularization levels for each output class. It thus moves away from the limitations of previous debiasing techniques based on binary models.

A pivotal aspect of this research involves an empirical analysis of the \textit{Bios} dataset, which contains LinkedIn biographies and the associated occupations. This investigation sheds light on the impact of the training dataset size on discriminatory biases, while also uncovering the deficiencies and inconsistencies of standard fairness metrics, particularly in the context of smaller datasets. The unpredictable nature of biases, and their reliability on the selected metrics underscore the current limitations of fairness measures in fully apprehending the spectrum of biases inherent in AI systems. This awareness has led to explorations in the field of explainable AI, with a view to a more complete understanding of biases, where traditional metrics are limited.

A central achievement of this thesis is the creation of COCKATIEL, an innovative, model-agnostic post-hoc explainability method for NLP models. This innovative approach distinctively integrates the discovery of concepts, their ranking, and interpretation, harmonizing effectively with explanations as conceptualized by humans, while still staying true to the foundational principles of the models. 
The experiments conducted in single and multi-aspect sentiment analysis tasks showed COCKATIEL's superior ability to discover concepts that align with Humans on Transformer models without any supervision.

Moreover, the thesis contributes to bridge the gap between fairness and explainability by introducing TaCo, a novel method to neutralize bias in Transformer model embeddings.
By using the concept-based explainability strategy of COCKATIEL, this approach effectively identifies and eliminates concepts predominantly influencing sensitive variable prediction, thus producing less biased embeddings. This method exemplifies the dual role of explainability as a tool for understanding and as a mechanism for enhancing fairness in AI models.

In conclusion, this thesis constitutes a significant interdisciplinary endeavor that intertwines explicability and fairness to challenge and reshape current NLP paradigms. The methodologies and critiques presented herein contribute profoundly to the ongoing discourse on fairness in machine learning, offering actionable solutions and insights for crafting more equitable and responsible AI systems. The far-reaching implications of this research are set to influence future research trajectories and guide the development of more just and accountable NLP technologies.

%% file: Introduction_fr.tex
\chapter{Introduction en français}

Au cours des dernières années, l'Intelligence Artificielle (IA) s'est imposée comme un outil incontournable dans de nombreux domaines de notre société. Si la maîtrise technique derrière ces avancées est souvent mise en avant, ce sont les implications sociales, et en particulier la notion d'équité, qui soulignent l'urgence d'un examen académique rigoureux dans ce champ.

Les algorithmes, de par leur nature, ne sont pas intrinsèquement neutres, justes ou équitables \citep{verma2018fairness}. Cette absence d'équité intrinsèque devient préoccupante lorsque les algorithmes commencent à manifester des biais, notamment à l'encontre de groupes identifiés par des attributs sensibles tels que l'orientation sexuelle, le genre ou la race. Ces biais algorithmiques ne sont pas que dus à de simples anomalies techniques, mais peuvent refléter des problèmes sociétaux plus profonds \citep{kleinberg2018discrimination}, nécessitant une compréhension globale et une approche proactive envers l'équité algorithmique. Il est moralement, mais aussi aujourd'hui légalement, impératif de s'attaquer à ces biais, en proposant un examen statistique et computationnel approfondi de l'équité dans les algorithmes. Notre objectif est de comprendre comment les biais se manifestent dans les décisions algorithmiques, leur impact sur divers groupes de population, et de développer des stratégies pour atténuer ces biais. Ce faisant, ce travail contribue à un corpus de connaissances en expansion qui cherche à équilibrer les avancées technologiques avec la responsabilité éthique et l'équité sociale.

\section{Enjeux sociaux et économiques de l'équité}

L'importance de l'équité dans l'apprentissage automatique découle souvent du fait que ces algorithmes opèrent sur des données historiques, susceptibles de contenir des biais discriminatoires et des représentations inégales de différents groupes. Si ces biais ne sont pas contrôlés, ils peuvent perpétuer ou même exacerber les inégalités existantes \citep{besse2022survey}.

Par exemple, dans le domaine de la finance, des modèles de notation de crédit biaisés peuvent perpétuer les disparités économiques. Certaines études, telles que \citep{hurlin2022fairness, besse2022survey}, explorent comment l'apprentissage automatique peut involontairement prolonger les biais historiques dans les pratiques de prêt (et comment tester l'existence d'une différence statistiquement significative entre les groupes protégés et non protégés).

Dans le domaine de la santé, des modèles d'apprentissage automatique non équitables peuvent conduire à un traitement inégal des patients. \cite{rajkomar2018ensuring} fournissent une analyse de la manière dont un algorithme de prédiction de risque de santé largement utilisé a montré un biais racial significatif. Cet article décrit comment la conception du modèle, les biais dans les données et les interactions des prédictions du modèle avec les cliniciens et les patients peuvent exacerber les disparités de soins de santé pour les populations ayant déjà subi des biais humains et structurels.
Une autre étude de \cite{cirillo2020sex} a exploré comment les différences et les biais de genre peuvent affecter le développement et l'utilisation de l'IA dans le domaine de la bio-médecine et des soins de santé. L'article discute des conséquences potentielles de ces différences et biais, y compris un accès inégal aux soins de santé et des diagnostics médicaux inexacts.

Dans le domaine de la justice pénale, l'équité dans les algorithmes est cruciale pour éviter de renforcer les biais contre certains groupes démographiques. Un exemple canonique provient d'un outil utilisé par les tribunaux aux États-Unis pour prendre des décisions de libération conditionnelle. Le logiciel, Correctional Offender Management Profiling for Alternative Sanctions (COMPAS), mesure le risque qu'une personne commette à nouveau un crime. Les juges utilisent COMPAS pour décider de libérer un délinquant ou de le maintenir en prison. De nombreuses enquêtes sur le logiciel ont révélé un biais contre les Afro-Américains \citep{compas2016}.
Une telle étude \citep{skeem2015risk} a montré que les algorithmes d'IA peuvent produire des résultats biaisés, en particulier lorsqu'ils sont entraînés sur des ensembles de données non représentatifs comme COMPAS. Cela peut entraîner des taux d'incarcération plus élevés pour certains groupes, tels que les minorités raciales, et perpétuer le racisme systémique dans le système de justice pénale.

Un autre domaine important dans lequel les biais algorithmiques peuvent impacter la société est le cas des publicités en ligne. Le ciblage publicitaire basé sur des facteurs démographiques tels que l'origine géographique, le genre et l'âge plutôt que sur les intérêts ou les comportements peut perpétuer des stéréotypes négatifs et entraîner une discrimination en limitant l'accès à des annonces d'emploi ou de logement pour certains groupes.
Par exemple, les algorithmes de diffusion d'annonces de Facebook, en optimisant pour un engagement maximal, peuvent conduire à des résultats biaisés qui entraînent l'amplification de certains groupes ou messages par rapport à d'autres. Cela peut conduire à une discrimination contre certains groupes, car les annonceurs peuvent cibler leurs publicités sur des démographies spécifiques ou empêcher certains groupes de voir leurs annonces de logement et d'emploi, comme le soulignent des études telles que \citep{ali2019discrimination} et \citep{sapiezynski2019algorithms}. Des discriminations étaient également présentes dans un algorithme qui diffusait des publicités promouvant des emplois dans les domaines des sciences, de la technologie, de l'ingénierie et des mathématiques (STEM) \citep{lambrecht2019algorithmic}. Cette publicité était conçue pour diffuser des annonces de manière neutre en termes de genre, cependant moins de femmes que d'hommes ont vu l'annonce, car il était devenu plus coûteux de montrer ces publicités aux femmes.

\section{Questions juridiques relatives à l'équité}

L'évolution rapide de l'IA et de la science des données a incité les autorités mondiales, notamment dans l'Union européenne, à élaborer une législation proactive pour faire face aux défis et implications sans cesse croissants de ces technologies de pointe. Le Règlement Général sur la Protection des Données (RGPD)\footnote{\url{https://gdpr-info.eu}}, adopté en mai 2018 par l'Union Européenne, constitue un cadre législatif avant-gardiste dans le domaine de la protection des données et, par extension, peut être perçu comme une première incursion dans la réglementation de l'IA en Europe. Ce règlement, bien qu'axé principalement sur la protection des données personnelles, a jeté les bases, de manière indirecte, pour la gouvernance des technologies d'IA, en raison de leur dépendance essentielle à l'égard de grandes quantités de données. Les principes du RGPD relatifs à la minimisation des données, au consentement, à la transparence et à la responsabilité revêtent une importance particulière dans le contexte de l'IA, ces technologies traitant souvent les données personnelles de manière complexe et parfois peu transparente.

L'influence du RGPD sur l'IA est diverse. Il a obligé les développeurs et utilisateurs d'IA à intégrer la confidentialité dès la conception des systèmes, promouvant ainsi une culture de « confidentialité par la conception » et de « confidentialité par défaut ». En outre, les dispositions du règlement, telles que le droit à l'explication et les restrictions sur la prise de décision automatisée, ont des répercussions directes sur l'IA, favorisant une plus grande transparence et responsabilité dans les systèmes d'IA. Ainsi, le RGPD peut être considéré comme la première étape significative de l'Europe dans la réglementation de l'IA, établissant un précédent pour la prise en compte des préoccupations éthiques et de confidentialité dans le développement et la mise en œuvre de l'IA.

S'appuyant sur les fondements établis par le RGPD, l'évolution du cadre juridique en matière de réglementation de l'IA a gagné en complexité et en importance, en particulier dans le domaine de l'équité de l'apprentissage automatique. Cette évolution, marquée par les initiatives réglementaires récentes de la Commission européenne, témoigne d'un engagement renforcé à relever les défis subtils posés par les technologies d'IA. Ces initiatives ont été fortement influencées par une prise de conscience croissante dans le discours juridique sur l'équité de l'IA : bien que les systèmes d'apprentissage automatique soient révolutionnaires, ils ne sont pas naturellement exempts de biais. Cette prise de conscience découle de cas très médiatisés, comme l'algorithme de récidive COMPAS évoqué dans la section précédente, qui ont mis en lumière les risques de discrimination algorithmique. La proposition de la Commission européenne pour un règlement sur les systèmes d'IA\footnote{\url{https://eur-lex.europa.eu/legal-content/EN/TXT/?uri=CELEX\%3A52021PC0206}}, qui vise à lutter contre les biais algorithmiques, est au centre de cette problématique. Notamment, son article 10 impose aux fournisseurs de systèmes d'IA d'utiliser des données d'entraînement pertinentes, représentatives, exemptes d'erreurs et complètes, s'attaquant ainsi à une source majeure de biais algorithmique : les données d'entraînement défectueuses. La proposition met l'accent sur la gouvernance des données d'entraînement, une étape cruciale pour assurer une plus grande équité des systèmes d'IA.

De plus, la proposition de la Commission autorise le traitement de catégories spéciales de données lorsqu'il est nécessaire pour détecter et corriger les biais, sous réserve de mesures rigoureuses de protection de la vie privée et de sécurité. Cette disposition constitue une avancée significative dans la lutte contre les biais algorithmiques, reconnaissant la nécessité d'inclure des données sensibles dans les évaluations d'équité tout en préservant des normes élevées de protection des données. La réglementation européenne met également l'accent sur la nécessité de propriétés statistiques appropriées concernant les personnes ou groupes cibles des systèmes d'IA à haut risque. Cette exigence implique une vigilance accrue envers les groupes susceptibles d'être discriminés, assurant leur prise en compte dans la conception des systèmes d'IA. En outre, la proposition détaille les modalités d'évaluation de la conformité des systèmes d'IA, en s'appuyant sur des métriques et seuils probabilistes. Toutefois, le choix de ces métriques dépasse la simple expertise technique et requiert une collaboration entre juristes et technologues pour s'assurer de leur conformité avec la conception juridique de la non-discrimination.

En somme, la proposition de la Commission européenne représente une initiative pionnière dans la formulation juridique du concept d'équité en IA. Elle souligne l'importance d'une approche multidisciplinaire, alliant perspectives techniques, juridiques et éthiques, pour garantir que les systèmes d'IA progressent technologiquement tout en respectant les principes d'équité et de non-discrimination. Pour une mise en œuvre et une conformité réussies avec ces cadres juridiques en évolution sur l'équité en IA, le rôle des chercheurs en mathématiques et en informatique est essentiel. Ces chercheurs apportent une perspective cruciale dans la traduction des normes juridiques d'équité en solutions technologiques pratiques. Leur participation active est indispensable pour assurer que les technologies d'IA respectent les mandats légaux et contribuent à promouvoir une société numérique plus équitable et juste.

Conformément à la loi sur l'IA de la Commission européenne, un mouvement mondial émergeant vers la réglementation de l'IA se dessine, avec divers pays et régions adoptant leurs propres mesures. Ces réglementations, couvrant un éventail d'applications d'IA, incluent notamment, dans le contexte des pratiques d'emploi, la ville de New York. Une nouvelle loi, entrant en vigueur en juillet 2023, exige des audits pour détecter les biais dans les Outils de Décision Automatisés pour l'Emploi (AEDTs)\footnote{\url{https://www.shrm.org/resourcesandtools/hr-topics/technology/pages/new-york-city-clarifies-who-what-covered-under-ai-bias-law.aspx}}, marquant ainsi un effort international croissant pour assurer l'équité dans les applications d'IA au-delà des frontières européennes.

Cette loi impose aux employeurs de réaliser des audits de leurs systèmes technologiques aux services des Ressources Humaines (RH) pour identifier les biais et de publier les résultats. Elle s'applique spécifiquement aux employeurs et agences d'emploi pour les postes situés à New York. Les AEDTs sont définis comme des outils aidant ou remplaçant la prise de décision dans les processus d'embauche, y compris les algorithmes d'analyse de CV et les chatbots pour les entretiens. La loi exige que ces audits évaluent l'impact potentiellement disparate sur le sexe, l'origine géographique et l'ethnicité. Les employeurs sont tenus de garantir la conformité avec cette loi et de publier les résultats des audits, qui doivent inclure divers détails sur le processus d'audit et les conclusions. Cette loi représente un effort significatif pour aborder les biais dans les pratiques d'emploi pilotées par l'IA.

\section{Définitions de l'équité}

Il est essentiel de reconnaître que le terme \textit{équité} n'est pas univoque ; il englobe une pluralité de définitions, chacune portant ses implications théoriques et conséquences pratiques distinctes. \cite{kim2020fact} mettent en lumière la diversité et l'incompatibilité potentielle de ces définitions, soulignant ainsi la complexité intrinsèque à l'atteinte de l'équité dans les systèmes algorithmiques.

Pour appréhender les subtilités de l'équité, il convient d'introduire deux notions fondamentales : l'\textit{attribut sensible} et le \textit{groupe protégé}. Un attribut sensible désigne une caractéristique susceptible de fonder une discrimination, incluant le genre, l'ethnie, la religion ou l'orientation sexuelle. Le groupe protégé regroupe des individus qui, de par leurs attributs sensibles, sont susceptibles de subir des discriminations. À titre d'exemple, on peut citer le genre (femme) ou l'ethnie (personne racisée).

Une approche intuitive de l'équité est celle de l'\textit{équité par ignorance}, telle que décrite par \cite{dwork2012fairness}. Ce concept suggère qu'un algorithme peut être considéré comme équitable s'il n'emploie pas explicitement des attributs sensibles dans son processus décisionnel. Toutefois, cette approche est limitée, car elle ne prend pas en compte la discrimination indirecte. Par exemple, l'élimination d'un attribut sensible tel que le genre ne garantit pas l'équité si un autre attribut non sensible, mais corrélé (tel que la taille) peut révéler indirectement l'information sensible, conduisant à des décisions biaisées.

Les définitions de l'équité se répartissent largement en deux catégories : l'\textit{équité de groupe} (non causale) et l'\textit{équité causale} (également nommée \textit{équité individuelle}, bien que ce terme ne couvre pas toutes les définitions de l'\textit{équité causale}). Chacune repose sur le principe d'égalité dans une métrique donnée $M$, avec la condition
$M(A = 0) = M(A = 1)$, où $A$ représente une variable binaire indiquant l'appartenance au groupe protégé.

\paragraph{Équité de Groupe} Les concepts d'\textit{équité de groupe} se concentrent sur des métriques qui ne sont pas basées sur des relations causales. Ils sont généralement plus simples à implémenter, car ils s'appuient sur des corrélations statistiques plutôt que sur des liens de causalité.

Une des principales métriques en matière d'équité non causale est la \textit{parité statistique}, définie par \cite{dwork2012fairness}. Cette métrique insiste sur l'égalité de probabilité d'un résultat prédit entre différents groupes. Un concept connexe est la \textit{parité statistique conditionnelle}, proposée par \cite{verma2018fairness}, qui requiert une probabilité égale de résultat prédit entre les groupes, compte tenu d'un ensemble de caractéristiques légitimes.

En outre, \cite{hardt2016equality} abordent des métriques telles que l'\textit{égalité des opportunités} (ou \textit{écart du taux de vrais positifs}), visant à assurer que la probabilité d'être prédit dans une classe donnée soit la même pour les individus protégés et non protégés ; ou
la \textit{parité prédictive}, qui vise à garantir que, pour tous les individus prédits appartenant à une classe donnée, la probabilité que ces individus soient réellement dans cette classe soit la même pour les individus protégés et non protégés.

La condition de \textit{calibration} \citep{chouldechova2017fair} est respectée si, pour tout score de probabilité prédit, la probabilité d'appartenir effectivement à la classe positive est la même pour les individus protégés et non protégés. Elle est semblable à la \textit{parité prédictive} mais plus exigeante, en ce sens que l'égalité doit être maintenue pour chaque niveau de confiance.

L'\textit{équilibre pour la classe positive/négative}, introduit par \cite{kleinberg2016inherent}, met en avant que le score de probabilité prédit moyen pour des résultats positifs/négatifs devrait être identique pour les groupes protégés et non protégés. L'objectif est de s'assurer que l'algorithme ne manifeste pas une confiance excessive ou insuffisante dans ses prédictions pour un groupe spécifique.

\paragraph{Équité Causale} L'\textit{équité causale} va au-delà de l'aspect superficiel des décisions algorithmiques, explorant des situations où les individus possèdent des attributs différents de ceux qu'ils ont en réalité. Cette approche s'intéresse aux implications plus profondes et aux chemins empruntés par les processus décisionnels.

La définition la plus simple de l'\textit{équité causale} est celle de l'\textit{absence d'effet total}, mise en évidence dans l'étude de \cite{makhlouf2020survey}. Cette métrique vise à assurer une probabilité égale de résultat, indépendamment de toute intervention sur un attribut sensible. Elle examine l'impact global de la modification d'un attribut sensible sur la décision finale ou le résultat de l'algorithme. C'est la version causale de la \textit{parité statistique}.

\textit{Pas d'effet du traitement sur les traités} \citep{makhlouf2020survey} implique que le résultat de la décision reste inchangé, quelles que soient les modifications apportées à un attribut sensible, en supposant que l'attribut ait initialement une certaine valeur. Cette idée représente un aspect fondamental de l'équité, mettant en exergue l'importance de résultats cohérents en présence d'interventions potentielles sur des caractéristiques sensibles. Il s'agit d'une approche directe mais essentielle de l'équité basée sur l'analyse contrefactuelle.

Une métrique clé en \textit{équité causale} est l'\textit{équité contrefactuelle}, telle que décrite par \cite{kusner2017counterfactual}. L'\textit{équité contrefactuelle} représente une forme nuancée du principe de \textit{pas d'effet du traitement sur les traités}. Elle prend en compte tous les attributs pertinents dans son évaluation. Reconnue dans les études sur l'\textit{équité causale}, cette approche assure que les résultats des décisions restent cohérents dans différents scénarios, mettant l'accent sur l'équité du traitement indépendamment des attributs variables.

L'\textit{effet causal moyen équitable} - et respectivement l'\textit{effet causal moyen équitable sur les traités} - est la version attendue de l'\textit{absence d'effet total} - et respectivement \textit{pas d'effet du traitement sur les traités} - introduite par \cite{khademi2019fairness}. Ces deux concepts sont fondamentaux pour la compréhension de l'\textit{équité causale}, se concentrant sur la cohérence des attentes dans la prise de décision.

Les \textit{effets directs et indirects} \citep{zhang2018fairness}, sont également essentiels dans l'évaluation de l'\textit{équité causale}. Cette métrique évalue l'équité en se basant sur les voies directes et indirectes par lesquelles les attributs sensibles influencent les résultats. Elle permet de décomposer les manières complexes par lesquelles divers attributs contribuent à la décision finale, garantissant une approche exhaustive de l'équité dans les décisions algorithmiques.

\paragraph{}La nature complexe et multivalente des définitions de l'équité dans les contextes algorithmiques met en relief les défis liés à la conception de systèmes véritablement équitables. Chaque définition présente ses forces et ses limites, et le choix parmi elles reflète souvent les priorités éthiques spécifiques et les considérations pratiques d'une application donnée.

\paragraph{} En pratique, il est quasiment impossible de respecter la condition $M(A=0)=M(A=1)$. On examine alors plutôt : $M\_Gap = M(A=0) - M(A=1)$, qui représente l'écart entre le groupe protégé et les autres. Il est courant de fixer un seuil $\tau$ tel que si $|M(A=0) - M(A=1)| < \tau$, alors le modèle sera considéré comme équitable.

\section{L'équité dans le Traitement Automatique des Langues}

Les modèles de Traitement Automatique des Langues (TAL) représentent une catégorie d'IA conçue pour gérer des tâches associées au langage humain. Ces tâches comprennent la réponse aux questions, la classification de textes, la traduction, la synthèse de contenu et l'extraction d'informations significatives.
Pour exécuter ces tâches, les modèles traitent et analysent d'importantes quantités de données en langage naturel. Bien que l'idée que ces modèles «comprennent» le langage puisse être sujet à débat sur le plan philosophique, leur efficacité pratique est manifeste, résultant de la fusion des principes de la linguistique et de l'informatique. Le développement de ces modèles, initialement orienté par des tâches spécifiques telles que celles évoquées dans les travaux initiaux de \cite{bengio2000neural}, soulève des interrogations quant à la «compréhension» dans le contexte de leurs progrès rapides.

Avec la croissance importante des modèles de TAL, en particulier des versions avancées telles que le modèle génératif de ChatGPT, les préoccupations liées à l'équité ont pris de l'ampleur. Des modèles antérieurs tels que BERT \citep{devlin2018bert} et RoBERTa \citep{liu2019roberta}, formés sur d'abondants textes issus d'Internet, avaient déjà révélé des biais potentiels dus à leurs tâches d'apprentissage non supervisées (telles que la prédiction du mot suivant, de la phrase suivante et du mot masqué), qui manquaient souvent de révision d'échantillons de données humaines. De plus, les ajustements basés sur les préférences humaines dans ces modèles pourraient introduire des biais supplémentaires. Ces biais, hérités du matériel de formation, peuvent entraîner des résultats inéquitables, notamment dans des applications critiques comme le filtrage de CV et le recrutement. Par exemple, un modèle principalement formé sur des données reflétant des stéréotypes de genre pourrait, sans le vouloir, perpétuer ces biais dans les processus de sélection de CV. L'utilisation étendue du TAL dans des domaines sensibles met en exergue la nécessité d'une considération rigoureuse de l'équité et des biais dans ces algorithmes.

Par ailleurs, avec l'émergence de nouveaux modèles génératifs, l'accessibilité des technologies de TAL s'est nettement améliorée, les rendant plus facile à utiliser. Cette facilité d'utilisation augmente la probabilité d'une adoption généralisée. Il est donc essentiel de veiller à ce que ces systèmes soient appliqués de manière éthique et équitable pour favoriser leur développement durable et leur acceptation plus large. La simplicité et l'accessibilité de ces modèles doivent être contrebalancées par une démarche consciencieuse visant à atténuer les biais, afin de garantir que leurs avantages se concrétisent sans causer de dommages involontaires ou perpétuer des inégalités.

Compte tenu des avancées significatives apportées par les modèles de TAL à l'IA et à l'interaction homme-machine, il est crucial de s'attaquer aux questions d'équité. Assurer une application éthique et équitable de ces systèmes est indispensable pour leur progression continue et leur acceptation dans divers secteurs de la société.

Dans cette thèse, nous nous attacherons à examiner la question de l'équité des algorithmes de TAL de toutes sortes, en mettant particulièrement l'accent sur les architectures Transformers (telles que BERT ou GPT) appliquées aux tâches de classification. Ce type de tâches est au cœur de l'automatisation des décisions dans des domaines critiques tels que le recrutement, entre autres.

%% file: Introduction.tex
\chapter{Introduction}\label{introduction}

In recent years, artificial intelligence has become an unavoidable tool in many areas of our society. While the technical mastery behind these advancements is often in the spotlight, it is the social implications, particularly the concept of fairness, that underscores the urgency for rigorous academic scrutiny in this field. 

Algorithms, by their nature, are not intrinsically neutral, fair, or equitable \citep{verma2018fairness}.  This lack of inherent fairness becomes a source of concerns when algorithms begin to exhibit biases, especially against groups identified by sensitive attributes like sexual orientation, gender, or race. Such biases in algorithms are not merely technical anomalies but reflections of deeper societal issues \citep{kleinberg2018discrimination}, necessitating a comprehensive understanding and proactive approach towards algorithmic fairness. It is imperative to address these biases, proposing a thorough statistical and computational examination of fairness in algorithms. Our goal is to understand how biases manifest in algorithmic decisions, their impact on diverse population groups, and to develop strategies for mitigating these biases. In doing so, this work contributes to a growing body of knowledge that seeks to balance the scales of technological advancement with ethical responsibility and social equity.

\section{Social and economic issues of Fairness}

The importance of Fairness in machine learning often stems from the fact that these algorithms often operate on historical data, which may contain discriminatory biases and unequal representations of different groups.
If unchecked, these biases can perpetuate or even exacerbate existing inequalities \citep{besse2022survey}.

For instance, in the field of finance, biased credit scoring models can perpetuate economic disparities. Some studies like \citet{hurlin2022fairness, besse2022survey} explore how machine learning can inadvertently extend historical biases in lending practices (and how to test whether there exists a statistically significant difference between protected and unprotected groups).

In healthcare, ML models that are not fair can lead to unequal treatment of patients. \cite{rajkomar2018ensuring} provide an analysis of how a widely used healthcare risk prediction algorithm exhibited significant racial bias. This article describes how model design, biases in data, and the interactions of model predictions with clinicians and patients may exacerbate health care disparities for populations that have experienced human and structural biases in the past.
Another study by \cite{cirillo2020sex} explored how gender differences and biases can affect the development and use of artificial intelligence in the field of bio-medicine and healthcare. The paper discussed the potential consequences of these differences and biases, including unequal access to healthcare and inaccurate medical diagnoses.

In criminal justice, fairness in predictive policing algorithms is crucial to prevent reinforcing biases against certain demographic groups. A canonical example comes from a tool used by courts in the United States to make pretrial detention and release decisions. The software, Correctional Offender Management Profiling for Alternative Sanctions (COMPAS), measures the risk of a person to recommit another crime. Judges use COMPAS to decide whether to release an offender, or to keep him or her in prison. Many investigations into the software found a bias against African-Americans \citep{compas2016}. 
One such study \cite{skeem2015risk} found that AI algorithms produce biased outcomes, particularly when trained on non-representative data sets like COMPAS. This can result in higher incarceration rates for certain groups, such as racial minorities, and perpetuate systemic racism in the criminal justice system.

Another important area in which algorithmic biases can impact society is the case of online advertisements. Ad targeting based on demographic factors such as race, gender, and age rather than interests or behaviors can perpetuate negative stereotypes and result in discrimination by limiting access to job or housing announcements for certain groups.
For example, Facebook's ad delivery algorithms, by optimizing for maximum engagement, can lead to biased outcomes that result in the amplification of certain groups or messages over others. This can lead to discrimination against certain groups, as advertisers may target their ads to specific demographics or exclude certain groups from seeing their housing and employment advertising, as highlighted by studies such as \cite{ali2019discrimination} and \cite{sapiezynski2019algorithms}.
Discriminatory behavior was also evident in an algorithm that would deliver advertisements promoting jobs in Science, Technology, Engineering, and Math (STEM) fields \citep{lambrecht2019algorithmic}. This advertisement was designed to deliver advertisements in a gender-neutral way. However, less women compared to men saw the advertisement due to gender-imbalance which would result in younger women being considered as a valuable subgroup and more expensive to show advertisements to.

\section{Legal issues of Fairness}

Rapid advancements in AI and data science has prompted authorities worldwide, particularly in the European Union, to implement new proactive legislation, addressing the evolving challenges and implications of these cutting-edge technologies.
The General Data Protection Regulation (GDPR)\footnote{\url{https://gdpr-info.eu}}, implemented in May 2018 by the European Union, stands as a pioneering legislative framework in the realm of data protection and, by extension, can be viewed as an initial foray into the regulation of Artificial Intelligence (AI) in Europe. This regulation, while primarily focused on the protection of personal data, inadvertently laid the groundwork for the governance of AI technologies, given their intrinsic reliance on vast quantities of data. The GDPR's principles of data minimization, consent, transparency, and accountability have become particularly relevant in the context of AI, as these technologies often process personal data in complex and sometimes opaque ways.

The GDPR's impact on AI is multifaceted. It has compelled AI developers and users to consider privacy at the onset of system design, promoting a culture of 'privacy by design' and 'privacy by default'. Furthermore, the regulation's provisions, such as the right to explanation and the restrictions on automated decision-making, have direct implications for AI, pushing towards greater transparency and accountability in AI systems. In this sense, the GDPR can be seen as Europe's first significant step in regulating AI, setting a precedent for the consideration of ethical and privacy concerns in AI development and deployment.

Building upon the foundation established by the GDPR, the evolving legal landscape in AI regulation has gained complexity and significance, particularly in the realm of machine learning (ML) fairness. This progression, marked by the European Commission's recent regulatory initiatives, reflects a deepening commitment to address the nuanced challenges posed by AI technologies. These initiatives have been significantly influenced by an emerging recognition within the legal discourse on ML fairness: that while machine learning systems are transformative, they are not inherently free of biases. This understanding stems from high-profile cases, such as the COMPAS recidivism algorithm presented in the previous section, which have highlighted the risks of algorithmic discrimination. Central to addressing these issues is the European Commission's proposal for a regulation on AI systems \footnote{\url{https://eur-lex.europa.eu/legal-content/EN/TXT/?uri=CELEX\%3A52021PC0206}}, which offers a comprehensive approach to combating algorithmic bias. The proposal, notably in its Article 10, mandates that providers of AI systems use training data that is relevant, representative, error-free, and complete. This requirement addresses a primary source of algorithmic bias: flawed training data. Whether under or over-represented, certain population groups can significantly skew the system's fairness. The proposal's focus on training data governance is a crucial step in ensuring AI systems are more equitable.

Moreover, the Commission's proposal allows for the processing of special categories of data when necessary for detecting and correcting biases, subject to strict privacy and security measures. This provision marks a significant advancement in the fight against algorithmic bias, acknowledging the need for sensitive data to be included in fairness evaluations while maintaining high data protection standards. The European regulation also emphasizes the need for appropriate statistical properties regarding the people or groups for whom the high-risk AI system is intended. This requirement necessitates increased vigilance towards potentially discriminated groups, ensuring their consideration in the design of AI systems. In addition, the proposal outlines the modalities for assessing AI system conformity, leveraging probabilistic metrics and thresholds. The choice of these metrics, however, transcends mere technical expertise, requiring a collaboration between legal experts and technologists to ensure they align with the legal conception of non-discrimination.

In essence, the European Commission's proposal represents a pioneering effort in legally framing the concept of fairness in AI. It underscores the importance of a multidisciplinary approach, blending technical, legal, and ethical perspectives to ensure that AI systems not only advance technologically but also adhere to principles of fairness and non-discrimination. To ensure the successful implementation and compliance with these evolving legal frameworks on AI fairness, the role of mathematical and computer science researchers becomes indispensable. These researchers bring a critical perspective in translating legal standards of fairness into practical, technological solutions. The active involvement of these experts is crucial in ensuring that AI technologies not only comply with legal mandates but also contribute to fostering a more equitable and just digital society.

In line with European Commission's AI Act, there is an emerging global movement towards the regulation of AI, with various countries and regions adopting their own measures. While these regulations address a range of AI applications, a prominent example in the context of employment practices is New York City. 
A new law, effective July 2023, mandates audits for bias in Automated Employment Decision Tools (AEDTs) \footnote{\url{https://www.shrm.org/resourcesandtools/hr-topics/technology/pages/new-york-city-clarifies-who-what-covered-under-ai-bias-law.aspx}}, signifying an expanding international effort to ensure fairness in AI applications beyond European borders.

This law requires employers to audit their HR technology systems for bias and publish the results. It applies specifically to employers and employment agencies for jobs located in New York City. AEDTs are defined as tools that assist or replace decision-making in employment decisions, including algorithms for analyzing resumes and chatbots for interviews. The law mandates that bias audits assess the potential disparate impact on sex, race, and ethnicity. Employers are responsible for ensuring compliance with this law and for publishing audit results, which must include various details about the audit process and findings. This law represents a significant effort to address biases in AI-driven employment practices.

\section{Fairness definitions}\label{intro:fairnessmetrics}

It's crucial to acknowledge that "fairness" is not a monolithic term; rather, it encompasses a multitude of definitions, each with its own theoretical implications and practical consequences. \cite{kim2020fact} highlight the diversity and potential incompatibility of these definitions, underscoring the complexity inherent in achieving fairness in algorithmic systems.

To understand the nuances of fairness, it is essential to introduce two key concepts: the "sensitive attribute" and the "protected group". A sensitive attribute is a characteristic that can be the basis of discrimination, encompassing gender, race, religion, or sexual orientation. The protected group consists of individuals who, due to their sensitive attributes, are at risk of experiencing discrimination. Examples include gender (“woman”) or race (“people of color”).

One intuitive approach to fairness is the idea of "Fairness through Unawareness", as described by \cite{dwork2012fairness}. This concept posits that an algorithm can be considered fair if it does not explicitly use sensitive attributes in its decision-making process. However, this approach is limited as it overlooks the potential for indirect discrimination. For instance, removing a sensitive attribute like gender does not guarantee fairness if another correlated, non-sensitive attribute (e.g., height) can indirectly reveal the sensitive information, leading to biased decisions.

Fairness definitions are broadly categorized into "group fairness" (which can be described as non-causal) and 'causal fairness' (also called individual fairness, but this term does not encompass all existing definitions of causal fairness). Each derives from the principle of equality in a chosen metric $M$, with the condition
$M(A = 0) = M(A = 1)$, where $A$ is a binary variable representing membership in the protected group.

\paragraph{Group Fairness} Group fairness concepts focus on metrics not based on causal relationships. These are generally simpler to implement, as they rely on statistical correlations rather than causal linkages. 

One of the main metrics in non-causal fairness is Statistical Parity, as defined by \cite{dwork2012fairness}. This metric emphasizes the equal probability of a predicted outcome across different groups. Another related concept is Conditional Statistical Parity, proposed by \cite{verma2018fairness}, which requires equal probability of a predicted outcome across groups, given a set of legitimate features. 

In addition, \cite{hardt2016equality} discuss metrics like Equality of Opportunity (or True Positive Rate gap), which aims to guarantee for all individuals in a chosen class, the probability of being predicted in this class is the same for protected and unprotected individuals, or 
Predictive Parity, which aims to ensure that, for all individuals predicted to belong to a chosen class, the probability that these individuals are actually in this class is the same for both protected and unprotected individuals.

The condition of Calibration \citep{chouldechova2017fair} is verified if, for any predicted probability score, the probability of actually belonging to the positive class is the same for protected and unprotected individuals. It is similar to Predictive parity yet stronger, in the sense that the equality should hold for any single level of confidence.

Balance for Positive/Negative Class, a concept introduced by \cite{kleinberg2016inherent}, emphasizes that the average predicted probability score for positive/negative outcomes should be the same for both protected and unprotected groups. The goal here is to ensure that the algorithm does not exhibit undue confidence or lack thereof in its predictions for any specific group.

\paragraph{Causal Fairness} Causal fairness goes beyond the surface level of algorithmic decisions, exploring situations where individuals possess different attributes than they do in reality. This approach delves into the deeper implications and pathways of decision-making processes. 

The simplest approach to causal fairness is the definition of No Total Effect, highlighted in the survey by \cite{makhlouf2020survey}. This metric focuses on ensuring an equal probability of an outcome, irrespective of any intervention on a sensitive attribute. It examines the overall impact of altering a sensitive attribute on the final decision or outcome of the algorithm. It is the causal version of the definition of Statistical Parity.

The concept of 'No Effect of Treatment on the Treated' \citep{makhlouf2020survey} in causal fairness implies that the decision outcome remains constant regardless of changes made to a sensitive attribute, assuming the attribute initially has a certain value. This idea represents a fundamental aspect of fairness, emphasizing the importance of consistent outcomes in the presence of potential interventions on sensitive characteristics. It's a straightforward yet crucial approach to fairness based on counterfactual analysis.

A key metric in causal fairness is Counterfactual Fairness, as described by \cite{kusner2017counterfactual}. Counterfactual fairness represents a nuanced form of the 'No effect of treatment on the treated' principle. It considers all relevant attributes in its assessment. Widely recognized in causal fairness studies, this approach ensures that decision outcomes remain consistent across different scenarios, emphasizing fairness in treatment regardless of varying attributes.

The 'fair on average causal effect' - and respectively 'fair on average causal effect on treated'-  is the expected version of the 'no total effect' -and respectively 'no effect of treatment on the treated'-  introduced by \cite{khademi2019fairness}. Both concepts are central to understanding causal fairness, focusing on expectation consistency in decision-making.

Direct and Indirect Effects \citep{zhang2018fairness}, are also crucial in evaluating causal fairness. This metric assesses fairness based on both the direct and indirect pathways through which sensitive attributes influence outcomes. It helps to dissect the complex ways in which various attributes contribute to the final decision, ensuring a comprehensive approach to fairness in algorithmic decisions.

\paragraph{}The multifaceted nature of fairness definitions in algorithmic contexts underscores the challenges in designing truly equitable systems. Each definition has its strengths and limitations, and the choice among them often reflects the specific ethical priorities and practical considerations of a given application.

\paragraph{} In practice, it's virtually impossible to validate the condition $M(A=0)=M(A=1)$. Instead, we look at $M\_Gap = M(A=0) - M(A=1)$, which represents the gap between the protected group and the others. It's common to set a threshold $\tau$ such that if $|M(A=0) - M(A=1)| < \tau$, then the model will be considered fair.

\section{Natural Language Processing Algorithms and Fairness Concerns}

Natural Language Processing (NLP) models are types of artificial intelligence designed to handle tasks that involve human language. These tasks include answering questions, categorizing text, translating languages, summarizing content, and extracting meaningful information. 
To perform these tasks, the models process and analyze large quantities of natural language data. While the concept of these models 'understanding' language is philosophically debatable, their practical effectiveness is evident thanks to the combination of principles from linguistics and computer science. The development of these models, initially driven by specific tasks like those highlighted in early works by \cite{bengio2000neural}, raises questions about 'understanding' amidst their rapid advancements.

With the increasing prominence of NLP models, particularly advanced versions like the generative model on ChatGPT, fairness concerns have gained attention. Earlier models like BERT \citep{devlin2018bert} and RoBERTa \citep{liu2019roberta}, trained on extensive internet text data, already hinted at potential biases due to their unsupervised training tasks (like next word, next sentence, and masked word prediction), which often lacked human data sample review. Furthermore, human preference adjustments in these models also could introduce additional biases. These inherited biases in training material can lead to unfair outcomes, especially in critical applications like resume screening and job recruitment. For instance, a model predominantly trained on data reflecting gender stereotypes might inadvertently perpetuate these biases in CV selection processes. The widespread application of NLP in sensitive domains underscores the imperative for careful consideration of fairness and bias in these algorithms.

Moreover, with the advent of new generative models, the usability of NLP technologies has significantly improved, making them more accessible and easy to use. This ease of use increases the likelihood of widespread adoption. Thus, ensuring that these systems are applied ethically and equitably is essential for their sustainable advancement and broader acceptance. The simplicity and accessibility of these models must be balanced with a conscientious approach to mitigating biases, ensuring that their benefits are realized without inadvertently causing harm or perpetuating inequalities.

Given the significant advances NLP models bring to artificial intelligence and human-computer interaction, addressing fairness issues is crucial. Ensuring ethical and equitable application of these systems is essential for their continued advancement and acceptance in various spheres of life.

In this thesis, we will focus on the issue of fairness in various types of NLP algorithms, with a particular emphasis on Transformer architectures (such as BERT or GPT) for classification tasks. These tasks are central to the automation of decision-making in critical areas such as recruitment, among others.

%% file: Preliminaries.tex
\chapter{Preliminaries}

In the introduction, we established the significance of fairness, especially in modern NLP models, highlighting the challenges and ethical considerations inherent in this rapidly evolving field. This chapter builds upon that foundation, focusing on the historical evolution of NLP models, the intricate nature of biases within these systems, and the overarching importance of datasets in NLP research.

The historical progression of NLP models is crucial for understanding how and why certain biases are embedded in these technologies. By tracing this development  in the section \ref{nlpstory}, we aim to uncover the roots of current challenges in achieving fairness and to identify opportunities for improvement.

We will also focus on the dual nature of biases in NLP in the section \ref{biaisnlp}. While some biases are integral to effective language processing, others can lead to unfair outcomes. This chapter seeks to differentiate between these types of biases and explore strategies for mitigating the negative impacts, thereby aligning NLP practices with ethical standards.

Lastly, the chapter will address the importance of having high-quality data in NLP, followed by an introduction to the specific datasets used in this thesis in the section \ref{nlpdataset}. This discussion will highlight the challenges and considerations in selecting and curating datasets for NLP research. By examining these datasets, we aim to illustrate how they contribute to the research and findings of this thesis, emphasizing the role of data in shaping the outcomes of NLP models.

Through this exploration, the chapter aims to provide a comprehensive understanding of the complexities in NLP, guiding towards more equitable and responsible applications in the field.

\section{NLP models}\label{nlpstory}

The evolution of Natural Language Processing (NLP) technologies has been characterized by significant advancements, each addressing specific challenges in the quest to enhance machine understanding of human language. 

The Bag of Words (BoW) model \citep{harris1970distributional}, one of the earliest NLP techniques, represented text by counting the frequency of each word, treating words as independent features. This approach, while foundational, was limited in its inability to capture the order or contextual nuances of language. Building on BoW, the Term Frequency-Inverse Document Frequency (TF-IDF) \citep{salton1988term} method emerged. 
TF-IDF is a statistical measure used to evaluate the importance of a word in a document, which is part of a corpus, by increasing proportionally to the number of times a word appears in the document and offset by the frequency of the word in the corpus. This method helped in distinguishing the significance of words in a document, allowing for a more nuanced text analysis.

The introduction of Word Embeddings, with techniques like Word2Vec \citep{mikolov2013efficient} and GloVe \citep{pennington2014glove}, marked a significant leap in NLP. Before the advent of these techniques, constructing lexical meaning spaces over large corpora existed but involved a labor-intensive, anual process. This required specifying how each element in the content word lexicon was affected by context-dependent processes of polysemy resolution, an approach that was not feasible for the thousands of items in the lexicon due to its enormity and complexity \citep{baroni2014frege}.
The Word Embeddings methods represented words in dense vector spaces based on their contextual usage, capturing complex word relationships and similarities.  
This was a substantial improvement over the word context insensitivity of BoW and TF-IDF. Although in any case, all its approaches are still insensitive to the context of the task, as they are created independently of the task being viewed.

Building upon the foundation laid by Word Embeddings, the development of the Continuous Bag of Words (CBoW) \citep{mikolov2013distributed} and Skip-gram \citep{guthrie2006closer} models represented further refinements in the domain of word representations. CBoW, by focusing on predicting a specific word based on its contextual words, provided a more nuanced approach to understanding word usage in different contexts. The Skip-gram model reversed this process, aiming to predict the context surrounding a given target word. Both CBoW and Skip-gram enhanced the contextual sensitivity of Word Embeddings.

The transition to Word Embeddings was crucial for the development and effectiveness of Recurrent Neural Networks (RNNs) \citep{medsker2001recurrent}. 
RNNs are a class of artificial neural networks designed to recognize patterns in sequences of data, such as text, genomes, handwriting, or spoken words. Unlike feedforward neural networks, where information moves in only one direction from the input layer to the output layer, RNNs have connections that form directed cycles. This architecture allows them to retain a form of memory, as they use their internal state (memory) to process sequences of inputs.
Although RNNs can also be run on more basic representation types (such as BoW, TF-IDF), it's the context of each word in relation to the others in a sentence that has performed well on tasks such as language modeling, text generation and sentiment analysis.  
Word embeddings provided a way to feed rich, contextually informed representations of words into RNNs and other neural architectures, enabling these models to process sequences of data more effectively. However, they faced challenges with long-range dependencies due to the vanishing gradient problem. Long Short-Term Memory (LSTM) \citep{hochreiter1997long} networks, a variant of RNNs, addressed this issue with their unique gating mechanisms - the input, output, and forget gates. These gates regulated the flow of information, allowing LSTMs to maintain important inputs over long sequences. Crucially, LSTMs introduced a form of attention mechanism. This attention in LSTMs was not as explicit or central as in later models but was implicit in how the gates could learn to retain or forget information, focusing the model's memory on certain parts of the input sequence. This was a foundational step towards more sophisticated attention mechanisms.

The Transformer model, introduced in "Attention Is All You Need" by \cite{vaswani2017attention}, represented a paradigm shift. 
This model diverged from previous methodologies that primarily generated word representations based on the immediate context of words or sentences. Instead, the Transformer adopted a task-oriented approach to word representation, enabling dynamic adjustments of word representations tailored to the specific requirements of tasks such as translation, summarization, or question answering. 
Distinguishing itself from LSTMs, Transformers processed input data in parallel using self-attention mechanisms. This self-attention allowed each position in the input sequence to attend to all positions in the previous layer simultaneously, a stark contrast to the sequential attention in LSTMs. Transformers also employed multi-head attention, allowing the model to capture various aspects of the input simultaneously. This parallel processing and the ability to directly model relationships between all words in a sequence led to significant improvements in efficiency and effectiveness.

\section{Bias in NLP}\label{biaisnlp}

In the context of the evolution of Natural Language Processing (NLP) technologies, understanding the role of biases in these models is crucial. Biases in machine learning, particularly in NLP, are integral for models to make accurate predictions. However, the nature and source of these biases are what determine their impact on the model's performance and fairness.

Biases that contribute positively to a model's predictive power are considered beneficial. For example, in a recruitment scenario for an IT position, if a model uses a candidate's previous experience in a similar role as a significant factor in its decision-making process, this represents a logical and beneficial bias. Such biases are based on causal relationships that are relevant to the task at hand. Having relevant experience is likely to make a candidate more suitable for the position, and thus, the model's reliance on this factor is justified and aids in accurate prediction. Conversely, detrimental biases are those that a model should not use for predictions as they do not have a causal impact on the outcome but are merely correlated with it. In the same recruitment scenario, if the model starts favoring candidates based on gender, assuming men are more suitable for IT roles due to historical data trends, it introduces a harmful bias. This type of bias is not based on a logical or ethical foundation but arises from correlations in the training data that reflect societal biases or historical inequalities, leading to discriminatory outcomes.

As NLP models have evolved from simpler frameworks like Bag of Words to more complex ones like Transformers, their ability to incorporate and utilize a larger number of biases has increased. This increased capacity for bias is a double-edged sword \citep{devlin2018bert,karita2019comparative}. On one hand, sophisticated models like Transformers can capture a wide array of nuances in language, leading to highly accurate predictions. On the other hand, their complexity means they are more susceptible to absorbing and perpetuating detrimental biases present in the training data, hence the trade-off between fairness and performance often discussed in the fairness literature \citep{kleinberg2016inherent, holstein2019improving}.

To illustrate this point, we compare a naive model with a Transformer model on a classification task: predicting occupation from LinkedIn biographies using the \textit{Bios} dataset (as detailed in section \ref{biosdataset}). For this experiment, we utilize two models: a simple one-versus-all logistic regression with L2 regularization that predicts occupation based on a Bag of Words (BoW) representation, and a Transformer model, specifically the DistilBERT \citep{sanh2019distilbert} model (described in section \ref{fairnessmetrics:subset:model}). The naive BoW model achieves an accuracy of 0.791, while the Transformer model achieves a higher accuracy of 0.86. However, when we consider a well-known fairness metric, the TPR (True Positive Rate) Gender Gap (explained in section \ref{intro:fairnessmetrics} and formalized in section \ref{fairnessmetrics:subset:metric}), we find that the naive BoW model performs better in terms of fairness. For instance, the TPR gender gap for the surgeon class is -0.178 for the naive model, compared to -0.223 for the Transformer model (the closer the metric is to 0, the fairer the model). 

The challenge in modern NLP is to harness the predictive power of these advanced models while constraining and mitigating the detrimental biases. This requires a multifaceted approach, including careful curation of training data, continuous monitoring for biases, and the implementation of fairness-aware machine learning techniques \citep{mehrabi2021survey}. The objective is to maintain the high level of performance offered by models like Transformers, but in a manner that is ethically responsible and socially fair, ensuring that the biases they leverage enhance the decision-making process rather than compromise it.

\section{NLP datasets and fairness}\label{nlpdataset}

Data serves as the foundation of NLP, influencing the efficacy and reliability of NLP systems across a spectrum of applications. Good quality data, characterized by its accuracy, diversity, and representativeness, is essential for training NLP algorithms. Whether for language translation, sentiment analysis, or more complex tasks, the data's quality determines the success and limitations of these systems.

As NLP technologies advance, the challenges associated with data management become increasingly complex. The trend towards using larger datasets for training sophisticated models, such as large language models (LLMs), introduces significant challenges in ensuring data quality. The vastness of these datasets makes it difficult to thoroughly inspect and understand every aspect of the data. This lack of transparency can lead to the inadvertent introduction and perpetuation of biases within the system. Biases in data, whether cultural, linguistic, or contextual, can lead to skewed outputs and discriminatory practices in NLP applications. Therefore, the need for robust methodologies to assess and correct data biases is more crucial than ever. 

Gaining a deep understanding of training data is crucial for revealing and addressing potential biases in NLP models. This insight is particularly important in the context of large and complex datasets used in current NLP research. By closely examining the nature and composition of these datasets, researchers can identify latent biases that might affect the model's performance and outputs. This process of critical evaluation is not just about improving model accuracy but also about ensuring fairness and impartiality in automated language processing. The following section presents the datasets used in this thesis, highlighting their significance and role in shaping the outcomes of the NLP models under study.

\subsection{Bios dataset}\label{biosdataset}

\begin{figure}[h]
    \centering
\includegraphics[width=1.0\linewidth]{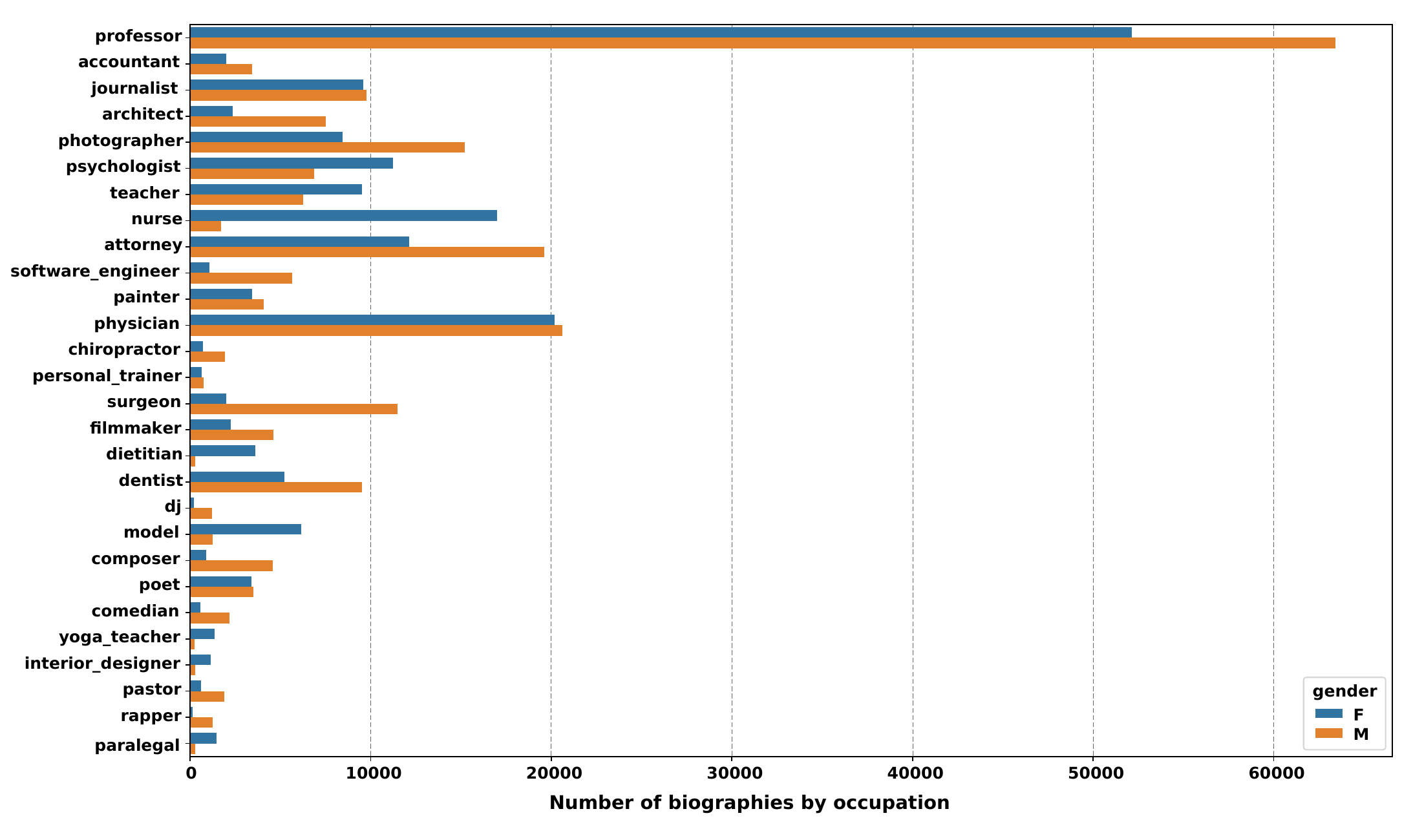}
    \caption{Number of biographies for each occupation by gender on the total \textit{Bios} dataset \cite{de2019bias}.}
    \label{preliminaries:fig:distribution}
\end{figure}

The \textit{Bios} dataset, detailed in \citep{de2019bias}, represents a significant effort in compiling a comprehensive resource for fairness research in NLP. This dataset includes around 400,000 biographies sourced from Common Crawl, a vast repository of web data. It focuses on biographies written in English, identifiable by a name-like pattern followed by an occupational title based on the Bureau of Labor Statistics Standard Occupation Classification system. Concentrating on the twenty-eight most frequently mentioned occupations, the dataset was created by processing WET files from sixteen different crawls conducted between 2014 and 2018, ensuring a diverse collection of biographical texts.

Each biography in the \textit{Bios} dataset is annotated with two key pieces of information: the gender of the individual (male or female) and their occupation (one of the 28 categories). This dataset is particularly notable for its heterogeneous representation of various occupations, highlighting significant gender imbalances in certain fields. While professions like professors and journalists show more balanced gender distributions, others such as nursing and software engineering demonstrate pronounced disparities (see Figure \ref{preliminaries:fig:distribution}).

The importance of the \textit{Bios} dataset in fairness research, especially in the domain of NLP, is profound. It provides an exceptional opportunity to examine how regularization strategies in machine learning can address gender biases, particularly in occupation prediction. This is crucial for developing fair and unbiased AI systems in scenarios like job candidate evaluation. Renowned in the fair learning community for its size and explicit labeling of gender, the \textit{Bios} dataset is ideal for multi-class classification studies in NLP. The varied distribution of biographies across different occupations and the variable gender proportions within these occupations offer a fertile ground for analyzing and understanding gender biases in automated systems. For this reason, this dataset will be central in illustrating the various methods developed during this thesis.

Moreover, the task associated with the \textit{Bios} dataset mirrors real-world applications, such as employment recommendation systems. In these systems, reducing bias is not just theoretical but a practical necessity for equitable treatment of candidates. Insights from studying and mitigating biases in models trained on the \textit{Bios} dataset can directly impact real-life scenarios, where gender biases in job recommendations can have significant societal consequences. Therefore, the \textit{Bios} dataset serves as a crucial tool in advancing the understanding and mitigation of gender biases in NLP models. 

\begin{figure}
    \centering
\includegraphics[width=0.7\linewidth]{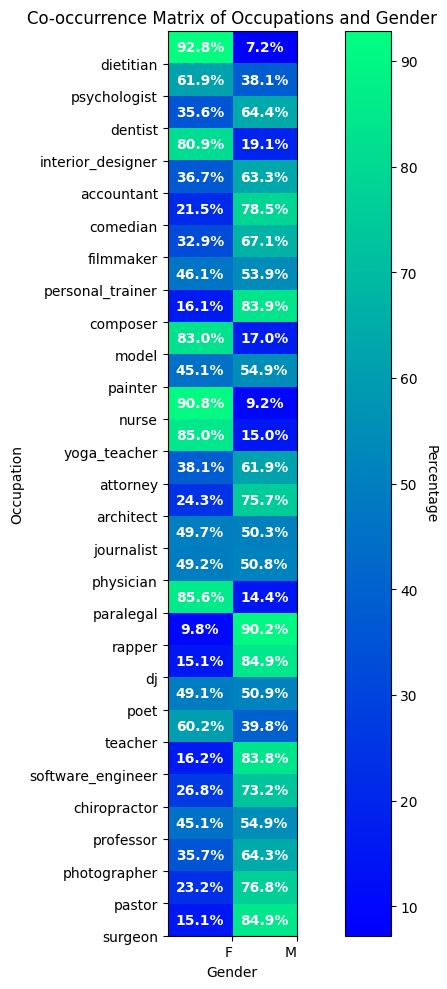}
    \caption{Co-occurrence of occupation and gender for \textit{Bios} dataset \cite{de2019bias}. Solely predicting the gender from the occupation yields a minimum of 62\% accuracy, which is the minimum baseline for an accurate but unbiased model. Any classifier with lower accuracy on gender prediction must have accuracy on occupation prediction lower than $100\%$.}
    \label{preliminaries:fig:cooccur}
\end{figure}

In addition, The \textit{Bios} dataset reveals significant correlations between gender and occupation, as illustrated in Figure \ref{preliminaries:fig:cooccur}. This is not a mere statistical anomaly but a reflection of real-world societal trends, where certain professions are more commonly associated with a specific gender. For example, the dataset shows a higher representation of women in nursing and men in surgical roles. These patterns are indicative of broader cultural and societal influences that shape career choices and opportunities along gender lines.

This nuanced understanding of gender-occupation correlations is pivotal in our quest to reduce discriminatory biases in AI. When developing models to predict gender based on occupation data, as found in the \textit{Bios} dataset, we achieve a baseline accuracy of 62\%. This statistic is significant; it demonstrates that completely abstracting gender from the equation would inevitably lead to a reduction in the accuracy of our models. Therefore, the challenge lies in creating AI systems that are not only fair and unbiased but also sensitive to the inherent correlations present in the data.

In essence, the \textit{Bios} dataset underscores the complexity of mitigating biases in AI. It demands a balanced approach where we neither ignore the realities of gender-occupation correlations nor allow them to perpetuate biases. By carefully navigating this balance, we can develop more equitable and accurate AI systems that reflect the diversity and realities of professional roles across genders. This approach is essential in advancing the field of NLP and AI, moving towards systems that are truly reflective of and responsive to the nuances of our society.

It should also be noted that while our methods in this manuscript aim to mirror the realities of contemporary society, they are versatile enough to be applicable to different distributions. For instance, in a hypothetical parallel world where gender discrimination are reversed, our methods could be readily adapted to accommodate constraints in the opposite direction without any significant issues.

\paragraph{Cleaning dataset}
To make the best use of the \textit{Bios} dataset throughout this thesis, we need to clean up the dataset. To do this, we take all the biographies and apply these modifications: \begin{itemize}
    \item we remove email addresses and URLs, 
    \item we remove the dots after pronouns and acronyms, 
    \item we remove double "?", "!" and ".", 
    \item we truncate all the biographies at 512 tokens by checking that they are cut at the end of a sentence and not in the middle.
\end{itemize}

\subsection{Others datasets used}

\begin{figure}[h]
    \centering
\includegraphics[width=1\linewidth]{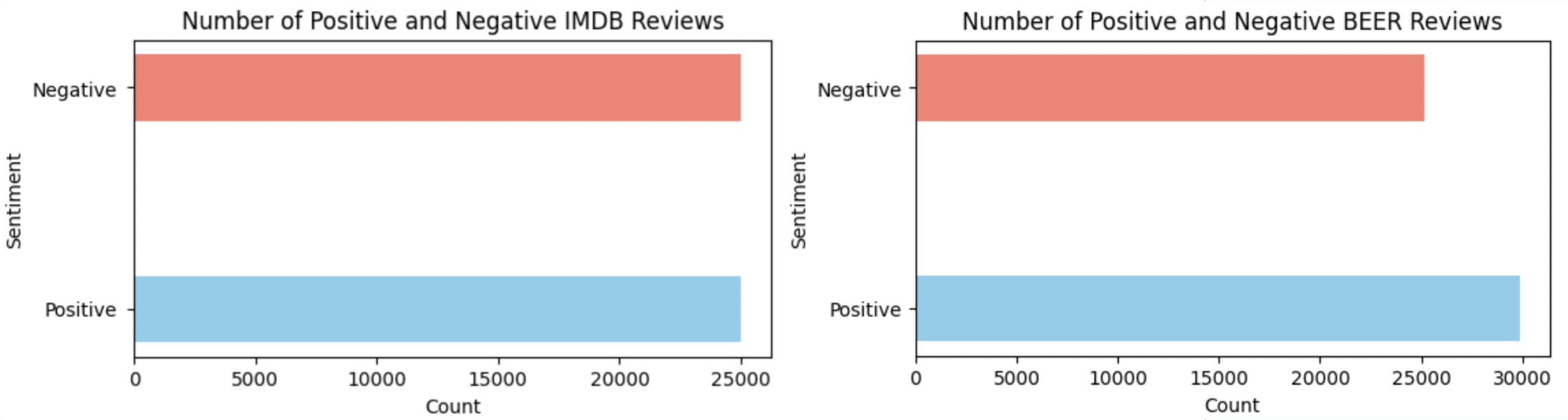}
    \caption{Number of reviews for each label (positive or negative) for \textit{IMDB} dataset (on the left) and for \textit{BEER} dataset (on the right).}
    \label{preliminaries:fig:barplot_imdb_beer}
\end{figure}

That although the \textit{Bios} dataset constitutes the primary data corpus for the experimental inquiries of this thesis, there has been an incorporation of two additional datasets, namely \textit{IMDB} \citep{maas-EtAl:2011:ACL-HLT2011} and \textit{BEER reviews} \citep{beer-reviews} datasets, specifically in the explainability segment. These datasets are not employed in the fairness section of the research, as they do not contain the sensitive variables needed for fairness analysis.

The \textit{IMDB} dataset, featuring 50,000 movie reviews from the Internet Movie Database, is structured to facilitate binary sentiment analysis: each review is classified as either positive or negative, with an even distribution of 25,000 reviews in each category. This balance in class distribution is crucial for unbiased model training in sentiment analysis.

On the other hand, the multi-aspect \textit{BEER reviews} dataset offers a more complex structure. It contains beer reviews that include ratings (from 0 to 5) on five different aspects: Appearance, Aroma, Palate, Taste, and Overall. The model's objective is to predict if the Overall score is greater than 3, signifying a positive review. However, the model does not have access to labels for the other aspects, adding a layer of complexity to the training process. Additionally, the dataset includes 994 reviews with specific annotations indicating the position of these aspects in the text, providing a rich source for explicability analysis in identifying how different aspects contribute to the overall sentiment of a review (see Figure \ref{preliminaries:fig:barplot_imdb_beer}).

While the \textit{Bios} dataset serves as the backbone for the primary experiments in this thesis, the \textit{IMDB} and \textit{BEER reviews} datasets enrich the explicability part, offering diverse perspectives and contexts.

\begin{figure}[h]
    \centering
\includegraphics[width=1\linewidth]{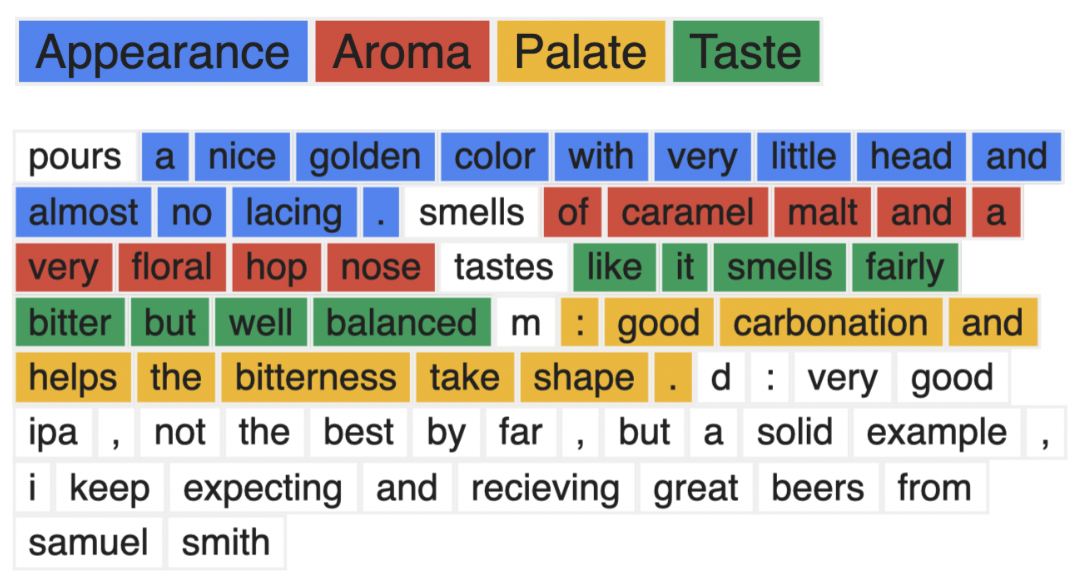}
    \caption{Example of a positive review on \textit{BEER} review with humans annotations.}
    \label{preliminaries:fig:barplot_imdb_beer}
\end{figure}

%% file: State_of_the_art.tex
\chapter{Fairness and Explainability, State of the Art}\label{SOTA}

In the rapidly advancing field of NLP, the balance between technical innovation and ethical responsibility is crucial. This chapter, building on the introductory discussion of fairness's ethical and legal aspects, delves into the latest methodologies in NLP aimed at addressing these concerns.

Here, we explore a range of strategies for mitigating bias, from initial data preparation to algorithm adjustments and final evaluations, presenting a comprehensive view of efforts to enhance fairness in NLP. We then transition to the concept of explainability in NLP, highlighting how making complex language models transparent can build trust and accountability. This chapter aims to present a clear and concise overview of the state-of-the-art in NLP.

\section{Bias mitigation techniques}

Bias in NLP systems has received a significant attention in recent years, with researchers and practitioners exploring methods for mitigating bias in NLP models on various task: machine translation \citep{vanmassenhove2019getting, stanovsky2019evaluating, savoldi2021gender, wisniewski2021biais}, hate speech detection \citep{park2018reducing, dixon2018measuring}, sentiment analysis \citep{kiritchenko2018examining, zellers2019hellaswag}, or text generation \citep{fleisig2023fairprism, wang2023toward}. Here, we focus on classification task in general.

The mitigation of bias in NLP for classification can be approached through various strategies, including \textit{pre-processing}, \textit{in-processing}, and \textit{post-processing} methods. More recently, we've seen the emergence of a fourth category, specific to Transformers architectures and close to the \textit{post-processing} approach, which can be called \textit{latent-space intervention} methods. Each of these strategies offers different advantages and limitations in addressing bias.

\subsection{Pre-processing methods}
The first approach to mitigate bias is to apply \emph{pre-processing} techniques to the data used to train the model. This method focuses on modifying the training data before the model learns from it, leading to more generalized and fairer models. Researchers like \cite{lu2020gender} highlight the advantages of this approach, emphasizing its capacity to tackle bias at its root. By altering the training data, the models developed are less likely to inherit biases, thereby promoting fairness and generalization.

We can reduce the bias directly in the text of the training dataset. For example, in the case of gender bias like the study in this paper, the most classic technique is to remove explicit gender indicators (like \textit{he}, \textit{she}, \textit{her}, etc.)\citep{de2019bias}. This technique is the one we will use to compare our proposed strategy to another one commonly used in industry. This technique is indeed simple to implement and makes it possible to reduce the bias, but in a partial and not very localized manner. However, modern Transformer models have been shown to be able to learn implicit gender indicators~\citep{devlin2018bert,karita2019comparative}.This essentially limits the impact that such a naive approach can have on this kind of models. As a matter of fact, a RoBERTa model trained without these gender indicators can still predict with upwards of 90\% accuracy the gender of a person writing a short LinkedIn bio (see Appendix~\ref{apx:genderpredrob}). 
Other classical techniques can be used, like identifying biased data in word embeddings, which represent words in a vector space. \cite{bolukbasi2016man, caliskan2017semantics, garg2018word} demonstrated that these embeddings reflect societal biases. There are also methods to show how these embeddings can be unbiased by aligning them with a set of neutral reference vectors \citep{zhao2018gender, zhao2018learning, bolukbasi2016man}.  However, \cite{gonen2019lipstick} show that although the de-biasing embedding methods can reduce the visibility of gender bias, they do not completely eliminate it. In fact, the bias would only be masked, and it would still be possible to find it again with embedding transformations. \citep{zhao2019gender} propose an alternative: de-biasing contextual word embedding instead. And they apply their de-biasing method to ELMo \citep{lee2018higher} embedding. More recently, \citep{rus2022closing} have done so on BERT \citep{devlin2018bert} embedding.

According to \cite{Shah2019Predictive} and \cite{Sun2019Mitigating}, \textit{pre-processing} methods are limited in detecting and neutralizing all types of bias, especially those deeply ingrained in language or context. The challenge lies in the complexity of biases, which are often subtle and contextually embedded in data.  
Gender biases are linked to gender detection (and therefore to its use by the model). Obvious forms of detection are linked to keywords or expressions used more by one gender than the other. It's these patterns that are detected and corrected by \textit{pre-processing} techniques. 
But language is more complex than that, and sentence construction, syntax and semantics can vary subtly from one genre to another. The most complete models are capable of understanding these subtleties, and we need to build more complex debiasing methods to understand them too.

\subsection{In-processing methods} A second approach to mitigate biases in AI is to use fairness-aware algorithms, which are specifically designed to take into account the potential for bias and to learn from the data in a way that reduces the risk of making biased decisions.  These are the \emph{in-processing} methods, which generally do not depend on the type of data input.  The advantages of in-processing methods are significant: They actively work to reduce bias during the training phase and are dynamic, adjusting to different forms of bias as the model learns \citep{lu2020gender}.

To achieve this, we explore diverse strategies such as adversarial learning, regularization, and fairness metric constraints, each offering a unique approach to mitigating bias.

We can use adversarial Learning by adjusting the discriminator. Adversarial learning involves training a model to make predictions while also training a second model to identify and correct any biases in the first model's predictions. By incorporating this technique into the training process, \cite{zhang2018mitigating, madras2018learning} demonstrate that it is possible to reduce the amount of bias present in machine learning models. Additionally, \cite{petrovic2022fair} introduces a similar approach that uses a gradient-based method to adaptively balance accuracy and fairness during training, showcasing its effectiveness on various datasets.

Also, a variety of regularization techniques are employed to constrain predictions in the realm of fairness in machine learning. One such method involves using gradient descent to optimize a conventional fairness metric, as discussed in \citep{chuang2021fair}. Beyond this classic approach, there exists a spectrum of regularization strategies aimed at reducing bias. For instance, one can minimize the mutual information \citep{shannon1949mathematical} between the model's output and the sensitive variable, a technique primarily applied in logistic regression classifiers \citep{kamishima2012fairness}. In a similar vein, the minimization of the Hilbert-Schmidt Independence Criterion (HSIC) \citep{gretton2005measuring} between prediction accuracy and sensitive attributes has been proposed by \cite{perez2017fair, li2019kernel}. HSIC is a statistical measure of dependence between two sets of variables; minimizing HSIC ensures that the model's predictions are independent of sensitive attributes. Furthermore, the application of optimal transport theory to minimize the Wasserstein-2 distance \citep{kantorovitch1939methodes} presents an innovative regularization approach for enhancing fairness in predictive models \citep{risser2022tackling}. The Wasserstein-2 distance measures the distance between two probability distributions; minimizing this distance helps in aligning the distribution of predictions across different demographic groups.
Additionally, \cite{do2022fair} proposes a regularization method that includes group-specific regularization terms, which penalize prediction errors differently across various demographic groups to reduce disparities and improve fairness in the model's performance.

Finally, \cite{zafar2017fairness, zafar2017fairness2} use fairness metrics constraints and solve the training problem subject to those constraints. Furthermore, \cite{zhao2022adaptive} presents a flexible framework that incorporates multiple fairness constraints simultaneously, allowing practitioners to tailor the fairness criteria to specific application needs while ensuring model compliance with these constraints.

All these \textit{in-processing} methods apply in the case of binary classification. There is indeed an \textit{in-processing} paper that proposes a method for multiclass classification for a computer vision task \citep{zhao2017men}, but this paper focuses on the regularization of the mean bias amplification and therefore does not deal with the classic fairness metrics.

However, the main limitations of these methods are that these algorithms can be complex to implement and may sometimes introduce new biases unintentionally. The complexity arises from the sophisticated nature of the algorithms and the need for careful calibration to ensure they do not inadvertently create new forms of bias \citep{Lohia2018Bias}.

\subsection{Post-processing methods} The last approach is to use \emph{post-processing} de-biasing methods. These methods are model-agnostic and therefore not specific to NLP since they modify the results of previously trained classifiers in order to achieve fairer results.  These methods are applied after a model has been trained, allowing them to adjust outputs to enhance fairness. This adaptability makes them suitable for use on any pre-trained model, providing a versatile tool in the pursuit of reducing bias in AI systems \citep{Lohia2018Bias}.

In this branch, different types of techniques emerged. \cite{menon2018cost} advanced a plugin methodology for the thresholding of predictions, wherein the determination of appropriate thresholds is facilitated through the estimation of class probabilities via logistic regression. Concurrently, \cite{fish2016confidence} developed a confidence-based paradigm, endorsing the assignment of a positive label to each prediction that surpasses a specified threshold, a technique subsequently integrated into AdaBoost \citep{fish2015fair}. This approach diverges from the conventional practice of employing a uniform threshold across all instances, advocating instead for the utilization of group-dependent thresholds, as evidenced in the literature \citep{chzhen2019leveraging, iosifidis2019fae} and presented at the survey \citep{hort2022bia}
In a related vein, \cite{kamiran2012decision, kamiran2018exploiting} introduced the concept of a 'reject option', a novel mechanism that alters the predictions for individuals situated proximate to the decision boundary. This approach is particularly noteworthy for its differential treatment of individuals based on group affiliation, whereby those from unprivileged groups are accorded a positive outcome, in contrast to their privileged counterparts who receive an unfavorable outcome. \cite{chiappa2019path} further enrich this landscape by examining causal models through a counterfactual lens. This approach seeks to align model predictions with a hypothetical scenario where individuals belong to different demographic groups, thus ensuring that the influence of protected attributes is appropriately corrected.

The exploration of \textit{post-processing} methods extends to various contexts, including binary classification \citep{hardt2016equality, sikdar2022getfair} and multiclass classification \citep{denis2021fairness}. While these methods are adept at making outputs fairer, it's important to remember that they tend to address the symptoms rather than the root causes of bias. This means that while they can adjust outputs to be more equitable, they may not eliminate the underlying biases in the training data or the model itself \citep{biswas2020machine}.

\subsection{Latent-space intervention}\label{sota:fairness_LS}

One way of improving the fairness of a model is to employ an approach widespread in information theory, which consists of eliminating information linked to sensitive variables to prevent the model from using them \citep{kilbertus2017avoiding}. We'll call this "\textit{sensitive variable information erasure}". So, the last approach consists in intervening in the model's latent space to remove sensitive information directly on this space.

To ensure that the model does not utilize sensitive variable information, one can measure the presence of sensitive information in the latent space using metrics such as mutual information \citep{shannon1949mathematical} or $\nu$-information \citep{Xu2020A}. Mutual information measures the amount of information shared between two random variables, indicating how much knowing one variable reduces uncertainty about the other \citep{shannon1949mathematical}. In contrast, $\nu$-information considers the computational constraints of the observer. It generalizes Shannon’s mutual information and other predictive notions, allowing for reliable estimation even in high-dimensional contexts. Unlike mutual information, $\nu$-information can be "created" through data processing \citep{Xu2020A}.

In~\citep{liang2020towards}, the authors introduce a technique based on decomposing the latent space using a PCA and subtracting from the inputs the projection into a bias subspace that carries the most information about the target concept. Another vein focuses on modifying the latent space with considerations coming from the field of information theory~\citep{Xu2020A}. Indeed, the $\nu$-information has been used as a means to carefully remove from the model's latent space only the information related to the target concept. In particular, popular approaches include the removal of information through linear adversarials on the last layer of the model~\citep{ravfogel2020null,RLACE}, through Assignment-Maximization adversarials~\citep{shao2023erasure}, and through linear models that act on the whole model~\citep{belrose2023leace}. However,  \citet{ravfogel2023loglinear} recently discovered a limitation of theses methods. The limitation presented in \citep{ravfogel2023loglinear} comes from the practical use of the $\nu$-information. This measure updates the mutual information, considering the expressivity of functions measuring information in the latent space. Their paper explains that by protecting information using a linear model, other, more expressive models can still decode this protected information. 

Clother than this, \cite{frye2020asymmetric,dorleon2022,galhotra2022} use methodology from the field of explainable AI (XAI) to perform feature selection to reduce the effect of algorithmic bias in a human-understandable yet principled manner.

\subsection{From Fairness to Explainability}
Ensuring fairness is not solely about neutralizing biases in the data or the model; it is also about understanding and communicating how the algorithm reaches its decisions. This is where the role of explainability comes into play.  

Explainability, or the ability to elucidate the decision-making process of a machine learning model, stands at the confluence of trustworthiness and transparency. It is the bridge that links the abstract world of algorithmic calculations to the tangible concerns of human users and stakeholders. Without explainability, it becomes challenging to scrutinize or contest a model's decisions, let alone discern whether it is operating fairly or perpetuating harmful stereotypes.

Furthermore, as ML models, especially those in NLP, become more sophisticated and intricate, the opacity of their inner workings intensifies. The "black box" nature of these algorithms not only hampers our ability to identify and rectify biases but also hinders the acceptance and broader adoption of such technologies in critical sectors. To establish trust and ensure the ethical deployment of ML models, it is indispensable to have mechanisms that can elucidate their operations.

\section{Explainability}\label{sec:sota:xai}

In this section, I will explore the state of the art in explainability methods for NLP models for classification task. Classification models, often critical in decision-support systems, play a significant role in ensuring fairness and accountability in applications such as hiring, loan approval, and medical diagnosis. Explainability in this context means providing insights into how and why a model makes specific predictions. This involves uncovering the underlying factors and decision-making processes that lead to particular classifications. 


\subsection{Attribution methods}

The most straightforward approach analyzes how each part of the input influences the model's prediction to create heat maps. These methods are known as attribution methods. 

\subsubsection{Perturbation-based attribution}
The perturbation-based approach relies on perturbations involving the strategic modification of input data and the observation of resultant changes in the model's predictions. 

The best-known and most natural method in this category is occlusion \citep{zeiler2014visualizing}. It consists of masking/removing each word/token (or other type of feature) by examining the impact on the model's output. In the same paper, \cite{zeiler2014visualizing} employed a visualization technique known as Deconvolutional Network (DeconvNet) to project feature activations back to the input space, thus enabling the visualization of specific input features that activate certain network features. More recently, in the same vein, \cite{zhao2024reagent} developed ReAGent, a method inspired by occlusion but optimized for text generation, which also applies to classification tasks.

The Local Interpretable Model-agnostic Explanations (LIME) technique \citep{ribeiro2016model}, for instance, employs this approach to approximate the model’s behavior locally with an interpretable model, thereby shedding light on the model's operation in specific instances.

Perturbation-based attribution methods can rely on sensitivity analysis to offer a robust approach to understanding model behavior. One notable technique \citep{fel2021sobol} involves Sobol indices \citep{sobol1993sensitivity}. Sobol indices measure the contribution of each input variable to the output variance by decomposing the total variance into parts attributable to individual variables and their interactions. They provide first-order indices that represent the effect of individual variables and total-order indices that capture the effect of variables along with their interactions. Another effective approach \citep{novello2022making} utilizes the Hilbert-Schmidt Independence Criterion (HSIC) \citep{gretton2005measuring}. HSIC quantifies the statistical dependence between input features and the model output, providing a non-linear measure of dependence that captures complex relationships. 


\subsubsection{Gradient-based attribution}
Gradient-based attribution methods provide insight into how models make predictions by assessing the contribution of each input feature. 

Saliency maps \citep{saliency}, one of the earliest approaches, compute the gradient of the output with respect to the input features, highlighting the regions that most influence the prediction. The input × gradient \citep{saliency} method refines this by multiplying the input feature values with their corresponding gradients, offering a more nuanced attribution by considering the magnitude of the inputs. 

Integrated Gradients (IG) \citep{IntegratedGradients}, a more advanced technique, computes the average gradient of the model output as the inputs vary along a path from a baseline to the actual input. Variations of IG have been developed to enhance its applicability to language models, such as Discretized Integrated Gradients \citep{sanyal2021discretized}, which adapts IG for discrete token spaces, and Sequential Integrated Gradients \citep{enguehard2023sequential}, which sequentially applies IG to capture interactions between tokens in a structured manner. 

DeepLIFT (Deep Learning Important Features) \citep{shrikumar2017learning} further extends these concepts by comparing the activation differences between the actual input and a baseline, attributing the differences in the output to each input feature.


\subsubsection{Internal-based attribution}
Internal-based attribution methods, which depend on the architecture of models, play a significant role in the explainability of NLP models, particularly with the advent of Transformer architectures. One prominent approach within this category is Attention Weight Attribution \citep{bahdanau2014neural}. This method interprets the model's predictions by analyzing the attention weights assigned to input tokens, under the assumption that higher weights indicate greater importance. However, this method has been met with controversy. Studies by \cite{jain2019attention,pruthi2019learning,serrano2019attention} have challenged the reliability of attention weights as explanations, arguing that attention mechanisms may not always align with human intuition and can be manipulated without significantly altering model outputs. To address these concerns, alternative methods such as Attention Flow \citep{abnar2020quantifying} have been proposed.  This approach seeks to track the flow of attention throughout the layers of the Transformer, providing a more comprehensive and potentially more faithful representation of how information is processed within the model. 


\subsubsection{SHAP methods}
\cite{lundberg2017unified} introduced SHapley Additive exPlanations (SHAP), a novel approach that integrates the estimation of Shapley values \citep{shapley1953value} with widely recognized explainability techniques such as LIME \citep{LIME} with KernelSHAP, or DeepLIFT \citep{shrikumar2017learning} with DeepSHAP (also introduced by \cite{lundberg2017unified}). Originating from game theory, Shapley Values were initially devised as a method for equitably allocating rewards among participants who collectively contribute to a specific result. SHAP has been adapted and expanded in various ways, detailled in the survey of \cite{mosca2022shap}. \cite{ghorbani2020neuron} tailored SHAP specifically for elucidating predictions from neural networks. Meanwhile, \cite{messalas2019model} and \cite{chen2018shapley} developed alternative methods to approximate Shapley values, enhancing their efficiency and accuracy. Fundamentally, SHAP serves as a framework for local feature-attribution explanations, assigning scores to input features based on their individual contributions at the instance level. However, other methods based on SHAP diverge in their objectives, offering different types of explanations. These range from global explanations \citep{covert2020understanding}, to counterfactual explanations \citep{singal2019shapley} and even concept-based explanations \citep{yeh2020completeness}.

\subsubsection{Weaknesses}
However, these approaches have weaknesses. 
\cite{wang2020gradient} highlighted a significant vulnerability of gradient-based methods to adversarial manipulation, where minor, often imperceptible perturbations can dramatically alter the interpretations provided by these methods. Also, \cite{adebayo2018sanity} pointed out the issue of partial input recovery in some interpretability methods. Their research suggested that certain methods, especially those reliant on heat maps (like attribution methods), might not fully capture all influential input features, leading to incomplete or skewed interpretations. Lastly, the question of stability in interpretability methods has been raised by \cite{ghorbani2019interpretation}. They demonstrated that minor variations in input, which do not significantly change the model's output, could lead to drastically different interpretations. This lack of stability raises concerns about the consistency and trustworthiness of the interpretations derived from these methods.

\subsection{Concept-based explanations}

Concept-based explainability is a growing, focused on generating human-understandable concepts which explaining the decisions made by machine learning models. These concepts can be: \begin{itemize}
    \item (\textit{supervised}) predefined and selected by humans, and methods are sought to determine how important they are to the model;
    \item or (\textit{unsupervised}) discovered by unsupervised methods directly in the model, and made understandable to humans.
\end{itemize}

\subsubsection{Supervised concept-based XAI}

Testing with Concept Activation Vectors (TCAV) \citep{kim2018interpretability} framework is the pioneer method on supervised concept-based XAI. TCAV quantifies the influence of predefined concepts on model predictions by analyzing the directional derivatives of the model's output with respect to the internal activations corresponding to these concepts. 

After TCAV, other significant contributions to supervised concept-based XAI include Concept Bottleneck Models (CBMs) \citep{koh2020concept}. CBMs operate by training models in a manner where intermediate bottlenecks represent specific, human-interpretable concepts, ensuring that the final predictions are directly dependent on these interpretable representations. \cite{yuksekgonul2022post} extend this approach -- with Post-hoc Concept Bottleneck Models (PCBMs) -- by allowing the introduction of concept bottlenecks into pre-trained models, making it feasible to retrofit existing models with interpretable bottlenecks without requiring retraining from scratch. This post-hoc modification significantly enhances the versatility and applicability of concept bottlenecks in real-world scenarios. PCBMs also leverage multimodal models to transfer concepts from other datasets or natural language descriptions, enhancing their flexibility and accessibility.bottlenecks in real-world scenarios.

However, all these supervised concept-based approach relies on Human inputs, as it requires the user to manually specify the concepts to be tested. This can be time-consuming and may not always produce the most comprehensive explanations \citep{ghorbani2019towards}.

\subsubsection{Unsupervised concept-based XAI}

Another approach, ACE~\citep{ghorbani2019towards}, aims to automate the concept extraction process. ACE uses a clustering algorithm to identify interpretable concepts in the model's activations, without the need for Human input. While this approach has the potential to greatly reduce the time and effort required for concepts extraction, the authors criticize their own reliance on pre-defined clustering algorithms, which may not always produce the most relevant or useful concepts.

An alternative uses matrix factorization techniques, such as non-negative matrix factorization (NMF)~\citep{lee1999learning}, to identify interpretable factors in the data~\citep{zhang2021invertible,fel:etal:2023}. \cite{fel:etal:2023}  
created a method to discover automatically pertinent concepts inside the network's activation space with NMF and to find the parts of the input space that most align with each concept with Sobol Indices \citep{sobol1993sensitivity}. They successfully tested the meaningfulness and the capacity of these explanations to help Humans understand the model's behavior through psychological experiments. Then, \cite{fel2023holistic} developed a framework for generating global and local explanations to unify automatic concept extraction and concept importance estimation techniques.

However, these approachs have only been applied to convolutional neural networks for image classification tasks so far. The only similar approach to be found in the literature that is totally model agnostic and also applied to textual data is \textit{Concept}SHAP \citep{yeh2020completeness}, a variation of SHAP \citep{lundberg2017unified} that discovers concepts.

\subsection{Explaining through rationalization}

For NLP applications, the techniques closest to those based on concepts are those in rationalization. These involve creating a self-interpretable neural architecture capable of simultaneously generating a prediction on classification tasks and its concept-based explanation. 

Researchers in the field of rationalization have proposed specific architectures to extract excerpts from whole inputs and predict a model's output based on these \emph{rationales}~\citep{lei2016rationalizing,jain2020learning,chang2020invariant,yu2019rethinking,bastings2019interpretable,paranjape2020information}. These rationales can be seen as explanations that are sufficiently high-level to be easily understood by humans. 

Finding rationales in text refers to the process of identifying expressions that provide the key reasons or justifications that are provided for a particular claim or decision about that text.  
\citet{lei2016rationalizing} defined rationales as "a minimal set of text spans that are sufficient to support a given claim or decision". They should satisfy two desiderata: they should be interpretable, and they should reach nearly the same prediction as the original output. To do so, they use a generator network that finds interesting excerpts and an encoder network that generates predictions based on them.  However, their scheme requires the use of reinforcement learning~\citep{reinforce} for the optimization procedure. \citet{bastings2019interpretable} proposed to include a reparametrization trick to allow for better gradient estimations without the need for reinforcement learning techniques, and a sparsity constraint to encourage the retrieval of minimal excerpts.

\citet{yu2019rethinking} and \citet{paranjape2020information} studied the problem of producing adequate rationales from a game-theoretic point of view.  However, these models can be quite complex to train, as they either require a reparametrization trick or a reinforcement learning procedure. \citet{jain2020learning} proposed to solve this problem by introducing a support model capable of producing continuous importance scores for instances of the input text, that the rationale extractor can  use to decide whether an excerpt will make  a good rationale or not.

All these rationales will serve as an explanation for single instances, but won't explain how models predict whole classes. \citet{chang2019game} introduced a rationalization technique that allows for the retrieval of rationales for factual and counterfactual scenarios using three players. 

\citet{bouchacourt2019educe} learned concepts without supervision from excerpts using a bidirectional LSTM during the training phase of the model, and the predictions are only based on the presence or absence of the individual concepts in the input sentences.  Going further, \citet{antognini-faltings-2021-rationalization} introduced ConRAT, a technique that includes orthogonality, cosine similarity and knowledge distillation constraints, as well as a concept pruning procedure to improve on both the quality of the extracted concepts and the model's accuracy.

However, rationalization techniques require to train an entirely new model. Only one rationale can also be found per input text, when  there might intuitively be several predictions for a given prediction. Finally, these approaches are not model agnostic and use architectures that have mostly been left behind since the introduction of the Transformer architectures, due to their inferior predictive capabilities.

\subsection{Other methods}
In this section, we address the final group of XAI methods that have not yet been discussed in this thesis, aiming to achieve the most exhaustive state-of-the-art overview possible.

\subsubsection{Example-based explainability}
Among these, example-based explainability stands out, utilizing sample-based techniques for generating explanations. This means the explanation’s format is a data point (an example) \citep{poche2023natural}.
The first type, factuals \citep{aamodt1994case}, are similar examples close to the query, helping to understand or justify the model's predictions. For broader insights, counterfactuals and semi-factuals are used. Counterfactuals \citep{wachter2017counterfactual} are the nearest samples with different predictions, while semi-factuals \citep{aryal2023even} are the farthest with similar predictions, both offering a view of the decision boundary. However, to understand how training data alterations might change the model's decisions, influential instances \citep{koh2017understanding} are examined. These are training samples that significantly impact the model's behavior. This approach falls under the umbrella of global explanations. Also falling into this category are Prototypes \citep{molnar2020interpretable}, comprising a selection of samples that represent either the entire dataset or specific classes within it. Typically chosen independently of the model, Prototypes offer an overview of the dataset. However, certain models are crafted with the aid of prototypes and thus can be inherently explained through their design.

\subsubsection{Logical explainability}
Within the framework of logical explainability, which pioneered XAI field, we find deductive and abductive reasoning (then came counterfactual explanations and contrastive explanations, which we'll detail in the next section).

Deductively valid explanations are essentially proofs characterized by logical soundness. This means that the phenomenon to be explained (the explanandum) is a logical consequence of the premises, or the explanans. This concept of the deductive-nomological method was introduced by Hempel in the 1950s. Abductive explanations can be regarded as a type of deductive explanation with more constraints. Abductive explanations are not only deductively valid but also involve a specific methodology for selecting the premises or explanans. This methodology can vary and is chosen by the creator of the explanations to match their constraints. 
This type of reasoning is mainly used by formal explicability methods \citep{ignatiev2019abduction, marques2023logic}. It provides a solid foundation of rigor, owing to its logical basis. However, in the case of modern models studied in NLP, such as Transformers, which have ever more parameters, these methods show their limits and will therefore not be relevant in the context of this thesis. Consequently, a compromise is necessary, opting for approaches that, while offering fewer guarantees, are more suited to the models in question.

\subsection{Counterfactual and Contrastive explanations}

Counterfactual and contrastive explanations offer nuanced perspectives in the field of XAI, providing new explanations possibilities to traditional methods. Traditional XAI reasoning typically focus on answering the question, "\textit{Why did the model make this prediction?}" In contrast, counterfactual explanations address the question, "\textit{What would need to change for the model to make a different prediction?}" while contrastive explanations ask, "\textit{Why did the model make this prediction instead of another?}".

In theory, all the types of methods discussed above can create counterfactual or contrastive explanations. 

Some methods, such as TCAV \citep{kim2018interpretability}, can be seen as providing counterfactual explanations. It is the case of TCAV because they identify concepts that significantly influence a model's predictions. By altering these concepts, one can generate hypothetical scenarios to see how the model's output changes. For example, in image classification, if the presence of a certain texture leads to a specific classification, TCAV can highlight this texture's importance and suggest how modifying it could change the prediction.
There are also methods, such as Polyjuice \citep{wu2021polyjuice}, which generate counterfactual examples directly -- thanks to GPT-type architectures in particular -- so that we can test the robustness of a model on multiple variations of the same dataset. 

For contrastive explanation side, very classical techniques such as attribution methods have been modified to create contrastive explanations \citep{jacovi:etal:2021, yin:neubig:2022}. In another branch, \cite{ross2020explaining} present Minimal Contrastive Editing (MICE), which generates minimal modifications to input texts to produce different model predictions, highlighting the exact elements that influence decisions.
It is even possible to use formal methods to make contrastive explanations as \citep{ignatiev2020contrastive}.

In conclusion, counterfactual and contrastive explanations offer a novel and complementary approach to traditional XAI approach. By addressing specific questions about model predictions, these methods provide deeper insights into how models make decisions and facilitate more robust and systematic error analysis.

%% file: Fairness_reg.tex
\chapter{Fairness Regularization}\label{Regularization}


This chapter initiates my doctoral research, focusing on developing a new method to mitigate biases in an area that has not yet been fully addressed: multi-class classification tasks in modern NLP algorithms, particularly Transformers. This work builds upon the initial understanding of fairness methodologies from earlier stages of my thesis, addressing a distinct gap in the current literature. In this chapter, I present my work on this critical issue, which culminated in the publication of the paper titled "\textit{How Optimal Transport Can Tackle Gender Biases in Multi-Class Neural Network Classifiers for Job Recommendations}" \citep{jourdan2023optimal} in the journal Algorithms. 

The paper introduces a novel optimal transport strategy, designed to mitigate undesirable algorithmic biases in multi-class neural network classification. This strategy is unique in its model-agnostic approach, making it applicable to any multi-class classification neural network model. With an eye towards the future certification of recommendation systems employing textual data, we applied this strategy to the \textit{Bios} dataset. This dataset facilitates the learning task of predicting the occupations of individuals based on their LinkedIn biographies, with a specific focus on gender representation. Our findings demonstrated that our approach effectively reduces undesired algorithmic biases in this context, achieving lower levels of bias compared to standard strategies.

\section{Introduction}\label{sec:intro}


Ensuring fairness is essential to ensure an ethical application of NLP algorithms in society. We have developed the methodology in this chapter with the aim of controlling  undesirable algorithmic biases on recommendation systems that exploit textual information from personal profiles on social networks, such as job or housing offers. 

Motivated by the future certification of AI systems (presented on chapter \ref{introduction}) based on black-box neural-networks against discrimination, this chapter expands on the work of \citep{risser2022tackling} to address algorithmic biases observed in NLP-based multi-class classification. The main methodological novelties of this chapter are: the extension of \citep{risser2022tackling} to multi-class classification, and a demonstration of how to apply it to NLP data in an application ranked as \emph{High Risk} by the \emph{AI act}. The bias mitigation model proposed in this chapter involves incorporating a regularization term, in addition to a standard loss, when optimizing the parameters of a neural network. This regularization term mitigates algorithmic bias by enforcing the similarity of prediction or error distributions for two groups distinguished by a predefined binary sensitive variable (e.g. males and females), measured using the 2-Wasserstein distance \citep{kantorovitch1939methodes} between the distributions. Note that  \citep{risser2022tackling} is the first paper that demonstrated how to calculate pseudo-gradients of this distance in a mini-batch context, enabling the use of this method to train deep neural-networks with reduced algorithmic bias.

To extend \citep{risser2022tackling} to multi-class classification with deep neural-networks, we need to address a key problem: estimating the Wasserstein-2 distance between multidimensional distributions (where the dimension equals the number of output classes) requires numerous neural-network predictions, leading to slow training.  In order to solve this problem, we will redefine the regularization term to apply it to predicted classes of interest, making the bias mitigation problem numerically feasible. 

Our secondary main contribution from an end-user perspective is to demonstrate how to mitigate algorithmic bias in a text classification problem using modern Transformer-based neural network architectures like RoBERTa small \citep{liu2019roberta}. It is important to note that our regularization strategy is model-agnostic and could be applied to other text classification models, like those based on LSTM architectures. We evaluate our method using the \emph{Bios} dataset \citep{de2019bias}, which includes over 400,000 LinkedIn biographies, each with an occupation and gender label. 

\section{Related work}


\subparagraph{Bias mitigation in NLP}

Bias in NLP systems has received a significant attention in recent years, with researchers and practitioners exploring various methods for mitigating bias in NLP models. In the chapter \ref{SOTA}, we reviewed some of the existing work on bias mitigation in NLP for \emph{pre-processing}, \emph{in-processing} and \emph{post-processing} techniques.

The approach used in this chapter to mitigate biases in AI is to use fairness-aware algorithms, which are specifically designed to take into account the potential for bias and to learn from the data in a way that reduces the risk of making biased decisions.  These are the \emph{in-processing} methods, which generally do not depend on the type of data input, either. To achieve this,  we can use adversarial Learning by adjusting the discriminator. Adversarial learning involves training a model to make predictions while also training a second model to identify and correct any biases in the first model's predictions. By incorporating this technique into the training process, \cite{zhang2018mitigating, madras2018learning} demonstrate that it is possible to reduce the amount of bias present in machine learning models. Another technique is to constrain the predictions with a regularization technique like \citep{kamishima2012fairness}, but this technique was only used on a logistic regression classifier. On the other hand, \cite{manisha2018fnnc} mitigate fairness specifically in neural networks.
Finally, \cite{zafar2017fairness, zafar2017fairness2} use fairness metrics constraints and solve the training problem subject to those constraints.
All these \textit{in-processing} methods apply in the case of binary classification. There is indeed an \textit{in-processing} paper that proposes a method for multiclass classification for a computer vision task \citep{zhao2017men}, but this paper focuses on the regularization of the mean bias amplification and therefore does not deal with the classic fairness metrics.

\subparagraph{Research implications} We want to emphasize that the \emph{pre-processing} and \emph{post-processing} methods are complementary to \emph{in-processing} methods. 
The 3 types of method need to be optimized and combined to further reduce bias. We have chosen here to concentrate on \emph{in-processing} methods.
In this context, our methodology tackles an issue which was still not addressed in the fair learning literature, as far as the authors know: we tackle algorithmic biases on multi-class neural-network classifiers and not on binary classifiers or on non neural-network classifiers. We believe that the potential of such a strategy is high for the future certification of commercial AI systems. The key methodological contribution of our work is  to show how to extend \citep{risser2022tackling} to multi-class classification for regularized  mini-batch training of neural-networks. As described in Section \ref{extended_W2reg},  extending this optimal transport regularization strategy to multi-class classification requires tackling an important technical lock related to the algorithmic cost of the procedure, which is the heart of this  methodological contribution. 
Note that this regularization strategy applies to any type of multi-class classification neural network model.  Thus, the method is particularly flexible and can be applied in various industrial classification problems. From a practical perspective, our secondary contribution is to showcase how using the proposed technique to the future certification of neural-network application ranked as \textit{High Risk} by the European Commission. We use in this chapter an NLP application, and thoroughly describe the procedure to correct strong biases. 

\section{Methodology}


The bias mitigation technique proposed in this chapter extends the regularization strategy of \cite{risser2022tackling} to multi-class classification. In this section, we first introduce our notation, then describe the regularization strategy of \cite{risser2022tackling} for binary classifiers, and then extend it to multi-class classifiers. This extension is the methodological contribution of our manuscript.

\subsection{General notations}\label{ssed:GeneralNotations}

\subparagraph{Input and output observations}
Let $(x_i, y_i)_{i = 1,\dots, n}$ be the training observations, where $x_i \in \mathbb{R}^p$ and $y_i \in \{0,1\}^K$ are the input and output observations, respectively. The value $p$ represents the inputs dimension or equivalently the number of input variables. It can for instance represent a number of pixels if $x_i$ is an image or a number of words in the dictionary if $x_i$ is a word embedding. The value $K$ represents the output dimensions. In a binary classification context, \emph{i.e.} if $K=1$, the fact that $y_i=0$ or $y_i=1$ specifies the class of the observation $i$. In a multi-class classification context, \emph{i.e.} if $K>1$, a common strategy consists in using one-hot vectors to encode the class $c$ of observation $i$: All values $y_i^k$, $k \in \{1,\ldots,K\}$ are equal to $0$, except the value $y_i^{c}$, which is equal to $1$. We will use this convention all along this manuscript.

\subparagraph{Prediction model}
A classifier $f_{\theta}$ with parameters $\theta$ is trained so that the predictions $\hat{y_i} \in \{0,1\}^K$ it indirectly makes based on the outputs $f_{\theta}(x_{i}) \in [0,1]^K$, are \emph{in average} 
as close as possible to the true output observations $y_i$ in the training set.
The link between  the model outputs $f_{\theta}(x_{i})$ and the prediction $\hat{y_i}$ depends on the classification context:
In binary classification,  $f_{\theta}(x_{i})$ is the predicted probability that $\hat{y_i}=1$, so it is common to use $\hat{y_i}=1_{f_{\theta}(x_{i})>0.5}$. Now, by using one-hot-encoded output vectors in multi-class classification, an output $f_{\theta}(x_{i}) = \left( f_{\theta}^1(x_{i}) , f_{\theta}^2(x_{i}), \ldots ,f_{\theta}^K(x_{i}) \right)$ represents the predicted probabilities that the observation $i$ is in the different classes $k \in \{1,2,\ldots,K\}$.
As a consequence, $\sum_k f_{\theta}^k(x_{i}) = 1$. More interestingly for us, 
the predicted class is the one having the highest probability, so $\hat{y_i}$ is a vector of size $K$ with null values everywhere, except at the index $\argmax_k f_{\theta}^k(x_{i})$, where its value is $1$. 

\subparagraph{Loss and empirical risk}
In order to train the classification model, an empirical risk $\mathcal{R}$ is minimized with respect to the model parameters $\theta$
\begin{equation}\label{regularization:eq:empiricalRisk}
\mathcal{R}(\theta):= \mathbb{E}[\ell (\hat{Y}:=f_{\theta} (X), Y)] \,,
\end{equation}
or empirically $R(\theta) = \frac{1}{n} \sum_{i=1}^n \ell (\hat{y_i}:=f_{\theta} (x_i), y_i)$, where the loss function $\ell$ represents the price paid for inaccuracy of predictions. This optimization problem is almost systematically solved by using variants of the stochastic (or mini-batch) gradient descent
\citep{Bottou_SIAMrev2018} in the machine learning literature.

\subparagraph{Sensitive variable}
An important variable in the field of \emph{fair learning} is the so-called \emph{sensitive variable}, which we will denote $S$. This variable is often binary and distinguishes two groups of observations $S_i \in \{0,1\}$. For instance, $S_i=0$ or $S_i=1$ can indicate that the person represented in observation $i$ is either a male or a female.
A widely used strategy to quantify that a prediction model is fair with respect to the variable $S$ is to compare the predictions it makes on observations in the groups $S=0$ and $S=1$, using a pertinent \emph{fairness metric} (see references of Section \ref{intro:fairnessmetrics}). From a mathematical point of view, this means that the difference between the distributions $(X,Y,\hat{Y})_{S=0}$ and $(X,Y,\hat{Y})_{S=1}$, quantified by the fairness metric,  should be below a given threshold. Consider for instance a binary prediction case where  $\hat{Y_i}=1$ means that the individual $i$ has access to a bank loan, $\hat{Y_i}=0$ means that the bank loan is refused, and that $S_i$ equals $0$ or $1$ refers to the fact that the individual $i$ is  a male or a female. In this case, one can use the difference between the empirical probabilities of obtaining the bank loan for males and females, as a fairness metric, \textit{i.e.} $P(\hat{Y}=1 | S=1) - P(\hat{Y}=1 | S=0)$. More advanced metrics  may also take into account the input observation $X$, the true outputs $Y$, or the prediction model outputs $f_{\theta}(X)$ instead of their binarized version $\hat{Y}$.  

\subsection{W2reg approach for binary classification}

\subsubsection{Regularisation strategy}

We now give an overview of the \emph{W2reg} approach, described in \citep{risser2022tackling}, to temper algorithmic biases of binary neural-network classifiers. The goal of  \emph{W2reg} is to ensure that the treated binary classifier $f_{\theta}$ generates predictions $\hat{Y}$ for which the  distributions in groups $S=0$ and $S=1$ do not deviate too much from pure equality. 
To achieve this, the similarity metric used in \citep{risser2022tackling} is the Wasserstein-2 distance \citep{kantorovitch1939methodes} between the distribution of the predictions in the two groups: 
\begin{equation}\label{regularization:eq:w2regstar}
\mathcal{W}_2^2 (\mu_{\theta,0},\mu_{\theta,1}) = \int_0^1   \left( {\mathcal{H}_{\theta,0}}^{-1} (\tau) - {\mathcal{H}_{\theta,1}}^{-1} (\tau) \right)^2 d \tau \,.
\end{equation}
where $\mu_{\theta,s}$ is the probability distribution of the predictions made by $f_{\theta}$ in group $S=s$, and ${\mathcal{H}_{\theta,s}}^{-1}$ is the inverse of the corresponding  cumulative distribution function. 
Note that $\mu_{\theta,s}$ is mathematically equivalent to the histogram of the model outputs $f_{\theta}(X)$ for an infinity of observations in the group $S=s$, after normalization so that the histogram integral is $1$.  
We remark that this metric is also based on the model outputs $f_{\theta}(X) \in [0,1]$ and not the discrete predictions $\hat{Y} \in \{0,1\}$ (see Section \ref{ssed:GeneralNotations}-\textit{Prediction model} for the formal relation), so the probability distributions $\mu_{\theta,s}$ are continuous. 
Ensuring that this metric remains low makes it possible to control the level of fairness of the neural-network model $f_{\theta}$ with respect to $S$. As specifically modelled by Eq.~\eqref{regularization:eq:w2regstar}, this is made by penalizing the average squared difference between the quantiles of the predictions in the two groups. 

In order to train neural-network which simultaneously make accurate and fair decisions, the strategy of  \citep{risser2022tackling} then consists in optimizing the parameters $\theta$ of the model $f_{\theta}$ such that:
\begin{equation}\label{regularization:eq:minimizedEnergy}
\widehat{\theta} = \argmin_{\theta \in \Theta} \left\{ \mathcal{R}(\theta) + \lambda \mathcal{W}_2^2 (\mu_{\theta,0},\mu_{\theta,1}) \right\} \,,
\end{equation}
where $\Theta$ is the space of the neural-network parameters (\textit{e.g.} the values of the weights, the bias terms and the convolution filters in a CNN). As usual when training a neural-network, the parameters $\theta$ are optimized using a gradient-descent approach, where the gradient is approximated at each gradient-descent step by using a mini-batch of observations. 

\subsubsection{Gradient estimation}

We compute the gradient of Eq.~\eqref{regularization:eq:minimizedEnergy} using the standard back-propagation strategy \citep{LeCunNC1989}. For the empirical risk part of Eq.~\eqref{regularization:eq:minimizedEnergy}, this requires computing the derivatives of the  losses $\ell (f_{\theta}(x_i), y_i)$ with respect to the neural-network outputs $f_{\theta}(x_i)$, something routinely done by packages like PyTorch, TensorFlow or Keras,  for all  mainstream losses. For the Wasserstein-2 part of Eq.~\eqref{regularization:eq:minimizedEnergy}, the authors of \citep{risser2022tackling} proposed to use a mathematical strategy to compute pseudo-derivatives of $\mathcal{W}_2^2 (\mu_{\theta,0},\mu_{\theta,1})$ with respect to the neural-network outputs $f_{\theta}(x_i)$.
Specifically, to compute the pseudo-derivative of a discrete and empirical approximation of $\mathcal{W}_2^2 (\mu_{\theta,0},\mu_{\theta,1})$ with respect to a mini-batch output $f_{\theta}(x_i)$, the following equation was used: 
\begin{equation}\label{regularization:eq:discreteDerivative}
\displaystyle{
\Delta_{\tau} \left[
 \mathds{1}_{s_i=0}
 \frac{     f_{\theta}(x_i) - cor_1(f_{\theta}(x_i))   }{ n_0 \left(H_{0}^{j_i+1} - H_{0}^{j_i} \right)}
 -  
 \mathds{1}_{s_i=1}
 \frac{     cor_0(f_{\theta}(x_i)) - f_{\theta}(x_i) }{ n_1 \left(H_{1}^{j_i+1} - H_{1}^{j_i} \right)}
\right]
} \,,
\end{equation}
where
$n_s$ is the number of observations in class $S=s$, the $H_s^{j}$ are discrete versions of the cumulative distribution functions $\mathcal{H}_{\theta,s}$ defined on a discrete grid of possible output values:
\begin{equation}\label{regularization:eq:PbDiscretization}
\eta^j = \min_i(f_{\theta}(x_i)) + j \Delta_{\eta} \,,\, j=1, \ldots, J_{\eta} \,,
\end{equation}
where $\Delta_{\eta} = J_{\eta}^{-1} (\max_i(f_{\theta}(x_i)) - \min_i(f_{\theta}(x_i)) )$, and $J_{\eta}$ is the number of discretization steps. We denote $H_s^j = H_s (\eta^j)$ and $j_i$ is defined such that $\eta^{j_i} \leq f_{\theta}(x_i) < \eta^{j_i+1}$. Finally  $cor_s(f_{\theta}(x_i)) = H_{s}^{-1}(H_{|1-s|}(f_{\theta}(x_i)))$. 

\subsubsection{Distinction between mini-batch observations and the  observations for $H_0$ and  $H_1$}

As shown in Eq.~\eqref{regularization:eq:discreteDerivative}, computing the pseudo-derivatives of Wasserstein-2 distance $W_2^2 (\mu^n_{\theta,0},\mu^n_{\theta,1})$ with respect to model predictions $f_{\theta}(x_i)$ requires computing the discrete cumulative distribution functions $H_{s}$, with $s \in \{0,1\}$. Computing the  $H_{s}$ would ideally require computing $f_{\theta}(x_i)$ for all $n$ observations $x_i$ of the training set, which would be computationally bottleneck. To solve this issue, \cite{risser2022tackling} proposed to approximate the $H_{s}$ at each mini-batch iteration, where Eq.~\eqref{regularization:eq:discreteDerivative} is computed, using a subset of all training observations. This observation subset is composed of $m$ randomly drawn observations in group $S=0$, $m$ other randomly drawn observations in group $S=1$, and the mini-batch observations. This guaranties that there are at least $m$ observations to compute either $H_0$ or $H_1$, and that the impact of each mini-batch observation is represented in $H_0$ and $H_1$. Note that these additional $2m$ predictions do not require to backpropagate any gradient information, so their computational burden is limited in terms of memory resources. Although it is also reasonable in terms of computational resources, the amount of $2m$ additional predictions should remain relatively small to avoid slowing down significantly the gradient descent. In previous experiences on images, $m=16$ or $m=32$ often appeared as reasonable, as it allowed to mitigate undesirable algorithmic biases and slowed down the whole training procedure by a factor of less than $2$. Finally  preserving the amount of such additional predictions to something reasonable at each gradient descent step will be at the heart of our methodological contribution when extending \emph{W2reg} to multi-class classification.

\subsection{Extended W2reg for multi-class classification}\label{extended_W2reg}

As discussed in Section \ref{sec:intro}, our work is motivated by the need for bias mitigation strategies in NLP applications where the neural-network predicts that an input text belongs to a class among more $K$ output classes, where $K>2$. We show in this section how to take advantage of the properties of \citep{risser2022tackling} to address this practical problem. We recall that the regularization strategy of \citep{risser2022tackling} is model agnostic, so the fact that we treat NLP data will only be discussed in the results section. In terms of methodology, the main issue to tackle is that the model outputs   $f_{\theta}(x_i)$ are in dimension $K>2$ and not one dimensional, which would require to compare multi-variate point clouds following the optimal transport principles which were modelled by Eq.~\eqref{regularization:eq:w2regstar} for 1D outputs. As we will see below, this generates algorithmic problems to keep the computational burden reasonable and to preserve the representativity of the pertinent information. Solving them requires extending \citep{risser2022tackling} with strong algorithmic constraints.

\subsubsection{Reformulating the bias mitigation procedure for multi-class classification}

The strategy proposed by \citep{risser2022tackling} to mitigate undesired biases is to train optimal decision rules $f_{\theta}$ by optimizing Eq.~\eqref{regularization:eq:minimizedEnergy}, where the Wassertein-2 distance between the prediction distributions $\mu_{\theta,0}$ and $\mu_{\theta,1}$ (\textit{i.e.} the distribution of the predictions $f_{\theta}$ for observations in groups $S=0$ and group $S=1$) is given by Eq.~\eqref{regularization:eq:w2regstar}. 
As described in Section \ref{ssed:GeneralNotations}, the predictions $f_{\theta}(x_i)$ are now a vector of dimension $K>2$ in a multi-class classification context (specifically $f_{\theta}(x) \in [0,1]^K$). Their distributions $\mu_{\theta,0}$ and $\mu_{\theta,1}$ are then multivariate. In this context, Eq.~\eqref{regularization:eq:w2regstar}  does not hold, and another optimal transport metric such as the multivariate Wasserstein-2 Distance or the Sinkhorn Divergence should be used \citep{ChizatEtAl20}. Note that different implementations of these metrics exist and are compatible with our problem---\textit{e.g.}, those of  \citep{flamary2021pot,feydy2019interpolating}.
This however opens a critical issue related to the number of observations needed to reasonably penalize the differences between two multivariate point clouds, representing the observations in groups $S=0$ and $S=1$. If the dimension $K$ of the compared data gets large, the number of observations required to reasonably compare the point clouds at each gradient descent step  explodes. This problem is very similar to the well known \emph{curse of dimensionality} phenomenon in machine learning, where the  amount of data needed to generalize accurately the predictions grows exponentially as the number of dimensions grows. 

This issue therefore leads us to think about which problem we truly need to solve when tackling undesired algorithmic bias in multi-class classification. From our application perspective, a discrimination appears when there the prediction model $f_{\theta}$ is significantly more accurate to predict a specific output in one of the two groups represented by $S \in \{0,1\}$. For instance, suppose that someone looks for \textit{Software Engineer} jobs and that an automatic prediction model $f_{\theta}$ is used to recommend job candidates to an employer. For a given job candidate $x_i$, the prediction model  will return a set of $K$ probabilities, each of them indicating whether $x_i$ is recommended for the job class $k$. Now $k$ will denote the class of jobs $x_i$ is looking for, \textit{i.e.} \textit{Software Engineer}.
The prediction model will be considered as unfair if male  profiles in this class (here, \textit{Software Engineer}) are on average clearly more often recommended by $f_{\theta}$ than female profiles also in this class, when an unbiased oracle would lead to equal opportunities, \textit{i.e.} 
\begin{equation}\label{regularization:eq:TPRg}
|P(\hat{Y}^k=1 | Y^k=1 , S=1) - P(\hat{Y}^k=1 | Y^k=1 , S=0) | > \tau \\,
\end{equation}
where the left-hand term is denoted the \emph{True Positive Rate  gap} (TPRg), and $\tau$ is a threshold above which the TPRg is considered as unfairly discriminating. As shown  in Section \ref{sec:results}, such situations can occur in automatic job profile recommendation systems using modern neural-networks.

Now that we have clarified the problem we need to tackle, we can reformulate the regularized multi-class model training procedure as follows:
\begin{itemize}
\item
We first train and test a non-regularized multi-class classifier $f_{\theta^{bl}}$. We will denote it the \emph{baseline classifier}.
\item
We define a threshold $\tau$ under which all occupations with predictions $k\in \{1,\ldots,K\}$ should have a TPRg (see Eq.~\eqref{regularization:eq:TPRg}). We denote $\{c_1,\ldots,c_C\}$ the classes for which this condition is broken, where each of these classes takes its values in $\{1,\ldots,K\}$.
\item
We then retrain the multi-class classifier $f_{\theta}$ with regularization constrains on the classes $\{c_1,\ldots,c_C\}$ only. The regularization strategy will be developed below in Section~\ref{regularization:ssec:regStrategy}.
\end{itemize}

By using this procedure, the number of observations required at each mini-batch step will be first limited to observations in the groups $\{c_1,\ldots,c_C\}$ only, which is a first step towards an algorithmically reasonable regularized training procedure. We also believe that this also avoids to over-constrain the training procedure, which often penalizes its convergence.  

\subsubsection{Regularization strategy}\label{regularization:ssec:regStrategy}

We now push further the algorithmic simplification of the regularization procedure by focusing on the properties of the mini-batch observations. In this subsection, we  suppose that $x_i$ is an input mini-batch observation, and recall  that we want to penalize large TPRg for specific classes $\{c_1,\ldots,c_C\}$ only. In this mini-batch step, the observations $x_i$ related to true output predictions $y_i^k=1$ for which  $k \notin \{c_1,\ldots,c_C\}$ are not concerned by the regularization, when computing the multivariate cumulative distribution function $H_0$ or $H_1$. 
At each mini-batch step, it therefore appears as appealing to only consider the dimensions out of $\{c_1,\ldots,c_C\}$, for which at least one true output observation respects $y_i^k=1$, with $k \in \{c_1,\ldots,c_C\}$.
This would indeed allow to further reduce the amount of additional predictions made in the mini-batch. The dimension of $H_0$ or $H_1$ would however vary at each mini-batch step, making potentially the distance estimation unstable if fully considering $C$-dimensional distributions.

To  take into account the fact that not all output dimensions $\{c_1,\ldots,c_C\}$ should be considered at each gradient descent step, we then make a simplification hypothesis: we neglect the relations between the different dimensions when comparing the output predictions in groups $S=0$ and $S=1$. This hypothesis is the same as the one made when using Naive Bayes classifiers \citep{IdiotBayes2001,rish2001empirical}. We believe that this hypothesis is particularly suited for one-hot-encoded outputs, as they are constructed to ideally have a single value close to 1 and all other values close to 0. We then split the multi-variate regularization strategy into a multiple one-dimensional strategy and optimize:
\begin{equation}\label{regularization:eq:minimizedEnergy_extension}
\widehat{\theta} = \argmin_{\theta \in \Theta} \left\{ \mathcal{R}(\theta) + 
\sum_{l=1}^{C}
\lambda_{c_l} \mathcal{W}_2^2 (\mu_{\theta,0}^{c_l},\mu_{\theta,1}^{c_l}) \right\} \,,
\end{equation}
where $\mathcal{W}_2^2$ is the metric of Eq.~\eqref{regularization:eq:w2regstar}, $\lambda_{c_l}$ is the weight given to regularize the TPR gaps in class $c_l$, and the $\mu_{\theta,s}^{c_l}$ are the distributions of the output predictions on dimension $c_l$, \textit{i.e.} the distribution of the $f_{\theta}^{c_l}(x)$, when the true prediction is $c_l$,  \textit{i.e.} when $y^{c_l}=1$. 
For a mini-batch observation $x_i$ related to an output prediction in a regularized class $k \in \{c_1,\ldots,c_C\}$, 
the impact of a mini-batch output $f_{\theta}^k(x_i)$ on the empirical approximation of $\mathcal{W}_2^2 (\mu_{\theta,0}^{k},\mu_{\theta,1}^{k})$ can then be estimated by following the same principles as in \citep{risser2022tackling}. We can then extend Eq.~\eqref{regularization:eq:discreteDerivative} with: 
\begin{equation}\label{regularization:eq:discreteDerivative_extension}
\displaystyle{
\Delta_{\tau} \left[
 \mathds{1}_{s_i=0}
 \frac{     f_{\theta}^k(x_i) - cor_1(f_{\theta}^k(x_i))   }{ n_{k,0} \left(H_{k,0}^{j_i+1} - H_{k,0}^{j_i} \right)}
 -  
 \mathds{1}_{s_i=1}
 \frac{     cor_0(f_{\theta}^k(x_i)) - f_{\theta}^k(x_i) }{ n_{k,1} \left(H_{k,1}^{j_i+1} - H_{k,1}^{j_i} \right)}
\right]
} \,,
\end{equation}
where the $H_{k,s}$ are discrete and empirical versions of the cumulative distribution functions of the prediction outputs on dimension $k$, \textit{i.e.} the  $f_{\theta}^k(x)$, when class $k$ should be predicted and the observations are in the group $s$. 
Note too that Eq.~\eqref{regularization:eq:discreteDerivative} contains $n_s$, which is the number of observations in class $S=s$. In order to manage unbalanced output classes in the multi-class classification context, we also use a normalizing term $n_{k,s}$ in Eq.~\eqref{regularization:eq:discreteDerivative_extension}. It quantifies the number of training observations in group $s \in \{0,1\}$ and class $k \in \{1,\ldots,K\}$. Other notations are the same as in Eq.~\eqref{regularization:eq:discreteDerivative}. 

In a mini-batch step, suppose finally that only need to take into account the classes
$\{\widehat{c_1}, \ldots , \widehat{c_{D}}\}$ among the $\{c_1,\ldots,c_C\}$. These selected classes are those for which at least a $y_i^{c_j}=1$, with $j \in \{1,\ldots,C\}$ and, $i$ is an observation of the mini-batch $B \subset \{1,\ldots,n\}$. We will then have to only sample  two times $m$ predictions, for each of the selected $D$ classes, to compute the $H_{\widehat{c_d},0}$ and $H_{\widehat{c_d},1}$ required in Eq.~\eqref{regularization:eq:discreteDerivative_extension}. This makes the computational burden to regularize the  neural-network training procedure reasonable, as the number of additional predictions to make only increases linearly with the number of treated classes at each mini-batch iteration. Note that no additional prediction will also be needed when a mini-batch contains no observation related to a regularized output class. This will naturally be often the case, when the number of classes $K$ gets large and/or the mini-batch size $\#B$ is small.

\subsubsection{Proposed training procedure}

The proposed strategy to train multi-class classifiers with mitigated algorithmic biases on specific classes prediction was motivated by the future need of certifying that automatic decision models are not discriminatory. In order to make absolutely clear our strategy, we detail it in algorithm \ref{alg:BTP_NN_W2_mutli_class}.

\begin{algorithm}
\caption{Procedure to train bias mitigated multi-class neural-network classifiers}
\label{alg:BTP_NN_W2_mutli_class}
\begin{algorithmic}[1]
\State \textbf{Input:} Training observations $(x_i,s_i,y_i)_{i=1,\ldots,n}$, where $x_i\in \mathbb{R}^p$, $s_i \in \{0,1\}$, and $y_i \in \{0,1\}^K$, plus a multi-class neural-network model $f_{\theta}$.
\State [\textit{Detection of the output classes with discriminatory predictions}]
\State Train the baseline parameters $\theta^{bl}$ of $f$ on $(x_i,s_i,y_i)_{i=1,\ldots,n}$ with no specific regularization.
\State Find the output classes $\{c_1,\ldots,c_C\}$ on which the model $f_{\theta^{bl}}$ has unacceptable True Positive Rate gaps (TPRg) using Eq.~\eqref{regularization:eq:TPRg}.
\State [\textit{Multi-class W2reg training}]
\State Re-initialize the training parameters $\theta$.
\For{$e$ in Epochs}
    \For{$b$ in Batches}
        \State Draw the batch observations $(x_i,s_i,y_i)_{i \in B}$, where $B$ is a subset of $\{1,\ldots,n\}$.
        \State Compute the mini-batch predictions $f_{\theta}(x_i)$, $i\in B$.
        \State Detect the output classes $\{\widehat{c_1}, \ldots , \widehat{c_{D}}\}$ among the $\{c_1,\ldots,c_C\}$ for which at least one $y_i^{c_j}=1$, with $j \in \{1,\ldots,C\}$ and $i\in B$.
        \State For each $j \in \{1,\ldots,D\}$, pre-compute $H_{\widehat{c_{j}},0}$ and $H_{\widehat{c_{j}},1}$ using $m$ output predictions $f_{\theta}(x_i)$, where $i\notin B$ and $y_i^{\widehat{c_{j}}}=1$. 
        \State Compute the empirical risk and its derivatives with respect to $f_{\theta}(x_i)$, $i\in B$.
        \State Compute the pseudo-derivatives of the discretized $W_2^2 (\mu^{\widehat{c_j}}_{\theta,0},\mu^{\widehat{c_j}}_{\theta,1})$ with respect to the pertinent mini-batch outputs $f_{\theta}(x_i)^{\widehat{c_j}}$ using Eq.~\eqref{regularization:eq:discreteDerivative_extension}.
        \State Backpropagate the risk derivatives and the pseudo-derivatives of the $W_2^2$ terms.
        \State Update the parameters $\theta$.
    \EndFor
\EndFor
\State \textbf{Output:} Trained neural network $f_{\theta}$ with mitigated biases.
\end{algorithmic}
\end{algorithm}

\section{Experimental protocol}

\subsection{Data}

We assess our methodology using the \textit{Bios} \citep{de2019bias} dataset, which  contains about 400K biographies, the gender of its author as well as its occupation (28). For more details on this dataset, please see section \ref{biosdataset}.

\subsection{Neural-network model and baseline training strategy}\label{regularization:ssec:baselinetraining}

Our task is to predict the occupation using only the textual data of the biography. We do it by using a RoBERTa model \citep{liu2019roberta}, which is based on Transformer architectures and is  pretrained with the Masked language modeling (MLM) objective. %
We specifically used a RoBERTa base model pretrained by Hugging Face. All information related to how it was trained can be found in  \citep{liu2019roberta}. It can be remarked, that a very large training dataset was used to pretrain the model, as it was composed of five datasets: \emph{BookCorpus} \citep{moviebook}, a dataset containing 11,038 unpublished books; \emph{English Wikipedia} (excluding lists, tables and headers); \emph{CC-News} \citep{ccnews2020} which contains 63 millions English news articles crawled between September 2016 and February 2019; \emph{OpenWebText} \citep{radford2019language}  an open-source recreation of the WebText dataset  used to train GPT-2; \emph{Stories} \citep{trinh2018simple} a dataset containing a subset of CommonCrawl data filtered to match the story-like style of Winograd schemas. Pre-training was performed on these data, by randomly masking 15\% of the words in each of the input sentences and then trying to predict the masked words 
After pre-training RoBERTa parameters on this huge dataset, we then trained it on the 400.000 biographies of the  \emph{Bios} dataset. Training was performed with PyTorch on 2 GPUs (Nvidia Quadro RTX6000 24GB RAM) for 10 epochs with a batch size of 8 observations and a sequence length of 512 words. The optimizer was Adam with a learning rate of 1e-5, $\beta_1 = 0.9$, $\beta_2 = 0.98$, and $\epsilon = 1e6$. Computational times required about 36 hours for each run. We want to emphasize that 5 runs of the training procedure were performed to evaluate the stability of the accuracy and the algorithmic biases. For each of these runs, we the split dataset in 70\% for training, 10\% for validation and 20\% for testing. We will denote, as baseline models, the neural-networks trained using this procedure. 

\subsection{Evaluating the impact of a gender-neutral dataset}\label{subs:genderneutral}

In order to evaluate the impact of a classic gender unbiasing strategy, we reproduced the baseline training protocol of Section \ref{regularization:ssec:baselinetraining} on two \emph{apparently} unbiased versions of the \emph{Bios} dataset.
This classic method for debiasing consists of removing explicit gender indicators ({\em i.e.}{\em 'he', 'she', 'her', 'his', 'him', 'hers', 'himself', 'herself', 'mr', 'mrs', 'ms', 'miss'} and first names).
For a BERT model type, however, we could not just remove words because the model is sensitive to sentence structure, not just lexical information. 
We therefore adjusted the method by replacing all the first names by  a neutral first name\footnote{We can take any first name because, since we change all the first names of the dataset by this one, it will necessarily be neutral.} (\textit{Camille}) and by choosing only one gender for all datasets (e.g., for all individuals of gender g, we did nothing; for the others, we replaced explicit gender indicators with those of g).
We then created two datasets with only female or male gender indicators, and the only first name \textit{Camille}.

Note that by using a fully trained model on our dataset, setting all gender indicators to either feminine or masculine should naively not change anything, since the model would only "know" one gender (which  would therefore be neutral). We however used a pre-trained model on gendered datasets. It is therefore important to verify that fine-tuning this model with a male gendered dataset is equivalent to training it on a female gendered dataset.
To assess this, we carried out several student tests. One between the accuracy of the trained model on the female gendered dataset and the accuracy of the male gendered one. One on the TPR gender gap for each of the professions between the two models. 
None of these tests had a statistically significant difference. We will then only present in Section~\ref{sec:results} the results obtained on the model trained on the female gendered data set.

\subsection{Training procedure for the regularized model}

We now follow the procedure summarized in Algorithm~\ref{alg:BTP_NN_W2_mutli_class} to train bias mitigated multi-class neural-network classifiers on the textual data of the \emph{Bios} dataset.
We first consider the 5  baseline models of Section~\ref{regularization:ssec:baselinetraining}, which were trained on the original  \emph{Bios} dataset (and not one of the unbiased datasets of Section~\ref{subs:genderneutral}).
As shown in Fig.~\ref{regularization:fig:confmat}-(left), where the diagonal of the confusion matrices differences between males and females represents the TPR gap of all output classes, two classes have TPR gaps above $0.1$ or under $-0.1$: \emph{Surgeon} (in favor of males) and \emph{Model} (in favor of females). We then chose to regularize the predictions for these two occupations. Other occupations could have been considered (see Fig.~\ref{regularization:fig:confmat_appendix}) but they did not contain enough statistical information to be properly treated. For instance, although the whole training set contains about 400.000 observations, it contains less than 100 female \emph{dj}s and less than 100 male \emph{paralegal}s. 

After having selected these two occupations, we trained 5 regularized models by minimizing Eq.~\eqref{regularization:eq:minimizedEnergy_extension}.
We chose a single $\lambda$ parameter for the regularization (the same for both classes, but we could have taken one per class), by using cross validation, with the goal to effectively reduce the TPR gaps on regularized classes without harming the accuracy too much. The best performance/debiasing compromise we found was $\lambda = 0.0001$. An amount of $m=16$ additional observations was used at each mini-batch step to compute each of the discrete cumulative histograms $H_{k,s}$ of the regularization terms pseudo-derivatives Eq.~\eqref{regularization:eq:discreteDerivative_extension}. The rest of the training procedure was the same as in Section~\ref{regularization:ssec:baselinetraining}. Computational times required about 70 hours for each run.

\subsection{Overview of the classifiers compared in Section~\ref{sec:results}}

\begin{figure}
    \centering
    \includegraphics[width=1\linewidth]{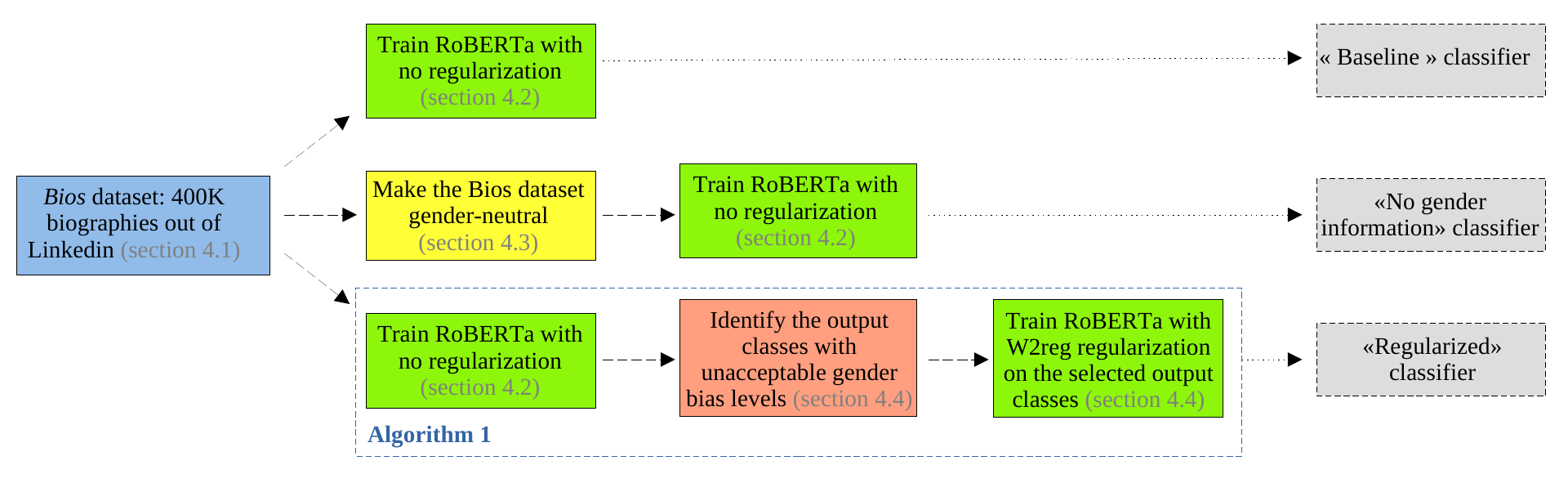}
    \caption{Pipelines followed to define  the three classifiers compared in  Section~\ref{sec:results}.}
    \label{regularization:fig:experimental_protocol_pipeline}
\end{figure}

In order to assess the impact of the in-processing bias mitigation technique proposed in this chapter, we will compare it in Section~\ref{sec:results} to  a non-regularized strategy (denoted \emph{baseline}), and a pre-processing strategy where the biographies were made neutral (denoted \emph{no gender information}). We recall that an amount of 5 models were trained in each case to evaluate the statility of the training proceedures and their biases. An overview of the pipeline used in each case is shown in Fig~\ref{regularization:fig:experimental_protocol_pipeline}.

\section{Results and discussion}\label{sec:results}

\begin{figure}
    \centering
    \includegraphics[width=1\linewidth]{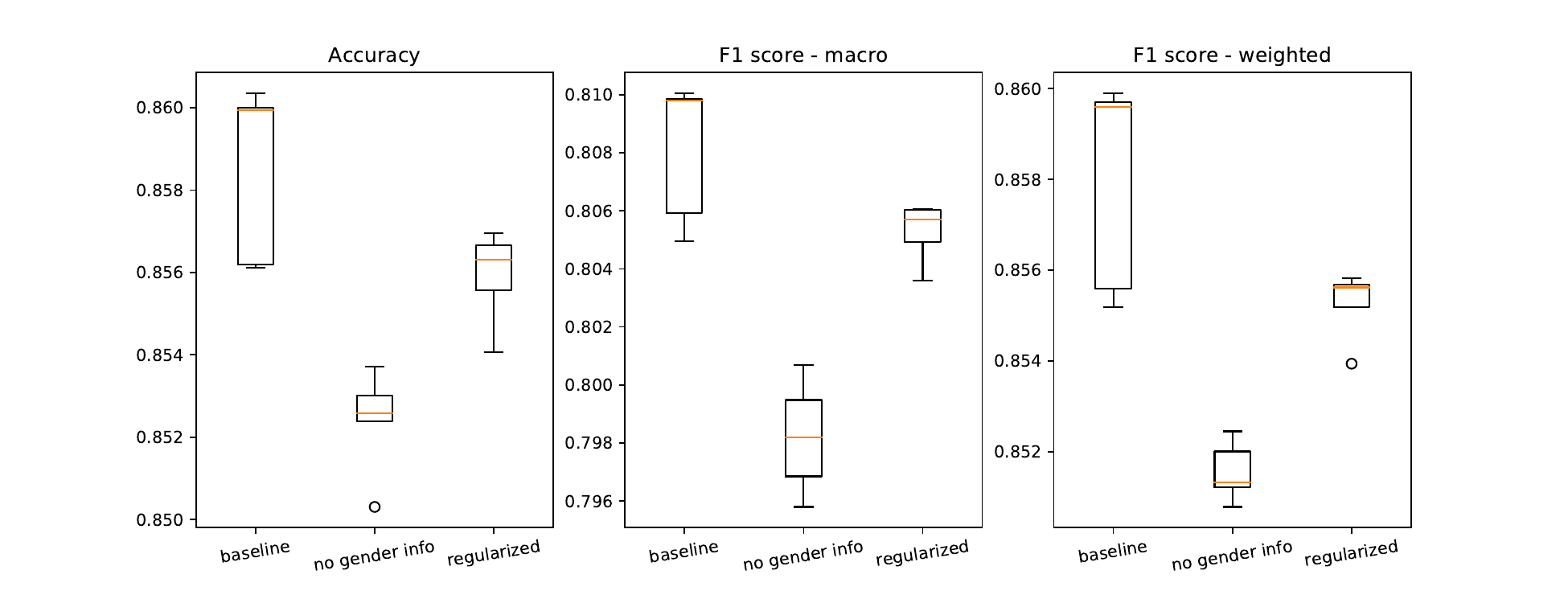}
    \caption{\textbf{(Left to Right)} Box plots of the Accuracy, unweighted F1 scores and weighted F1 scores for the baseline models with raw biographies, the baseline models with unbiased biographies, and the  regularized models with raw biographies.}
    \label{regularization:fig:acc_F1scores}
\end{figure}

In commercial applications, fair prediction algorithms will be obviously more popular and useful if they remain accurate. Thus, we made sure that our regularization technique did not have a strongly negative impact on the prediction accuracy. We then quantified different accuracy metrics: First the average accuracy and then two variants of the F1 score, as it is very appropriate for a multiclass classification problem like ours. These two variants are the so-called “macro” F1-score, where we calculate the metric for each class, then we average it without taking into account the number of individuals per class; and the “weighted” F1-score where the means are weighted using the classes representativeness.
We can draw similar conclusions for these three metrics, as shown in Fig.~\ref{regularization:fig:acc_F1scores}: our regularization method is certainly a little below the baseline in terms of accuracy, but it is more stable. In addition, it is clearly more accurate than the gender neutralizing technique of Section~\ref{subs:genderneutral}.

\begin{figure}
    \centering
\includegraphics[width=1\linewidth]{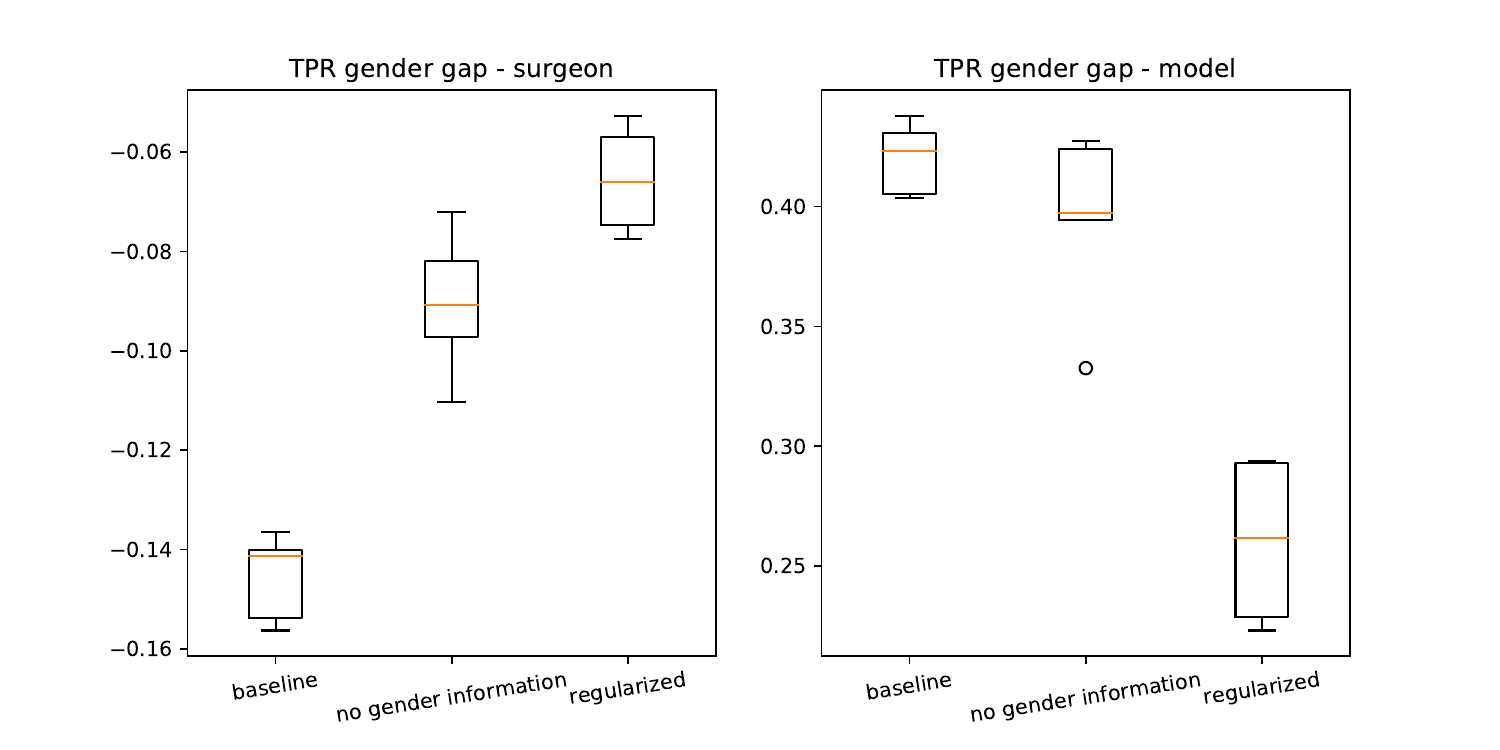}
    \caption{Box plots of the True Positive Rate (TPR) gender gaps for the output classes \textit{Surgeon} and \textit{Model} obtained using the baseline models with raw biographies, the baseline models with unbiased biographies, and the  regularized models with raw biographies. Note that there is a bias in favour of females or males if a TPR gender gap is positive or negative, respectively.}
    \label{regularization:fig:TPRGGap}
\end{figure}

We then specifically observed the impact of our regularization strategy in terms of TPR gap on the two regularized classes \emph{Surgeon} and \emph{Model}. Boxplots of the TPR gap for these output classes are shown in  Fig.~\ref{regularization:fig:TPRGGap}. They  confirm that the algorithmic bias has been reduced for these two classes. For the class \emph{Surgeon}, removing gender indicators had a strong effect, but the regularization strategy further reduced the biases.  For the class \emph{Model}, removing gender indicators had little effect, and the regularization strategy reduced the biases by almost a factor two.

\begin{figure}
    \centering
\includegraphics[width=1\textwidth]{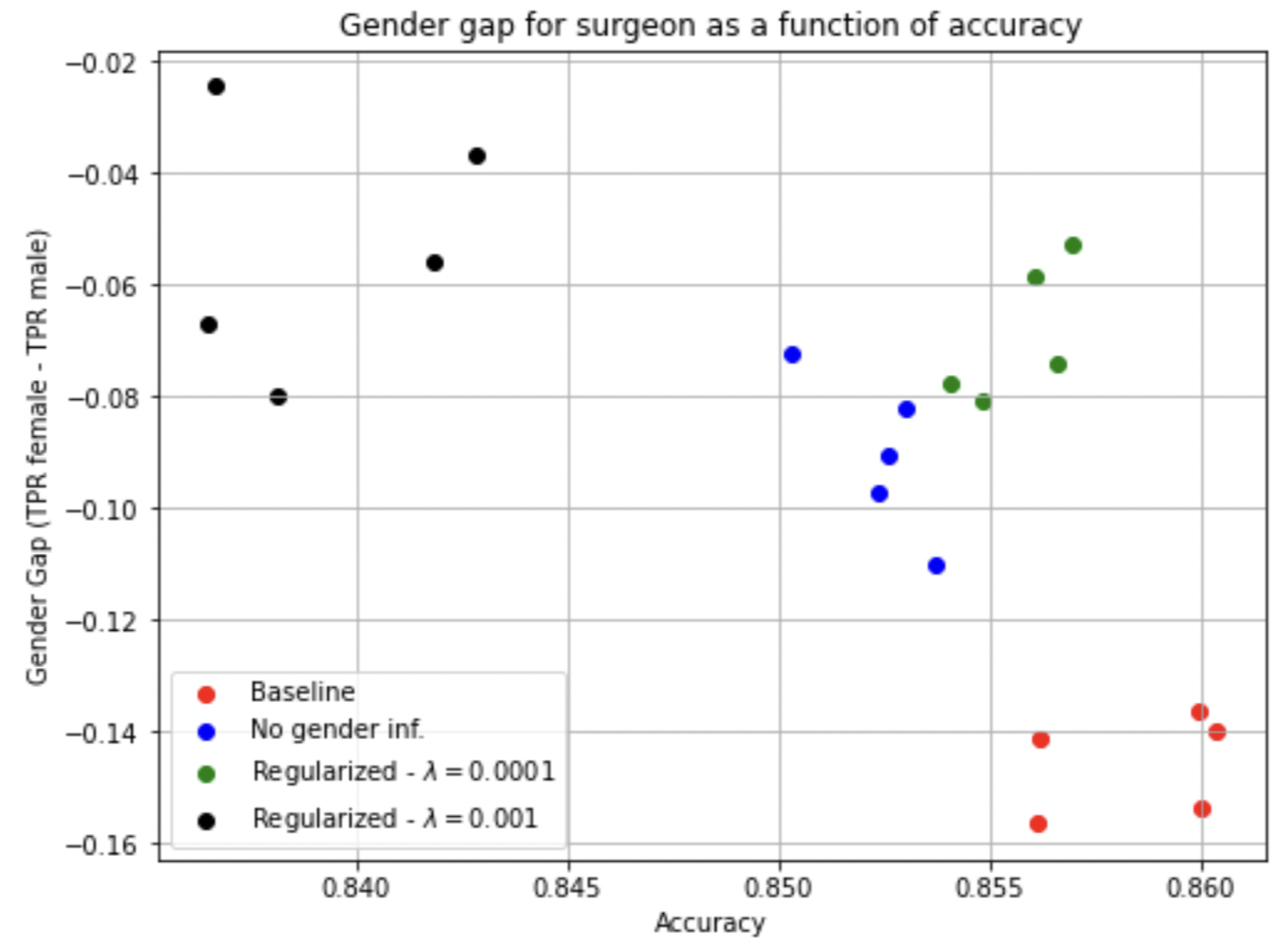}
\includegraphics[width=0.99\textwidth]{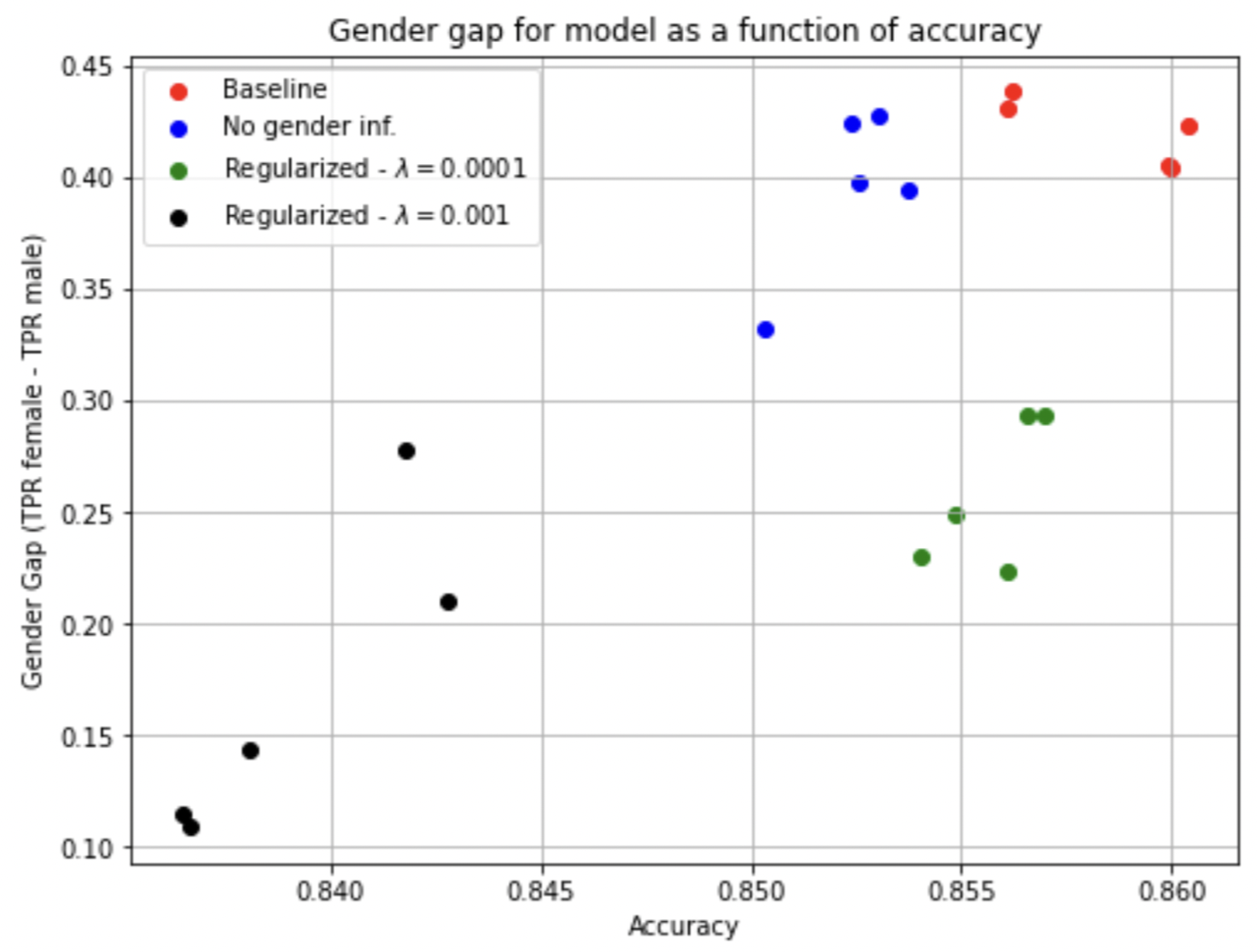}
    \caption{True Positive Rate (TPR) gender gaps for the output classes \textit{Surgeon} (Upper) and \textit{Model} (Lower) obtained using the baseline models with raw biographies, the baseline models with unbiased biographies, and the  regularized models with raw biographies for $\lambda = 0.0001$ (ours) and $\lambda = 0.001$ (variation) as a function of accuracy for each runs.}
    \label{regularization:fig:TPRacc}
\end{figure}

In Figure \ref{regularization:fig:TPRacc}, we analyze the TPR gender gap for the two regularized classes (\textit{Surgeon} and \textit{Model}) as a function of accuracy across different models: the baseline model, the model trained on the dataset without explicit gender indicators, and the regularized model across all runs (5). We also compare our selected $\lambda=0.0001$ value with an alternative well-chosen $\lambda=0.001$. When $\lambda=0.001$, the TPR gender gap for both classes is better compared to $\lambda=0.0001$; however, the difference is not substantial (3 out of 5 runs show similar levels for surgeon and 2 out of 5 for model). Conversely, the accuracy is clearly lower with $\lambda=0.001$. Therefore, the choice of $\lambda=0.0001$ seems to us to be a better trade-off between achieving a favorable TPR gender gap and maintaining good accuracy. In the other figures, we will retain only our $\lambda=0.0001$.

\begin{figure}
    \centering
\includegraphics[width=1.0\linewidth]{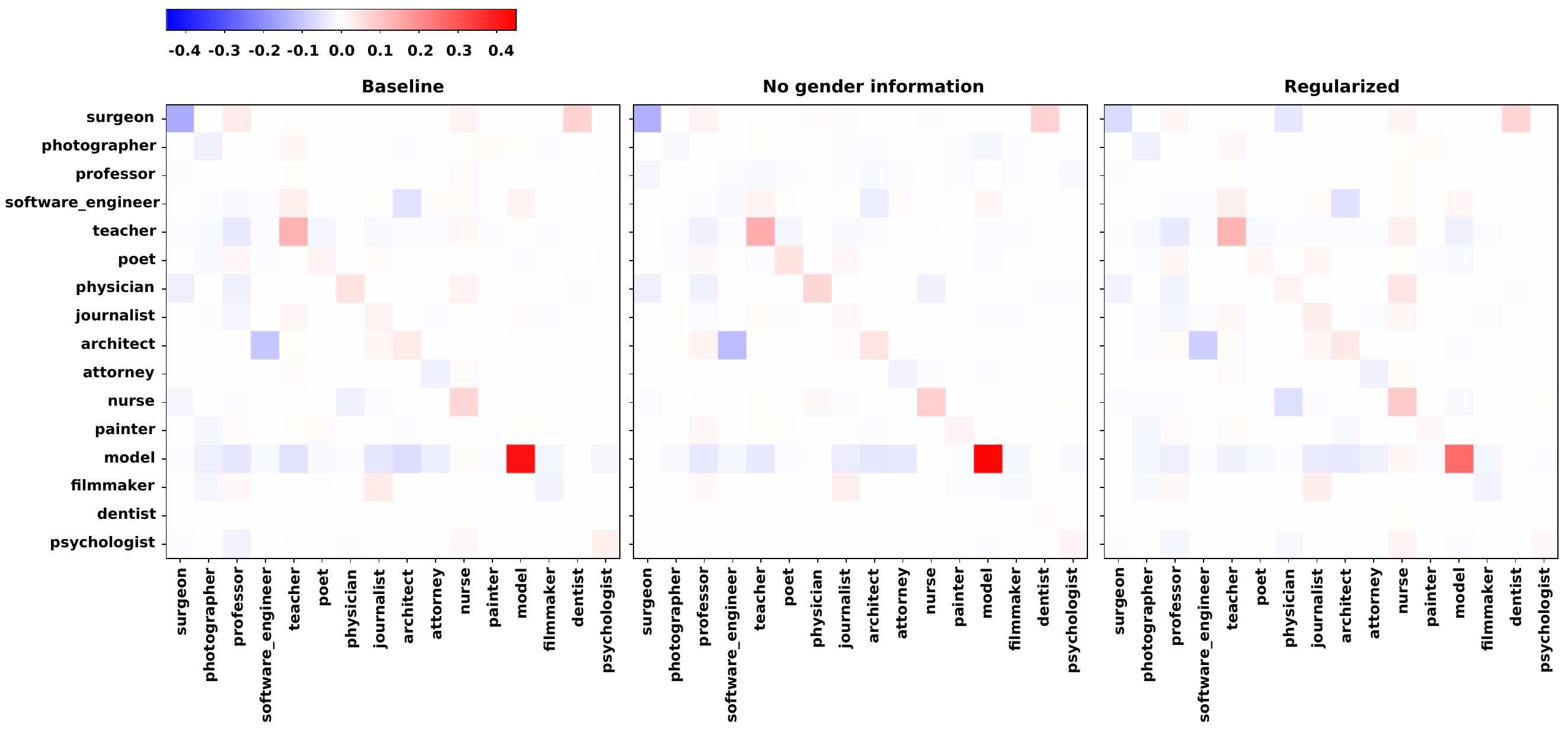}
    \caption{Average difference between the confusion matrices of the predicted outputs $\hat{y}$ versus the true outputs $y$  obtained for females and males. Note that the diagonal values of these matrices correspond to average TPR gaps. The confusion matrices were also normalized over the true (rows) conditions. The redder a value the stronger  the bias in favour of females, and the bluer a value the stronger the bias in favour of males.}
    \label{regularization:fig:confmat}
\end{figure}

\begin{figure}
    \centering
\includegraphics[width=1\linewidth]{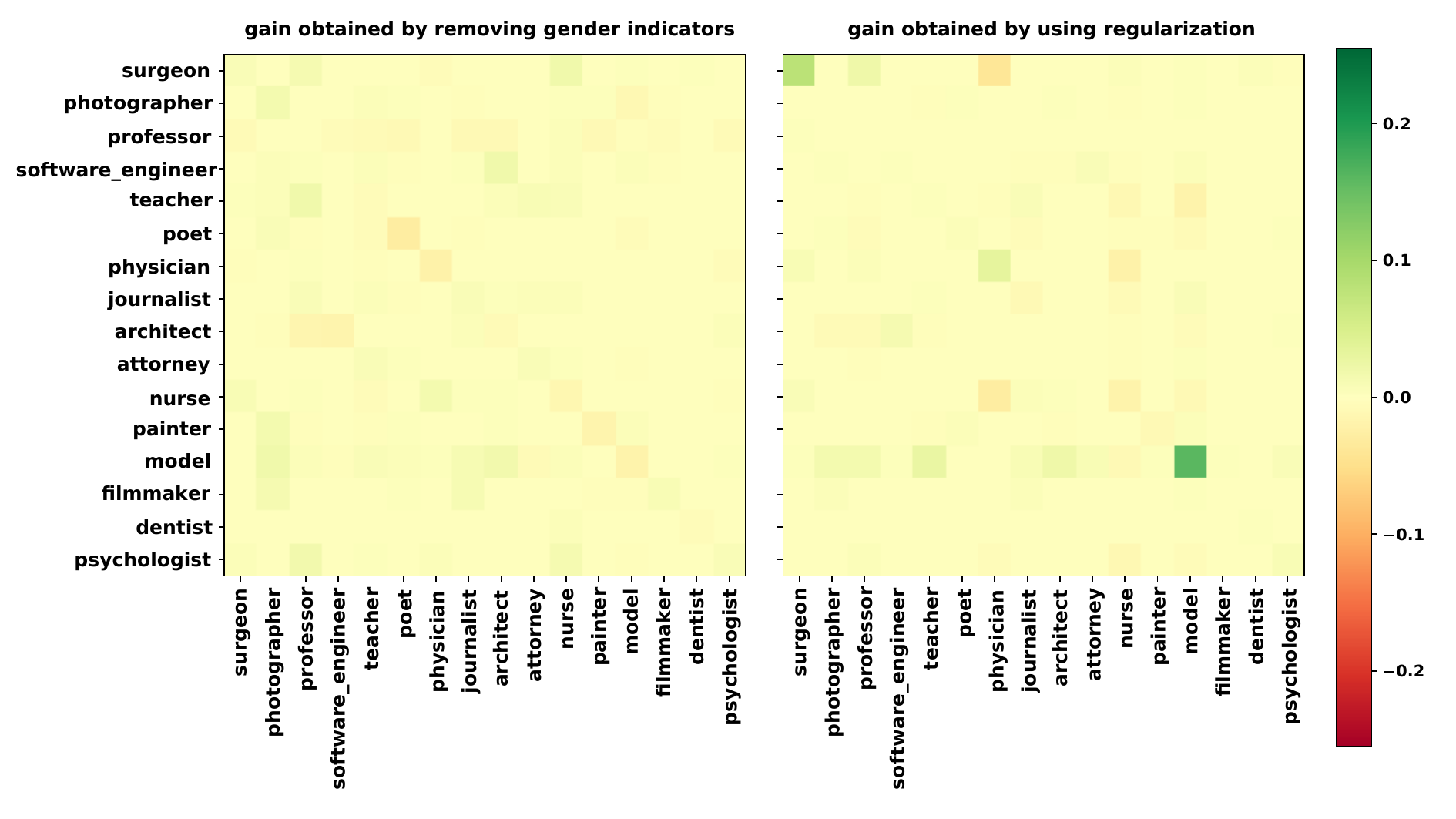}
    \caption{\textbf{(Left)} Difference between the baseline matrix of Fig.~\ref{regularization:fig:confmat}, and the one obtained with an unbiased training set. \textbf{(Right)} Difference between the baseline matrix of Fig.~\ref{regularization:fig:confmat}, and the one obtained using regularized optimization. The greener the values, the more the technique has reduced the bias between men and women. The redder the values, the more the bias has been amplified.}
    \label{regularization:fig:confmat_gains}
\end{figure}

We finally wanted to make sure that reducing the unacceptable biases on these two classes would not be at the expense of newly generated biases. We then measured the difference between the average (on the 5 models) confusion matrix for females and males only.

In Fig.~\ref{regularization:fig:confmat}, we see the evolution of our biases according to the selected method. 
Note first that the diagonal of these matrix differences corresponds to the TPR gaps.
Remark too that we only represent the results obtained on the 16 most frequent occupations for visibility concerns, but  are the complete matrices are show in the appendix.
On our two regularized classes, we are getting closer to white (i.e. non-bias), and for the other classes, we also observe a decrease in bias in general, and no outlier point.
For a finer analysis and more clarity, we represent in Fig.~\ref{regularization:fig:confmat_gains}  the difference between the absolute values of the baseline matrix of Fig.~\ref{regularization:fig:confmat} and each of the compared matrices (\emph{i.e.} with neutralized genders and regularization). 
This clearly represents us the “gains” of these two bias reduction methods to compare them. Figure \ref{regularization:fig:confmat_gains} confirms our intuition given in Fig.~\ref{regularization:fig:confmat}: in the case where the gender indicators are removed, the gain is rather slight and depends on the class. In the case of our regularization, the two regularized classes obtain a very clear positive gain, and there is no marked negative gain on the rest of the matrix.

\newpage

\section{Conclusion}
In this chapter, we have defined a strategy to address the critical need for certifying that commercial prediction models present moderate discrimination biases.  We specifically defined a new algorithm to mitigate undesirable algorithmic biases in multi-class neural-network classifiers, and applied it to NLP application that is ranked as \emph{High risk} by E.U. regulations. Our method was shown to successfully temper algorithmic biases in this application, and outperformed a classic strategy both in terms of prediction accuracy and mitigated bias. In addition, computational times were only reasonably increased compared with a baseline training method. 
The state of the art of \textit{in-processing} unbiasing methods mainly focuses on binary models, and our approach addresses the multiclass problem. The possibility of choosing which classes to regularize and of applying a different $\lambda$ for each class gives a wide range of application of the method.

We want to make clear two potential difficulties that we anticipate for future users of our method:
(1) The W2reg method allows the user to choose a specific $\lambda$ values for each  regularized class. As mentioned in the previous paragraph, this makes the method very flexible and gives a control on the level of regularization required to obtain reasonable biases for each output class. Finding optimal regularization weights can however be time consuming for the user. Note that this compromise to find between accuracy and regularization is however extremely common in engineering science.  (2) As thoroughly discussed in Appendix A,  the regularization strategy is also effective when the training set has a reasonable amount of observations in a treated class. Using our method in order to later certify that a multi-class classification neural network is not biased for a specific output class will then require to have enough training data in this class. This phenomenon is related to the generalization properties of any decision model trained on reference data. More observations will have to be acquired in these classes otherwise.

Now that \citep{risser2022tackling} has been extended to multi-class classification, a natural perspective would be to also use it for the regression case. We expect this extension to be methodologically straightforward, as the binary and multi-class W2reg strategies for classification already regularize continuous outputs (specifically, the probabilities of belonging into a class). A more challenging perspective would be to adapt and eventually to reformulate W2reg to other popular decision models, as all applications of AI are not based on neural-networks. Adapting W2reg to other fairness criteria would finally make it more versatile.

We finally want to emphasize that although our method was applied to NLP data, it can be easily applied to any multi-class neural-network classifier. We also believe that it could be simply adapted to other fairness metrics. Our regularization method is implemented to work as a loss in PyTorch and is compatible with PyTorch-GPU. It is freely available on GitHub\footnote{\url{https://github.com/lrisser/W2reg}}.

%% file: Fairness_metrics.tex
\chapter{Stability of fairness metrics}

We move from a focused exploration of bias reduction methods via fairness measures in the previous chapter to a broader and more critical examination of these metrics themselves in this chapter. This shift in perspective allows for a more comprehensive understanding and critique of the established metrics used to assess discrimination bias in machine learning. Central to this chapter is the paper "\textit{Are fairness metric scores enough to assess discrimination biases in machine learning?}" \citep{jourdan2023fairness} accepted for publication at the Third Workshop on Trustworthy Natural Language Processing, ACL 2023. This paper proposes an experimental protocol for analyzing the shortcomings of the most commonly used fairness metrics from a statistical standpoint. 


This paper presents novel experiments shedding light on the shortcomings of current metrics for assessing biases of gender discrimination made by machine learning algorithms on textual data. We focus on the \emph{Bios} dataset, and our learning task is to predict the occupation of individuals, based on their biography. Such prediction tasks are common in commercial NLP applications such as automatic job recommendations. 
We address an important limitation of theoretical discussions dealing with group-wise fairness metrics: they focus on large datasets, although the norm in many industrial NLP applications is to use small to reasonably large linguistic datasets for which the main practical constraint is to get a good prediction accuracy. We then question how reliable are different popular measures of bias when the size of the training  set is simply sufficient to learn reasonably accurate predictions.
Our experiments sample the \emph{Bios} dataset and learn more than 200 models on different sample sizes. This allows us to statistically study our results and to confirm that common gender bias indices provide diverging and sometimes unreliable results when  applied to relatively small training and test samples.  This highlights the crucial importance of variance calculations for providing sound results in this field.

\section{Introduction}

Potential biases introduced by  Artificial Intelligence (AI) systems are now both an academic concern, but also a critical problem for industry, as countries plan to regulate AI systems that could adversely affect individual users.  The so-called \emph{AI act}\footnote{\url{https://eur-lex.europa.eu/legal-content/EN/TXT/HTML/?uri=CELEX:52021PC0206&from=EN}} will require AI systems sold in the European Union to have good statistical properties with regard to any potential discrimination they could engender (more details in chapter \ref{introduction}).  

These regulatory advances have made it a pressing issue to define which  metrics are appropriate for evaluating whether machine learning models can be considered fair algorithms in various industrial settings. In this context, we believe that these articles open at least two issues:
(1) Each fairness metric quantifies the fairness of a model in a different way and not all metrics are compatible with each other, as already discussed in \citep{kleinberg2016inherent, chouldechova2017fair, pleiss2017fairness}. It is therefore easy to optimize its algorithm according to a single metric to claim fairness while overlooking all the other aspects of fairness measured by other metrics.
(2) Given that contemporary, theoretical discussions of fairness focus on large datasets but that the norm in many industrial NLP applications is to use small linguistic datasets \citep{ezen2020comparison}, one can wonder how reliable different popular measures of bias when the size of the training and validation sets is simply sufficient to learn reasonably accurate predictions. 
In general, this leads us to pose two questions, which are central to this chapter: 
Are fairness metrics always reliable on small samples, which are common in industrial contexts? How do they behave when applying standard debiasing techniques?

To answer these questions, we propose a new experimental protocol to expose gender biases in NLP strategies, using variously sized subsamples of the \textit{Bios} dataset \citep{de2019bias}.  We create 50 samples for each sample size (10k, 20k, 50k, and 120k) and train a model on each of the 200 samples. This gives us a mean and a variance on our results for all sample sizes to be able to compare them from a statistical point of view.
We study the biases in these samples  using three metrics; each sheds light on specific properties of gender bias. 

Our study shows how bias is related to the training set size on a standard NLP dataset by revealing three points:
First, commonly accepted bias indices appear unreliable when computed on ML models trained on relatively small training sets. Moreover, our experiments reveal that the group parity gender gap metric (\ref{fairnessmetrics:subset:metric}) appears to be more reliable than other metrics on small samples. 
Second, in the tested standard and large training sets, results are not homogeneous across professions and across the measures: sometimes there is gender bias against males, and sometimes against females in professions where one would expect something different.  Finally, the most traditional de-biasing methods, which consist in removing gender-sensitive words or replacing them with neutral variants, makes different metrics yield surprising and sometimes seemingly incompatible bias effects. We explain this phenomenon by the definitions of the metrics. In light of these findings, we think that one should use the main fairness metrics jointly to look for biases in smaller datasets and run enough models to have a variance. Such bootstrapping procedures appear essential to robustly analyze how fair a prediction model is.

Our chapter is structured as follows.  Section \ref{fairnessmetrics:sec:relatedwork} surveys related work. Section \ref{fairnessmetrics:sec:protocol} introduces our experimental setup.  Section \ref{fairnessmetrics:sec:results} discusses our results, with conclusions coming in Section \ref{fairnessmetrics:sec:conclusion}. Section \ref{fairnessmetrics:sec:limitations} discusses some of the limitations of our work.

\section{Related Work}\label{fairnessmetrics:sec:relatedwork}

We presented the state-of-the-art in mitigation bias in detail in Chapter \ref{SOTA}.  However, the work mentioned in that chapter focused only on single, large datasets. Recently, a growing literature has started to propose to leverage statistical properties of fairness metrics, thus providing both sophisticated analysis and practically useful algorithms~\citep{lum2022biasing, diciccio2020evaluating}. In particular, a more rigorous statistical approach for BERT models was introduced in~\citep{sellam2021multiberts}.

In this chapter, we investigate the pertinence of different fairness metrics on samplings of different sizes out of a large dataset. We apply our principled statistical procedure and we present the results of these measures, along with their standard deviation and properties coming from Student's t-tests. 
In addition to our scientific contribution, we have paid particular attention to the clarity of our explanations and the simplicity of our proposed protocol to allow small players to easily employ them for their real-world use cases.
Finally, our results attest to the importance of applying techniques of statistical analysis to Fairness problems, and we hope that the guarantees gained through them provide a convincing argument for its more generalized application in the field.

\section{Experimental protocol}\label{fairnessmetrics:sec:protocol}

In this section, we detail the various components of our experimental setup.  Section \ref{fairnessmetrics:subset:model} describes the general type of model used to train the 200 models on the \textit{Bios} dataset.  Section \ref{fairnessmetrics:subset:debiaising} introduces our debiasing technique used to illustrate our protocol. Section \ref{fairnessmetrics:subset:sampled} explains the sampling procedure and gives guarantees on the representativeness of the samples. Finally, Section \ref{fairnessmetrics:subset:metric} describes the different fairness metrics that we will compare and we justify these choices.

\subsection{DistilBERT model}\label{fairnessmetrics:subset:model}

The \textit{Bios} dataset \citep{de2019bias}  contains about 400K biographies, the gender of its author as well as its occupation (28). For more details on this dataset, please see section \ref{biosdataset}. 

Our task is to predict the occupation using only the textual data of the biography. This task is relevant in the case of our study because job prediction from LinkedIn biographies is used for job recommendation. It is therefore easy to imagine the consequences of gender discrimination in this context.

For this task, we will use the DistilBERT architecture.  DistilBERT \citep{sanh2019distilbert} is a Transformer architecture derivative from but smaller and faster than the original BERT \citep{devlin2018bert}. 
This model is commonly used to do text classification. DistilBERT is trained on BookCorpus \citep{moviebook} (like BERT), a dataset consisting of 11,038 unpublished books and English Wikipedia (excluding lists, tables and headers), using the BERT base model as a teacher. 

We have fine-tuned  DistilBERT  to adapt it to our text classification task. 
In our protocol, only the datasets were intervened on while keeping other factors the same in each model.
We used 5 epochs,  a batch size of 16 observations, an AdamW optimizer with a learning rate of 2e-5, and a cross-entropy loss when training the  model. 

\subsection{De-biasing methodology}\label{fairnessmetrics:subset:debiaising}
In this section, we describe the debiasing technique used to illustrate our experimental protocol. It is important to note that this technique is very basic and serves solely to demonstrate our protocol. The same protocol can be applied using any more advanced debiasing technique.

For this illustration, we remove explicit gender indicators such as \textit{'he'}, \textit{'she'}, \textit{'her'}, etc. The detailed methodology for this approach is explained in section \ref{subs:genderneutral}.


\subsection{Sampled training and test sets}\label{fairnessmetrics:subset:sampled}
We tested the robustness of our model with respect to the various bias measures on training sets of different sizes. We randomly sampled 50 different training sets containing 10K, 20K, 50K, and 120K biographies out of the 400K of \citep{de2019bias}.
We trained a model on each of these 200 samples. Each of these models has the same architectures and the same hyper-parameters stated previously. To guarantee the representativeness of the sample, we ensured that each sample had the same percentage of each gender for each occupation as in the initial data set.  For example, given 2002 female surgeons out of 388862 persons in the initial dataset (0.51 \%), we  randomly picked 51 women surgeons for a sample with 10000 individuals (0.51 \%). For the split between the train and test sets, we  respectively used 70\% and 30\% of the dataset. 

Creating these 200 different models and observations makes it possible to quantify the variability of the results obtained using each size of subsampled training sets. This will additionally allow us to ensure that all differences discussed in our results are statistically significant using Student's t-tests.
Our experimental protocol, therefore, gives us more guarantees than traditional protocols based on a single model.

\subsection{Gender bias metrics}\label{fairnessmetrics:subset:metric}

Let $\hat{Y}$ and $Y$ be the predicted and the true target labels (\textit{i.e.}, the occupations), respectively. Let $G$ be a random variable representing the binary gender of the biography’s subject. 

For each model, we quantified the gender bias by using the following metrics: 
Group Parity (GP), True Positive Rate (TPR), and Predictive parity (PP) presented on section \ref{intro:fairnessmetrics}. They are  defined as:

\begin{eqnarray}
  GP_{g,y} = P(\hat{Y} = y | G = g) \,,\\
  TPR_{g,y} = P(\hat{Y} = y | G = g, Y=y) \,,\\
  PP_{g,y} = P(Y = y | \hat{Y} = y, G = g) \,.
\end{eqnarray}

To measure the gender gap with these metrics, we computed the difference between binary genders $g$ and $\tilde g$ — for each occupation $y$:

\begin{equation}\nonumber
    M\_Gap_{g,y} = M_{g,y} - M_{\tilde g,y} \,,
\end{equation} 
where $M$ is $GP$, $TPR$ or $PP$.

We've chosen to look at these three metrics specifically because GP is the most classic and well-known, TPR is the most widely used in NLP and PP is similar to Calibration (within groups), and widely used in equity to compare with other metrics.




\section{Results and discussion}\label{fairnessmetrics:sec:results}

\begin{figure*}[!h] 
    \centering
    \includegraphics[width=1\linewidth]{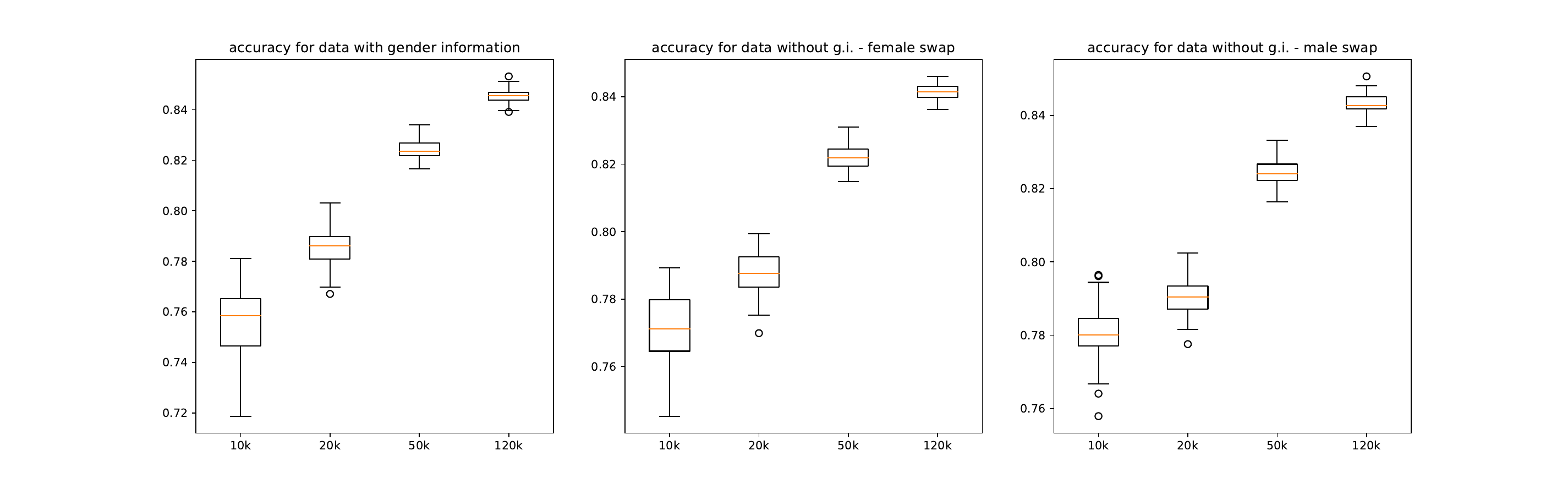}
    \caption{Boxplots representing the variations of prediction accuracy for all sampling sizes}
    \label{fairnessmetrics:fig:Results_accuracies}
\end{figure*}

As shown in Figure~\ref{fairnessmetrics:fig:Results_accuracies}, all the models we trained reached a prediction accuracy ranging from $0.72$ to $0.86$, as in \citep{de2019bias}, which we consider as good since the classification problem involved 28 different occupations.  

All comparisons in this part were considered as significant by using Student's t-tests (p-value of $0.05$).

We created two datasets without gender information: one version with all female gender indicators and the other with all male gender indicators. Gender, therefore, has no impact on the finetuning part of our model. However, since we are starting from a pre-trained DistilBERT model (without a gender-neutral dataset), we had to check that the pre-training had no impact on the prediction. We therefore also made a Student's test between the predictions of one model trained on the dataset with all the female gender indicators, and of another trained on the dataset with all the male gender indicators. The difference was not statistically significant; using one model or the other makes no difference.

The analysis of the results of our protocol is made in two steps: a specific part and a general part.
Below in Section \ref{fairnessmetrics:subset:specific}, we analyze biases on two specific occupations, {\em Surgeon} and {\em Physician}. These two occupations are socially very interesting and their male/female distribution is very different, which is something we wanted to study.  In Section \ref{fairnessmetrics:subset:general}, we also observe the biases across the gamut of occupations in {\em Bios}.  All the results found in this preliminary study remain valid in a generalized case where we look at all the classes of the model. \\
Dividing our study like this allows us to discuss various details which support our key message without weighing down the chapter in the specific part while guaranteeing that our analysis is global and applies to the other classes of the model in the general part.

\subsection{Results and discussion for the classes Surgeon and Physician} \label{fairnessmetrics:subset:specific}

\begin{figure*}[!h] 
    \centering
    \includegraphics[width=1\linewidth]{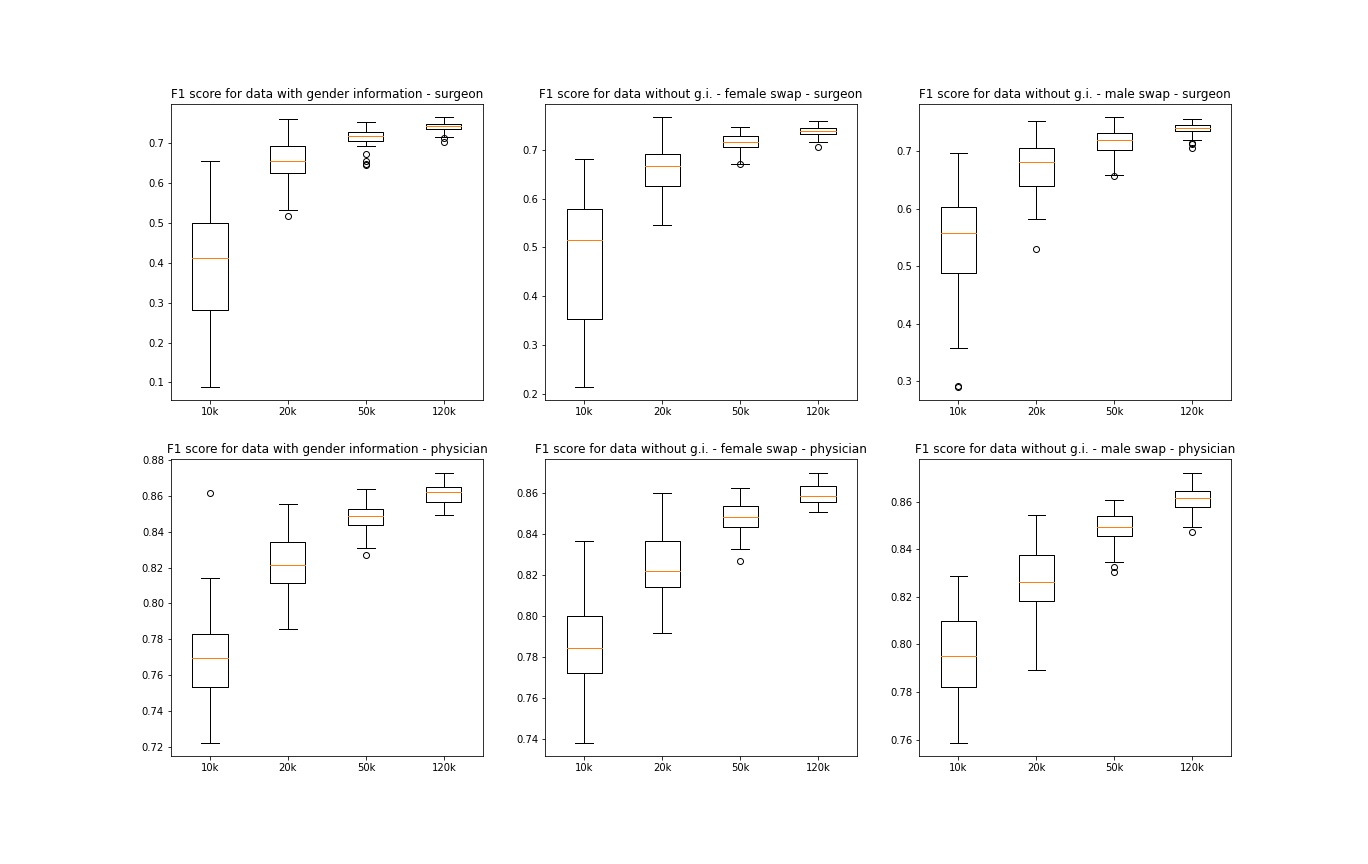}
\caption{Boxplots representing the variations of prediction F1-scores for all sampling sizes for surgeon (top) and physician (bottom)}
    \label{fairnessmetrics:fig:Results_F1score}
\end{figure*}

\begin{figure*}[!h] 
    \centering
        \includegraphics[width=1\linewidth]{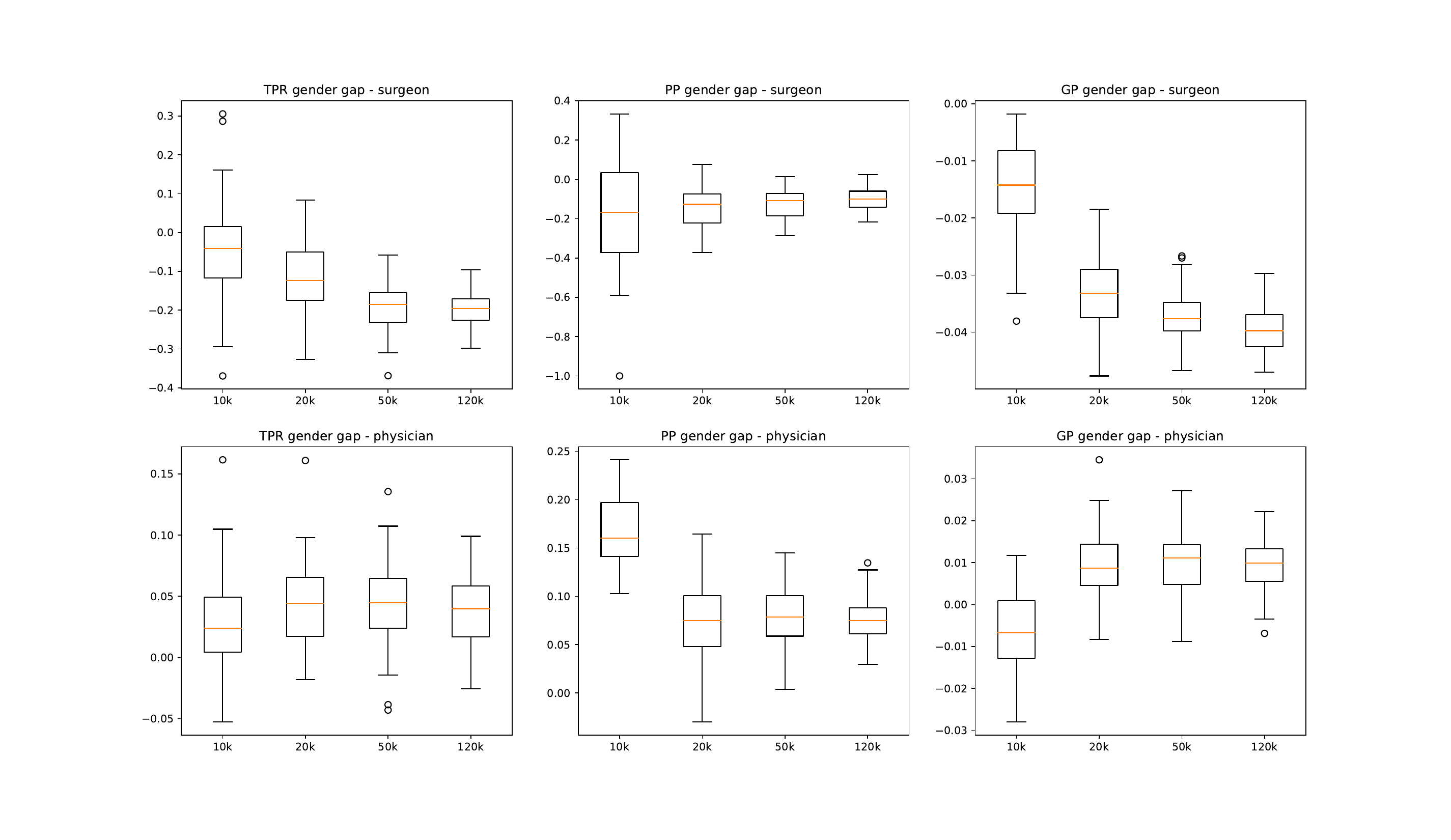}
\caption{Boxplots of the gender gaps obtained using  10K, 20K, 50K, and 120K randomly sampled observations (50). \textbf{(Left)} True Positive Rate (TPR) gender gaps for surgeons and physicians; \textbf{(Middle)} Predictive Parity (PP) gender gaps for surgeons and physicians; \textbf{(Right)} Group Parity (GP) gender gaps for surgeons and physicians.}
    \label{fairnessmetrics:fig:ResultsBERT}
\end{figure*}

Although the model is trained to predict  the occupations of bio authors from the 28 possible choices, we focus, in our study, on the analysis of the biases on two specific occupations:  \emph{Surgeon} versus the 27 remaining occupations, and  \emph{Physician} versus the other occupations. We chose these professions so that we could compare an occupation with an imbalanced gender distribution and one with balanced a gender distribution. The occupation \emph{Physician} is well balanced in the training set between males and females (49,5\% female), while the training set for \emph{Surgeon} contains many more males than females (15\% female). 

\begin{figure*}[!h] 
    \centering
    \includegraphics[width=1\linewidth]{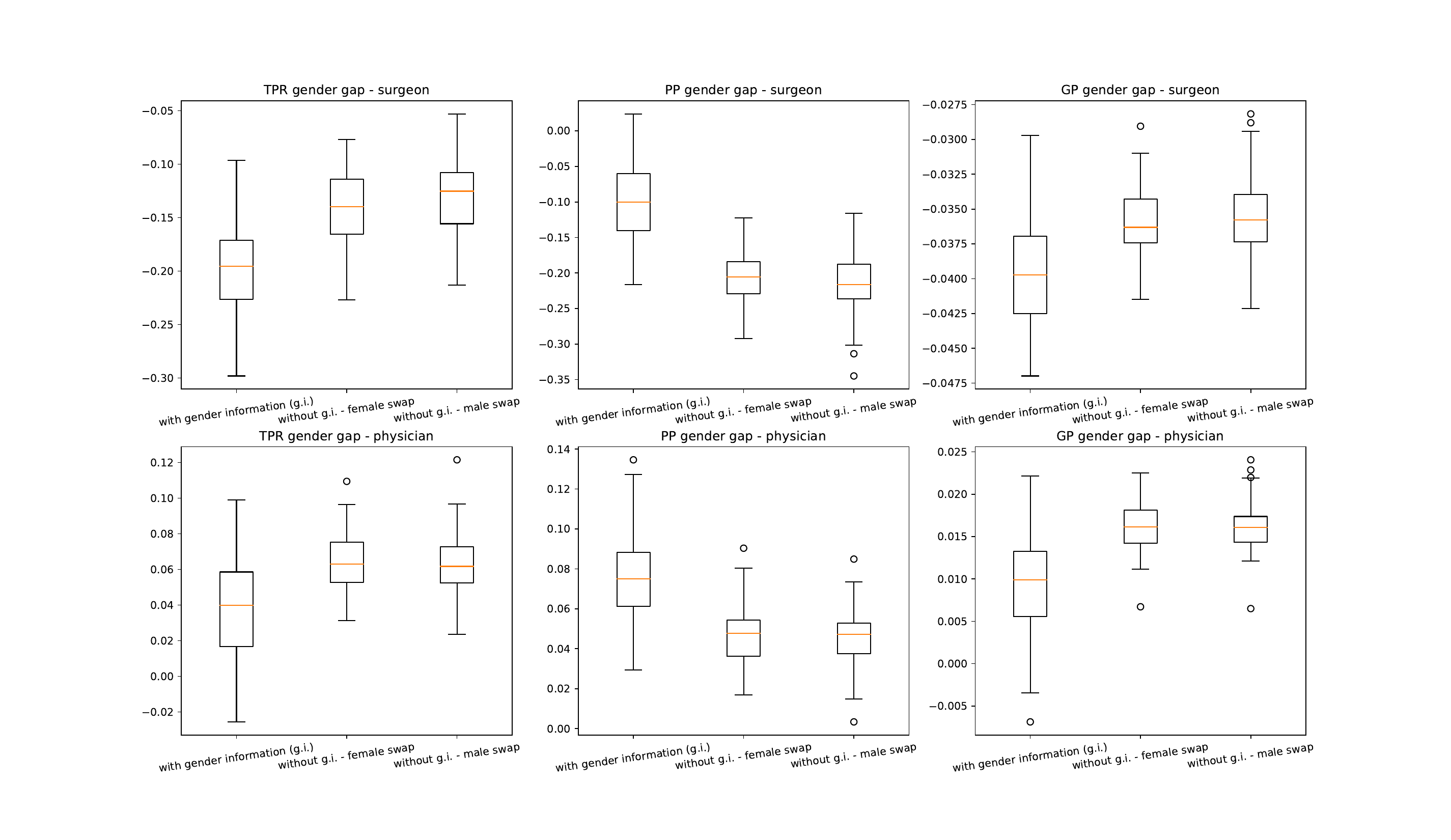}
    \caption{Boxplots representing the fairness metrics for surgeon (top) and physician (bottom) for 120k samplings for the base model, model with only female indicators, and model with only male indicators}
    \label{fairnessmetrics:fig:Results_models}
\end{figure*}

We computed F1-scores in Figure~\ref{fairnessmetrics:fig:Results_F1score}, which are good to reasonable, except for the 10K samples for surgeons, which appear as too small for our predictive task. Quantitative results related to the fairness metrics are shown in Figures \ref{fairnessmetrics:fig:ResultsBERT} and \ref{fairnessmetrics:fig:Results_models}. Each box-plot contains the TPR, GP, PP Gender Gaps obtained on the test set for \emph{surgeons} and \emph{physician}. Negative (positive) gender gaps mean that there is discrimination against females (males). 

\subsubsection{Results on small data samples} 

Our experiments clearly show that the lower the amount of observations in the training set, the more the fairness metrics vary in the test set. The samples with 10K and 20K observations present particularly unstable biases.

For example, most TPR (resp. GP) Gender Gaps are negative (resp. positive) for \emph{surgeon} (resp. \emph{physician}) but some samples yield positive TPR (resp. negative GP) Gender Gaps. This is problematic since we cannot deduce a priory that a particular sample should produce discrimination one way or the other.

In addition, the average biases also depend on the sample size. Again, we obtained unstable average biases for small samples (10K, 20K). 
The bias indicators are estimated on the minority class: an amount of 41, 115, 334, and 903 predicted surgeons were obtained in the test set for the 10K, 20K, 50K, and 120K sampling sizes. Hence, their estimation is unstable for small samples.

However, GP appears more stable than the other metrics in our experiments, in particular when there were few observations. Its variance was indeed close to $0.01$, which is much lower than the variances of $0.1$ and $0.2$ for GP and PP, respectively. We explain this because on our dataset, for TPR and PP, they do not use all predicted surgeons (unlike GP), but only the predicted surgeons who are also real surgeons (in 10k sampling, there are 41 predicted surgeons vs. 30 real surgeons and predicted surgeons, which is an information loss of 26,8\%). 
We model this intuition mathematically:

\paragraph{Intuition} Let $\hat{Y}$ and $Y$ be the predicted and the true target labels, respectively. Let $G$ be a random variable representing the binary gender and let $n$ the number of all individuals.
We can write the estimators of Group Parity, True Positive Rate and Predictive Parity metrics like this:
\begin{align*}
    \hat{GP}_{g,y} &= \frac{\sum_{i=1}^n1_{\{ \hat{Y}_i = y \ \cap \ G_i = g\}}}{\sum_{i=1}^n1_{\{G_i = g\}}} \\
    \hat{TPR}_{g,y} &= \frac{\sum_{i=1}^n1_{\{ \hat{Y}_i = y \ \cap \ Y_i=y  \ \cap \ G_i = g\}}}{\sum_{i=1}^n1_{\{Y_i=y \ \cap \ G_i = g\}}} \\
    \hat{PP}_{g,y} &= \frac{\sum_{i=1}^n1_{\{ \hat{Y}_i = y \ \cap \ Y_i=y  \ \cap \ G_i = g\}}}{\sum_{i=1}^n1_{\{\hat{Y}_i = y \ \cap \ G_i = g\}}}
\end{align*}

We set $A = \{ \hat{Y}_i = y \ \cap \ G_i = g\}$ and $B = \{ Y_i = y \}$. 
By definition, $\#(A \ \cap \ B) \leq \#A$ where $\#$ is the cardinal of the set. So we have $\#\{ \hat{Y}_i = y \ \cap \ Y_i = y \ \cap \ G_i = g\} \leq \# \{ \hat{Y}_i = y \ \cap \ G_i = g\}, \forall i=1,...,n$.

We can define $n_{GP}$, $n_{TPR}$, $n_{PP}$  the number of individuals respectively looked by the estimator of Group Parity, True Positive Rate and Predictive Parity metrics and we have: 
\begin{align*}
    n_{GP} &= \sum_{i=1}^n\# (\{ \hat{Y}_i = y \ \cap \ G_i = g\} \ \cap \ \{ G_i = g\} ) = \sum_{i=1}^n\#\{ \hat{Y}_i = y \ \cap \ G_i = g\}, \\
n_{TPR} &= \sum_{i=1}^n\# (\{ \hat{Y}_i = y \ \cap \ Y_i=y \ \cap \ G_i = g\} \ \cap \ \{ Y_i=y \ \cap \ G_i = g\} ) \\
&= \sum_{i=1}^n\#\{ \hat{Y}_i = y \ \cap \ Y_i=y \ \cap \ G_i = g\}, \\
n_{TPR} &= \sum_{i=1}^n\# (\{ \hat{Y}_i = y \ \cap \ Y_i=y \ \cap \ G_i = g\} \ \cap \ \{ \hat{Y}_i = y \ \cap \ G_i = g\} ) \\
&= \sum_{i=1}^n\#\{ \hat{Y}_i = y \ \cap \ Y_i=y \ \cap \ G_i = g\}.
\end{align*}
Then : $n_{TPR} = n_{PP} \leq n_{GP}$. \\


Considering this result, we recommend using a simpler indicator like GP gender gap for small-size sets.

\subsubsection{Bias analysis with different metrics}
\paragraph{General results}
Even for large samples with 120K observations, biases sometimes differed from what we expected.  For the occupation {\em surgeon} (15\% of females) the gender gap was negative for all metrics, which was expected. 
For {\em physician} (49,5\% of females), we also expected to have a negative or zero gender gap (see  \citep{bolukbasi2016man}). However, the gender gaps were  positive for all metrics, which means that the models discriminated against males. 
This example shows that intuitions of model-builders about biases are not always correct and this awareness should influence model construction and testing.

\paragraph{Results with debiasing}

Intuitively, removing explicit gender indicators should reduce the bias \citep{de2019bias}. As shown Figure~\ref{fairnessmetrics:fig:Results_models}, however, our experiments show that this is not necessarily the case. Using  TPR and GP Gender Gaps, we see a bias initially in favor of women (resp. men) and increases (resp. decreases) for the \textit{physician} (resp. \textit{surgeon}) class after debiasing. Removing gender indicators thus favored women in these two occupations. 

PP Gender Gap shows different effects for debiasing: For \textit{physician} (resp. \textit{surgeon}), the initial bias  in favor of women (resp. men) decreases (resp. increases) after debiasing. Removing gender indicators thus favored men in these two occupations.

To explain this phenomenon, we  can remark that removing gender indicators allowed us to predict more women than before in the two professions. The metrics interpret this differently. By definition, $PP_{f,y}= P\left(Y = y | \hat{Y} = y, G = f \right)$  decreases when the number of $\hat{Y}$ increases. In addition, \\
$TPR_{f,y}= P\left(\hat{Y} = y | Y = y, G = f\right)$  and $GP_{f,y} = P\left(\hat{Y} = y | G = f \right)$  increases when the number of $\hat{Y}$ increases. 

Using either GP/TPR gender gap or PP gender gap amounts to choosing between focusing on the number of people predicted in the discriminated group (parity) or  focusing on the people in the discriminated group who are well predicted (truth). This explains the different interpretations of these indicators.

\subsection{Results  and discussion for all classes}\label{fairnessmetrics:subset:general}

In this section, we confirm our analysis of the specific occupations of {\em Surgeon} and {\em Physician} from a global point of view on all the classes of the model.

\begin{figure*}[!h] 
    \centering
    \includegraphics[width=1\linewidth]{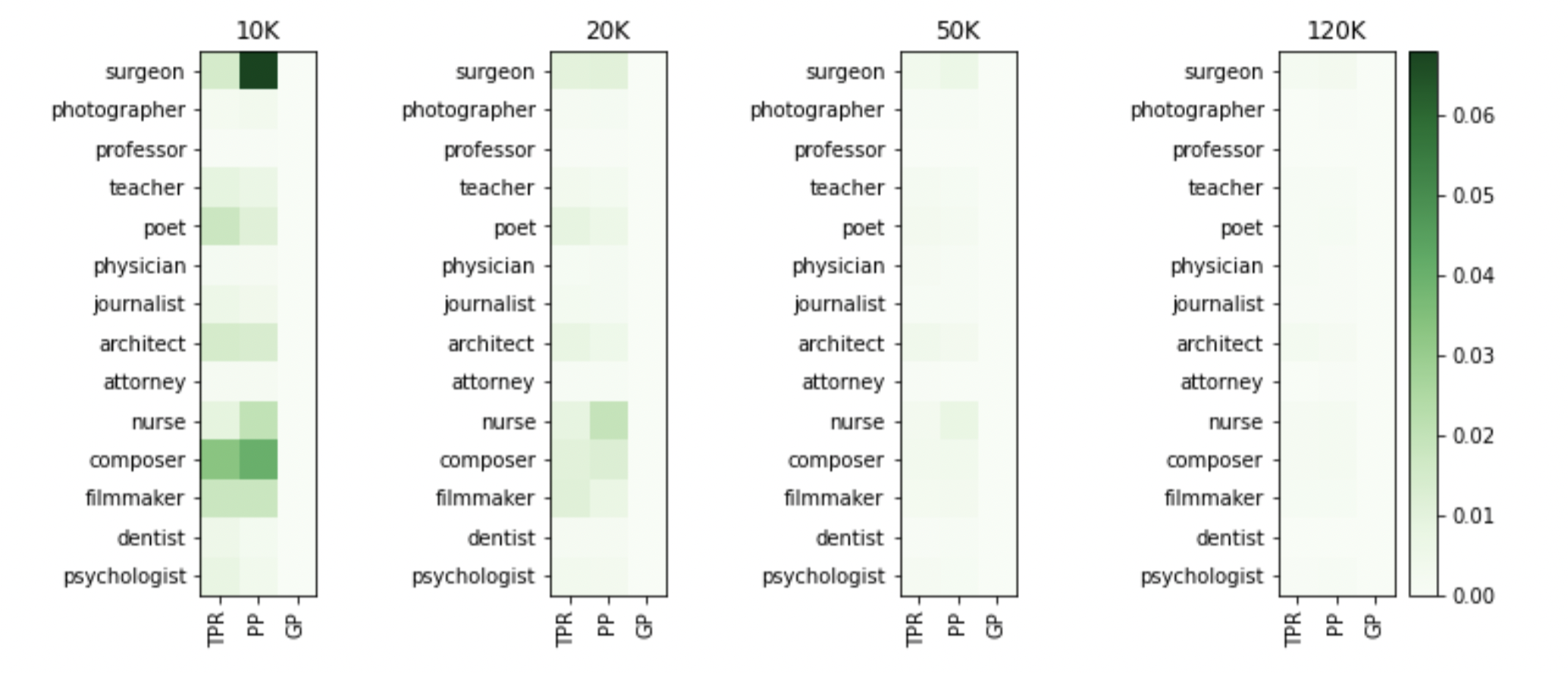}
    \caption{\textbf{Variance of TPR/PP/GP gender gap for all occupations} for model training on the classic dataset for all sample sizes. The higher the variance, the darker the green. We have 50 sampling for each sample size. We kept only professions that have at least one prediction per gender for all samplings. So we had to remove \textit{paralegal}, \textit{dj}, \textit{rapper}, \textit{pastor}, \textit{chiropractor}, \textit{software engineer}, \textit{attorney}, \textit{yoga teacher}, \textit{painter}, \textit{model}, \textit{personal trainer}, \textit{comedian}, \textit{accountant}, \textit{interior designer}, and \textit{dietitian}}
    \label{fairnessmetrics:fig:TPR_PP_GP_variance}
\end{figure*}

\begin{figure*}[!h] 
    \centering
    \includegraphics[width=1\linewidth]{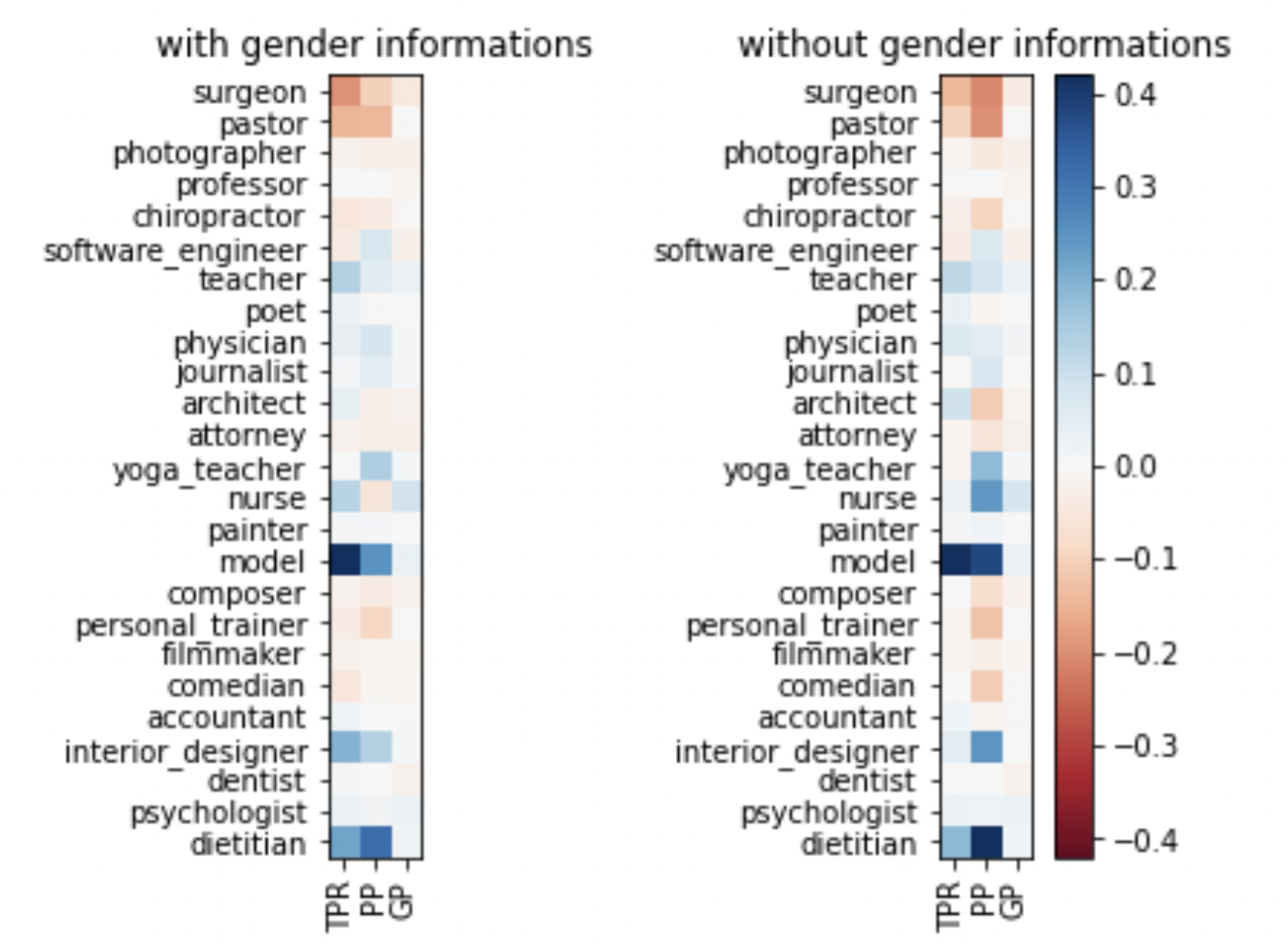}
    \caption{\textbf{Mean of TPR/PP/GP gender gap for all occupations} for model trained on 120K samplings. On the right, the model was trained on the classic dataset, and on the left, the model was trained on the dataset without gender indicators. The more it is red, the more it is biased in favor of males, the more it is blue, the more it is biased in favor of females.
    We kept only professions with more than 10 predictions per gender. So we had to remove \textit{paralegal}, \textit{dj} and \textit{rapper}.}
\label{fairnessmetrics:fig:TPR_PP_GP_all_occupations}
\end{figure*}

The general results on all occupations confirm the analysis we made on the two occupations previously: \begin{enumerate}
    \item In Figure \ref{fairnessmetrics:fig:TPR_PP_GP_variance}, we have more and more important deviations on the variance of the metrics as the size of the data set decreases. And that on most trades. As explained before, the GP gender gap is more stable, because it has more data. 
    \item In the first table of Figure \ref{fairnessmetrics:fig:TPR_PP_GP_all_occupations} (for the classic model), the metrics give inconsistent results for several occupations: depending on the metric bias in favor of men or women for the same profession and the same model. This is particularly visible for the occupations: \textit{software engineer}, \textit{poet}, \textit{architect}, \textit{attorney}, and \textit{nurse}.
    \item By comparing the two tables in Figure \ref{fairnessmetrics:fig:TPR_PP_GP_all_occupations}, we confirm that depending on the metric we are looking at, the basic debiasing technique used will not have the same effects on the bias. In several professions, we see that the bias on the TPR gender gap in favor/against women increases when on the bias on the PP gender gap decreases and vice versa. This is evident in the professions: \textit{surgeon}, \textit{pastor}, \textit{photographer}, \textit{chiropractor}, \textit{teacher}, \textit{journalist}, \textit{architect}, \textit{attorney}, \textit{nurse}, \textit{composer}, \textit{personal trainer}, \textit{comedian}, \textit{interior designer}, and \textit{dietetitan}.
\end{enumerate}

These results give us guarantees on the generalization of our analysis carried out on the two classes previously. We find the same problems with the metrics and the size of the sample, regardless of the occupation being looked at.

\section{Conclusion}\label{fairnessmetrics:sec:conclusion}
This chapter used the \textit{Bios} dataset to study the influence of the training set size on discriminatory biases. Our results shed light on new phenomena: (1) fairness metrics did not converge to stable results for small sample sizes, which precluded any conclusions about the nature of the biases;  (2) even on large training samples, the biases discovered were not always those expected and varied according to the metrics for several occupations; (3) a simple debiasing method, which consists in removing explicit gender indicators, had an unstable impact in our results depending on the metrics, though our analysis of the metrics can explain the instability. 
These results give two clear messages to data scientists who  must design NLP applications with a potential social impact. They should first be particularly careful, as the decision rules they train may have unexpected discriminatory biases.  In addition, a bias metric not only returns a score but has a strong practical meaning and may be unreliable, in particular when working with small training sets.  So multiple metrics should be considered and statistical methods to obtain the variance of the observed metrics are necessary to guarantee the fairness of a model.

\section{Limitations}\label{fairnessmetrics:sec:limitations}

A limitation of our conclusions is that although it is necessary to use several fairness metrics to be able to properly quantify the bias, this is not enough. These metrics must be well chosen according to the context and the task being looked at. The expertise of a person working in the field is therefore always necessary to have the most complete possible interpretation of the bias. 
More specifically, the different fairness metrics measure distinct properties, and the fact that they are often incompatible, has been a core part of the fair ML conversation from the beginning \citep{barocas2017fairness}. 
Thus, suggesting to choose a different metric depending on the sample size may sometimes be inappropriate, since this choice may depend on the meaning of the metric in a given application.
We must therefore be very careful and see the notion of robustness as additional necessary information and not as a replacement for the metric's meaning.

We also did not reduce the bias using advanced strategies because this paper focuses more on the analysis intended for a population closer to the law than to machine learning. In this vein, it is interesting to note that more and more tools are available to reduce the bias. In particular, \cite{sikdar2022getfair}  makes it possible to reduce the bias according to several fairness metrics, therefore to remain in our logic of taking several metrics.

The main problem raised by our article comes from the fact that fairness indices are not stable when they are calculated. We should consider them as random variables and look at their law. The first step is to look at the mean and the variance as done in this paper but having the full distribution would be more interesting. Works that compute the asymptotic law can be taken as an example like \cite{ji2020can, besse2022survey}.

%% file: Explainability.tex
\chapter{Explainability}\label{Cockatiel}

In the previous chapter, we examined the shortcomings of conventional fairness metrics, prompting a shift in focus towards a more nuanced perspective: explainability.
This transition to explainability addresses the crucial need for understanding how AI models make decisions, particularly in complex domains like NLP, to better understand their biases. By shifting our focus to explainability, we aim to unravel the 'black box' of AI, providing insights into the reasoning behind model predictions. This is especially important in high-stakes applications like healthcare, recruitment, or legal matters, where understanding the basis of a decision is as critical as the decision itself.

This chapter presents the work carried out as part of my own doctoral research, encapsulated in the paper "\textit{COCKATIEL: COntinuous Concept ranKed ATtribution with Interpretable ELements for explaining neural net classifiers on NLP tasks.}" \cite{jourdan2023cockatiel}, accepted for publication at Findings of ACL 2023.
I authored this paper as the primary author, sharing equal contribution with Agustin Picard. We divided the writing responsibilities, with my focus being on the methodology and results sections, and collaboratively conducted various experiments to validate our findings. Thomas Fel also provided significant support in the experimental part of the paper, contributing his valuable expertise in explainability within the field of vision. 


The paper, a cornerstone of my doctoral work and now a pivotal part of this thesis, presents COCKATIEL, a novel, model-agnostic XAI approach, introduced as a response to the limitations of traditional interpretability tools like attention maps and attribution methods, whose shortcomings are well-documented in the literature (e.g., \cite{pruthi2019learning, brunner2019identifiability}). This post-hoc, concept-based technique is adept at extracting meaningful explanations from the final layer of neural networks trained for NLP classification tasks. It leverages Non-Negative Matrix Factorization (NMF) to uncover the concepts that these models utilize for making predictions and employs Sensitivity Analysis for an accurate assessment of the relevance of these concepts.



We conduct experiments in single and multi-aspect sentiment analysis tasks and we show COCKATIEL's superior ability to discover concepts that align with humans' on Transformer models without any supervision, we objectively verify the faithfulness of its explanations through fidelity metrics, and we showcase its ability to provide meaningful explanations in two different datasets.

\section{Introduction}

\begin{figure}[h]
    \centering
\includegraphics[width=1\linewidth]{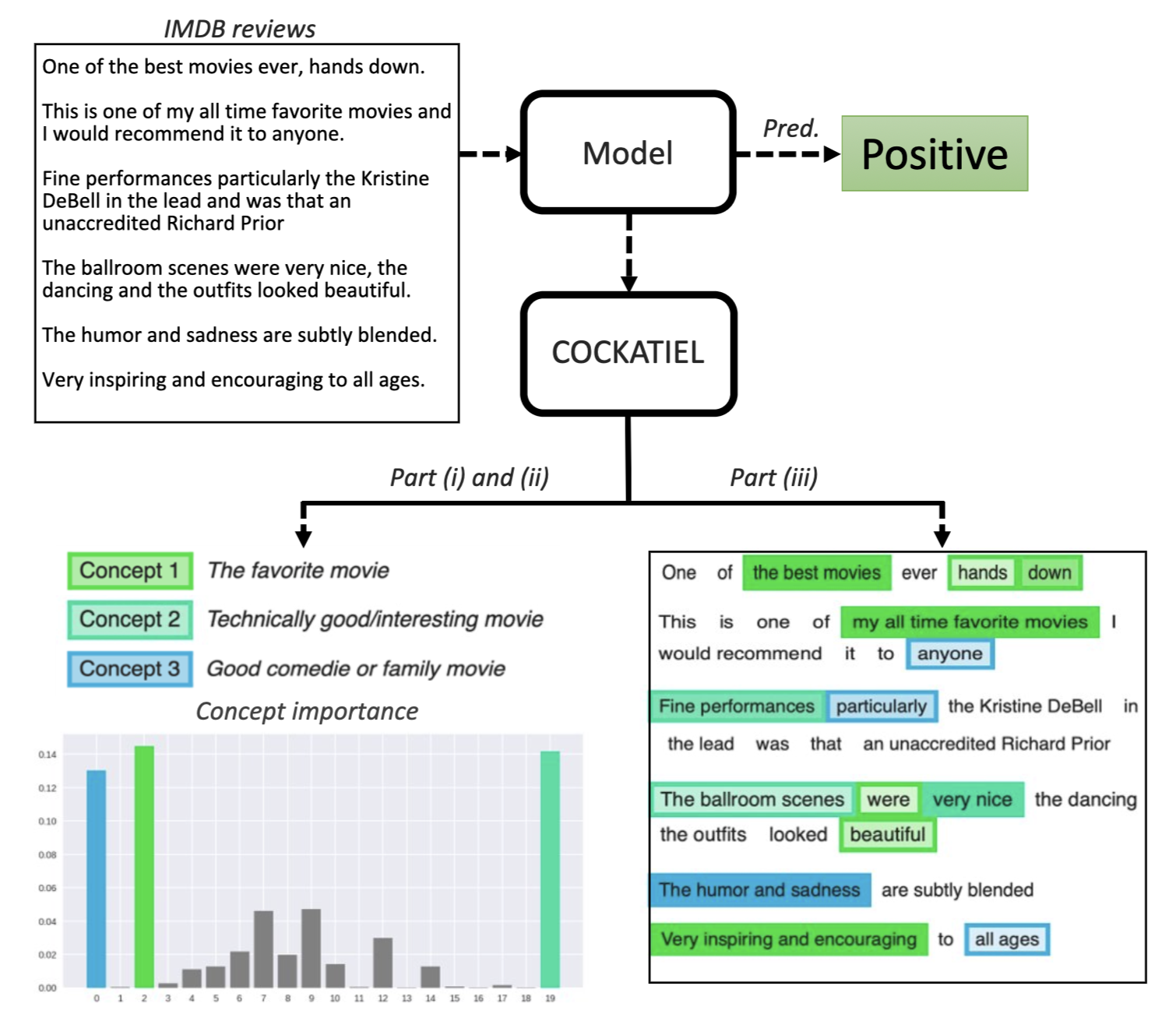}
\caption{\textbf{An illustration of COCKATIEL}. Given some sentences of IMDB reviews, COCKATIEL \textit{(i)} identifies concepts for prediction, \textit{(ii)} ranks them, and \textit{(iii)} gives the most important elements for each concept (to help us interpret the concept).}
\label{cockatiel:fig:visu_positivelabel}
\end{figure}

NLP models have undeniably gotten increasingly more complex since the introduction of the Transformer architectures~\citep{vaswani2017attention,devlin2018bert,liu2019roberta}. This trend, which is also occurring in the domain of Computer Vision, has brought about a need for understanding how these models make their predictions. The presence of bias in these models could indeed be prejudicial in applications where the user's lives are at stake~\citep{de2019bias}. 
Humans should be able to comprehend the reasons behind the model's decisions if these models are to gain general acceptance. Also, companies need to ensure that they are deploying algorithms which are free of harmful biases and that the explanations that they are obligated to issue are easily understandable by employees and end-users alike~\citep{kop2021eu}.  Intelligibility by humans has then become a key topic in explainable AI systems. As AI systems become more sophisticated and are deployed in increasingly complex environments, the ability to provide clear and concise explanations of their decisions becomes more pressing. Researchers have proposed multiple solutions to address this challenge, presented on section \ref{sec:sota:xai}.

In this chapter, we present COCKATIEL, a novel technique for generating reliable and meaningful explanations for NLP neural architectures for classification problems.  It extends CRAFT~\citep{fel:etal:2023} and our contributions can be summarized as follows:

\begin{itemize}
    \item We introduce a post-hoc explainability technique that is applicable to any neural network architecture containing non-negative activation functions. The technique is capable of explaining predictions of individual instances as well as providing insights of the model's general behavior. 
    \item We measure COCKATIEL's ability to discover concepts that align with those that Humans would employ in a sentiment analysis application. 
    Although we did not train the model on data annotated with these human concepts, COCKATIEL's explanations find them with high accuracy.
    \item  We demonstrate that in addition to generating meaningful concepts for Humans, these explanations are faithful to the models: An explanation $X$ provided by method $C$ is faithful to a model $M$ just in case if $X$ is returned as a putative explanation of $M$'s behavior by $C$, the $X$ plays a causal role in $M$'s behavior. 
    \item We provide examples of explanations on fine-tuned RoBERTa models~\citep{liu2019roberta} and bidirectional LSTMs trained from scratch to show how the concept decomposition can be used to understand the inner workings of complex models.
\end{itemize}

\section{Related Work}

We summarize here the points explained in detail in chapter \ref{SOTA} that are useful for understanding the chapter's results. 

\subsection{Explaining through rationalization}

Finding rationales in text involves identifying key justifications for a specific claim or decision. \citet{lei2016rationalizing} described rationales as minimal text segments essential for supporting a claim, requiring interpretability and similar prediction capacity as the original text. This process uses a generator to find excerpts and an encoder for predictions, but relies on reinforcement learning \citep{reinforce} for optimization. \citet{bastings2019interpretable} improved this by introducing a reparametrization trick for better gradient estimations and a sparsity constraint for minimal excerpt retrieval. Game-theoretic approaches by \citet{yu2019rethinking} and \citet{paranjape2020information} added complexity, while \citet{jain2020learning} introduced a simpler model for importance scoring. 
All these rationales will serve as an explanation for single instances, but won't explain how models predict whole classes. \citet{chang2019game} introduced a rationalization technique that allows for the retrieval of rationales for factual and counterfactual scenarios using three players.

However, all these techniques are not model agnostic and require specific architectures, in particular rather simple architectures or LSTMs, and training procedures.  But these architectures have been shown to not produce optimal results.

\subsection{Concept-based explanations}

Concept-based explainability is a growing area of research in AI, focused on generating human-understandable explanations for the decisions made by machine learning models. Initially, methods like TCAV~\citep{kim2018interpretability} utilized gradient-based techniques for feature identification but required manual concept specification, making them time-consuming and less comprehensive. To enhance efficiency, approaches like ACE~\citep{ghorbani2019towards} automated the concept extraction process, albeit with limitations due to reliance on pre-defined clustering algorithms. Further advancements employed matrix factorization techniques, such as NMF~\citep{lee1999learning}, for interpretable factors identification, showing efficacy in image classification tasks~\citep{zhang2021invertible,fel:etal:2023}. In the NLP field, self-interpretable neural architectures~\citep{bouchacourt2019educe} have been developed, learning concepts autonomously but facing challenges in balancing concept relevance with prediction accuracy. Recent developments like ConRAT~\citep{antognini-faltings-2021-rationalization} integrate various constraints and concept pruning to improve both concept quality and model accuracy.

\begin{figure}[h]
    \centering
    \includegraphics[width=0.8\linewidth]{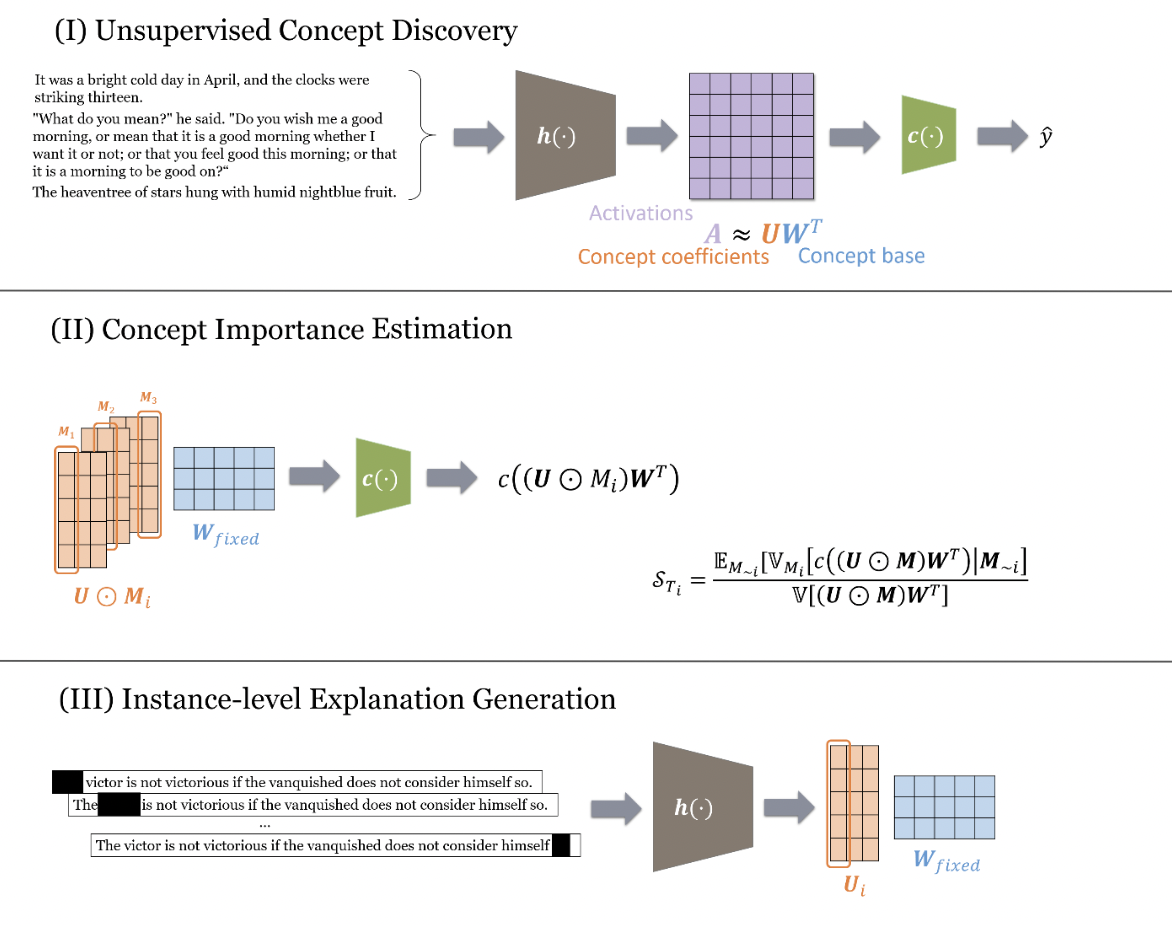}
    \caption{\textbf{Overview of our method:} COCKATIEL can be divided into three phases. \textit{(i)} The first step is assembling the concepts base. We propose to do this by constituting a database of whole or excerpts of input texts, projecting each one of these elements into the embedding of the model of our choice $h (\bm{x})$ and using the NMF algorithm to decompose the resulting non-negative matrix into two low-rank, non-negative matrices: $\vu$ and $\vw$. \textit{(ii)} Once $\vu$ and $\vw$ have been computed, we can compute the Total Sobol indices for the concept base's columns by masking the coefficients and by looking at their effect on the classifier's output: $c((\vu \odot \vm)\vw^T)$. \textit{(iii)} Finally, we propose to retrieve the influence of each word of the instance under study in each concept through Occlusion, that is, by applying masks to each word (or clause) in the input and quantifying the changes in each of the concept coefficients.}
    \label{cockatiel:fig:diagram}
\end{figure}

\section{COCKATIEL}\label{cockatiel:sec:cockatiel}

In this section, we describe COCKATIEL, our concept-based XAI technique for NLP models to generate human-understandable explanations.  It has three main components: \textit{(i)} it uses Non-Negative Matrix Factorization (NMF) to discover the concepts that the neural network under study leverages to make predictions; \textit{(ii)} it exploiting Sensitivity Analysis to estimate accurately the importance of each of these concepts for the model; and \textit{(iii)} it uses a black-box explainability technique to generate instance-wise explanations at a per-word and per-clause level.  Fig.~2 presents a schematic outline of COCKATIEL.

\paragraph{Notation}

In a supervised learning framework, we assume that a neural network model $\pred \colon \mathcal{X}^n \rightarrow \mathcal{Y}^n$ has already been trained for some classification task. We denote by $(\bm{x}_1, ..., \bm{x}_n) \in \mathcal{X}^n$ a set of $n$ input texts and $(y_1, ..., y_n) \in \mathcal{Y}^n$ their associated labels. 

We consider $\pred$ to be a composition of $h$, the last embedding of $\bm{x}$ (\textit{i.e.} the last layer of the feature extractor model), and $c$, the classification function, 
$\pred(\bm{x}) = c \circ h(\bm{x})$ with $ h(\bm{x}) \subseteq \mathbb{R}^p$ .

COCKATIEL will factorize $h$ through NMF, so we require $h$ to be non-negative -- i.e. $h(\bm{x}) \geq 0 \; \forall \bm{x} \in \mathcal{X}$. This constraint is typically verified when the last layer has an activation function such that $\bm{\sigma}(\bm{x}) \geq 0$, which is the case in (but it's not limited to) layers or blocks using \textit{ReLU}.

\subsection{Unsupervised concept discovery - "Concept part"} 

COCKATIEL discovers concepts without supervision by factorizing the neural network's intermediary activations by using a NMF algorithm.

Because we are factorizing $h$, we can generate explanations on embeddings without needing to deal with the complexities of attention layers~\citep{pruthi2019learning}; nor do we have to deal with the non-identifiability of Transformer models~\citep{brunner2019identifiability}. Thus, the concept extraction phase of our method does not depend on the specificities of attention. We will address this later on in Section~\ref{cockatiel:sec:local-explanations} to be able to generate our instance-level explanations.

\paragraph{NMF algorithm:}
We choose an excerpt-extraction function $\bm{\tau}_1$ to generate a database of excerpts coming from texts that the model places in the desired class $d_c$ -- i.e. $\bm{X}_i = \bm{\tau}_1 (x_i)$ such that $\pred(x_i) = d_c$. 

Then, we place ourselves at the model's last layer and we extract the activations $\bm{A} = h (\bm{X}_i)$ for each of the excerpts $\bm{X}_i$ in the database. With this information, we solve the constrained optimization problem engendered by the NMF algorithm:
\begin{equation}
    \label{cockatiel:eq:nmf}
    (\vu, \vw) = \argmin_{\vu \geq 0, \vw \geq 0} ~ \frac{1}{2}\|\va - \vu\vw^T\|^2_{F}, 
\end{equation}  
where $||\cdot||_F$ is the Frobenius norm.

This allows us to decompose the high-rank matrix containing all activations $\va \in \mathbb{R}^{n \times p}$ into two low-rank matrices $\vu \in \mathbb{R}^{n \times r}$ and $\vw \in \mathbb{R}^{p \times r}$. Intuitively, this corresponds to $\vw$ being a matrix whose columns represent the concepts that we will use to generate explanations, and $\vu$ is a matrix containing the coefficients quantifying the presence of each concept. These matrices are built so as to minimize the reconstruction error $\frac{1}{2}\|\va - \vu\vw\|^2_F$, enforcing the relevance of the concepts, and with a non-negative constraint for each matrix, thus encouraging sparsity in their elements.

It is important to note that these coefficients $u_{ij} \in \mathbb{R}_+$, so the presence of a concept can be determined by where its value stands in the concept's coefficients distribution. In practice, we have found that fixing a threshold at the quantile representing the 10\% highest values leads to accurate and easy to interpret explanations.

\paragraph{Choice of $\bm{\tau}_1$:} 
As we want the concepts to be descriptive enough to convey an abstraction but short enough to only contain one, we work with excerpts chosen by an excerpt-extraction function $\bm{\tau}_1$. The choice of $\bm{\tau}_1$, which should depend on the dataset and the text's format,  heavily impacts the type of explanations that we are able to generate.

We have identified 3 possible $\bm{\tau}_1$ functions: \textit{(i)} take all the full text ; \textit{(ii)} split the text into sentences (of at least 6 words) ; \textit{(iii)} split the text into clauses. Linguistically, it doesn't make sense to take smaller tokens like one or two words since their meaning is typically too unfocused to provide a real explanation.

We therefore chose $\bm{\tau}_1$ to respond specifically to each use-case. If we want to capture the mood of whole inputs, we can designate the inputs as the excerpts, and then interpret them by leveraging the local part of our method. If we instead wish to extract more simple but structured concepts, we can choose $\bm{\tau}_1$ to pick sentences of at least 6 words and ending in a full-stop. The first condition is necessary in the case of the \textit{beer review} dataset, which is composed of short sentences containing very simple descriptions.  For this dataset, using only very short excerpts would fail to convey the complexity of the ideas conveyed by the concepts. In this chapter, we present results using these two excerpt-extraction functions. 

\begin{figure}[h]
    \centering
\includegraphics[width=1\linewidth]
{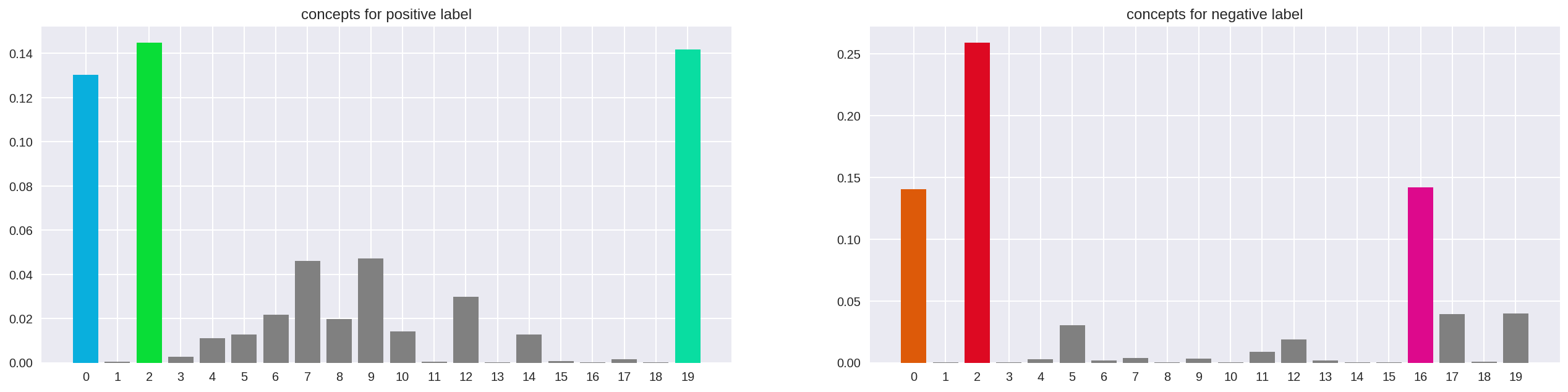}
    \caption{\textbf{Concept importance: } The global influence of the NMF concepts on the predictions on RoBERTa model is  measured using Sobol indices. There are different concepts for each class (positive and negative label).}
    \label{cockatiel:fig:importanceconcept_roberta}
\end{figure}

\begin{figure}[t]
    \centering
\includegraphics[width=1\linewidth]
{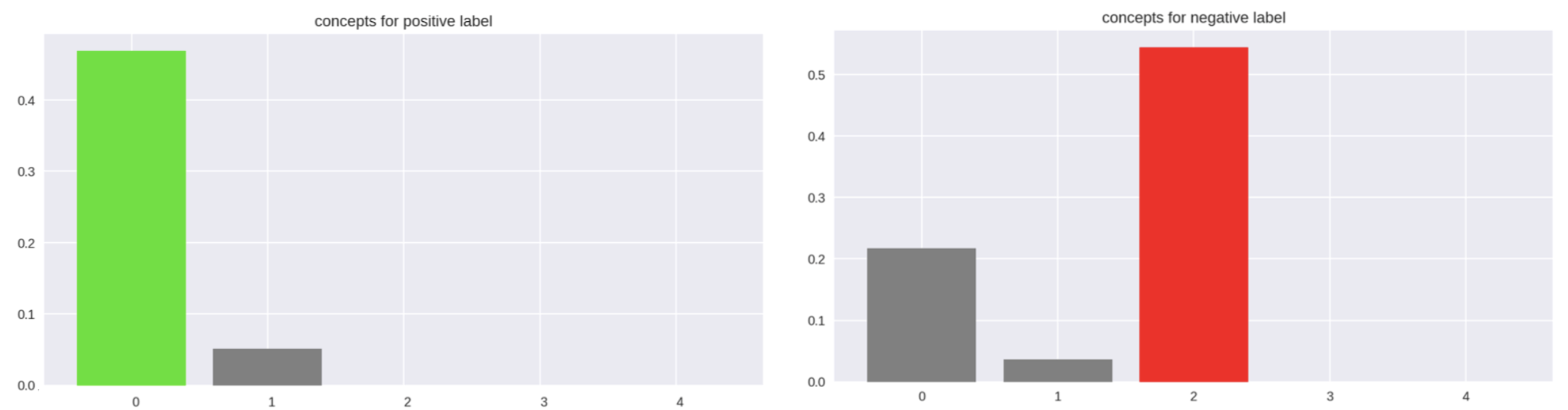}
    \caption{\textbf{Concept importance: } The global influence of the NMF concepts on the predictions on LSTM Model is then measured using Sobol indices. We have different concepts for each class (positive and negative label).}
\label{cockatiel:fig:importanceconcept_LSTM}
\end{figure}

\subsection{Concept importance estimation - "Ranking part"}

A common issue when utilizing concept extraction methods is the discrepancy between concepts deemed relevant by humans and those utilized by the model for classification. To mitigate the potential for confirmation bias during the concept analysis phase, we estimate the overall importance of the extracted concepts. To determine which concept has the most significant impact on the model output, we use a counterfactual reasoning \citep{peters2017elements, pearl2016causal}, and then use sensitivity analysis \citep{cukier1973study, iooss2015review}. A classic strategy in this area is the use of total Sobol indices \citep{sobol1993sensitivity}.
This method captures the importance of a concept, along with its interactions with other concepts, on the model output by calculating the expected variance that would remain if all the indices of the masks except $\vm_i$ were fixed. 

To estimate the importance of a concept $\vu_i$ for each fixed class, we measure the fluctuations of the model output $c(\vu \vw^T)$ in response to perturbations of the concept coefficient $\vu_i$. Specifically, we use a sequence of random variables $\vm$ to introduce concept fluctuations and reconstruct a perturbed activation $\tilde{\va} = (\vu \odot \vm)\vw^T$. We then propagate this perturbed activation to the model output $\vy = c(\tilde{\va})$. An important concept will have a large variance in the model output, while an unused concept will barely change it. 

\paragraph{Sobol indices details}
Let $(\Omega, \mathcal{A}, \mathbb{P})$ be a probability space of possible concept perturbations. To build these concept perturbations, we use $\boldsymbol{M}=\left(M_1, \ldots, M_r\right) \in \mathcal{M} \subseteq [0,1]^r$, i.i.d. stochastic masks on the original vector of concept coefficients $\boldsymbol{U} \in \mathbb{R}^r$. We define concept perturbation $\tilde{\boldsymbol{U}}=\boldsymbol{\pi}(\boldsymbol{U}, \boldsymbol{M})$ with the perturbation operator $\pi(\boldsymbol{U}, \boldsymbol{M})=\boldsymbol{U} \odot \boldsymbol{M}+(\mathbf{1}-\boldsymbol{M}) \mu$ with $\odot$ the Hadamard product and $\mu \in \mathbb{R}$ a baseline value, here zero. 

We denote the set $\mathcal{U}=\{1, \ldots, r\}$, $\boldsymbol{u}$ a subset of $\mathcal{U}$, its complementary $\sim \boldsymbol{u}$ and $\mathbb{E}(\cdot)$ the expectation over the perturbation space. We define $\boldsymbol{c}: \mathcal{A} \rightarrow \mathbb{R}$, the classification function, which takes the $\boldsymbol{U}$ perturbations of the last layer and returns the logit of the observed class, and we assume that $\boldsymbol{c} \in \mathbb{L}^2(\mathcal{A}, \mathbb{P})$ i.e. $|\mathbb{E}(\boldsymbol{c}(\boldsymbol{U}))|<+\infty$.

The Hoeffding decomposition gives $\boldsymbol{c}$ in function of summands of increasing dimension, denoting $\boldsymbol{c}_{\boldsymbol{u}}$ the partial contribution of the concepts $\boldsymbol{U}_{\boldsymbol{u}}=\left(U_i\right)_{i \in \boldsymbol{u}}$ to the score $\boldsymbol{c}(\boldsymbol{U})$ :

\begin{equation}
\label{cockatiel:eq:hoeffding}
\begin{aligned}
\boldsymbol{c}(\boldsymbol{U}) & =\boldsymbol{c}_{\varnothing}  +\sum_i^r \boldsymbol{c}_i\left(U_i\right) +\sum_{1 \leqslant i<j \leqslant r} \boldsymbol{c}_{i, j}\left(U_i, U_j\right) \\
& +\cdots +\boldsymbol{c}_{1, \ldots, r}\left(U_1, \ldots, U_r\right) \\
& =\sum_{\boldsymbol{u} \subseteq \mathcal{U}} \boldsymbol{c}_{\boldsymbol{u}}\left(\boldsymbol{U}_{\boldsymbol{u}}\right)
\end{aligned}
\end{equation}

Eq.~\ref{cockatiel:eq:hoeffding} consists of $2^r$ terms and is unique under the orthogonality constraint:
$$
\quad \mathbb{E}\left(\boldsymbol{c}_{\boldsymbol{u}}\left(\boldsymbol{U}_{\boldsymbol{u}}\right) \boldsymbol{c}_{\boldsymbol{v}}\left(\boldsymbol{U}_{\boldsymbol{v}}\right)\right)=0,  \forall(\boldsymbol{u}, \boldsymbol{v}) \subseteq \mathcal{U}^2 \text { s.t. } \boldsymbol{u} \neq \boldsymbol{v}
$$

Moreover, thanks to orthogonality, we have: $$\boldsymbol{c}_{\boldsymbol{u}}\left(\boldsymbol{U}_{\boldsymbol{u}}\right)=\mathbb{E}\left(\boldsymbol{c}(\boldsymbol{U}) \mid \boldsymbol{U}_{\boldsymbol{u}}\right)-\sum_{\boldsymbol{v} \subset \boldsymbol{u}} \boldsymbol{c}_{\boldsymbol{v}}\left(\boldsymbol{U}_{\boldsymbol{v}}\right),$$ and we can write model variance as:

\begin{equation}
\label{cockatiel:eq:var}
    \begin{aligned}
\mathbb{V}(\boldsymbol{c}(\boldsymbol{U})) & =\sum_i^r \mathbb{V}\left(\boldsymbol{c}_i\left(U_i\right)\right) 
+\sum_{1 \leqslant i<j \leqslant r} \mathbb{V}\left(\boldsymbol{c}_{i, j}\left(U_i, U_j\right)\right) \\
& +\ldots+\mathbb{V}\left(\boldsymbol{c}_{1, \ldots, r}\left(U_1, \ldots, U_r\right)\right) \\
& =\sum_{\boldsymbol{u} \subseteq \mathcal{U}} \mathbb{V}\left(\boldsymbol{c}_{\boldsymbol{u}}\left(\boldsymbol{U}_{\boldsymbol{u}}\right)\right)
\end{aligned}
\end{equation}

Eq.~\ref{cockatiel:eq:var} allows us to write the influence of any subset of concepts $\boldsymbol{u}$ as its own variance. This yields, after normalization by $\mathbb{V}(\boldsymbol{c}(\boldsymbol{U}))$, the general definition of Sobol' indices.

\begin{definition}[\textbf{Sobol indices}]
\textit{The sensitivity index $\mathcal{S}_{\boldsymbol{u}}$ which measures the contribution of the concept set $\boldsymbol{U}_{\boldsymbol{u}}$ to the model response $\boldsymbol{c}(\boldsymbol{U})$ in terms of fluctuation is given by:}
\begin{equation}
\label{eq:sobol}
    \begin{aligned}
\mathcal{S}_{\boldsymbol{u}} & =\frac{\mathbb{V}\left(\boldsymbol{c}_{\boldsymbol{u}}\left(\boldsymbol{U}_{\boldsymbol{u}}\right)\right)}{\mathbb{V}(\boldsymbol{c}(\boldsymbol{U}))}  =\frac{\mathbb{V}\left(\mathbb{E}\left(\boldsymbol{c}(\boldsymbol{U}) \mid \boldsymbol{U}_{\boldsymbol{u}}\right)\right)-\sum_{\boldsymbol{v} \subset \boldsymbol{u}} \mathbb{V}\left(\mathbb{E}\left(\boldsymbol{c}(\boldsymbol{U}) \mid \boldsymbol{U}_{\boldsymbol{v}}\right)\right)}{\mathbb{V}(\boldsymbol{c}(\boldsymbol{U}))}
\end{aligned}
\end{equation}
\end{definition}

Sobol indices provide a numerical assessment of the importance of various subsets of concepts in relation to the model's decision-making process. Thus, we have: $\sum_{\boldsymbol{u} \subseteq \mathcal{U}} \mathcal{S}_{\boldsymbol{u}}=1$.

Additionally, the use of Sobol' indices allows for the efficient identification of higher-order interactions between features. Thus, we can define the Total Sobol indices as the sum of of all the Sobol indices containing the concept $i: \mathcal{S}_{T_i}=\sum_{\boldsymbol{u} \subseteq \mathcal{U}, i \in \boldsymbol{u}} \mathcal{S}_{\boldsymbol{u}}$. 
So we can write:

\begin{definition}[\textbf{Total Sobol indices}]
\textit{The total Sobol index $\mathcal{S}{T_i}$, which measures the contribution of a concept $\vu_i$ as well as its interactions of any order with any other concepts to the model output variance, is given by:}
\begin{align}
        \label{cockatiel:eq:total_sobol}
           \mathcal{S}_{T_i} 
           & = \frac{ \mathbb{E}_{\bm{M}_{\sim i}}( \Var_{M_i} ( \vy | \bm{M}_{\sim i} )) }{ \Var(\vy) } \\
           & = \frac{ \mathbb{E}_{\bm{M}_{\sim i}}( \Var_{M_i} ( c((\vu \odot \vm)\vw^T) | \bm{M}_{\sim i} )) }{ \Var( c((\vu \odot \vm)\vw^T)) }.
    \end{align}
\end{definition}

There are already a plethora of different techniques that allow us to compute this index efficiently~\citep{saltelli2010variance, marrel2009calculations, janon2014asymptotic, owen2013better, tarantola2006random}. But concretely, we estimate the total Sobol indices using the Jansen estimator \citep{janon2014asymptotic}, a widely recognized efficient method \citep{puy2022comprehensive}. 
The Jansen estimator is commonly utilized in conjunction with a Monte Carlo sampling strategy, but we improve over Monte Carlo by using a Quasi-Monte Carlo sampling strategy. This technique generates sample sequences with low discrepancy, resulting in a more rapid and stable convergence rate \citep{gerber2015integration}.

\subsection{Instance-level explanation generation - "Interpretable elements part"}\label{cockatiel:sec:local-explanations}

In this part, we interpret the concepts found previously. To do this, we find which words and clauses are associated with each concept.

We adapt Occlusion~\citep{zeiler2014visualizing}: a black-box attribution method that works by masking each word looking at the impact on the model output. In this case, to get an idea of the importance of each word for a given concept, we mask words in a sentence and measure the effect of the new sentence (without the words) on the concept. This operation can be performed at word or clause level -- i.e. mask words or whole clauses -- to obtain explanations that are more or less fine-grained depending on the application.

\paragraph{Motivations:}

This choice has been shown to perform particularly well on NLP models~\citep{fel2021sobol} and doesn't suffer from the inefficiency of having to sample a considerable amount of masks for each explanation. 
Indeed, in ~\citep{fel2021sobol}, they compared Occlusion to other explainability techniques that are commonly used in NLP, and they showed that it is more faithful to the model than Saliency~\citep{saliency}, Grad-Input, SmoothGrad~\citep{smilkov2017smoothgrad}, Integrated Gradients~\citep{IntegratedGradients}, and their own Sobol method on both LSTM and BERT models.

In addition, in the case of Transformer models, using a black-box method such as Occlusion avoids manipulating the attention layers between the input and the activation matrix $A$, where our concepts are located. In doing so, we avoid the non-identifiability problem of Transformer models \citep{pruthi2019learning}.

\paragraph{Application:}
Empirically, we perform the following operations:

For a sentence $X_i$, $A_i = h(X_i)$. We have a fixed $W$ calculate with the NMF and $W_k$, the $k$ concept of $W$. As before, we get the importance of the sentence $X_i$ for the concept $k$:
\begin{equation*}
    U_i^k = \argmin_{U \geq 0} ~ \frac{1}{2}\| A_i - U W_k^T\|^2_{F}.
\end{equation*}  
Then, we remove the element $j$ from the sentence $i$: $\Tilde{X}_{i-j}$ (i.e. we replace the (tokenized) feature by a zero). So we have $\Tilde{A}_{i-j} = h(\Tilde{X}_{i-j})$, and:
\begin{equation*}
    \Tilde{U}_{i-j}^k = \argmin_{U \geq 0} ~ \frac{1}{2}\| \Tilde{A}_{i-j} - U W_k^T\|^2_{F}, 
\end{equation*}  

So, $\phi(k,i,j)$ quantifies the influence of the element $j$ in the sentence $i$ for the concept $k$:
\begin{equation*}
   \phi(k,i,j) =  U_i^k - \Tilde{U}_{i-j}^k, 
\end{equation*}

\begin{figure}
    \centering
    \includegraphics[width=1\linewidth]
    {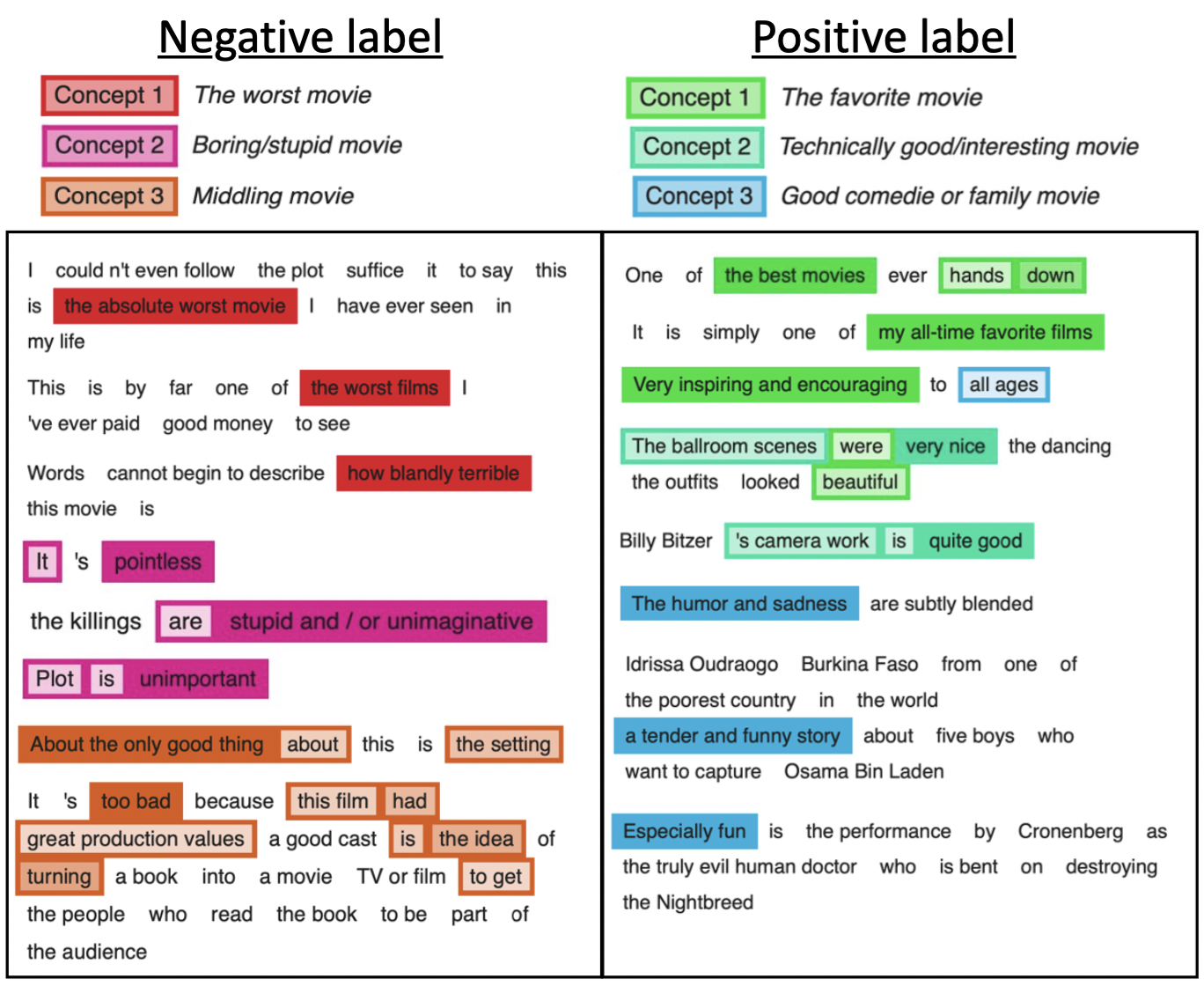}
    \caption{Concepts generated with $l = 20$ for a few sentences taken from IMDB reviews. The colored elements are those important for the concept of the corresponding color (calculated with part \textit{(iii)} of our method). The more colorful the element, the more important it is for the concept (continuously). We have selected the 3 most important concepts for each label (see Fig.~\ref{cockatiel:fig:importanceconcept_roberta}). The name of the concept is chosen manually in view of the important elements corresponding to the concepts. See more examples on Fig.\ref{cockatiel:fig:examplesannexe}.}
    \label{cockatiel:fig:examples}
\end{figure}

\begin{figure}
    \centering
\includegraphics[width=1\linewidth]
{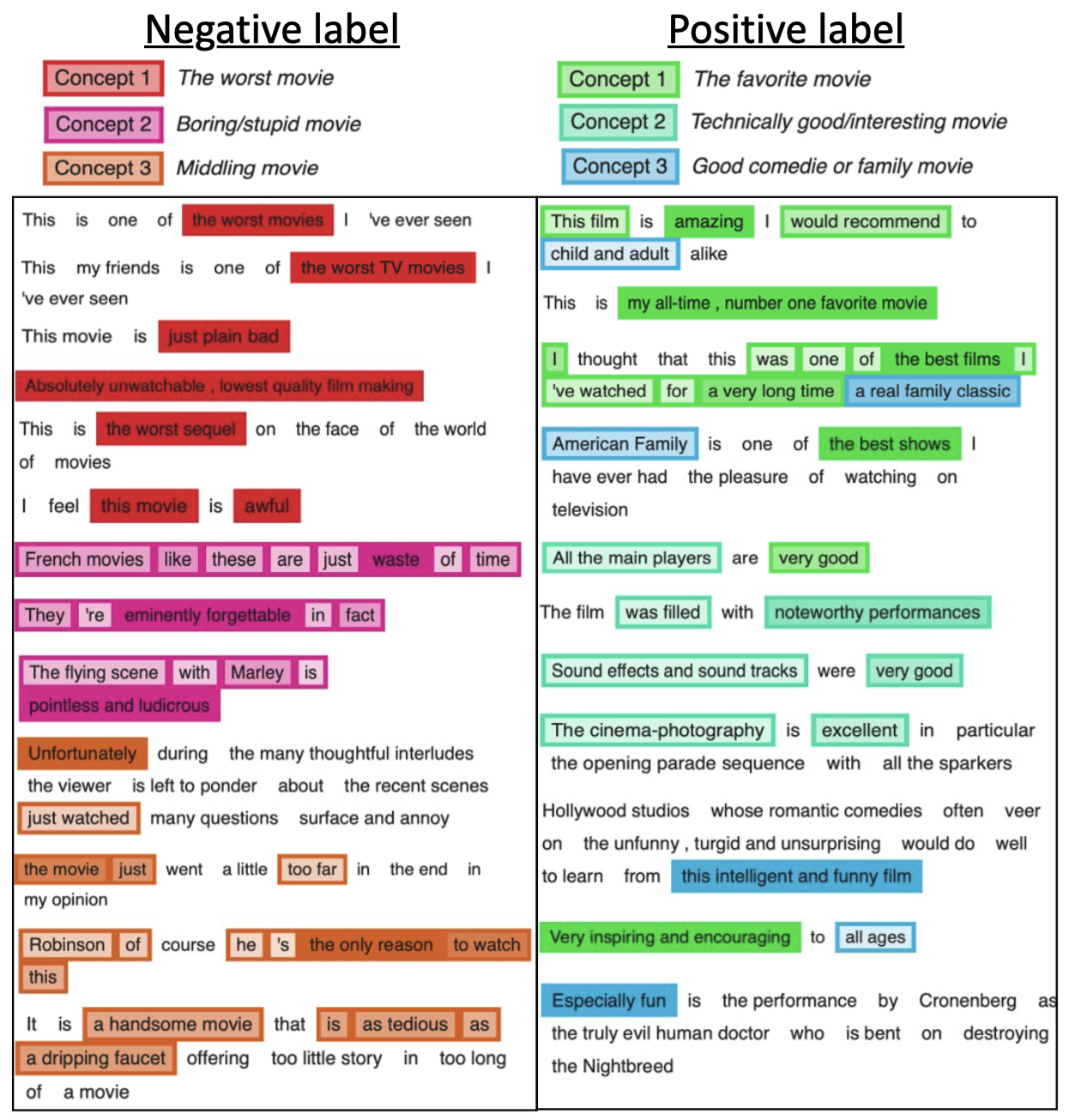}
    \caption{Concepts generated for a RoBERTa model with $l = 20$ for a few sentences taken out of IMDB reviews. The colored elements are those important for the concept of the corresponding color (calculated with part \textit{(iii)} of our method). The more colorful the element, the more important it is for the concept (continuously). We have selected the 3 most important concepts for each label (see Fig.~\ref{cockatiel:fig:importanceconcept_roberta}). The name of the concept is chosen manually in view of the important elements corresponding to the concepts.}
    \label{cockatiel:fig:examplesannexe}
\end{figure}

For the visualisations (see e.g. Fig.~\ref{cockatiel:fig:examples}), we color the element with the color of the concept for which it is most important. In addition, the darker the color, the more important the element is for the concept.

\begin{figure}
    \centering
\includegraphics[width=1\linewidth]
{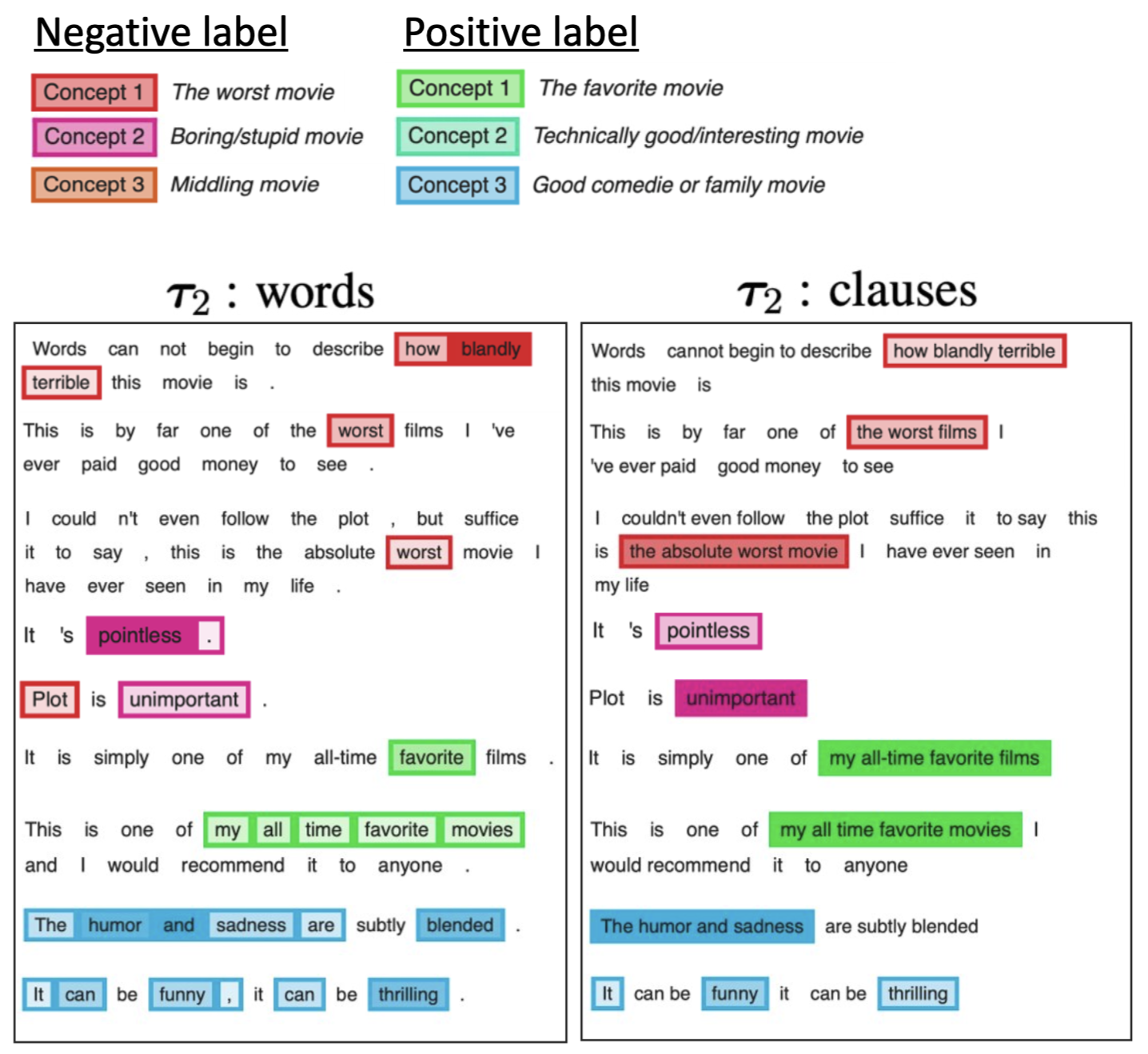}
    \caption{Concepts generated for a RoBERTa model with $l = 20$ for a few sentences taken out of IMDB reviews. 
    The excerpts chosen by an excerpt-extraction function $\bm{\tau}_1$ are sentences for both (so, we have same concepts). The colored elements are those that are considered to be the most important for the concept of the corresponding color (calculated with part \textit{(iii)} of our method). We compare the visualisations of the same sentences with two different excerpt-extraction functions $\bm{\tau}_2$:   words (on the left) and clauses (on the right). 
    We split the text into clauses for occlusion using the fair library’s SequenceTagger implementation.}
    \label{cockatiel:fig:visu_tau}
\end{figure}

\paragraph{Choice of $\bm{\tau}_2$:} Just like in the case of the NMF, the choice of the form of the elements of the input to occlude will have an impact on the understandability of the explanations. This can be generalized via another excerpt extraction function $\bm{\tau}_2$, whose optimal shape will depend on the dataset, the text's format and the learned concepts (i.e. Occlusion shouldn't be applied at a per-clause level if the concepts were learned using a $\bm{\tau}_1$ providing single words, so this first exceprt extraction function must be taken into consideration). There is a certain trade-off between the granularity and the interpretability of the explanations, as illustrated in Figure~\ref{cockatiel:fig:visu_tau} which contains some examples with different choices of $\bm{\tau}_2$. In general, we advise to try different combinations of $\bm{\tau}_i$ to find the desired level of granularity in the explanations for each use-case.

\section{Experimental evaluation}
For all of our results, we fine-tuned RoBERTa~\citep{liu2019roberta} based models on each dataset. We ensured the non-negativity of at least one layer of the model by adding a ReLU activation after the first layer of the 1-hidden-layer, dense MLP of the classification head. For the qualitative analysis, we  also tested COCKATIEL's performance on bidirectional LSTM models trained from scratch. 

We will first analyze the meaningfulness of the discovered concepts by measuring their alignment with human annotations on the different aspects of a multi or single-aspect sentiment analysis task. Then, we will ensure that our explanations are faithful to the model through an adaptation of the insertion and deletion metrics to concept-based XAI. Finally, we will showcase some examples of explanations and of applications for our method.

\subsection{Implementation details}
We trained 3 different models. For each model, we performed a single run and we split datasets in 70\% for train, 10\% for validation and 20\% for test. 

\subsubsection{Trained RoBERTa on Beer dataset}
We used a RoBERTa base pretrained on hugging face by \citep{liu2019roberta} (all the information on the pretrain can be found in the paper). 

We then trained the model on Beer dataset. The model was trained on 2 GPUs for 10 epochs with a batch size of 32 and a sequence length of 512. The optimizer was AdamW with a learning rate of 1e-5, $\beta_1 = 0.9$, $\beta_2 = 0.98$, and $\epsilon = 1e6$ 

\subsubsection{Trained RoBERTa on IMDB dataset}

We used a RoBERTa model already fine-tuned on IMDB from hugging face.
This model used the pre-training presented above, we fine-tuned it with 2 epochs, a batch size of 16, and an Adam optimizer with a learning rate of 2e-5, $\beta_1 = 0.9$, $\beta_2 = 0.999$ and $\epsilon = 1e-8$.

\subsubsection{Trained LSTM on IMDB dataset}

We created our LSTM with:
\begin{lstlisting}
SentimentRNN(
  (embedding): Embedding(1001, 512)
  (lstm): LSTM(512, 128, 
  num_layers=4, batch_first=True, 
  bidirectional=True)
  (dropout): Dropout(p=0.3, 
  inplace=False)
  (fc_1): Linear(in_features=128,
  out_features=128, bias=True)
  (relu): ReLU()
  (fc_2): Linear(in_features=128,
  out_features=2, bias=True)
  (sig): Softmax(dim=1))
\end{lstlisting}

Then, we trained it on the IMDB dataset. The model was trained on 2 GPUs for 5 epochs with a batch size of 128 and a sequence length of 512. The optimizer was Adam with a learning rate of 1e-4.

\subsection{Alignment with human concepts}

\begin{figure}[h]
    \centering
\includegraphics[width=1\textwidth]{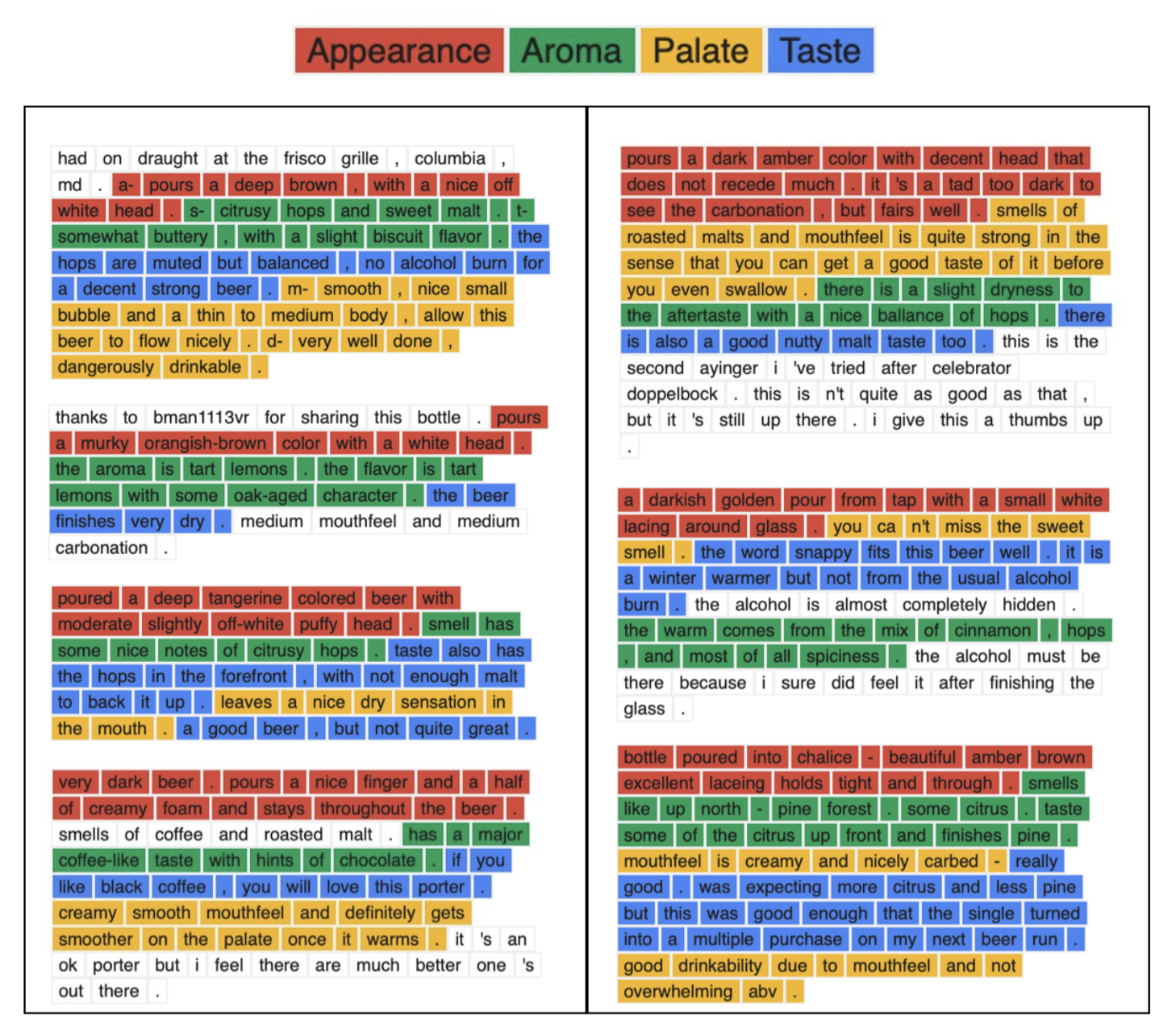}    
    \caption{Concepts generated with $l=20$ for a beer review. The colors depict the aspects for each annotate concept. COCKATIEL is trained only on the label and we use the NMF part of the method to find annotate concepts.} 
    \label{cockatiel:fig:beer_visu}
\end{figure}

Following the human-alignment evaluation in~\citep{antognini-faltings-2021-rationalization}, we perform beer task:

\paragraph{Beer Task}
We will measure the extent to which our concepts overlap the human annotations for the 4 different aspects of the multi-aspect \textit{beer reviews} dataset~\citep{beer-reviews}. This dataset contains reviews for beers with commentary and marks (from 0 to 5) on 5 different aspects: Appearance, Aroma, Palate, Taste and Overall. The model will be trained to predict whether the overall score is greater than 3 -- i.e. a positive review on the beer -- and will not have access to the labels for the other aspects. Additionally, it includes 994 reviews with annotations indicating the position of these aspects in the text. The objective of this evaluation is to look for concepts that align with these annotations and measure their capacity to predict the location of each different aspect. In particular, we searched across the whole annotated dataset for the concepts whose F1 score for the prediction of each aspect was maximal. It is important to note that this does not take into account to which extent they are important for the model to predict, but this only serves as an automatized test for determining whether the explainability technique is capable of generating understandable concepts.

We calculate the precision, recall and F1 scores for each aspect, and we do so with $l = 10$ and $l=20$ concepts. We remind the reader that, unlike the baselines, our method is a post-hoc technique, and thus, the model does not need to be re-trained, and that changing the number of concepts takes only a few minutes of compute on GPU.

In Table \ref{cockatiel:tab:annotation}, we present a comparison of our results to those obtained with some rationalization techniques: RNP~\citep{lei2016rationalizing}, RNP-3P~\citep{yu2019rethinking}, InvRAT~\citep{chang2020invariant} 
and ConRAT~\citep{antognini-faltings-2021-rationalization} for the task on \textit{Beer}. 
We demonstrate that not only our model achieves the highest accuracy, but also that it outperforms all the other methods in its ability to accurately recognize the human annotations, be it by its precision, recall or F1 score.

\begin{table}[]
    \centering
    \begin{tabular}{p{0.35cm}*{22}{@{\hskip.9mm}c@{\hskip.9mm}}}
& &  & \multicolumn{1}{c}{\textit{Average}} & \multicolumn{1}{c}{\textit{Appearance}} & \multicolumn{1}{c}{\textit{Aroma}} & \multicolumn{1}{c}{\textit{Palate}} & \multicolumn{1}{c}{\textit{Taste}} \\
\hline
& \textbf{Model} & \textbf{Acc.} & \textbf{F1} & \textbf{F1} & \textbf{F1} & \textbf{F1} & \textbf{F1} \\
\hline
\multirow{6}*{\rotatebox{90}{$l = 20$}}  & RNP & 81.1 & 24.9 & 26.5 & 21.5 & 20.4 & 20.9 \\
& RNP-3P & 80.5 & 23.3 & 27.8 & 19.8 & 11.1 & 34.5 \\
& Intro-3P & 85.6 & 19.1 & 26.6 & 14.3 & 17.9 & 17.4 \\
& InvRAT & 82.9 & 33.8 & 49.6 & 26.9 & 24.1 & 34.5 \\
& ConRAT & \underline{91.4} & \underline{40.9} & \underline{55.3} & \underline{33.6} & \underline{32.3} & \underline{42.4} \\
& \textbf{Ours} & \textbf{95.2} & \textbf{47} & \textbf{69.4} & \textbf{37.7} & \textbf{32.4} & \textbf{48.4} \\
\hline
\multirow{6}*{\rotatebox{90}{$l = 10$ }} & RNP & 84.4 & 19.5 & 18.5 & \underline{24.0} & \underline{20.6} & 15.07 \\
& RNP-3P & 83.1 & 17.8 & 26.3 & 15.9 & 12.7 & 16.1 \\
& Intro-3P & 80.9 & 16.1 & 34.4 & 12.8 & 12.9 & 4.1 \\
& InvRAT & 81.9 & 21.8 & \underline{36.3} & 20.8 & 12.1 & 17.9 \\
& ConRAT & \underline{91.3} & \underline{23.8} & 34.8 & 22.7 & 17.3 & \underline{20.3} \\
& \textbf{Ours} & \textbf{95.2} &\textbf{ 45.5} &  \textbf{59.7} & \textbf{38.9} & \textbf{32.5} & \textbf{50.9}\\
\hline
\end{tabular}

\vspace{5mm}

\begin{tabular}{p{0.35cm}*{12}{@{\hskip.9mm}c@{\hskip.9mm}}}
& & \multicolumn{2}{c}{\textit{Average}} & \multicolumn{2}{c}{\textit{Appearance}} & \multicolumn{2}{c}{\textit{Aroma}} & \multicolumn{2}{c}{\textit{Palate}} & \multicolumn{2}{c}{\textit{Taste}} \\
\hline
& \textbf{Model} & \textbf{Prec.} & \textbf{Rec.} & \textbf{P} & \textbf{R} & \textbf{P} & \textbf{R} & \textbf{P} & \textbf{R} & \textbf{P} & \textbf{R} \\
\hline
\multirow{6}*{\rotatebox{90}{$l = 20$}}  & RNP & 24.7 & 21.3 & 28.6 & 23.2 & 22.1 & 21.0 & 17.7 & 24.1 & 28.1 & 16.7 \\
& RNP-3P & 26 & 21.8 & 30.4 & 25.6 & 19.3 & 20.4 & 10.3 & 12.0 & 43.9 & 28.4 \\
& Intro-3P & 21 & 18.0 & 28.7 & 24.8 & 14.3 & 14.4 & 16.6 & 19.3 & 24.2 & 13.6 \\
& InvRAT & 37.5 & 31.6 & 54.5 & 45.5 & 26.1 & 27.6 & 22.6 & 25.9 & 46.6 & 27.4 \\
& ConRAT & \textbf{43.8} & \underline{39.7} & \underline{57.8} & \underline{53.0} & \underline{31.9} & \underline{35.5} & \textbf{29.0} & \underline{36.3} & \textbf{56.5} & \underline{33.9} \\
& \textbf{Ours} & \underline{40.6} & \textbf{58.4} & \textbf{67.5} & \textbf{71.4} & \textbf{34.1} & \textbf{42.3} & \underline{24.8} & \textbf{46.7} & 36.1 & \textbf{73.3} \\
\hline
\multirow{6}*{\rotatebox{90}{$l = 10$ }} & RNP & 32.7 & 14.5 & 40.1 & 12.0 & \underline{33.3} & \underline{18.7} & \underline{25.1} & \underline{17.4} & 32.3 & 9.8 \\
& RNP-3P & 28.4 & 13.2 & 41.8 & 19.2 & 22.2 & 12.4 & 16.5 & 10.4 & 33.2 & 10.6 \\
& Intro-3P & 24 & 12.2 & 51.0 & 26.0 & 18.8 & 9.7 & 16.5 & 10.6 & 9.7 & 2.6 \\
& InvRAT & 36.6 & 15.7 & \underline{59.4} & 26.1 & 31.3 & 15.5 & 16.4 & 9.6 & 39.1 & 11.6 \\
& ConRAT & \underline{38.2} & \underline{17.6} & 51.7 & \underline{26.2} & \textbf{32.6} & 17.4 & 23.0 & 13.8 & \textbf{45.3} & \underline{13.1} \\
& \textbf{Ours} & \textbf{39.5} & \textbf{58.4} & \textbf{63.3} & \textbf{56.4} & 27.3 & \textbf{67.4} & \textbf{26} & \textbf{43.5} & \underline{41.4} & \textbf{66.1} \\
\hline
\end{tabular}
    \caption{Objective performance of rationales for the multi-aspect beer reviews. All baselines are trained separately on each aspect rating, except for ConRAT~\citep{antognini-faltings-2021-rationalization}, which is trained on the \textit{Overall} label just like our method. Bold and underline denote the best and second-best results, respectively. (\textit{On the top}) table with Accuracy and F1 score for all the multi-aspect; (\textit{On the bottom}) table with Precision and Recall for all the multi-aspect.}
    \label{cockatiel:tab:annotation}
\end{table}

\subsection{Evaluation of Explanation Faithfulness}

We have demonstrated that we can generate concepts that greatly align with humans', but to legitimately serve as an explainability technique, we must also guarantee its faithfulness. 
This element is key, as the concepts leveraged by the model may not perfectly align with humans in every task, but we still want the explanation to reflect what the model is doing.
An XAI method is said to be faithful if its explanations faithfully convey the information that the model is using to generate its predictions.
In~\citep{ghorbani2019towards,zhang2021invertible}, they proposed to use an adaptation of the deletion and insertion explainability metrics to concept-based methods. In essence, they proposed to gradually mask/add the concepts (following their importance) and seeing the impact on the logits. If the concepts are indeed important for the model to predict, they should drastically decrease/increase as vital information for the prediction is progressively being erased/added.

To evaluate the explanation Faithfulness and present qualitative results, we used the IMDB dataset \citep{maas-EtAl:2011:ACL-HLT2011}. The IMDB dataset is a collection of 50K movie reviews from the Internet Movie Database (IMDB) website. For each review, IMDB specifies whether it is positive or negative (the label). The dataset is balanced, with 25K positive and 25K negative reviews.
We used a RoBERTa model to predict the label from the reviews.

\begin{figure}
    \centering
\includegraphics[width=1\textwidth]{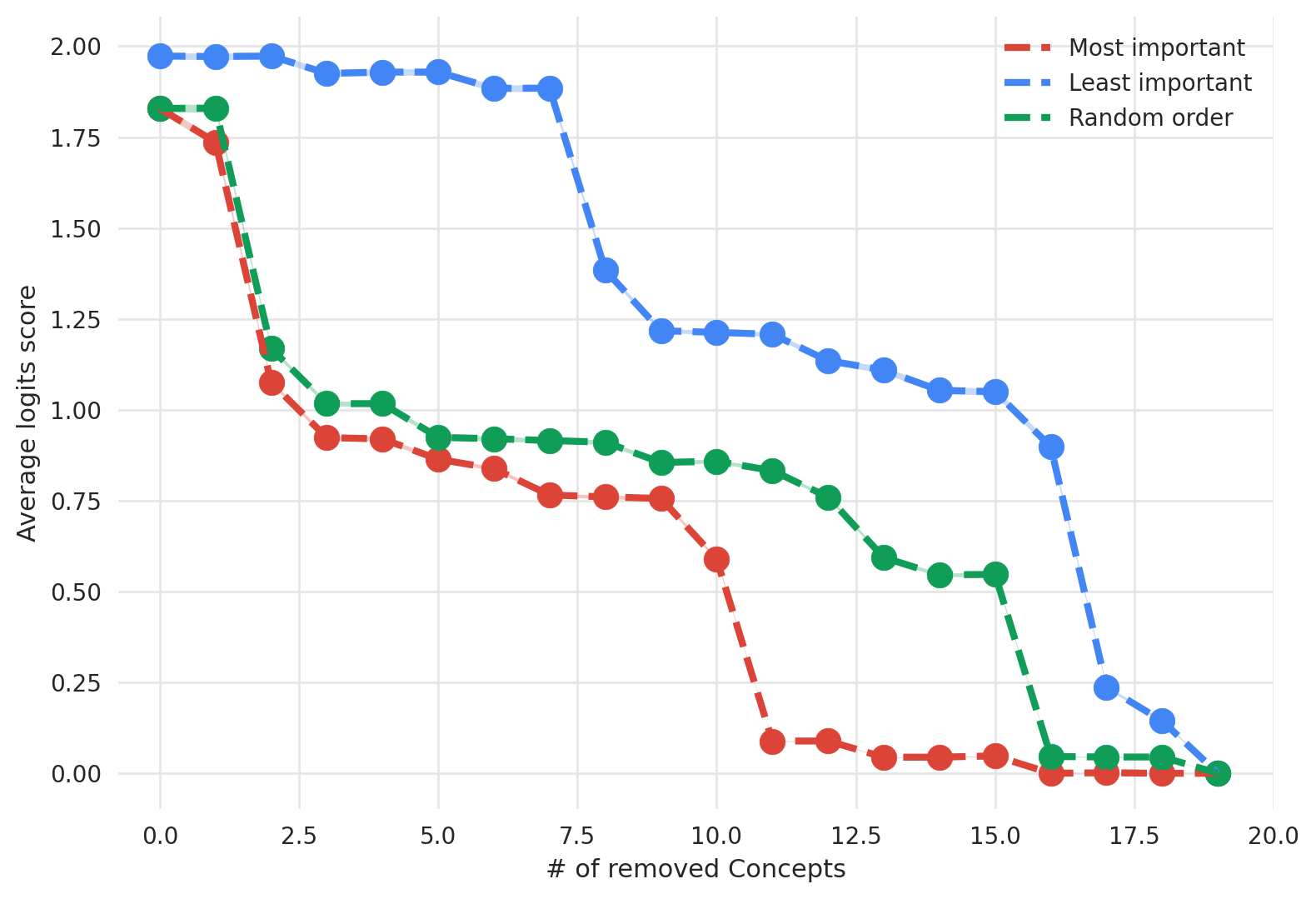}
\includegraphics[width=1\textwidth]{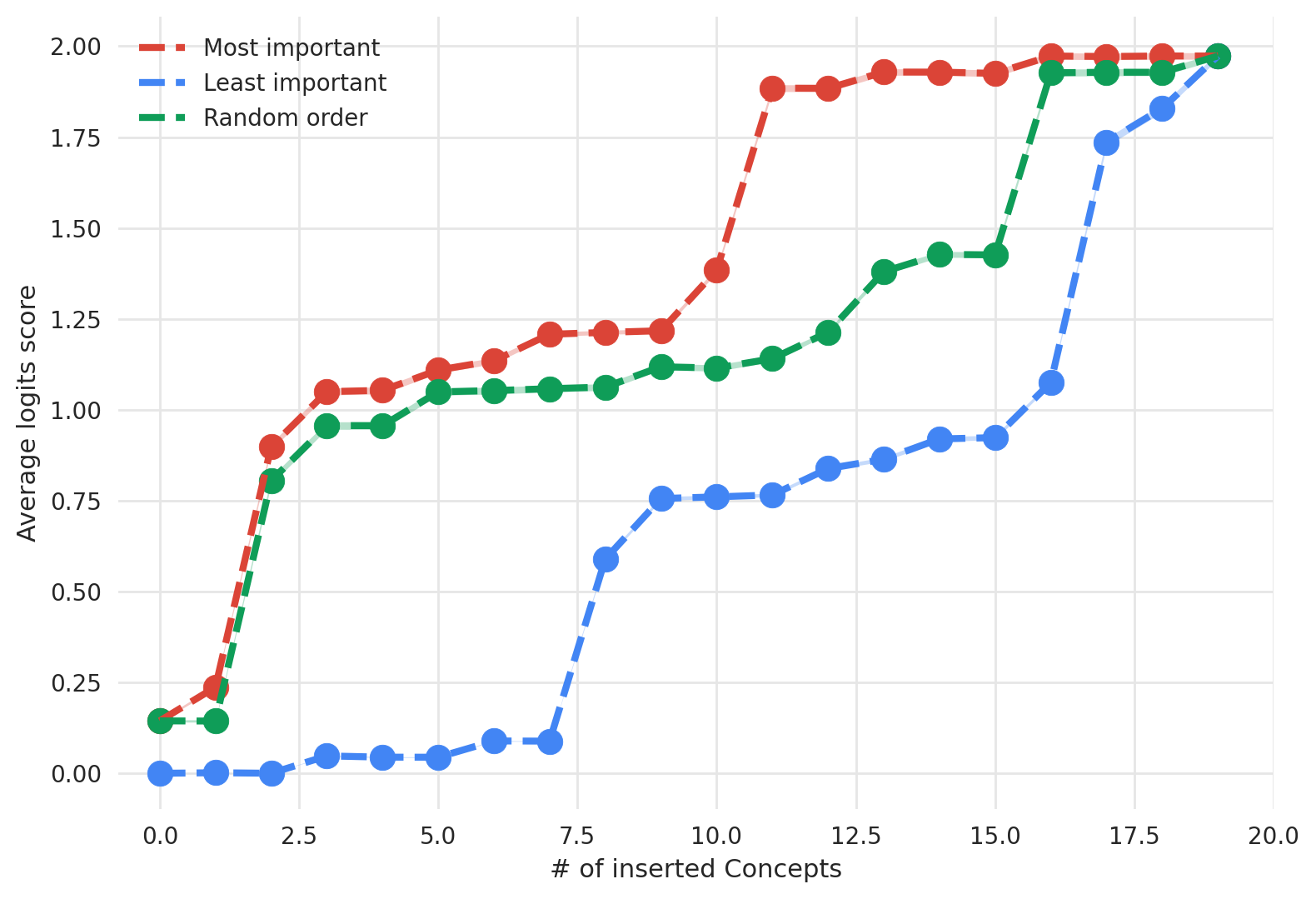}
    \caption{(Upper) Deletion curve for RoBERTa on \textit{IMDB Reviews} (lower is better). (Lower) Insertion
curve for RoBERTa on \textit{IMDB Reviews} (higher is better).}
    \label{cockatiel:fig:fidelity}
\end{figure}

In Fig.~\ref{cockatiel:fig:fidelity}, we showcase the plots for these two fidelity metrics on the \textit{IMDB Reviews} dataset.

We observe that the concepts are indeed important for the model's predictions. In the both plots, the curve corresponding to the concept ranked in order of importance according to our Sobol method is better than a random ranking of these concepts, and much better than if we had taken the order of Sobol importance in reverse. In particular, to obtain statistically significant results, we took 10 sets of 10k reviews, and computed the mean and standard deviation values for both of the metrics.

\begin{figure}
    \centering
\includegraphics[width=1\linewidth]
{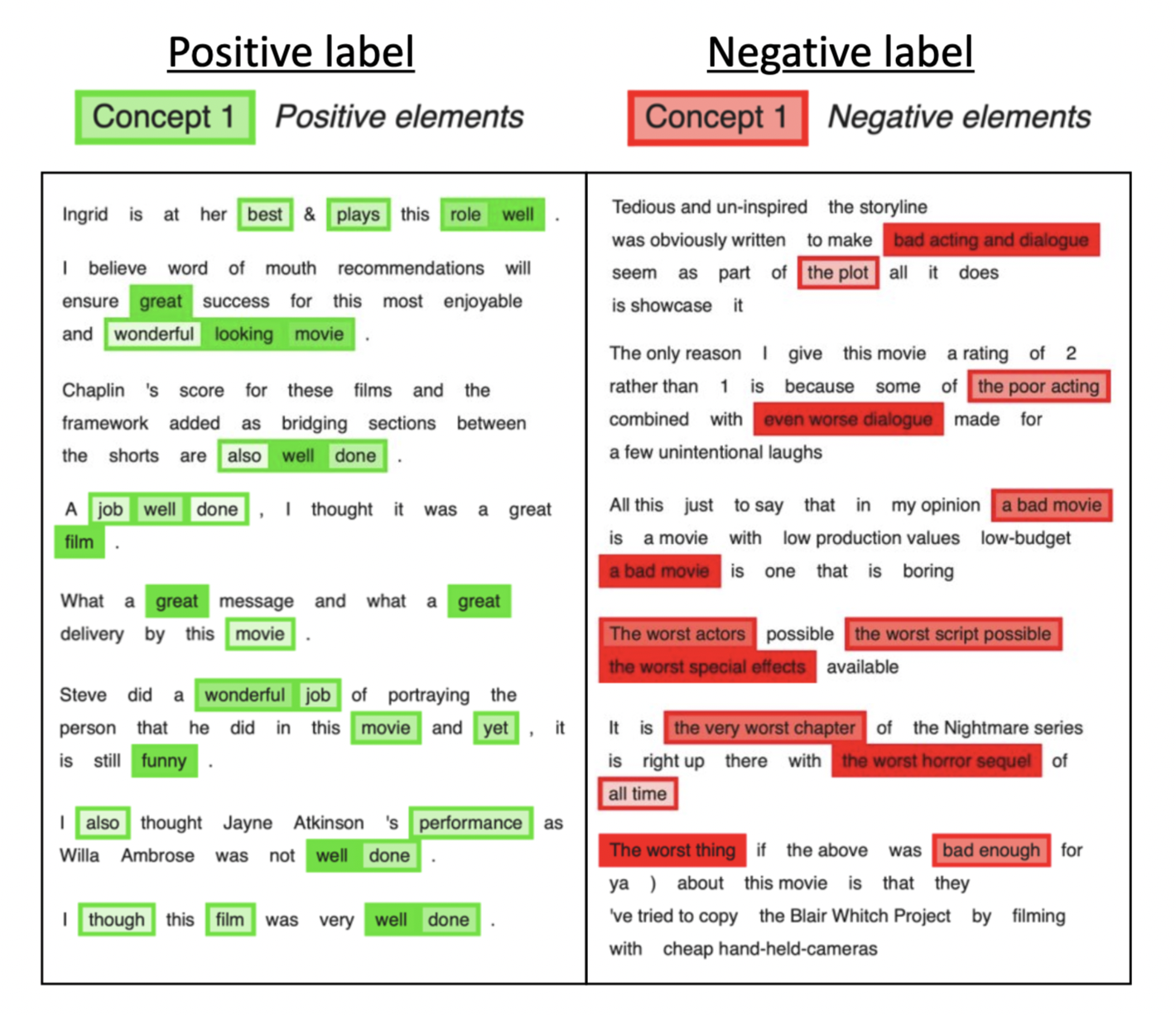}
    \caption{Concepts generated for a LSTM model with $l = 5$ for a few sentences out of  IMDB reviews. The colored elements are those important for the concept of the corresponding color (calculated with part \textit{(iii)} of our method). The more colorful the element, the more important it is for the concept (continuously). We have selected the most important concept for each label (see Fig.~\ref{cockatiel:fig:importanceconcept_LSTM}). The name of the concept is chosen manually in view of the important elements corresponding to the concepts.}
\label{cockatiel:fig:visu_LSTM}
\end{figure}

\subsection{Qualitative evaluation}

A model with a good accuracy like RoBERTa gives very good explanations.  Others like LSTM do not do so well and do not yield good explanations.  This is not a surprise; if the model predicts badly, necessarily the concepts it uses to predict will be bad. Similarly, if the model is very basic, it uses simple concepts to predict. 
The reviews in IMDB are also well written, so it is more comfortable to analyse sentences and words to properly call the concepts found by the NMF. 

In Fig.~\ref{cockatiel:fig:examples}, we can see the 3 most important concepts for each label class. Each of its concepts "\textit{the favorite movie}", "\textit{technically good/interesting movie}", "\textit{good comedie of family movie}" for the positive class or "\textit{the worst movie}", "\textit{middling movie}", "\textit{boring/stupid movie}" for the negative class are ideas that seem natural and which structures our vision of why a film would be positive or negative.

LSTMs are much less complex than RoBERTa, and as such, we can expect them to leverage less and much simpler concepts for their predictions. In contrast of RoBERTa, in the case of the LSTM (see figure \ref{cockatiel:fig:visu_LSTM}), COCKATIEL detected a single important concept per predicted class. For the positive class, this concept encompasses  \textit{the positive language elements} mostly, and for the negative class, \textit{the negative elements}. This is a much more basic view of the review classification problem, and COCKATIEL allows us to confirm our intuitions about the richness of the embedding learned by the LSTM.

\section{Conclusion}

In this chapter, we revisited concept-based explainability techniques and presented COCKATIEL, a post-hoc, model agnostic method capable of generating meaningful and faithful explanations for NLP models trained on classification tasks.
The method has three parts: \textit{(i) a concept part}, using Non-Negative Matrix Factorization to discover the concept, \textit{(ii) a ranking part}, using Total Sobol indices to measure the influence of each concept, and \textit{(iii) an interpretable elements part}, using a black-box attribution method to quantify the impact of each element out of each concept.

We measured COCKATIEL's ability to discover concepts that align with those humans and obtained better scores than state-of-the-art methods. We demonstrated that in addition to generating meaningful concepts for humans, these explanations are faithful to the models. Finally, we gave some qualitative examples of explanations for different models to understand the method "in practice".

Note that COCKATIEL is implemented to work in PyTorch and is compatible with Hugging Face models. It is freely available on GitHub\footnote{\url{https://github.com/fanny-jourdan/cockatiel}}.

\section*{Limitations}

We have demonstrated that COCKATIEL is capable of generating meaningful explanations that align with human concepts, and that they tend to explain rather faithfully the model. 

The concepts extracted of NMF are abstract and we interpret them using part 3 of the method. However, for the interpretation, we rely on our own understanding of the concept linked to the examples of words or clauses associated with the concept. This part therefore requires human supervision and will not be identical depending on who is looking. One way to add some objectivity to this concept labeling task would be to leverage topic modeling models to find a common theme to each concept.

In addition, $\bm{\tau}_1$ and $\bm{\tau}_2$ were chosen empirically to allow for an adequate concept complexity/human understandability trade-off in our examples. We recognize that this choice might not be optimal in every situation, as more complex concept may be advantageous in some cases, and more easily understandable ones, in others. We surmise that this choice might also depend on the amount of concepts and on the model's expressivity.

Finally, we have studied the meaningfulness and fidelity of our generated concepts, but ideally, the simulatability should also be tested. This property measures the explanation's capacity to help humans predict the model's behavior, and has recently caught the attention of the XAI community~\citep{fel2021cannot,shen2020useful,nguyen2018comparing,hase2020evaluating}. We leave this analysis for future works.





%% file: XAI_Fairness.tex
\chapter{Explainability for Fairness}

In this chapter, we build upon COCKATIEL, the effective explainablibility method developed in the previous chapter, utilizing it as a foundation for more effectively reducing biases in NLP models. Thanks to our XAI methodology, which reveals the concepts on which a model relies for prediction, we can identify concepts that are strongly linked to sensitive variables, such as gender, ethnicity or age, which are often the source of bias, in order to target them and mitigate these biases more effectively. This idea forms the crux of the preprint titled "\textit{TaCo: Targeted Concept Removal in Output Embeddings for NLP via Information Theory and Explainability}" \citep{jourdan2023taco} which is the focus of this chapter. As the first author, I led the experimental work, supported by Louis Béthune, and wrote the entire paper with the assistance of Agustin Picard. 

The paper revolves around the principle derived from information theory, which posits that for a model to be fair, it should not be able to predict sensitive variables. However, the implicit presence of information related to these variables in language makes identifying and mitigating biases a complex task. To address this challenge, we introduce a novel approach that functions at the embedding level of an NLP model, independent of its specific architecture. This method capitalizes on recent advancements in XAI techniques and implements an embedding transformation to eliminate the implicit information of a selected variable. By manipulating the embeddings in the final layer directly, our approach facilitates easy integration into existing models without necessitating extensive modifications or retraining. Through our evaluation, we demonstrate that this post-hoc approach effectively reduces gender-related associations in NLP models, while maintaining their overall performance and functionality. 

This chapter not only presents a new methodological advancement but also contributes significantly to the broader conversation about achieving fairness in machine learning and AI.

\section{Introduction}

Artificial Intelligence models are increasingly being deployed in impactful areas but often carry biases, leading to unfair and unethical consequences.
To guard against such consequences, we must identify and understand the biased information these models utilize. This links the task of ensuring the ethical behavior of a model with its interpretability or explainability: if we can show the real reasons or features why a model makes the predictions or recommendations it does, then we can control for biases among these features.    

In this chapter, we propose a method for finding factors that influence model predictions using an algebraic decomposition of the latent representations of the model into orthogonal dimensions. By using the method of 
Singular Value Decomposition (SVD), which is well suited to our particular test case, we isolate various dimensions that contribute to a prediction. As our case study, we study the use of gender information in predicting occupations in the {\em Bios} dataset \citep{de2019bias} by different Transformer encoder models (RoBERTa \citep{liu2019roberta}, DistilBERT \citep{sanh2019distilbert}, and DeBERTa \citep{he2020deberta}). By intervening on the latent representations to remove certain dimensions and running each model on the edited representations, we first show a causal connection between various of these dimensions and the models' predictions of gender and occupation. This enables us to eliminate dimensions in the decomposition that convey gender bias in the decision without having a significant impact on the models' predictive performance on occupations. Because SVD gives us orthogonal dimensions, we know that once we have removed a dimension, the information it contains is no longer present inside the model.

Crucially, we show how to link various dimensions in the SVD decomposition of the latent representation with concepts like gender and its linguistic manifestations that matter to fairness, providing not only a faithful explanation of the model's behavior but a humanly understandable -- or plausible explanation, in XAI terms -- one as well. Faithfulness provides a causal connection and produces factors necessary and sufficient for the model to produce its prediction given a certain input.  Plausibility has to do with how acceptable the purported explanation is to humans \citep{jacovi:goldberg:2020}.   

In addition, we demonstrate that our method can reach close to the theoretically ideal point of minimizing algorithmic bias while preserving strong predictive accuracy, and we offer a way of quantifying various operations on representations.  Our method also has a relatively low computational cost, as we detail below. 
In principle, our approach applies to any type of deep learning model; it also can apply to representations that are generated at several levels, as in modern Transformer architectures with multiple attention layers~\citep{vaswani2017attention}.  As such, it can serve as a diagnostic tool for exploring complex deep learning models.

\section{Related work}

In this chapter, we aim to enhance the fairness of our model by employing a prevalent approach in information theory, which involves eliminating information related to sensitive variables to prevent the model from utilizing it \citep{kilbertus2017avoiding}. This approach to fairness, which we'll call "\textit{sensitive variable information erasure}", has already been discussed in various forms in the literature, detailed in chapter \ref{SOTA}. In particular, many methods in line with our proposed method -- of intervening in the latent space of the model -- have been presented in section \ref{sota:fairness_LS}.


However,  \citet{ravfogel2023loglinear} recently discovered a limitation of theses methods. The limitation presented in \citep{ravfogel2023loglinear} comes from the practical use of the $\nu$-information. This measure updates the mutual information, considering the expressivity of functions measuring information in the latent space. Their paper explains that by protecting information using a linear model, other, more expressive models can still decode this protected information. 


Our method is not affected by this problem. Because, as we're effectively destroying information -- and not simply hiding it from linear models --, no model should be able to access it.

Specifically, we propose to use methodology from the field of explainable AI (XAI) to perform feature selection to reduce the effect of algorithmic bias~\citep{frye2020asymmetric,dorleon2022,galhotra2022} in a human-understandable yet principled manner. Indeed, our aim is to decompose the model’s embedding into an \textbf{interpretable base of concepts} -- i.e. our features --, select those that carry the most information about sensitive variables and the least about the actual task, and suppress them from the embedding. We would thus be guarding against sensitive variables without a specific parametric model, and thus, less surgically precise but more interpretable in our interventions than the literature: a trade-off that can be enticing in certain applications.

To do this, we'll be drawing on work in concept-based XAI, which proposes to search for human-understandable, high-level concepts in the model's latent space and determine their role in the predictions. Some notable methods include TCAV~\citep{kim2018interpretability} and CRAFT~\citep{fel:etal:2023} for Computer Vision, and COCKATIEL~\citep{jourdan2023cockatiel} for NLP -- on which we will rely to create understandable decompositions of the latent space.

\section{Preliminaries and definitions}

\begin{figure}[h]
    \centering
\includegraphics[width=0.7\linewidth]{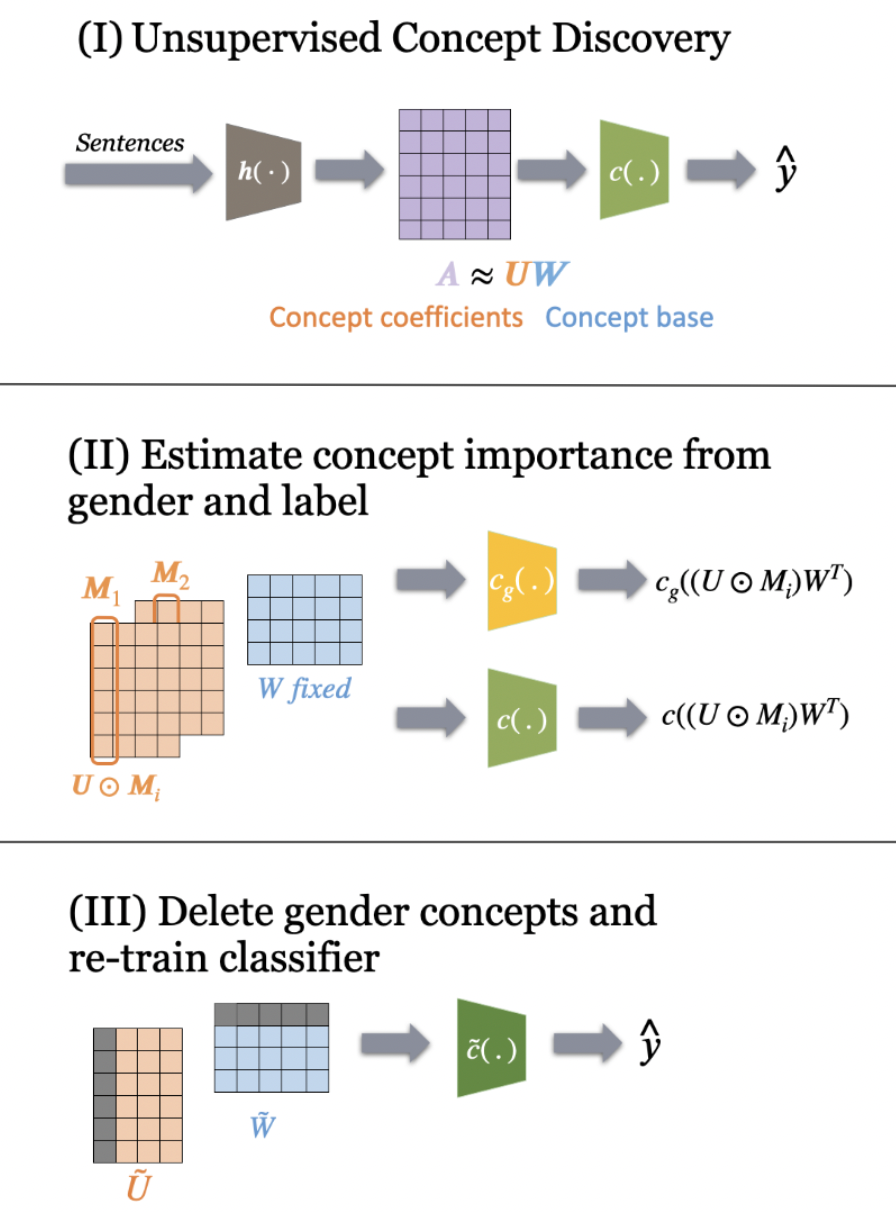}
    \caption{\textbf{Overview of our method.} The SVD decomposition \textbf{(I)} yields concepts, whose importance \textbf{(II)} with respect to gender and occupation are evaluated with Sobol method. Finally, some concepts are removed, which ''neutralizes'' the gender information and \textbf{(III)} produces a fairer classifier.}
    \label{taco:fig:diagram}
\end{figure}

The idea behind the method we propose for creating gender-neutral embeddings can be summed up in 3 components: \textit{(i)} Identification of occupation-related concepts within the latent space of an NLP model; \textit{(ii)} evaluation of the significance of each concept in predicting gender; and \textit{(iii)} removal of specific concepts to construct a gender-neutral embedding, followed by retraining a classifier to predict occupation based on this modified embedding. For a visual representation of the method, refer to Figure \ref{taco:fig:diagram}, which provides an explanatory diagram illustrating the different stages of the process.

To perform the proposed method, it is crucial to secure certain guarantees and engage in rigorous theoretical modeling. This ensures the identification and application of suitable, practical tools. This involves establishing clear criteria for selecting practical tools and methodologies, enabling the achievement of dependable and consistent results. These are defined in this section. 

However, it is important to note that there is no guarantee that our group of found concepts will clearly separate between those containing information on the output and those containing information on the sensitive variable. It would be possible, in an unfavorable case, to have only concepts containing information relative to both variables.

\subsection{Notation}
In the context of supervised learning, we operate under the assumption that a neural network model, denoted as $\pred \colon \mathcal{X} \rightarrow \mathcal{Y}$, has been previously trained to perform a specific classification task. Here, $\bm{x} \in \mathcal{X}$ represents an embedded input text (in our example, a LinkedIn biography), and $y \in \mathcal{Y}$ denotes their corresponding label (in our example, the occupation).  To further our analysis, we introduce an additional set, $g \in \{\text{Male, Female}\}$, representing the binary gender variable. This variable is crucial as it is the axis along which we aim to ensure fairness in our model. With $n$ individuals in the training set, we define $\vx = (\bm{x}_1,...,\bm{x}_n) \in \mathcal{X}^n$, $\vy = (y_1,...,y_n) \in \mathcal{Y}^n$, and  $\vg = (g_1,...,g_n) \in \{\text{Male, Female}\}^n$.

The model $\pred$ is conceptualized as a composition of two functions. First, $h$ encompasses all layers from the input $\bm{x}$ to a latent space, which is a transformed representation of the input data. Then, the classification layers $c$ classify these transformed data. The model can be summarized as:
\begin{equation}\label{taco:eq:general_pred_model}
\pred(\bm{x}) = c \circ h(\bm{x}) \,,
\end{equation}
where $h(\bm{x}) \in \mathbb{R}^d$. 
In this space, we obtain an output embedding matrix, denoted as $\va = h(\vx) \in \mathbb{R}^{n \times d}$, which is also referred to as the \texttt{CLS} embedding in the context of an encoder Transformer for a classification task.

\subsection{Causal underpinnings of the method}

A faithful reconstruction of $\pred$ should support a causal relation that can be formulated as follows: $A$ {\em and} $B$ and {\em had} $A$ {\em not been the case,} $B$ {\em would not have been the case either} \citep{lewis:1973,pearl2009causality,jacovi:goldberg:2020}. $A$ describes the causally necessary and ``other things being equal'' sufficient condition for $B$.  Faithfulness implies that a particular type of input is causally necessary for the prediction at a given data point $\bm{x}$. Such a condition could take the form of one or more variables having certain values or set of variables.  It can also take the form of a logical statement at a higher degree of abstraction.  

A counterfactual theory is simply a collection of such statements, and for every deep learning model, there is a {\em counterfactual theory} that describes it \citep{jacovi:etal:2021,asher:etal:2022, yin:neubig:2022}. 

The counterfactual theory itself encodes a counterfactual model \citep{lewis:1973} that contains: a set of worlds or ``cases'' $\mathscr{W}$, a distance metric $\|.\|$ over $\mathscr{W}$, and an interpretation function that assigns truth values to atomic formulas at $w \in \mathscr{W}$ and then recursively assigns truth values to complex formulas including counterfactuals {\em had A not been the case, B would not have been the case}, where $A$ and $B$ are Boolean or even first-order formulas describing factors that can figure in the explanans and predictions of the model. The cases in a counterfactual model represent the elements that we vary through counterfactual interventions. The distance metric tells us what are the minimal post-intervention elements among the cases that are most similar to the pre-intervention case.  We can try out various counterfactual models for a given function $\pred$ \citep{asher:etal:2022}.  

For instance, we can alternatively choose \textit{(i)} $\mathscr{W} = \mathcal{X}^n$ the model's input entries, \textit{(ii)} $\mathscr{W} = $ the set of possible attention weights in $\pred$ (assuming it's a Transformer model),
\textit{(iii)} $\mathscr{W} = $ the possible parameter settings for a specific intermediate layer of $\pred$ --e.g., the attribution matrix as in \citep{fel:etal:2023}, \textit{(iv)} $\mathscr{W} = $ 
the possible parameter settings for the final layer of an attention model as suggested in \citep{wiegreffe:pinter:2019}, or \textit{(iv)} sets of factors or dimensions of the final layer output matrix from the entire model \citep{jourdan2023cockatiel}.
Our explanatory model here follows the ideas of \citep{jourdan2023cockatiel}, though it differs importantly in details. For one thing, we note that our general semantic approach allows us to have Boolean combinations, even first-order combinations of factors, to isolate a causally necessary and sufficient {\em explanans}.

\subsubsection{Counterfactual intervention for concept ranking}

In part 2 of our method, we calculate the causal effect on the sensitive variable for each dimension $r$ of $\vu$ and $\vw$ of the $\va=\vu\vw$ decomposition created in part 1 of the method (more details on this decomposition in practice in section \ref{taco:sec:SVD}). We conduct a counterfactual intervention that involves removing the dimension or ``concept'' $r$. By comparing the behavior of the model with the concept removed and the behavior of the original model, we can assess the causal effect of the concept on the sensitive variable. As explained above, this assumes a counterfactual model where $\mathscr{W}$ corresponds to the possible variations of $\va=h(\vx)$ when reducing the dimensionality of its decomposition $\va=\vu \vw$.

In addition, this approach requires that we are really able to remove the targeted dimension, which means that the dimensions must be independent of each other. 

\subsection{Information theory guarantees}\label{taco:sec:inftheory}
As part of our counterfactual intervention, we calculate the "importance of a dimension" for the sensitive variable, so that our final integration $\va$ is \textit{as independent as possible} from the sensitive variable $\vg$. According to information theory, if we are unable to predict a variable $\vg$ from the data $\va$, it implies that this variable $\vg$ does not influence the prediction of $y$ by $\va$~\citep{Xu2020A}. This definition is widely used in causal fairness studies \citep{kilbertus2017avoiding}, making it a logical choice for our task.

We train a classifier $c_g$ for the gender $\vg$ prediction from the matrix $\va$ (see appendix~\ref{taco:modelsdetailsclassif} for details on its architecture). If $c_g$ cannot achieve high accuracy for gender prediction, it implies that $\va$ does not contain gender-related information. The most reliable approach is thus to remove each of the dimensions found by the decomposition one by one and re-train a classifier to predict gender from this matrix. This can be computationally expensive if we have many dimensions. An alternative is to train only one classifier to predict gender on the original matrix $\va$, and then use a sensitivity analysis method (section \ref{taco:sec:importance}) on these dimensions to determine their importance for prediction. We will choose the latter for our method. 

To validate the efficiency of this approach, we adopt $\nu$-information as our metric. This metric serves as a crucial tool in quantifying the degree to which our methodology successfully mitigates the influence of the sensitive variable within the model. In the results presented in section \ref{taco:sec:results-analysis-quantitative}, we showcase -- through this metric -- the significant reduction in the presence of sensitive variable information post-intervention.
The selection of $\nu$-information as our evaluative metric is not arbitrary but is rooted in its established relevance within the field. This metric was initially introduced in \citep{ravfogel2023loglinear}, and its utility by \citet{belrose2023leace}.
These studies underscore the metric's robustness and its suitability for our analysis.

\section{Methodology}

Now that we've established the constraints on how to implement our method, we define the tools used for each part of the method: \textit{(i)} For the decomposition that uncovers our concepts, we use the SVD decomposition. \textit{(ii)} For the gender importance calculation, we train a classifier for this task and then use Sobol importance. \textit{(iii)} To remove concepts, we simply remove the columns of the matrix that correspond to the target concept.

\subsection{Generating concepts used for occupation classification}\label{taco:sec:SVD}

Firstly, we look into the latent space before the last layer of the model, where the decisions are linear. We have an output embedding matrix (also called \texttt{CLS} embedding) of the Transformer model in this space.
The aim here is to employ matrix decomposition techniques to extract both a concept coefficients matrix and a concept base matrix for the latent vectors.

To illustrate this part, we discover concepts without supervision by factorizing the final embedding matrix through a Singular Value Decomposition. We choose this technique, but it is possible to do other decompositions, like PCA~\citep{wold1987principal}, RCA~\citep{bingham2001random} or ICA~\citep{comon1994independent}.

\paragraph{Singular Value Decomposition (SVD)} 

The embedding matrix $\va \in \mathbb{R}^{n \times d}$ can be decomposed using an SVD as:
$\va = U_0 \Sigma_0 V_0^\intercal = U_0W_0 \,,$
where $U_0 \in \mathbb{R}^{n \times n}$ contains the left singular vectors,  $V_0 \in \mathbb{R}^{d \times d}$  contains the right singular vectors, and $\Sigma_0 \in \mathbb{R}^{n \times d}$ is a rectangular diagonal matrix containing the singular values on its diagonal. Note that $U_0$ and $V_0$ are orthonormal matrices, which means that  $U_0^\intercal U_0=Id$ and $V_0^\intercal V_0=Id$.

An SVD can be used to explain the main sources of variability of the decomposed matrix -- here, $\va$. The matrix $W_0^{-1} =  V_0 \Sigma_0^{-1}$ is indeed considered as a linear projector of $\va$, and $U_0$ is the projection of $\va$ on a  basis defined by $V_0$. The amount of variability, or more generally, of information that has been captured in $\va$ by a column of $U_0$ is directly represented by the corresponding singular value in $\Sigma_0$. Using only the $r$ largest singular values of $\Sigma_0$, with $r<<d$, therefore allows capturing as much variability in $\va$ as possible with a reduced amount of dimensions. 
We then decompose $\va$ as: 
\begin{equation}\label{taco:eq:decomp_A}
\va \approx \vu \vw \
\end{equation}
where $\vw = \Sigma V^\intercal$ is a compressed version of $W_0$, in the sense that $\Sigma \in \mathbb{R}^{r,r}$ is the diagonal matrix containing the $r$ largest singular values of $\Sigma$ and  $V \in  \mathbb{R}^{d,r}$ contains the $r$ corresponding right singular vectors. The matrix $\vu \in \mathbb{R}^{n,r}$ consequently contains the $r$ right singular vectors which optimally represent the information contained in $\va$ according to the SVD decomposition (see \citep{EckartYoung1936} for more details). In this chapter, we refer to these vectors as the extracted \emph{concepts} of $\va$.

\paragraph{Implementation details}
We used the SVD implementation in \\
\texttt{sparse.linalg.svds} provided in the \texttt{Scipy} package. This function makes use of an incremental strategy to only estimate the first singular vectors of $\va$, making the truncated SVD scalable to large matrices. Note that this implementation turned out to obtain the most robust decompositions of $\va$ after an empirical comparison between  \texttt{torch.svd}, \texttt{torch.linalg.svd}, \texttt{scipy.linalg.svd}, and \\
\texttt{sklearn.decomposition.TruncatedSVD} on our data.

\subsection{Estimating concept importance with Sobol Indices}\label{taco:sec:importance}

We have seen in Eq. \eqref{taco:eq:general_pred_model} that $c$ is used to predict $\vy$ based on $\va$, and we discussed in section \ref{taco:sec:inftheory}, that the non-linear classifier $c_{g}$ also uses $\va$ as input but predicts the gender $\vg$. We also recall that $\va$ is approximated by $\vu\vw$ in Eq. \eqref{taco:eq:decomp_A}, and that the columns of $\vu$ represent the so-called \emph{concepts} extracted from $\va$.  

We then evaluate the concepts' importance by applying a feature importance technique on $\vg$ using  $c_{g}$, and on $\vy$ using $c$. More specifically, Sobol indices~\citep{sobol1993sensitivity} are used to estimate the concepts' importance, as in~\citep{jourdan2023cockatiel}. A key novelty of the methodology developed in this chapter is to simultaneously measure each concept's importance for the outputs $\vg$ and $\vy$, and not only $\vy$ as proposed in~\citep{jourdan2023cockatiel}.

\begin{figure}[ht]
    \centering
\includegraphics[width=1\linewidth]{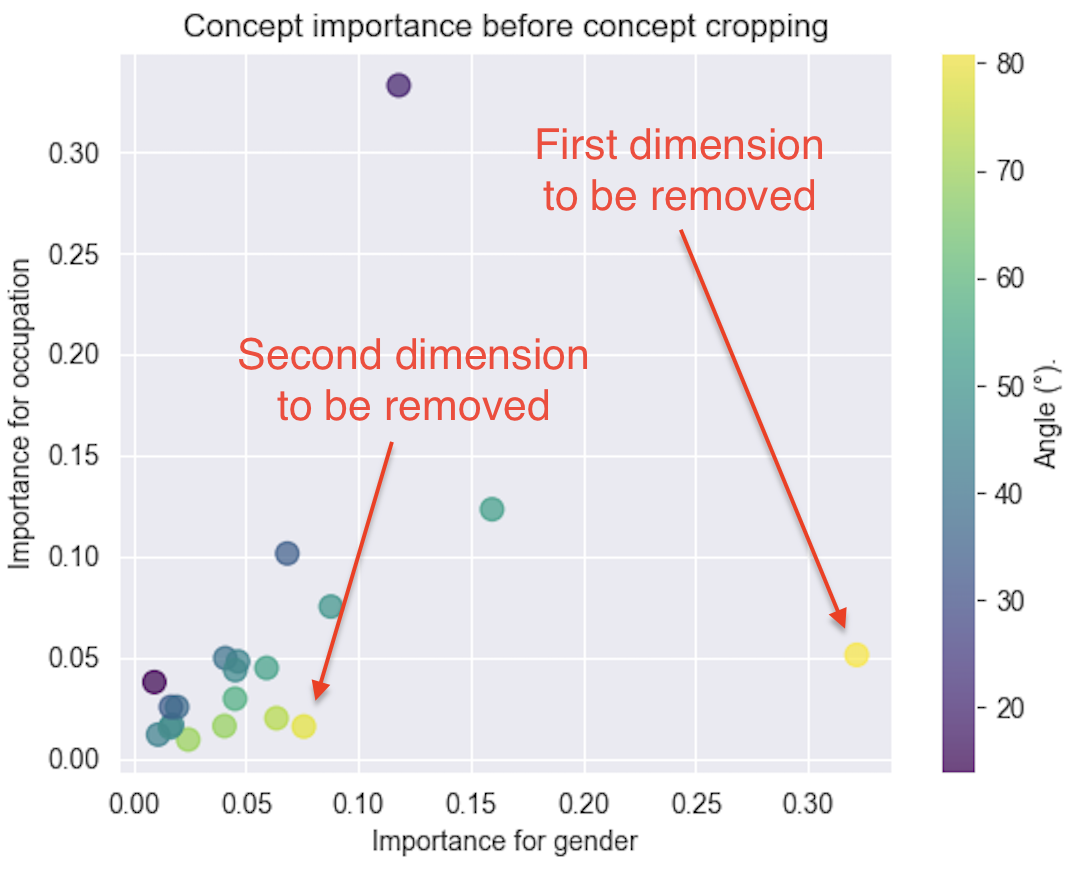}
    \caption{
    \textbf{Co-importance plot for $r=20$ dimensions with respect to \textit{Occupation} and \textit{Gender} labels, computed with the Sobol method on RoBERTa model.} The colour is based on  the angle $a=90-\frac{2}{\pi}\arctan{(\frac{y}{x})}$. The angles with extreme values correspond to concepts with high importance for gender, but comparatively low importance for occupation. Their removal yields a favourable tradeoff in accuracy/fairness.
    }
    \label{taco:fig:conceptimp}
\end{figure}

To estimate the importance of a concept $\vu_i$ for gender (resp. for the label), we measure the fluctuations of the model output $c_{g}(\vu \vw)$ (resp. $c(\vu \vw)$) in response to  perturbations of concept coefficient $\vu_i$ .

We then propagate this perturbed activation to the model output $\vg = c_{g}(\tilde{\va})$ (resp. $\vy = c(\tilde{\va})$). We can capture the importance that a concept might have as a main effect -- along with its interactions with other concepts -- on the model’s output by calculating the expected variance that would remain if all the concepts except the i were to be fixed. This yields the general definition of the total Sobol indices.
An important concept will have a large variance in the model's output, while an unused concept will barely change it.

Here, we define the classic Total Sobol indices and how they are calculated using our method. In practice, these indices can be calculated very efficiently~\citep{marrel2009calculations,saltelli2010variance,janon2014asymptotic} with Quasi-Monte Carlo sampling and the estimator explained below.

Let $(\Omega, \mathcal{A}, \mathbb{P})$ be a probability space of possible concept perturbations. To build these concept perturbations, we use $\boldsymbol{M}=\left(M_1, \ldots, M_r\right) \in \mathcal{M} \subseteq [0,1]^r$, i.i.d. stochastic masks where $M_i \sim \mathcal{U}(\left[  0,1 \right]),  \forall i \in 1,...,r$. We define concept perturbation $\Tilde{\vu}=\boldsymbol{\pi}(\vu, \boldsymbol{M})$ with the perturbation operator $\boldsymbol{\pi}(\vu, \boldsymbol{M})=\vu \odot \boldsymbol{M}$ with $\odot$ the Hadamard product (that takes in two matrices of the same dimensions and returns a matrix of the multiplied corresponding elements).

We denote the set $\mathcal{U}=\{1, \ldots, r\}$, $\boldsymbol{u}$ a subset of $\mathcal{U}$, its complementary $\sim \boldsymbol{u}$ and $\mathbb{E}(\cdot)$ the expectation over the perturbation space. We define $\phi: \mathcal{A} \rightarrow \mathbb{R}$, a function that takes the perturbations $\vu$ from the last layer and applies the difference between the model's two largest output logits -- i.e. $\phi(\vu) = l_{(1)}(c(\vu))- l_{(2)}(c(\vu))$ where $c$ is the classification head of the model (when we do Sobol for the occupation task, it is $c$, and when we do Sobol for the gender task, it is $c_g$) and $l_{(1)}$ and $l_{(2)}$ represent the highest and second highest logit values respectively. We assume that $\phi\in \mathbb{L}^2(\mathcal{A}, \mathbb{P})$ -- i.e. $|\mathbb{E}(\phi(\boldsymbol{U}))|<+\infty$.

The Hoeffding decomposition provides $\phi$ as a function of summands of increasing dimension, denoting $\phi_{\boldsymbol{u}}$ the partial contribution of the concepts $\boldsymbol{U}_{\boldsymbol{u}}=\left(U_i\right)_{i \in \boldsymbol{u}}$ to the score $\phi(\boldsymbol{U})$:

\begin{equation}
\label{taco:eq:hoeffding}
\begin{aligned}
\phi(\boldsymbol{U}) & =\phi_{\varnothing}  +\sum_{i=1}^r \phi_i\left(U_i\right) \\
& +\sum_{1 \leqslant i<j \leqslant r} \phi_{i, j}\left(U_i, U_j\right)+\cdots \\
& +\phi_{1, \ldots, r}\left(U_1, \ldots, U_r\right) \\
& =\sum_{\boldsymbol{u} \subseteq \mathcal{U}} \phi_{\boldsymbol{u}}\left(\boldsymbol{U}_{\boldsymbol{u}}\right)
\end{aligned}
\end{equation}

Eq.~\ref{taco:eq:hoeffding} consists of $2^r$ terms and is unique under the orthogonality constraint:
$$
\quad \mathbb{E}\left(\phi_{\boldsymbol{u}}\left(\boldsymbol{U}_{\boldsymbol{u}}\right) \phi_{\boldsymbol{v}}\left(\boldsymbol{U}_{\boldsymbol{v}}\right)\right)=0,  \  \forall(\boldsymbol{u},\boldsymbol{v}) \subseteq \mathcal{U}^2\text { s.t. } \boldsymbol{u} \neq \boldsymbol{v}$$
Moreover, thanks to orthogonality, we have $\phi_{\boldsymbol{u}}\left(\boldsymbol{U}_{\boldsymbol{u}}\right)=\mathbb{E}\left(\phi(\boldsymbol{U}) \mid \boldsymbol{U}_{\boldsymbol{u}}\right)-\sum_{\boldsymbol{v} \subset \boldsymbol{u}} \phi_{\boldsymbol{v}}\left(\boldsymbol{U}_{\boldsymbol{v}}\right)$ and we can write the model variance as:

\begin{equation}
\label{taco:eq:var}
    \begin{aligned}
\mathbb{V}(\phi(\boldsymbol{U})) & =\sum_i^r \mathbb{V}\left(\phi_i\left(U_i\right)\right) \\
& +\sum_{1 \leqslant i<j \leqslant r} \mathbb{V}\left(\phi_{i, j}\left(U_i, U_j\right)\right) \\
& +\ldots+\mathbb{V}\left(\phi_{1, \ldots, r}\left(U_1, \ldots, U_r\right)\right) \\
& =\sum_{\boldsymbol{u} \subseteq \mathcal{U}} \mathbb{V}\left(\phi_{\boldsymbol{u}}\left(\boldsymbol{U}_{\boldsymbol{u}}\right)\right)
\end{aligned}
\end{equation}

Eq.~\ref{taco:eq:var} allows us to write the influence of any subset of concepts $\boldsymbol{u}$ as its own variance. This yields, after normalization by $\mathbb{V}(\phi(\boldsymbol{U}))$, the general definition of Sobol' indices.

\paragraph{Definition}\textit{Sobol' indices \citep{sobol1993sensitivity}.}
The sensitivity index $\mathcal{S}_{\boldsymbol{u}}$ which measures the contribution of the concept set $\boldsymbol{U}_{\boldsymbol{u}}$ to the model response $f(\boldsymbol{U})$ in terms of fluctuation is given by:
\begin{equation}
\label{taco:eq:sobol}
\mathcal{S}_{\boldsymbol{u}} =\frac{\mathbb{V}\left(\phi_{\boldsymbol{u}}\left(\boldsymbol{U}_{\boldsymbol{u}}\right)\right)}{\mathbb{V}(\phi(\boldsymbol{U}))}= \frac{\mathbb{V}\left(\mathbb{E}\left(\phi(\boldsymbol{U}) \mid \boldsymbol{U}_{\boldsymbol{u}}\right)\right)-\sum_{\boldsymbol{v} \subset \boldsymbol{u}} \mathbb{V}\left(\mathbb{E}\left(\phi(\boldsymbol{U}) \mid \boldsymbol{U}_{\boldsymbol{v}}\right)\right)}{\mathbb{V}(\phi(\boldsymbol{U}))} 
\end{equation}

Sobol' indices provide a numerical assessment of the importance of various subsets of concepts in relation to the model's decision-making process. Thus, we have: $\sum_{\boldsymbol{u} \subseteq \mathcal{U}} \mathcal{S}_{\boldsymbol{u}}=1$.

Additionally, the use of Sobol' indices allows for the efficient identification of higher-order interactions between features. Thus, we can define the Total Sobol indices as the sum of of all the Sobol indices containing the concept $i: \mathcal{S}_{T_i}=\sum_{\boldsymbol{u} \subseteq \mathcal{U}, i \in \boldsymbol{u}} \mathcal{S}_{\boldsymbol{u}}$. 
So we can write:
\paragraph{Definition}\textit{Total Sobol indices.}
The total Sobol index $\mathcal{S}{T_i}$, which measures the contribution of a concept $\vu_i$ as well as its interactions of any order with any other concepts to the model output variance, is given by:
\begin{align}
        \label{taco:eq:total_sobol}
           \mathcal{S}_{T_i} 
           & = \frac{ \mathbb{E}_{\bm{M}_{\sim i}}( \Var_{M_i} ( \vy | \bm{M}_{\sim i} )) }{ \Var(\vy) } \\
           & = \frac{ \mathbb{E}_{\bm{M}_{\sim i}}( \Var_{M_i} ( \phi((\vu \odot \vm)\vw) | \bm{M}_{\sim i} )) }{ \Var( \phi((\vu \odot \vm)\vw)) }.
    \end{align}

In practice, our implementation of this method remains close to what is done in COCKATIEL \citep{jourdan2023cockatiel}, with the exception of changes to the $\phi$ function created here to manage concepts found in all the classes at the same time (unlike COCKATIEL, which concentrated on finding concepts in a per-class basis). 

\subsubsection{Classifier heads training}\label{taco:modelsdetailsclassif}

The importance estimation in part 2 of our method requires a classifier to be trained on top of the features. An ideal classifier will leverage the maximum amount possible of information about gender or occupation that can be realistically extracted from latent features, in the spirit of the maximum information that can be extracted under computational constraints~\citep{Xu2020A}. Hence, it is crucial to optimize an expressive model to the highest possible performance.  

In the latent space of the feature extractor, the decisions are usually computed with a linear classifier. However, when operating on latent space with removed concepts or when working with a task on which the Transformer head has not been fine-tuned, nothing guarantees that the task can be solved with linear probes. Hence, we chose a more expressive model, namely a two-layer perceptron with ReLU non-linearities, and architecture $768\rightarrow 128\rightarrow C$ with $C$ the number of classes. The network is trained by minimizing the categorical cross-entropy on the \textit{train set}, with \textit{Softmax} activation to convert the logits into probabilities. We use Adam with a learning rate chosen between $10^{-5}$ and $10^{-1}$ with cross-validation on the \textit{validation set}. Note that the chapter reports the accuracy on the \textit{test set}.  

The gender classifier $c_{g}$ and the occupation classifier $\tilde{c}$ are both trained with this same protocol.

\subsection{Neutralizing the embedding with importance-based concept removal} 

In this final section, we propose to eliminate $k$ concepts that exhibit significant gender dependence within the final embedding matrix, with the objective of improving its gender neutrality. This provides us with an improved embedding which can serve as a base to learn a fairer classifier to predict downstream tasks and, thus, a more gender-neutral model overall.

This concept-removal procedure consists of choosing the $k$ concepts that maximize the ratio between the importance for gender prediction and for the task. This graphically corresponds to searching for the concepts whose angle with respect to the line of equal importance for gender and task prediction are the highest (see Fig.~\ref{taco:fig:conceptimp}). This strategy allows us to work on a trade-off between the model's performance and its fairness.

In practice, we sort the concepts $\vu_1$,$\vu_2$,....$\vu_r$ according to the second strategy in descending order and delete the first $k$ concepts. This essentially leaves us with a concept base $\tilde{\vu} = (\vu_{k+1},..., \vu_r)$. This same transformation can be applied to $\vw$'s rows, yielding $\tilde{\vw}$.
From the matrix obtained from the matrix product $\tilde{\vu} \tilde{\vw}$, we retrain a classifier $\tilde{c}$ to predict $\vy$ (refer to Appendix~\ref{taco:modelsdetailsclassif} for more details).

It is interesting to note that, because we are causally removing information from the latent space, it is impossible for any model to recover it from the truncated space -- \textit{i.e.} we are not limited by the family of models that's intervening~\citep{ravfogel2023loglinear}. Thus, we're trading off the precision of the methods in the state-of-the-art~\citep{RLACE,belrose2023leace,ravfogel2020null,spectral-removal,shao2023erasure} for interpretability of the removed information.

\paragraph{Remark:} The more concepts we remove, the less we will be able to predict the gender, so the fairer the model will be. However, this also entails a decrease on the amount of information that is encoded in the embedding and that can be used for downstream tasks.
The parameter $k$ must therefore be chosen to comply with our specific application's constraints.

\section{Results}

\subsection{Dataset and models}

To illustrate our method, we use the \textit{Bios} dataset \citep{de2019bias}, which  contains about 440K biographies with labels for the genders of the authors, and their 28 occupations. For more details on this dataset, please see section \ref{biosdataset}. 
From the \textit{Bios} dataset, we also create a second dataset \textit{Bios-neutral} by removing all the explicit gender indicators. \textit{Bios-neutral} will be used as a baseline to which we can compare our method.

For both datasets, we train a RoBERTa model \citep{liu2019roberta}, a DistilBERT model \citep{sanh2019distilbert}, and a DeBERTa model \citep{he2020deberta} to predict the occupation task.

\paragraph{Creating the \textit{Bios-neutral} dataset}

From \textit{Bios}, we create a new dataset without explicit gender indicators, called \textit{Bios-neutral}. We extend the method defined in Section \ref{subs:genderneutral} as follows: 

At first, we tokenize the dataset, then: \vspace{-2mm} \begin{itemize}
    \item If the token is a first name, we replace it with "\textit{Sam}" (a common neutral first name). 
    \item If the token is a dictionary key (on the dictionary defined below), we replace it with its values. 
\end{itemize}

To determine whether the token is a first name, we use the list of first names:  \href{https://www.usna.edu/Users/cs/roche/courses/s15si335/proj1/files.php\%3Ff=names.txt&downloadcode=yes}{usna.edu}. The dictionary used in the second part is based on the one created in \citep{field2020unsupervised}.

\subsubsection{Models training}

\paragraph{RoBERTa base training}
We use a RoBERTa model \citep{liu2019roberta}, which is based on the Transformer architecture and is pre-trained with the Masked language modeling (MLM) objective. 
We specifically used a RoBERTa base model pre-trained by HuggingFace. All information related to how it was trained can be found in  \citep{liu2019roberta}. It can be remarked, that a very large training dataset was used to pre-train the model, as it was composed of five datasets: \emph{BookCorpus} \citep{moviebook}, a dataset containing 11,038 unpublished books; \emph{English Wikipedia} (excluding lists, tables and headers); \emph{CC-News} \citep{10.1145/3340531.3412762} which contains 63 millions English news articles crawled between September 2016 and February 2019; \emph{OpenWebText} \citep{radford2019language}  an open-source recreation of the WebText dataset used to train GPT-2; \emph{Stories} \citep{trinh2018simple} a dataset containing a subset of CommonCrawl data filtered to match the story-like style of Winograd schemas. Pre-training was performed on these data by randomly masking 15\% of the words in each of the input sentences and then trying to predict the masked words.

\paragraph{DistilBERT base training}
DistilBERT \citep{sanh2019distilbert} is a Transformer architecture derivative from but smaller and faster than the original BERT \citep{devlin2018bert}. 
This model is commonly used to do text classification. DistilBERT is trained on BookCorpus \citep{moviebook} (like BERT), a dataset consisting of 11,038 unpublished books and English Wikipedia (excluding lists, tables and headers), using the BERT base model as a teacher.

\paragraph{DeBERTa base training}
DeBERTa enhances the BERT and RoBERTa models by incorporating disentangled attention and an improved mask decoder. These enhancements enable DeBERTa to surpass RoBERTa in most Natural Language Understanding (NLU) tasks, utilizing 80GB of training data. In the DeBERTa V3 iteration (used here),  DeBERTa is optimized efficiently through ELECTRA-Style pre-training and Gradient Disentangled Embedding Sharing. This version, compared to its predecessor, shows significant performance gains in downstream tasks. 

The base model of DeBERTa V3 features 12 layers with a hidden size of 768. It possesses 86M core parameters, and its vocabulary includes 128K tokens, adding 98M parameters to the Embedding layer. This model was trained on 160GB of data, similar to DeBERTa V2.

\paragraph{Occupation prediction task}
After pre-training RoBERTa/DistilBERT/DeBERTa parameters on this huge dataset, we then trained it on the 400.000 biographies of the  \emph{Bios} dataset. 
The training was performed with PyTorch on 2 GPUs (Nvidia Quadro RTX6000 24GB RAM) for 10/3/3 epochs (we train 15 epochs, but we perform early stopping, giving the best results at 10/3/3 epochs) with a batch size of 8 observations and a sequence length of 512 words. The optimizer was Adam with a learning rate of 1e-6, $\beta_1 = 0.9$, $\beta_2 = 0.98$, and $\epsilon =1e6$. 
We split the dataset into 70\% for training, 10\% for validation, and 20\% for testing.

\subsection{Results analysis - Quantitative part}\label{taco:sec:results-analysis-quantitative}

\begin{figure}
    \begin{subfigure}{.46\textwidth}
        \centering
        \includegraphics[width=1\linewidth,left]{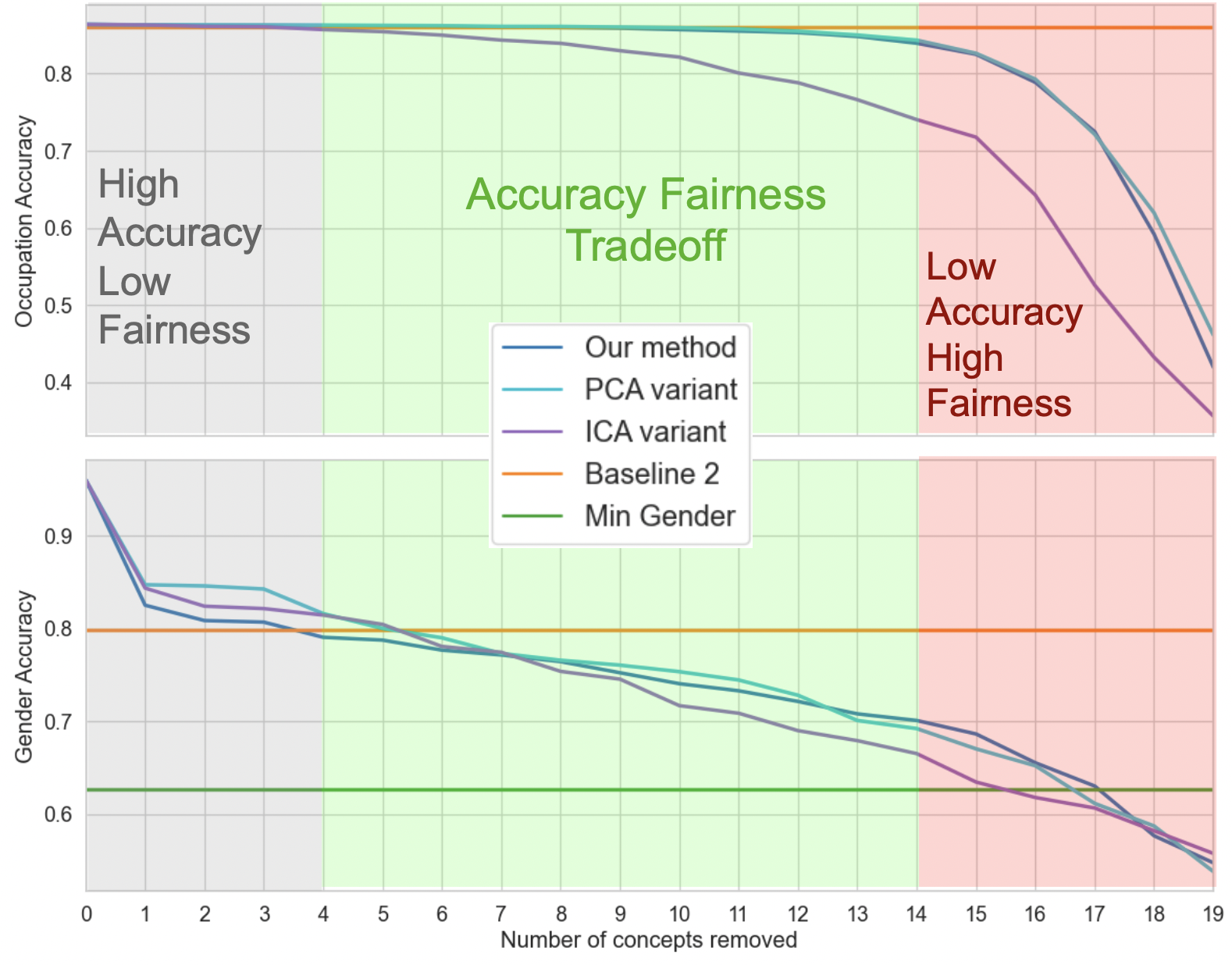}
        \caption{}
        \label{taco:fig:accresults}
    \end{subfigure}
    \hspace*{\fill}
    \begin{subfigure}{.54\textwidth}
        \centering
        \includegraphics[width=1\linewidth,right]{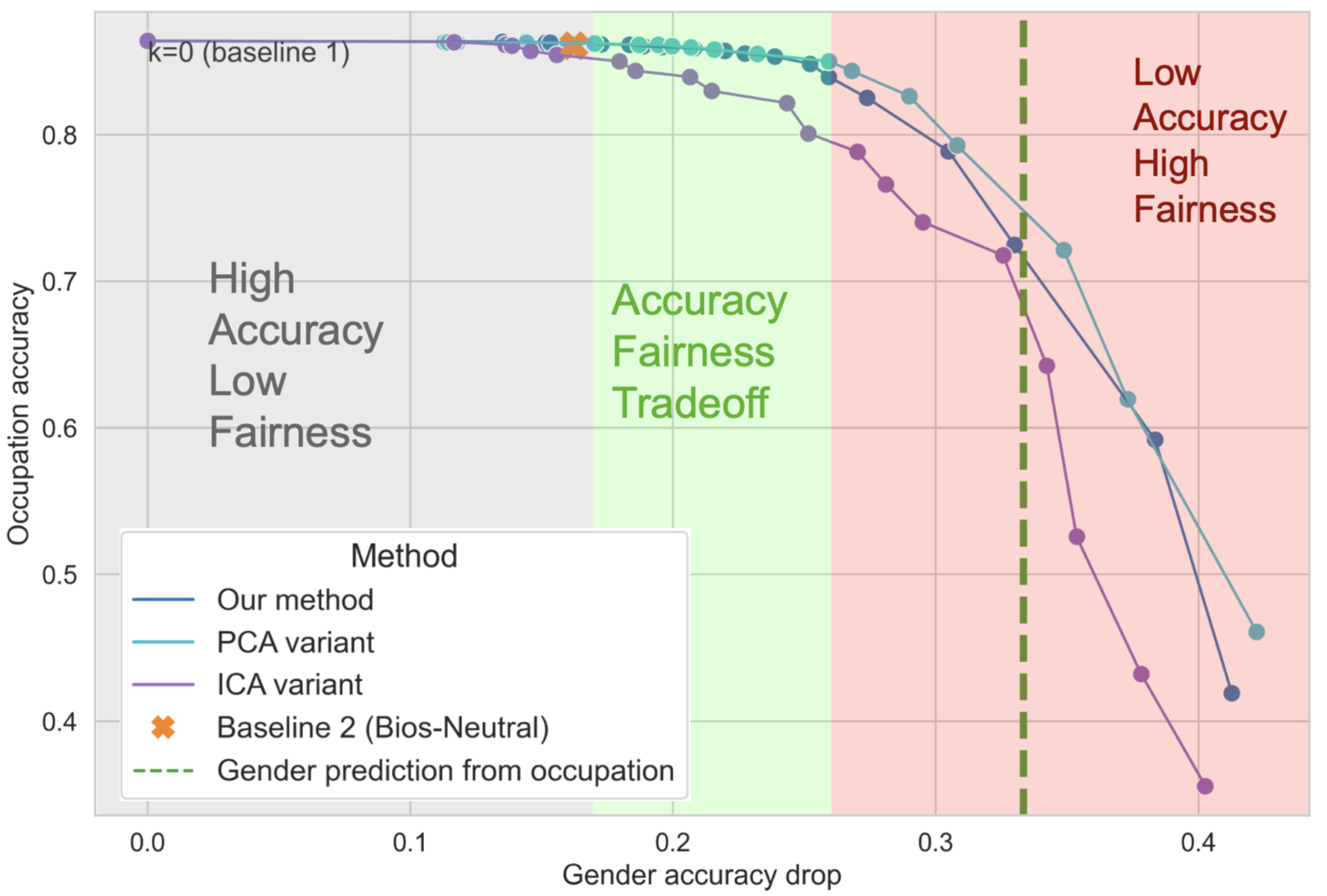}
        \caption{}
        \label{taco:fig:accpareto}
    \end{subfigure}
    \caption{\textbf{Accuracy drop when $k$ concepts are removed using our method with $r = 20$ dimensions on RoBERTa model.} In Figure~\ref{taco:fig:accresults}, (Top) the occupation accuracy, (Bottom) the gender accuracy.
    Figure~\ref{taco:fig:accpareto} shows occupation accuracy as a function of gender accuracy drop. The gender accuracy drop is reported relatively to the \texttt{Baseline1}, the classifier trained on \textit{Bios} with all the concepts.}
    \label{taco:fig:allresultsroberta}
\end{figure}

\begin{figure}
    \begin{subfigure}{.46\textwidth}
        \centering
        \includegraphics[width=1\linewidth,left]{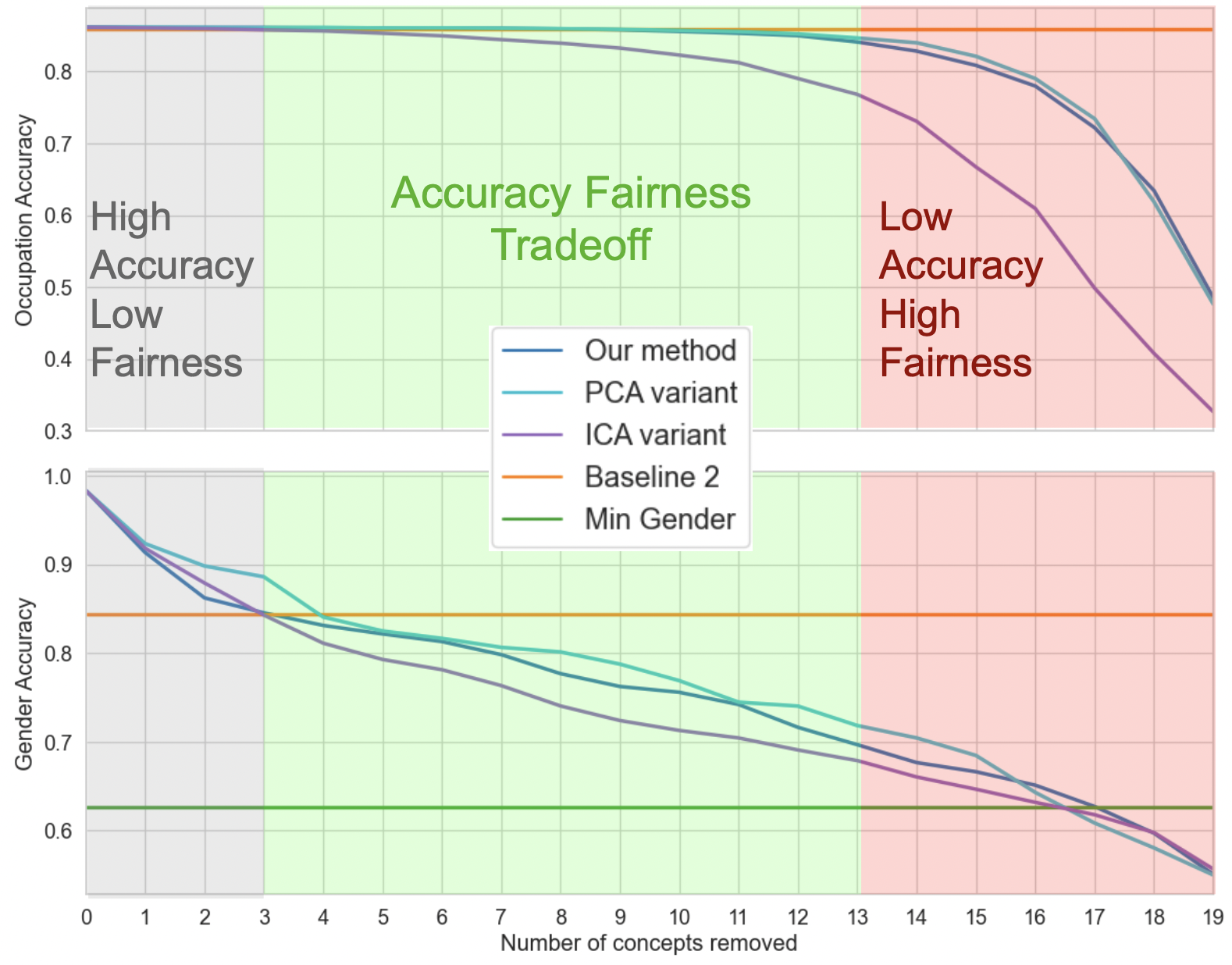}
        \caption{}
        \label{taco:fig:accresults_20distilbert}
    \end{subfigure}
    \hspace*{\fill}
    \begin{subfigure}{.54\textwidth}
        \centering
        \includegraphics[width=1\linewidth,right]{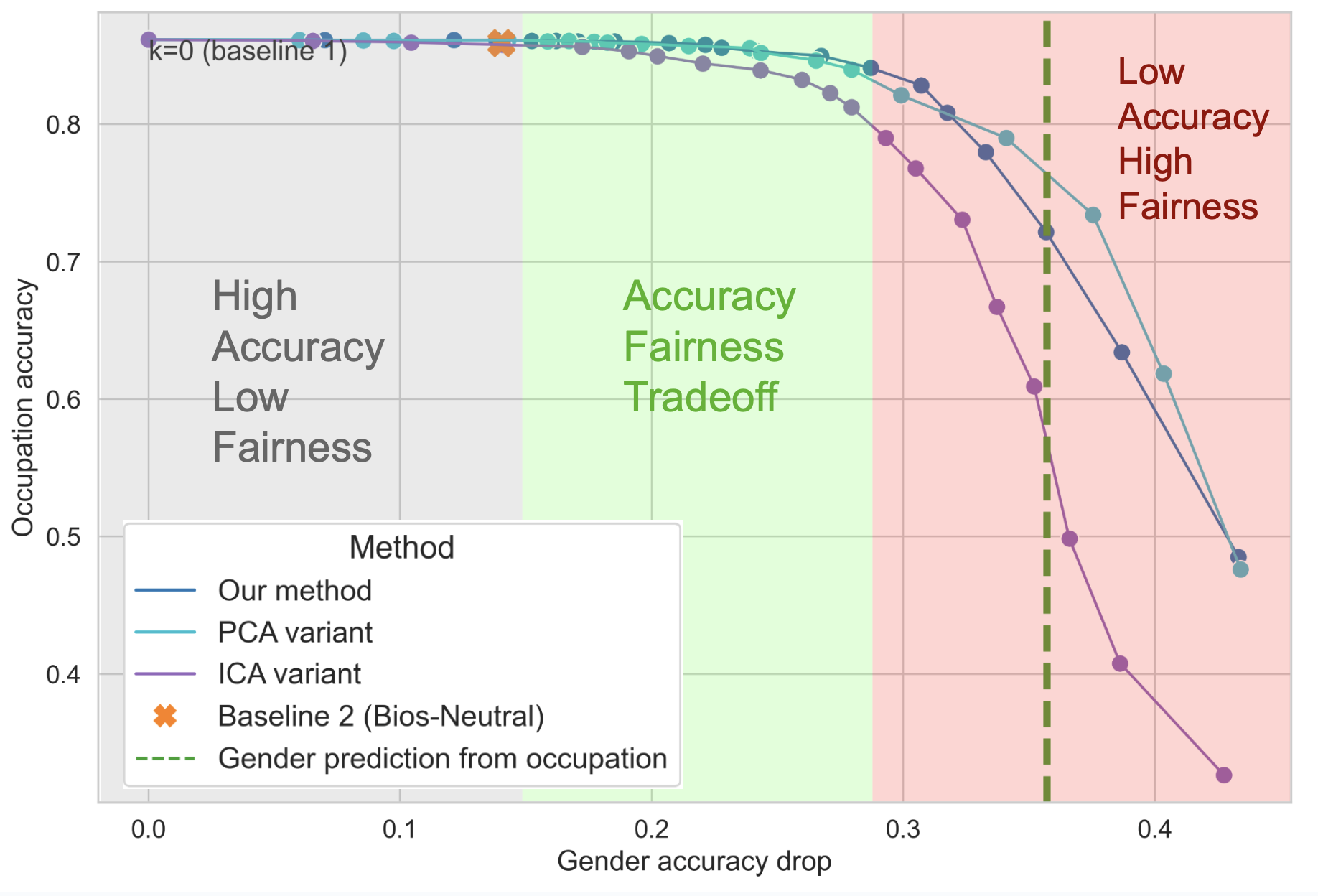}
        \caption{}
        \label{taco:fig:accpareto_20distilbert}
    \end{subfigure}
    \caption{\textbf{Accuracy drop when $k$ concepts are removed using our method with $r = 20$ dimensions on DistilBERT model.} In Figure~\ref{taco:fig:accresults_20distilbert}, (Top) the occupation accuracy, (Bottom) the gender accuracy. Figure~\ref{taco:fig:accpareto_20distilbert} shows occupation accuracy as a function of gender accuracy drop. The gender accuracy drop is reported relatively to the \texttt{Baseline1}, the classifier trained on \textit{Bios} with all the concepts.}
    \label{taco:fig:allresults_20distilbert}
\end{figure}

\begin{figure}
    \begin{subfigure}{.46\textwidth}
        \centering
        \includegraphics[width=1\linewidth,left]{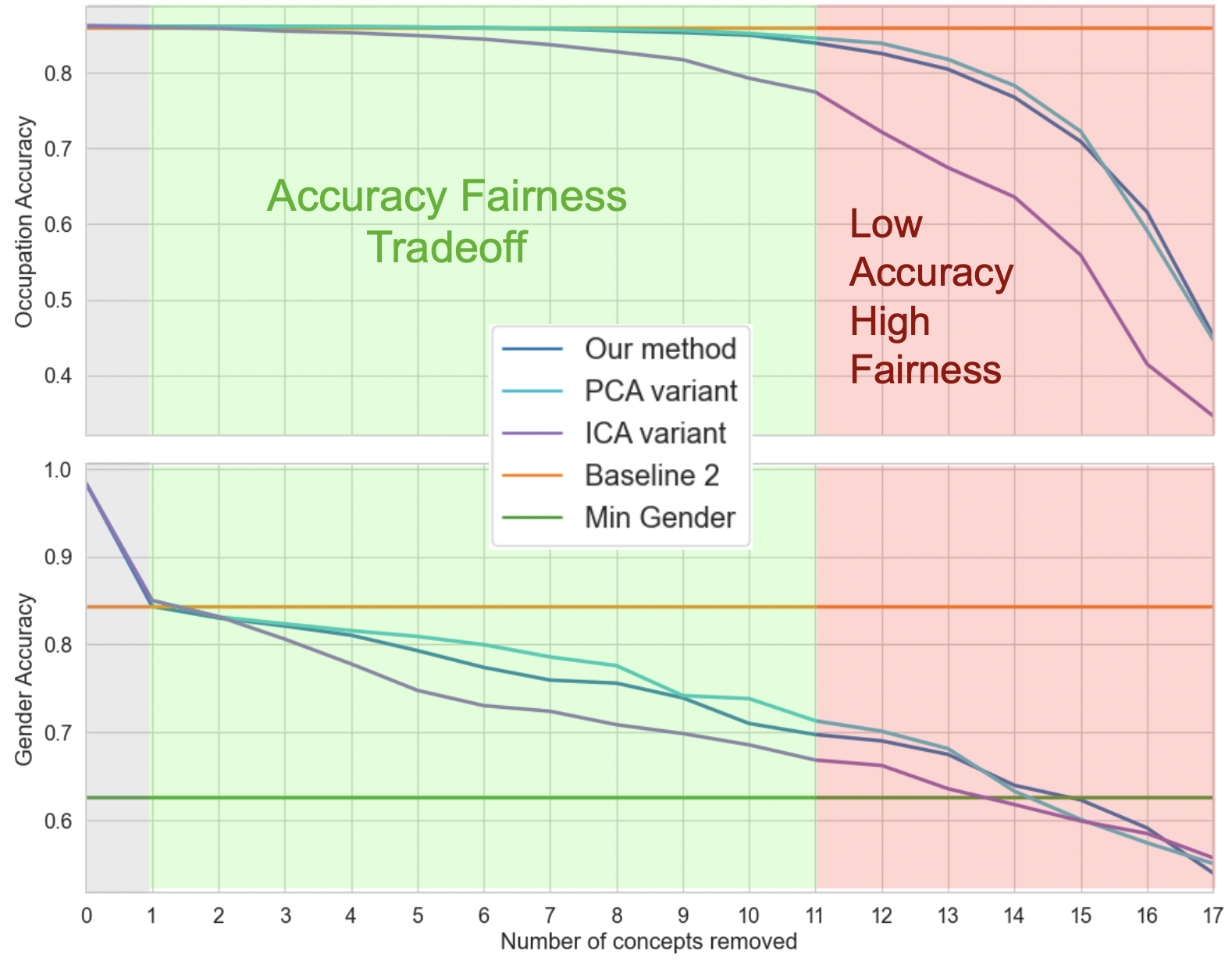}
        \caption{}
        \label{taco:fig:accresults_18distilbert}
    \end{subfigure}
    \hspace*{\fill}
    \begin{subfigure}{.54\textwidth}
        \centering
        \includegraphics[width=1\linewidth,right]{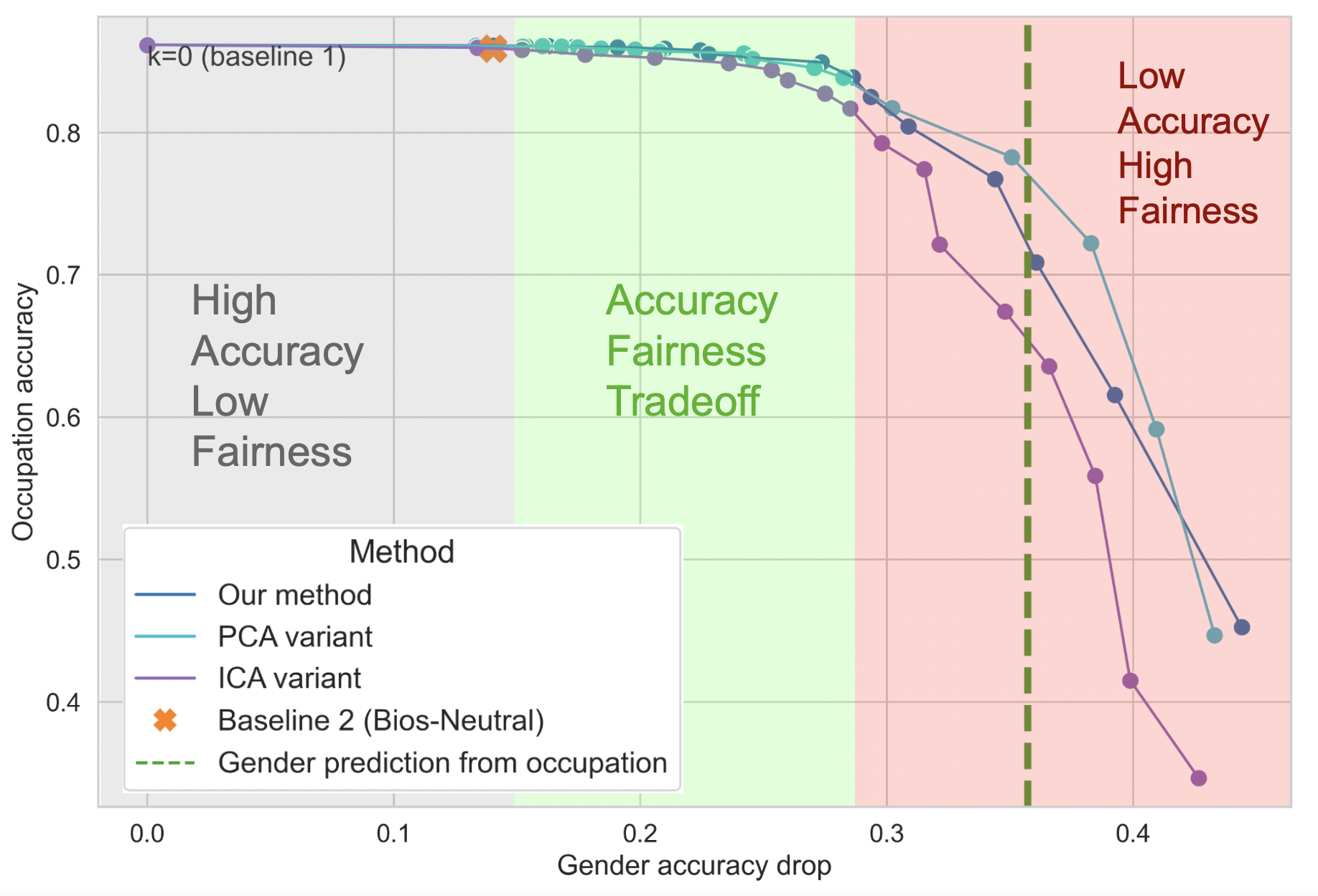}
        \caption{}
        \label{taco:fig:accpareto_18distilbert}
    \end{subfigure}
    \caption{\textbf{Accuracy drop when $k$ concepts are removed using our method with $r = 18$ dimensions on DistilBERT model.} In Figure~\ref{taco:fig:accresults_18distilbert}, (Top) the occupation accuracy, (Bottom) the gender accuracy. Figure~\ref{taco:fig:accpareto_18distilbert} shows occupation accuracy as a function of gender accuracy drop. The gender accuracy drop is reported relatively to the \texttt{Baseline1}, the classifier trained on \textit{Bios} with all the concepts.}
    \label{taco:fig:allresults_18distilbert}
\end{figure}

\begin{figure}
    \begin{subfigure}{.46\textwidth}
        \centering
        \includegraphics[width=1\linewidth,left]{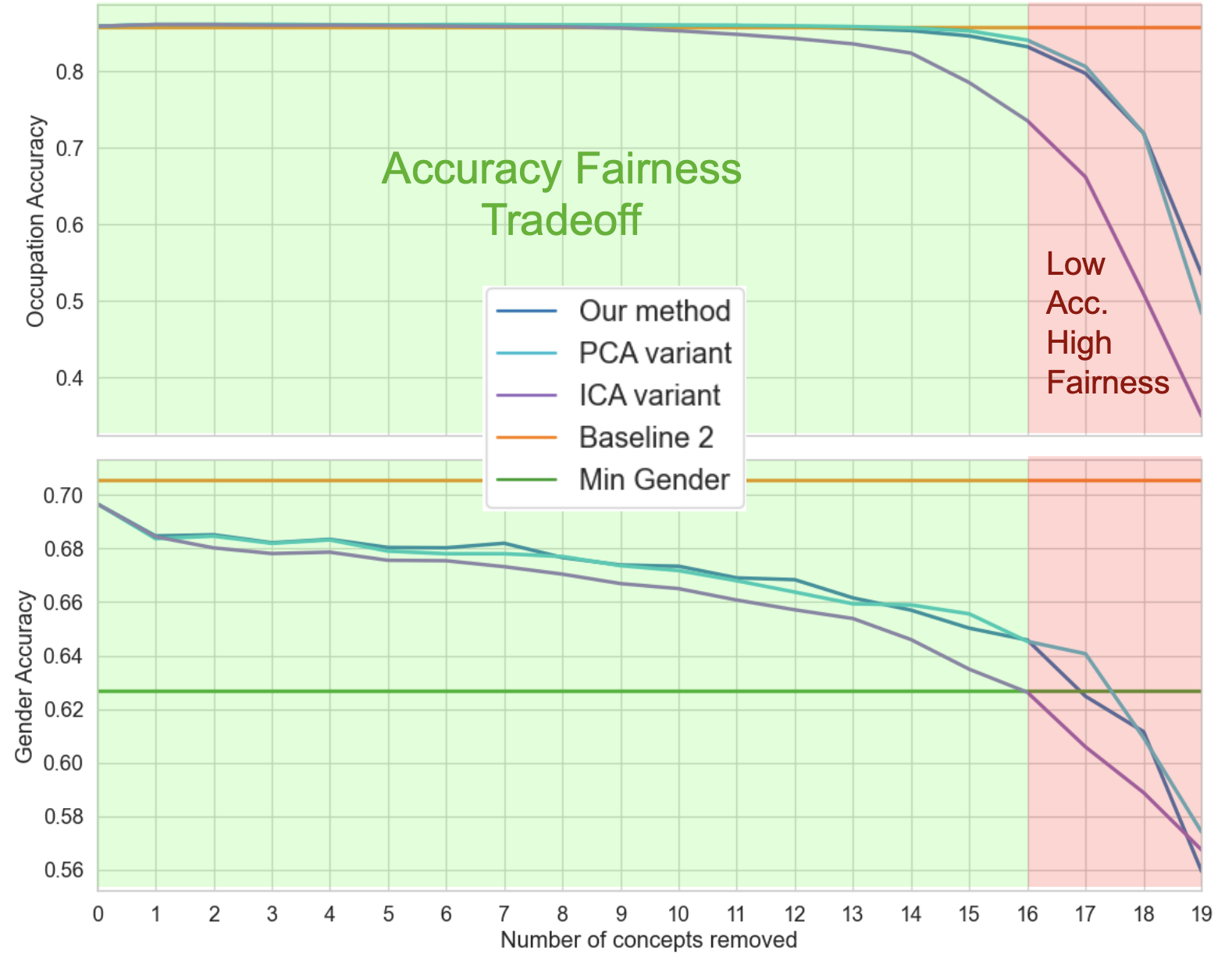}
        \caption{}
        \label{taco:fig:accresults_20deberta}
    \end{subfigure}
    \hspace*{\fill}
    \begin{subfigure}{.54\textwidth}
        \centering
        \includegraphics[width=1\linewidth,right]{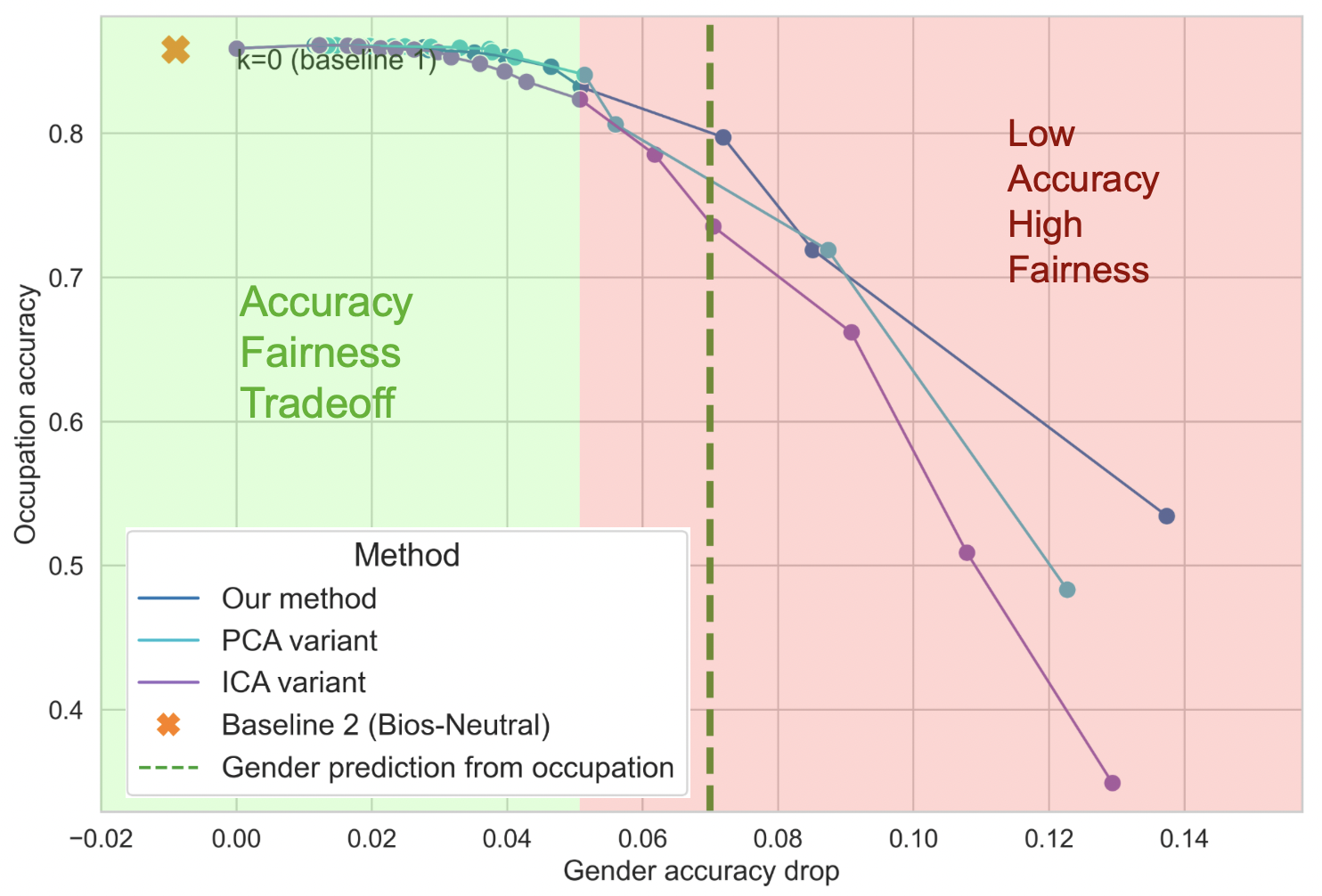}
        \caption{}
        \label{taco:fig:accpareto_20deberta}
    \end{subfigure}
    \caption{\textbf{Accuracy drop when $k$ concepts are removed using our method with $r = 20$ dimensions on DeBERTa model.} In Figure~\ref{taco:fig:accresults_20deberta}, (Top) the occupation accuracy, (Bottom) the gender accuracy. Figure~\ref{taco:fig:accpareto_20deberta} shows occupation accuracy as a function of gender accuracy drop. The gender accuracy drop is reported relatively to the \texttt{Baseline1}, the classifier trained on \textit{Bios} with all the concepts.}
    \label{taco:fig:allresults_20deberta}
\end{figure}

In this section, we show the results of our method for the RoBERTa model when its \texttt{CLS} embedding is decomposed with the SVD into $r=20$ dimensions (Figure \ref{taco:fig:allresultsroberta}), the DistilBERT model, when its \texttt{CLS} embedding is decomposed with the SVD into $r=20$ dimensions (Figure \ref{taco:fig:allresults_20distilbert}), then when it is decomposed into $r=18$ dimensions (Figure \ref{taco:fig:allresults_18distilbert}). And the results for the DeBERTa model when its \texttt{CLS} embedding is decomposed with SVD into $r=20$ dimensions (Figure \ref{taco:fig:allresults_20deberta}).

We have chosen to look at how our method varies according to the different models to check that the results of our method generalize well. We have also chosen to look at how the same model varies according to the number of dimensions chosen during the decomposition. While the theoretical SVD decomposition is deterministic, our choice of the Sparse SVD from \texttt{scipy} does produce different concepts depending on the choice of $r$. In this section, we also take this into account by trying different values of $r$.

Initially, we will examine the progression of our model's performance for each dimension removed (one by one) utilizing our methodology. Following the precepts of information theory, we will assess the effectiveness of our bias mitigation strategy by measuring the model's capacity to predict the gender and the task as we progressively remove concepts.

We showcase the outcome of removing the $k$ most important concepts for $k \in \llbracket 0, 19  \rrbracket$ for Figures \ref{taco:fig:allresultsroberta}, \ref{taco:fig:allresults_20distilbert} and \ref{taco:fig:allresults_20deberta}, and  $k \in \llbracket 0, 17 \rrbracket$ for Figure \ref{taco:fig:allresults_18distilbert}
in terms of different metrics: accuracy on gender prediction and on predicting occupation (downstream task).  
We compare our technique to two baselines: \texttt{Baseline1} -- models trained on the \textit{Bios} dataset (i.e. when $k=0$) --, and \texttt{Baseline2} -- models trained on the \textit{Bios-neutral} dataset. 

For a more comprehensive analysis, we extend our baseline by incorporating variations of our method with other decompositions. The purpose of this addition is to observe how our method performs when using different representations of the concepts on the latent space.
We include the Principal Component Analysis (PCA) \citep{wold1987principal} decomposition, a statistical technique that transforms the data into a set of orthogonal components, ordered by the amount of variance they capture. PCA shares the same orthogonal dimension properties as SVD. Additionally, we include the Independent Component Analysis (ICA) \citep{comon1994independent} decomposition, a computational technique for separating a multivariate signal into additive, independent components. Unlike PCA and SVD, ICA does not impose orthogonality constraints but offers statistically independent axes. For both additional decompositions, we retain the top $r$ most significant dimensions, as with SVD, to facilitate a better comparison with our standard method.

For each of these baselines and our method, we present the averaged results over five experimental runs.

From Figure \ref{taco:fig:allresultsroberta}, simply by removing one concept, we achieve significant gains in the gender neutrality of RoBERTa model, as it predicts gender with much lower accuracy (82.3\% compared to the initial 96\%) while maintaining its performance in occupation prediction (accuracy decreasing from 86.4\% to 86.3\%). Starting from the removal of the fourth dimension, the accuracy of gender prediction falls below the threshold of \texttt{Baseline2}, with 79.4\% accuracy compared to 79.8\% for \texttt{Baseline2}, while maintaining an accuracy of 86.3\% for occupation (compared to 86\% for \texttt{Baseline2}). Moreover, our method, in addition to being explainable, time-efficient, and cost-effective, exhibits greater effectiveness. After removing these first four concepts, we enter a regime where removing concepts continues to gain us fairness (as the accuracy of the gender task continues to fall), while the accuracy of the occupation remains unchanged. This regime is therefore the one with the best trade-off between accuracy and fairness (shown in green here). \textit{More precisely, the perfect trade-off in this case would be to remove the first 13 dimensions.}
From concept 14 onwards, the accuracy of occupation begins to decline, as we look more and more at concepts with a high importance for occupation (according to Sobol), and we enter a new regime where we gain more and more in fairness, but where the accuracy of our task collapses. It may be worthwhile deleting a fairly large number of concepts to get into this zone, but only if we are willing to pay the price in terms of prediction quality. 

Figures \ref{taco:fig:allresults_20distilbert} and \ref{taco:fig:allresults_18distilbert}, for the DistilBERT model (SVD decomposition with $r=20$ or $r=18$), give very similar results to those obtained for the RoBERTa model. In the same way, we find the 3 regimes, with firstly a high level of accuracy but with a level of fairness lower than that of \texttt{Baseline2}, then an ideal regime, with a high level of accuracy and a level of fairness better than \texttt{Baseline2}, and finally a regime where the level of fairness continues to increase as concepts are removed, but where accuracy collapses.

In the case of the DeBERTa model, it's a little different, since the \texttt{CLS} embedding already contains very little gender information, unlike the other models. There is, therefore, less room for improvement with our method since the \texttt{CLS} embedding is already not very biased according to our metric. What's more, \texttt{Baseline2} is not an interesting baseline in this case, since it is no better than the base model. However, we can see that our method still works on this model since, according to figure \ref{taco:fig:allresults_20deberta}, we are able to lower the accuracy for gender prediction without affecting the model's performance on the first 12 concepts removed. 

It is important to note that in this specific application, the optimal fairness-accuracy trade-off corresponds to accuracy on gender prediction of 62\% as that's the accuracy one gets when learning a model to predict the gender with the occupation labels as only information.

In Figures \ref{taco:fig:accpareto}, \ref{taco:fig:accpareto_20distilbert}, \ref{taco:fig:accpareto_18distilbert} and \ref{taco:fig:accpareto_20deberta}, we observe the Pareto front between task accuracy and the loss of gender accuracy. This allows us to better compare the various versions of our method according to the decomposition chosen for each model. The comparison among these three decompositions yields similar results regardless of the model considered. The variation of our method using ICA decomposition is less effective than the others: it is more challenging to reduce gender accuracy without significantly affecting task accuracy with this decomposition. The variation using PCA decomposition produces results similar to our method with SVD decomposition; however, SVD performs slightly better in the "Accuracy Fairness Tradeoff" regime, while PCA is marginally superior in the "Low accuracy High Fairness" regime. Nonetheless, it is more logical to choose our method with SVD as it better addresses the real need where maintaining task accuracy is crucial.

\subsection{Concepts removed analysis - Explainability part}

In this section, we interpret the dimensions of the Singular Value Decomposition (SVD) as valuable concepts for predicting our two tasks: gender and occupation. For this, we use the interpretability part of the COCKATIEL \citep{jourdan2023cockatiel} method on concepts/dimensions previously found with the SVD and ranked with the Sobol method by importance for gender and occupation.

\paragraph{Concept interpretability} COCKATIEL modifies the Occlusion method~\citep{zeiler2014visualizing}, which operates by masking each word and subsequently observing the resultant impact on the model’s output. In this instance, to infer the significance of each word pertaining to a specific concept, words within a sentence are obscured, and the influence of the modified sentence (devoid of the words) on the concept is measured. This procedure can be executed at either the word or clause level, meaning it can obscure individual words or entire clauses, yielding explanations of varying granularity contingent upon the application.

\begin{figure}
    \centering
\includegraphics[width=0.7\linewidth]{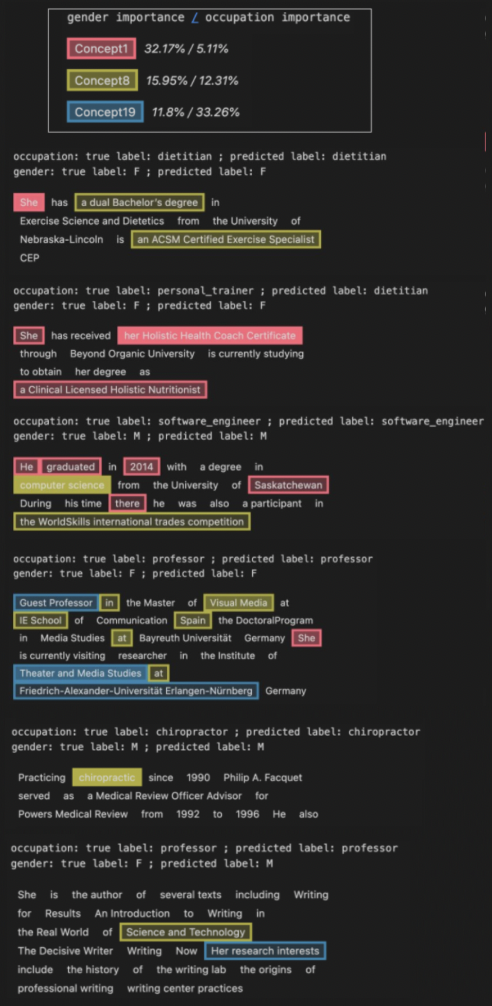}
    \caption{\textbf{Understanding concepts extracted from the model trained on \textit{Bios}.} Concept 1 prominently features explicit gender indicators like '\textit{she}' and '\textit{his}', crucial for gender identification but irrelevant for occupation. Concept 8 integrates information about gender-unbalanced professions, aiding in predicting both, an individual’s gender and occupation. Concept 19 focuses on information related to the professor occupation, a gender-balanced class and most represented in the dataset.}
    \label{taco:fig:explanation}
\end{figure}

\begin{figure}
    \centering
\includegraphics[width=0.7\linewidth]{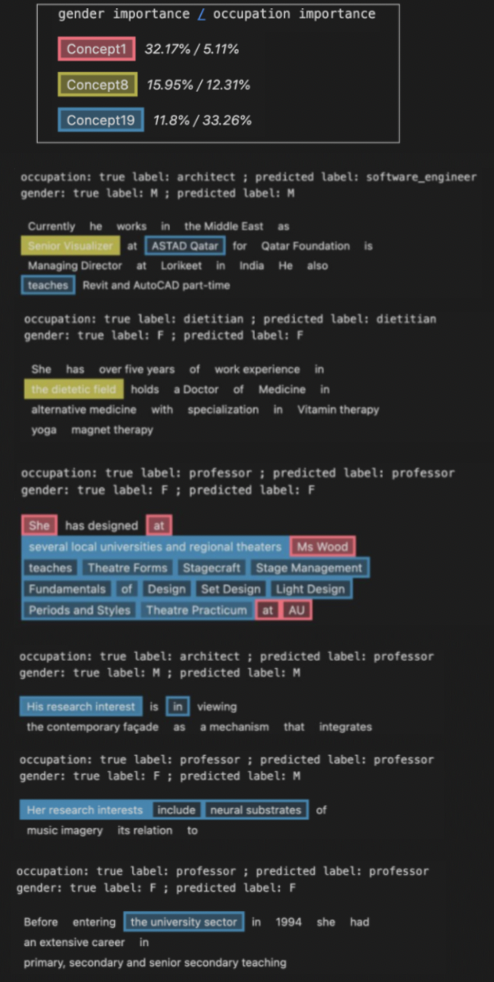}
    \caption{\textbf{Understanding concepts extracted from the model trained on \textit{Bios}.} Concept 1 prominently features explicit gender indicators like '\textit{she}' and '\textit{his}', crucial for gender identification but irrelevant for occupation. Concept 8 integrates information about gender-unbalanced professions, aiding in predicting both, an individual’s gender and occupation. Concept 19 focuses on information related to the professor occupation, a gender-balanced class and most represented in the dataset.}
    \label{taco:fig:explanation2}
\end{figure}

\paragraph{Concepts interpretation on RoBERTa model}
For the dimensions of the SVD to be meaningful for explanation, they must first be meaningful for the RoBERTa model's prediction, whether it be the RoBERTa model predicting gender or the one predicting occupation. We rank the concepts w.r.t their removal order, and we focus on three important ones: \\
\textbf{\textit{Concept 1}}, with a gender importance of 32\%, and an occupation importance of 5\%; 
\textbf{\textit{Concept 8}}, with a gender importance of 16\%, and an occupation importance of 12\%; 
\textbf{\textit{Concept 19}}, with a gender importance of 12\%, and an occupation importance of 33\%.

These three concepts are crucial for the prediction of at least one of our two tasks. Moreover, they are representative of our problem statement as \textbf{\textit{concept 1}} is more important for gender, \textbf{\textit{concept 8}} is as important to gender as occupation, and \textbf{\textit{concept 19}} is more important for occupation. We illustrate several examples of explanations for these three concepts in Figure \ref{taco:fig:explanation} to enable qualitative analysis.

According to the importance per dimension calculated in part 2 of our method, \textbf{\textit{concept 1}} is almost exclusively predictive of gender. The most significant words for the concept of this dimension are explicit indicators of traditional gender.

\textbf{\textit{Concept 8}} is capable of predicting both gender and occupation. Within its concept, we find examples of biographies predominantly related to unbalanced professions such as dietitian (92.8\% women), chiropractor (26.8\% women), software engineer (16.2\% women), and model (83\% women). It is logical that the concept allowing the prediction of these professions also enables the prediction of gender, as it is strongly correlated in these particular professions.

\textbf{\textit{Concept 19}}, which strongly predicts occupation, focuses on biographies predicting professors (constituting more than 29.7\% of all individuals). This class is gender-balanced (with 45.1\% women), providing little information on gender.

\begin{figure}
    \centering
\includegraphics[width=0.7\linewidth]{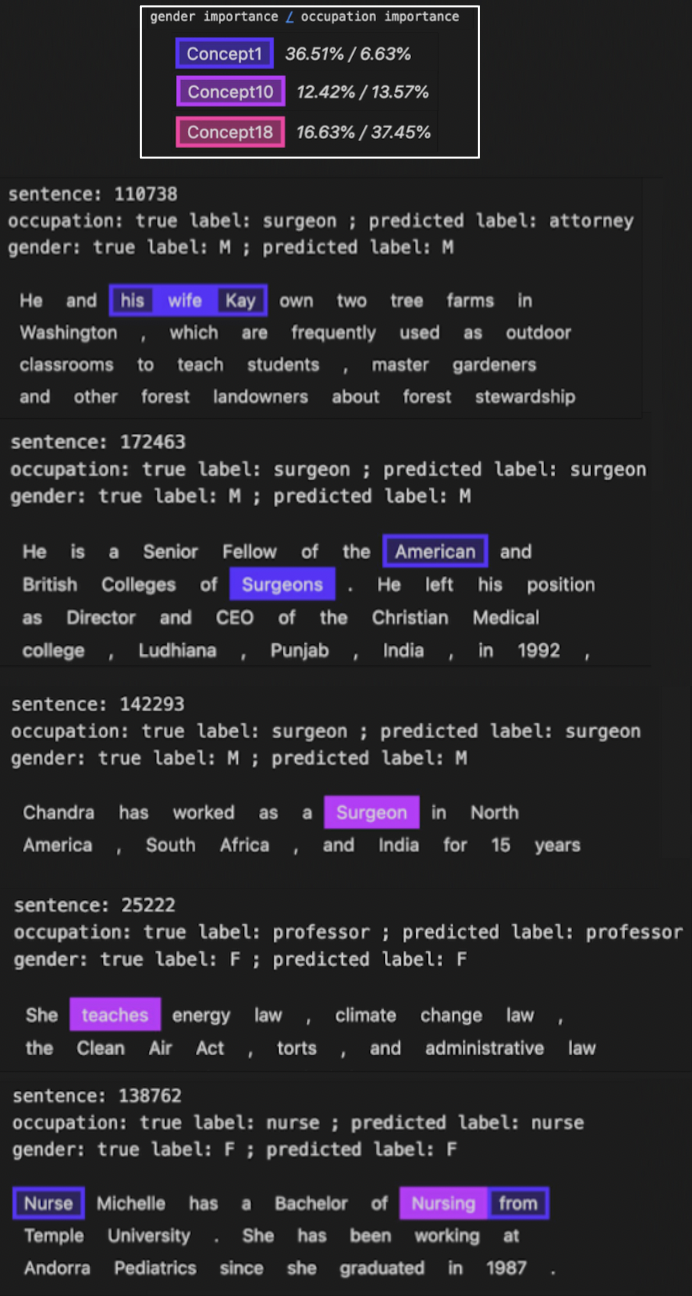}
    \caption{\textbf{Understanding concepts extracted from the DistilBERT model trained on \textit{Bios}} (for a $18$-concept SVD decomposition). \textbf{\textit{Concept 1}} prominently features explicit gender indicators like '\textit{she}' and '\textit{his}', crucial for gender identification but irrelevant for occupation, and some information on specific professions that are particularly gender imbalanced. \textbf{\textit{Concept 1O}} integrates information about gender-imbalanced professions, aiding in predicting both, an individual’s gender and occupation, and also some information on occupation in general. \textbf{\textit{Concept 18}} focuses on information related to the occupation in general.}
    \label{taco:fig:explanationdistilbert}
\end{figure}

\begin{figure}
    \centering
\includegraphics[width=0.7\linewidth]{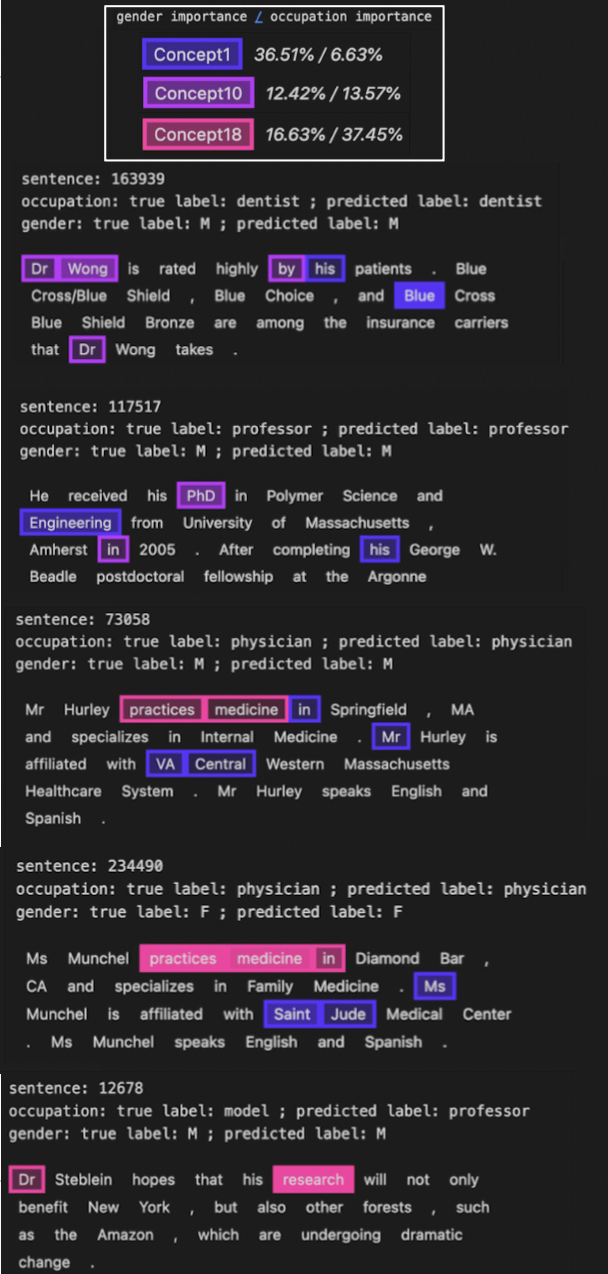}
    \caption{\textbf{Understanding concepts extracted from the DistilBERT model trained on \textit{Bios}} (for a $18$-concept SVD decomposition). \textbf{\textit{Concept 1}} prominently features explicit gender indicators like '\textit{she}' and '\textit{his}', crucial for gender identification but irrelevant for occupation, and some information on specific professions that are particularly gender imbalanced. \textbf{\textit{Concept 1O}} integrates information about gender-imbalanced professions, aiding in predicting both, an individual’s gender and occupation, and also some information on occupation in general. \textbf{\textit{Concept 18}} focuses on information related to the occupation in general.}
    \label{taco:fig:explanationdistilbert2}
\end{figure}

\paragraph{Concepts interpretation on DistilBERT model}
We also create explanations for the DistilBERT model for SVD decomposition with $r=18$ concepts. 

For the dimensions of the SVD to be meaningful for explanation, they must first be meaningful for the model's prediction, whether it be the model predicting gender or the one predicting occupation. We rank the concepts w.r.t. their removal order, and we focus on three important ones: 
\textbf{\textit{concept 1}}, with a gender importance of 36.51\%, and an occupation importance of 6.63\%; 
\textbf{\textit{concept 10}}, with a gender importance of  12.42\%, and an occupation importance of 13.57\%; 
\textbf{\textit{concept 18}}, with a gender importance of 16.63\%, and an occupation importance of 37.45\%.

These three concepts are crucial for the prediction of at least one of our two tasks. Moreover, they are representative of our problem statement as \textbf{\textit{concept 1}} is more important for gender, \textbf{\textit{concept 10}} is as important to gender as occupation, and \textbf{\textit{concept 18}} is more important for occupation. We illustrate several examples of explanations for these three concepts in Figure \ref{taco:fig:explanationdistilbert} to enable qualitative analysis. 

The dynamics of these 3 concepts are close to those observed for those of the RoBERTa model, but the differences between the concepts are less clear-cut.
For example, here \textit{\textbf{concept 1}}, which is the most important for gender but very unimportant for occupation, does indeed contain explicit gender indicators such as "he" or "she", but it also contains information on some very gender-unbalanced occupations such as nurse or surgeon. 
\textit{\textbf{Concept 10}}, which is about as important for gender as it is for occupation, contains information on most of the very gender-unbalanced occupations such as nurse or surgeon, but also some information on occupations in general, notably professor and physician, which are gender-balanced occupations and are particularly present in the dataset.
\textit{\textbf{Concept 18}}, which is the most important for occupation, less important for gender (but nonetheless not negligible, since it obtains a Sobol importance of 16.63\%), contains information on occupations in general. Mainly the most represented professions such as professor or physician, which are gender balanced, but also certain other more gender unbalanced professions.

Therefore, we can see that the concepts for the DistilBERT model are more porous and can display common keywords, unlike those for RoBERTa, which was more divided.

\subsection{Latent Space interpretation}

Uniform Manifold Approximation and Projection (UMAP) is a dimensionality reduction technique that is particularly effective for visualizing high-dimensional data in a low-dimensional space. By providing a two-dimensional compression of the latent space, UMAP helps us to directly appreciate the structure and relationships within the data encoded by the model. This visualization is crucial for understanding how different features and patterns are represented in the latent space, and can reveal insights that might not be apparent from quantitative metrics alone.

\begin{figure}
    \centering
    \begin{subfigure}[t]{1\textwidth}
        \centering
        \includegraphics[width=1\linewidth]{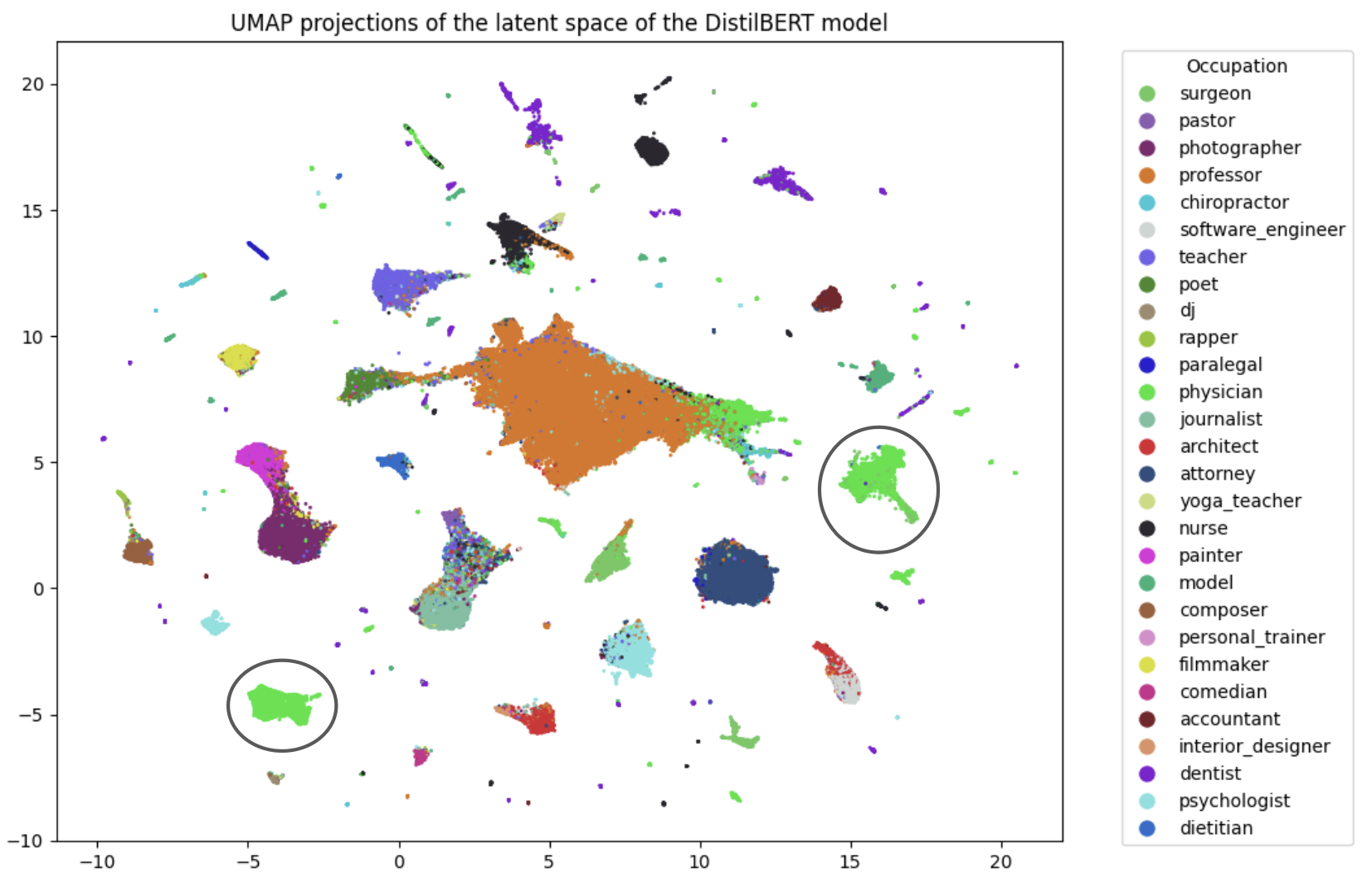}
        \label{taco:fig:UMAP_occ_distilbert}
    
    \begin{subfigure}[t]{1\textwidth}
        \centering

        \includegraphics[width=\linewidth]{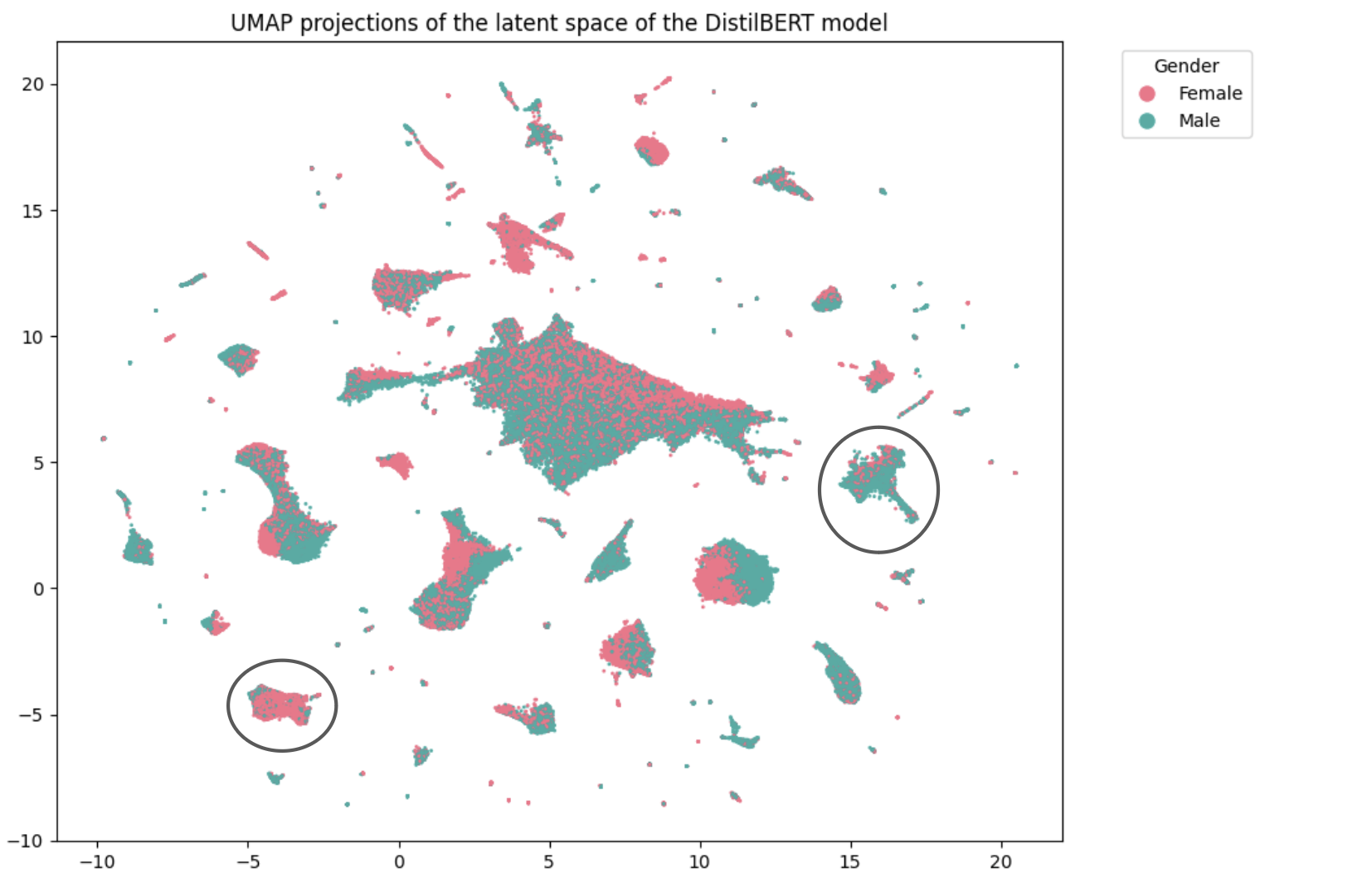}
        \label{taco:fig:UMAP_gen_distilbert}
    \end{subfigure}
        
    \end{subfigure}
    
    \caption{UMAP projections for occupation (top) and gender (bottom)}
    \label{taco:fig:UMAP_distilbert}
\end{figure}

In Figure \ref{taco:fig:UMAP_distilbert}, the projections reveal that the data are segmented into clusters based on occupation, while the gender distribution is more intermixed. This is expected, as the latent space was optimized by the DistilBERT model to enhance occupation classification. However, an intriguing observation is that the occupation of \textit{Physician} appears as two distinct clusters in the occupation-colored plot. When we examine the gender-colored plot, it becomes evident that these clusters correspond to gender, with one cluster representing \textit{Male Physicians} and the other \textit{Female Physicians}. For clarity, these clusters have been highlighted in the figures.
Furthermore, within the same occupation cluster, gender separation is often noticeable. This is particularly evident in the \textit{Attorney} cluster, marked in dark blue, where gender is clearly divided. These observations underscore the influence of both occupation and gender on the model's internal representations.

\section{Conclusion}

In this chapter, we introduced a method for neutralizing the presence of bias in Transformer model's embeddings in an interpretable and cost-efficient way through concept-based explainability. In this study, we present a method applied to gender-related information, yet inherently adaptable to other sensitive variables. Indeed, by removing concepts that are mostly influential for gender prediction, we obtain embeddings that rely considerably less on gender information for generating predictions and, at the same time, remain highly accurate at the task. Furthermore, by doing so via explainable AI, we perform interventions that can be easily interpreted by humans, and thus gain trust in the bias mitigation process and insights into what produced the bias in the first place. Finally, thanks to its low computation cost, it can be easily applied to pre-trained, fine-tuned, and even embeddings accessed through inference APIs, thus allowing our method to be applied to any model in a post-hoc manner.

Note that TaCo is implemented to work in PyTorch and is compatible with Hugging Face models. It is freely available on GitHub\footnote{\url{https://github.com/fanny-jourdan/TaCo}}.

%% file: Conclusion.tex
\chapter{Conclusion}

In this thesis, we have embarked on a comprehensive exploration of fairness in NLP, presenting novel methodologies and critiques that aim to advance the field towards more equitable and transparent machine learning models.

This thesis began with the development of an innovative algorithm designed to mitigate algorithmic biases in multi-class neural-network classifiers. Applied to high-risk NLP applications as per EU regulations, this method demonstrated its prowess in tempering biases while also outperforming traditional strategies in prediction accuracy and bias mitigation. The flexibility in adjusting regularization levels for each output class marks a significant advancement over existing binary model-focused unbiasing methods. However, challenges such as finding optimal regularization weights highlight the complex balance between accuracy and fairness.

The pivotal point in our research was the empirical analysis using the \textit{Bios} dataset, which revealed critical insights into the influence of training set size on discriminatory biases. This analysis brought to light the limitations and variability of traditional fairness metrics, particularly in smaller datasets. The unpredictable nature of biases and their dependency on chosen metrics underscored the inadequacy of current fairness metrics in capturing the complete picture of biases in AI systems. This led us to delve into the realm of explainability in AI, to gain a more comprehensive understanding of biases that go beyond what traditional metrics could decipher. Thus, our focus shifted towards developing methods that not only mitigate bias but also provide a deeper, more holistic understanding of it.

This shift in focus brought forth COCKATIEL, an innovative, model-agnostic post-hoc explainability method for NLP models. By effectively combining concept discovery, ranking, and interpretation, COCKATIEL aligns with human-conceptualized explanations and remains faithful to the underlying models. This method represents a significant stride in explainable AI, offering practical, qualitative examples for diverse models.

The thesis then ventured into bridging the domains of fairness and explainability through TaCo, our innovative approach to neutralizing bias in transformer model embeddings. By leveraging concept-based explainability, we were able to identify and remove concepts predominantly influential for gender prediction, thereby producing less biased embeddings. This method showcases the potential of explainability not just as a tool for understanding AI models but as a mechanism for enhancing fairness. By making interpretable interventions, we not only reduced bias but also facilitated a deeper understanding of its origins. This approach underscores the synergy between fairness and explainability, demonstrating how explainability can be effectively utilized to achieve fairness objectives in AI.

In summary, this thesis represents a significant interdisciplinary effort, intertwining explicability and fairness, to challenge and reshape the existing paradigms in NLP. The methodologies and critiques presented herein not only contribute to the ongoing discourse on fairness in machine learning but also offer practical solutions and insights that can be leveraged to build more equitable and responsible AI systems. The implications of this work have the potential to influence future research directions and inform the development of more just and accountable NLP technologies.


Future work stemming from this thesis will focus on deepening the understanding of concept-based XAI and interpretability in NLP models. The main limitation of this research area lies in the difficulty of interpreting these concepts; therefore, efforts will be concentrated on this aspect. Two main axes are linked to this limitation: the way these concepts are discovered (through various matrix decompositions) and how they are interpreted. First, we will create a framework to evaluate the relevance of concept decompositions (in terms of human understanding of the concepts and fidelity to the model) that underlie this type of conceptual explanation, in order to rank the best decompositions. Once the best method for discovering concepts is identified, it will be necessary to create visualizations of these concepts (which are currently just vectors) to make them interpretable by humans. The initial idea developed in this thesis was to examine the words in the most important sentences for the concept by highlighting them and then finding a common theme. This theme was identified manually; the first improvement would thus be to automate this search using a LLM to increase efficiency. We could also use visualization techniques such as word clouds to make the reading of the concept faster for humans. However, by focusing solely on words, we overlook one of the major strengths of these concepts, which is their ability to capture syntactic information. Therefore, we also wish to test whether the important sentences of a concept correspond to a specific structure within a list of syntactic structures, such as passive/active voice, the number of clauses per sentence, negation, etc. Finally, once the interpretation of the concepts is more advanced, we can return to creating methods addressing fairness, such as TaCo, using these new concepts.

%% file: Conclusion_fr.tex
\chapter{Conclusion en français}

Dans cette thèse, nous avons mené une étude approfondie sur l'équité dans le domaine du Traitement Automatique des Langues (TAL), en présentant des méthodologies innovantes et des analyses critiques visant à orienter le champ vers des modèles d'apprentissage automatique plus justes et transparents.

Cette thèse a débuté avec le développement d'une méthode conçue pour atténuer les biais algorithmiques dans les classificateurs neuronaux multiclasses. Utilisée dans des applications de TAL à haut risque conformément aux réglementations de l'UE, cette méthode a démontré son efficacité dans la modération des biais tout en surpassant les stratégies traditionnelles en termes de précision de prédiction et d'atténuation des biais. La flexibilité dans l'ajustement des niveaux de régularisation pour chaque classe de sortie marque une avancée significative par rapport aux méthodes de débiaisage existantes axées sur les modèles binaires. Cependant, des défis tels que la recherche de poids de régularisation optimaux soulignent l'équilibre complexe entre performance et équité.

Un moment clé de notre recherche a été l'analyse empirique du jeu de données \textit{Bios}, révélant des perspectives essentielles sur l'impact de la taille de l'ensemble d'entraînement sur les biais discriminatoires. Cette étude a mis en exergue les limites et la variabilité des métriques d'équités traditionnelles, en particulier pour les petits ensembles de données. La nature imprévisible des biais et leur dépendance vis-à-vis des métriques choisies ont mis en lumière les lacunes des métriques d'équités actuelles pour saisir pleinement l'étendue des biais dans les systèmes d'IA. Cela nous a incités à nous pencher sur l'explicabilité en IA, afin d'obtenir une compréhension plus globale des biais, allant au-delà des capacités des métriques traditionnelles. Ainsi, nous avons orienté nos efforts vers le développement de méthodes non seulement atténuant les biais, mais offrant également une compréhension plus approfondie et globale de ces derniers.

Ce changement d'orientation a conduit à la création de COCKATIEL, une méthode post-hoc d'explicabilité agnostique au modèle pour les modèles de TAL. En combinant efficacement la découverte de concepts, leur classement et leur interprétation, COCKATIEL s'aligne sur les explications conceptualisées par l'homme tout en restant fidèle aux modèles sous-jacents. Cette méthode marque une avancée significative dans l'IA explicable, offrant des exemples pratiques et qualitatifs pour divers modèles.

La thèse a ensuite exploré la jonction entre les domaines de l'équité et de l'explicabilité à travers TaCo, notre approche innovante pour neutraliser les biais dans les embeddings des modèles Transformers. En exploitant l'explicabilité basée sur les concepts, nous avons pu identifier et éliminer les concepts influençant principalement la prédiction de genre, produisant ainsi des embeddings moins biaisés. Cette méthode illustre le potentiel de l'explicabilité non seulement comme un outil de compréhension des modèles d'IA, mais aussi comme un mécanisme pour améliorer l'équité. En effectuant des interventions interprétables, nous avons réduit les biais tout en facilitant une compréhension plus profonde de leurs origines. Cette approche souligne la synergie entre l'équité et l'explicabilité, démontrant comment l'explicabilité peut être utilisée efficacement pour atteindre les objectifs d'équité en IA.

En résumé, cette thèse représente un effort interdisciplinaire significatif, entrelaçant explicabilité et équité, pour défier et remodeler les paradigmes existants en équité dans le TAL. Les méthodologies et critiques présentées ici contribuent non seulement au discours en cours sur l'équité dans l'apprentissage automatique, mais offrent également des solutions pratiques et des perspectives qui peuvent être exploitées pour construire des systèmes d'IA plus équitables et responsables. 

Dans la continuité directe de cette thèse, les travaux futurs se concentreront sur l'approfondissement de l'explicabilité basée sur les concepts et l'interprétabilité des modèles de TAL. La principale limitation de cette branche de recherche réside dans la difficulté d'interprétation de ces concepts ; par conséquent, l'effort sera mis sur cet aspect. Deux axes principaux sont liés à cette limitation : la manière dont ces concepts sont découverts (via les différentes décompositions matricielles) et la façon dont ils sont interprétés. Dans un premier temps, nous créerons un cadre permettant d'évaluer la pertinence des décompositions des concepts (en termes de compréhension humaine des concepts et de fidélité au modèle) à la base de ce type d'explications conceptuelles, afin de pouvoir classer les meilleures décompositions. Une fois la meilleure méthode de découverte des concepts identifiée, il sera nécessaire de créer des visualisations de ces concepts (qui ne sont actuellement que des vecteurs) pour les rendre interprétables par les humains. La première idée développée dans cette thèse a consisté à examiner les mots dans les phrases les plus importantes pour le concept en les surlignant puis en trouvant un thème commun. Ce thème était identifié manuellement ; la première amélioration consisterait donc à automatiser cette recherche à l'aide d'un LLM pour gagner en efficacité. Nous pourrions également utiliser des techniques de visualisation comme le nuage de mots pour faciliter la lecture des concepts par les humains. Cependant, en se concentrant uniquement sur les mots, nous négligeons une des grandes forces de ces concepts, qui est leur capacité à capturer des informations syntaxiques. Nous souhaitons donc aussi tester si les phrases importantes d'un concept correspondent à une structure spécifique dans une liste de structures syntaxiques, telles que la voix passive/active, le nombre de clauses par phrase, la négation, etc. Enfin, lorsque l'interprétation des concepts sera plus avancée, nous pourrons revenir à la création de méthodes traitant de l'équité comme TaCo en utilisant ces nouveaux concepts.

%% file: AppendixRoBERTatraining.tex
\chapter{Gender prediction with RoBERTa}\label{apx:genderpredrob}

In this section, we study the impact on gender prediction of removing explicit gender indicators with a Transformer model. To do this, we train a model on a dataset and we train a model with the same architecture on modified version of the dataset where we remove the explicit gender indicators. We train these models to predict the gender and we compare both of these model's accuracies.

In our example, we use the \textit{Bios} dataset \citep{de2019bias} explained in Section \ref{biosdataset}, which contains about 400K biographies (textual data). For each biography, we have the gender (M or F) associated. We use each biography to predict the gender, then we clean the dataset, and we create a second dataset without the explicit gender indicators.

\section{Dataset pre-processing protocol}

\paragraph{Cleaning dataset}
To clean the dataset, we take all biographies and apply these modifications: 
\begin{itemize}
    \item we remove email addresses and URLs, 
    \item we remove the dots after pronouns and acronyms, 
    \item we remove double "?", "!" and ".", 
    \item we truncate all the biographies at 512 tokens by checking that they are cut at the end of a sentence and not in the middle.
\end{itemize}

\paragraph{Creating the \textit{Bios-neutral} dataset}

From \textit{Bios}, we create a new dataset without explicit gender indicators, called \textit{Bios-neutral}. 

At first, we tokenize the dataset, then: \vspace{-2mm} \begin{itemize}
    \item If the token is a first name, we replace it with "\textit{Sam}" (a common neutral first name). 
    \item If the token is a dictionary key (on the dictionary defined below), we replace it with its values. 
\end{itemize}

To determine whether the token is a first name, we use the list of first names:  \href{https://www.usna.edu/Users/cs/roche/courses/s15si335/proj1/files.php\%3Ff=names.txt&downloadcode=yes}{usna.edu}. The dictionary used in the second part is based on the one created in \citep{field2020unsupervised}.

\section{Models training}

\paragraph{RoBERTa base training}
We use a RoBERTa model \citep{liu2019roberta}, which is based on the Transformers architecture and is pre-trained with the Masked language modeling (MLM) objective. 
We specifically used a RoBERTa base model pre-trained by HuggingFace. All information related to how it was trained can be found in  \citep{liu2019roberta}. It can be remarked, that a very large training dataset was used to pre-train the model, as it was composed of five datasets: \emph{BookCorpus} \citep{moviebook}, a dataset containing 11,038 unpublished books; \emph{English Wikipedia} (excluding lists, tables and headers); \emph{CC-News} \citep{10.1145/3340531.3412762} which contains 63 millions English news articles crawled between September 2016 and February 2019; \emph{OpenWebText} \citep{radford2019language}  an open-source recreation of the WebText dataset used to train GPT-2; \emph{Stories} \citep{trinh2018simple} a dataset containing a subset of CommonCrawl data filtered to match the story-like style of Winograd schemas. Pre-training was performed on these data by randomly masking 15\% of the words in each of the input sentences and then trying to predict the masked words.

\paragraph{Gender prediction task}
After pre-training RoBERTa parameters on this huge dataset, we then trained it on the 400.000 biographies of the  \emph{Bios} dataset to predict the gender (F or M). 
The training was performed with PyTorch on 2 GPUs (Nvidia Quadro RTX6000 24GB RAM) for 5 epochs with a batch size of 8 observations and a sequence length of 512 words. The optimizer was Adam with a learning rate of 1e-6, $\beta_1 = 0.9$, $\beta_2 = 0.98$, and $\epsilon =1e6$. 
We split the dataset into 70\% for training, 10\% for validation, and 20\% for testing.

\section{Results}
\begin{figure}[ht]
    \centering
\includegraphics[width=1.0\linewidth]{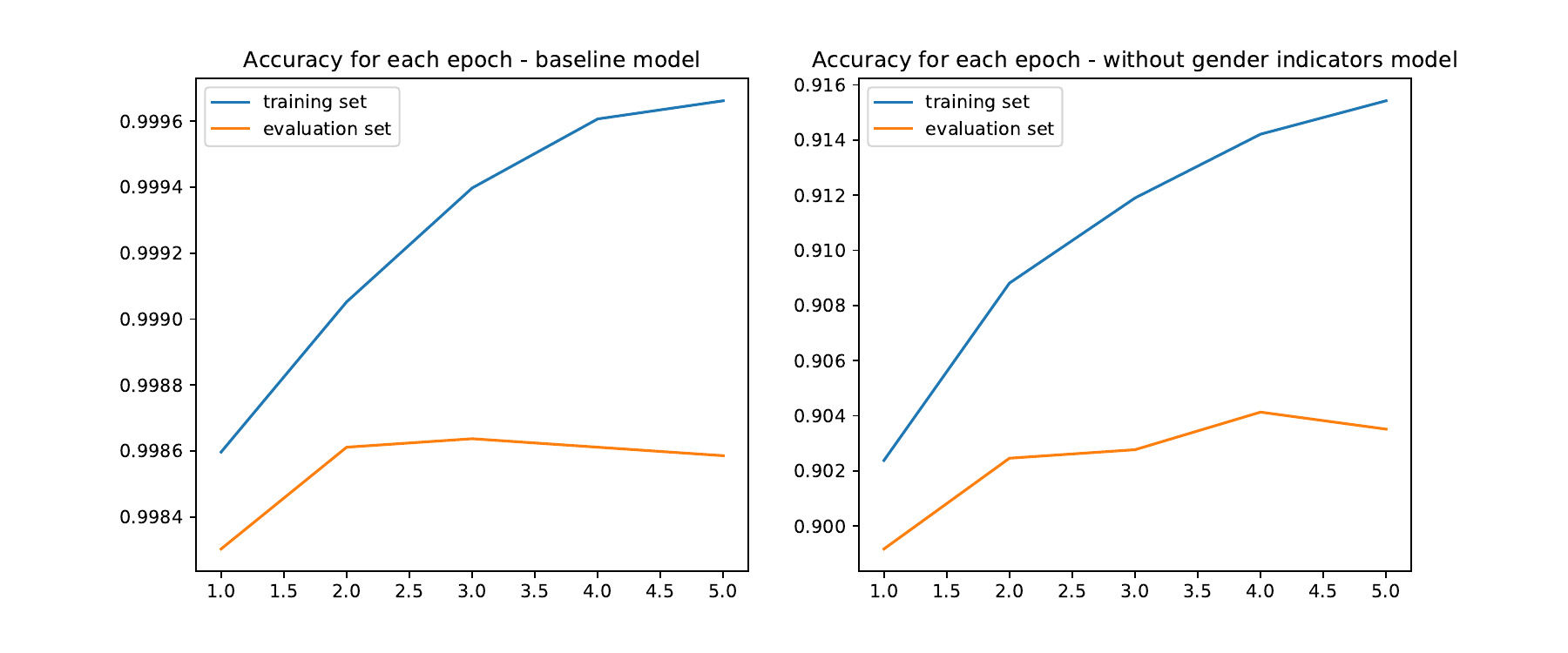}
    \caption{Convergence curves (accuracy) during training for the gender classification task, 5 epochs. (On the left) RoBERTa model trained on \textit{Bios}; (On the right): RoBERTa model trained on \textit{Bios} without explicit gender indicators.}
    \label{apx:fig:acccurves}
\end{figure}

In figure \ref{apx:fig:acccurves}, we go from 99\% accuracy for the baseline to 90\% accuracy for the model trained on the dataset without explicit gender indicators. Even with a dataset without explicit gender indicators, a model like RoBERTa performs well on the gender prediction task. 

%% file: AppendixA.tex
\chapter{Additional figures chapter \ref{Regularization}}\label{appendixA}

\section{Extension figures}
\begin{figure}
    \centering
\includegraphics[width=1.\linewidth]{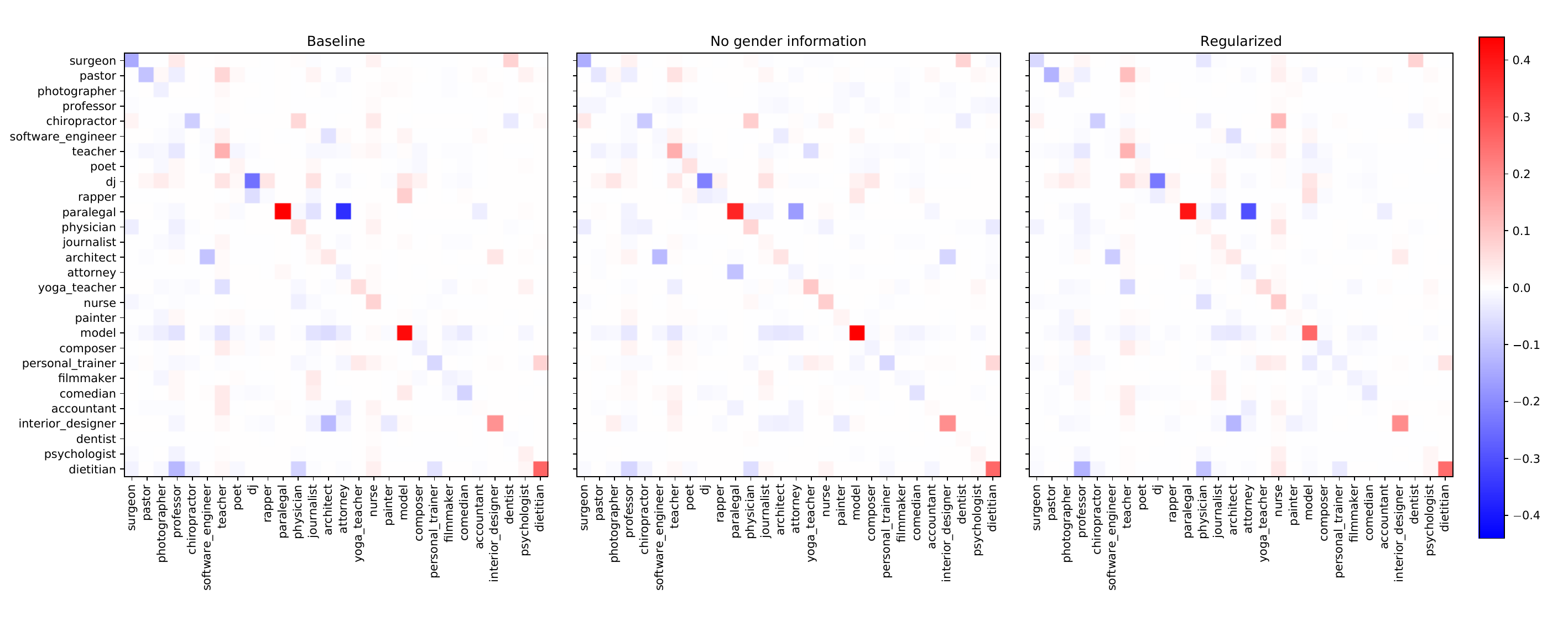}
    \caption{Extension of Fig.~\ref{regularization:fig:confmat} to all output classes of the Bios dataset and not only the most frequent ones, which were selected in Fig.~\ref{regularization:fig:confmat} for readability purposes.}
    \label{regularization:fig:confmat_appendix}
\end{figure}

\begin{figure}
    \centering
\includegraphics[width=1\linewidth]{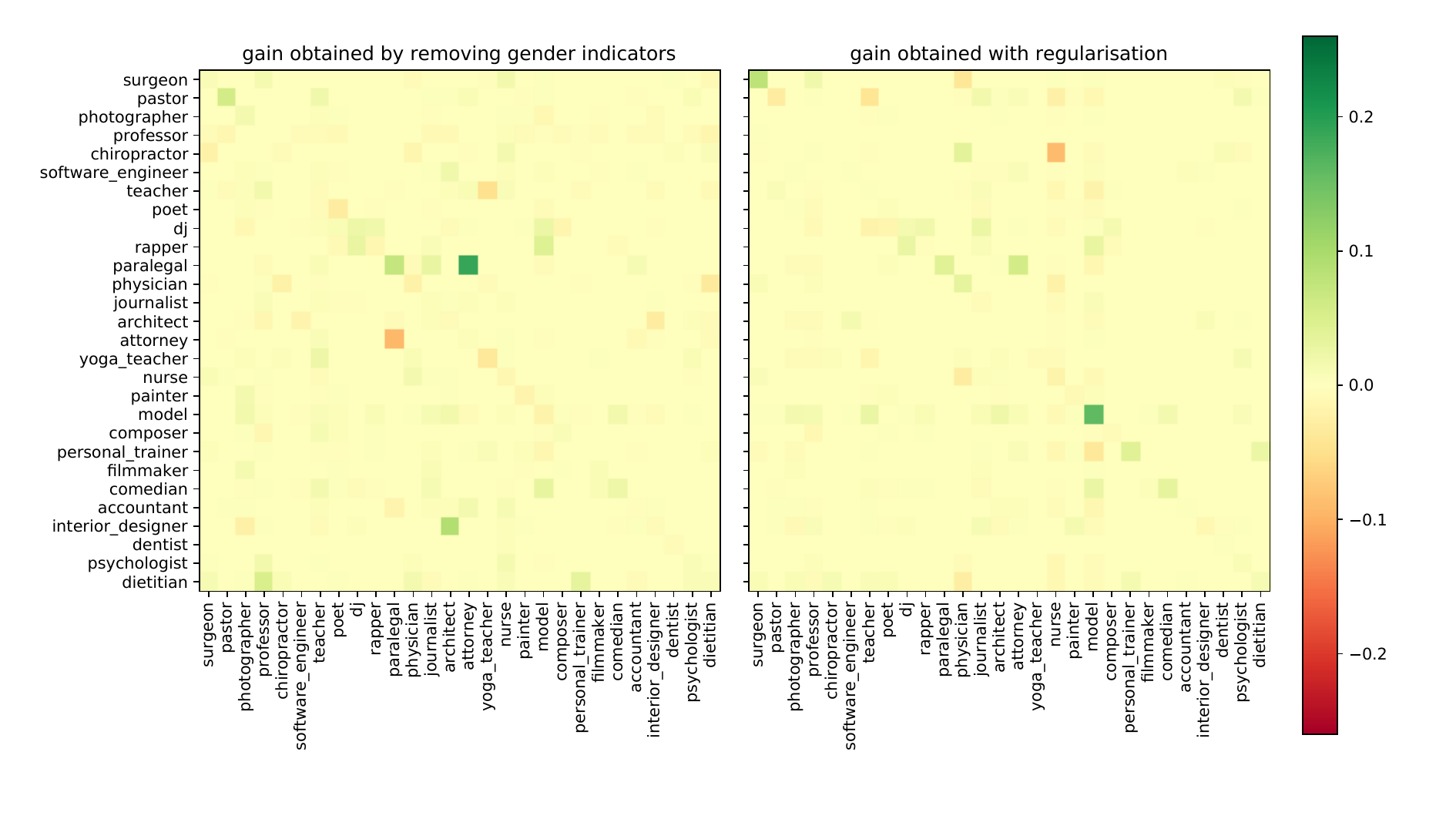}
    \caption{Extension of Fig.~\ref{regularization:fig:confmat_gains} to all output classes of the Bios dataset and not only the most frequent ones, which were selected in Fig.~\ref{regularization:fig:confmat_gains} for readability purposes.}
    \label{regularization:fig:confmat_gainsall}
\end{figure}

The results shown in Figs.~\ref{regularization:fig:confmat} and \ref{regularization:fig:confmat_gains} selected the most largely represented output classes for readability purposes. We show in this appendix their extensions, Figs.~\ref{regularization:fig:confmat_appendix} and \ref{regularization:fig:confmat_gainsall}, to all output classes of the \textit{Bios} dataset \citep{de2019bias}. It can be observed in these figures that other output classes than \textit{Model} and \textit{Surgeon} presented high gender biases, when using the baseline strategy: \textit{Paralegal}, \textit{DJ} and \textit{Dietician}. Although we used these output classes when training the prediction model to make the classification task complex, we voluntarily decided to not regularize them for statistical concerns:
These occupations are indeed first poorly represented in the \textit{Bios} dataset and are additionally strongly unbalanced between males and females. Although the whole training set contains more than 400,000 biographies, there are less than 100 biographies for female \textit{DJ}, male \textit{Dietician} and male \textit{Paralegal}. This makes their treatment with a statistically-sound strategy unreliable. When applied to statistically poorly represented observations, a constrained neural-network won't indeed learn to use generalizable features in the input biographies, but will instead overfit the specificities of each observation which is strongly highlighted by the constraint. We can however see that the tested bias mitigation strategies on the classes \textit{Model} and \textit{Surgeon} did not amplify the biases on the \textit{Paralegal}, \textit{DJ} and \textit{Dietician} classes.

From a certification perspective in the E.U., the \emph{AI act} will ask to clearly mention to end-users the cases for which the predictions may be unreliable or potentially biased. In this context, our strategy makes it possible to certify that mutli-class neural-network classifiers make unbiased decisions on output classes that would be biased using standard training, if the training data offer a sufficient representativity and variability of the characteristics in these classes. In the case where a company would desire to certify that poorly represented classes in the training set are free of biases, the certification procedure will naturally require acquiring more observations.

\section{Convergence curves}

\begin{figure}[ht]
    \centering
\includegraphics[width=1\linewidth]{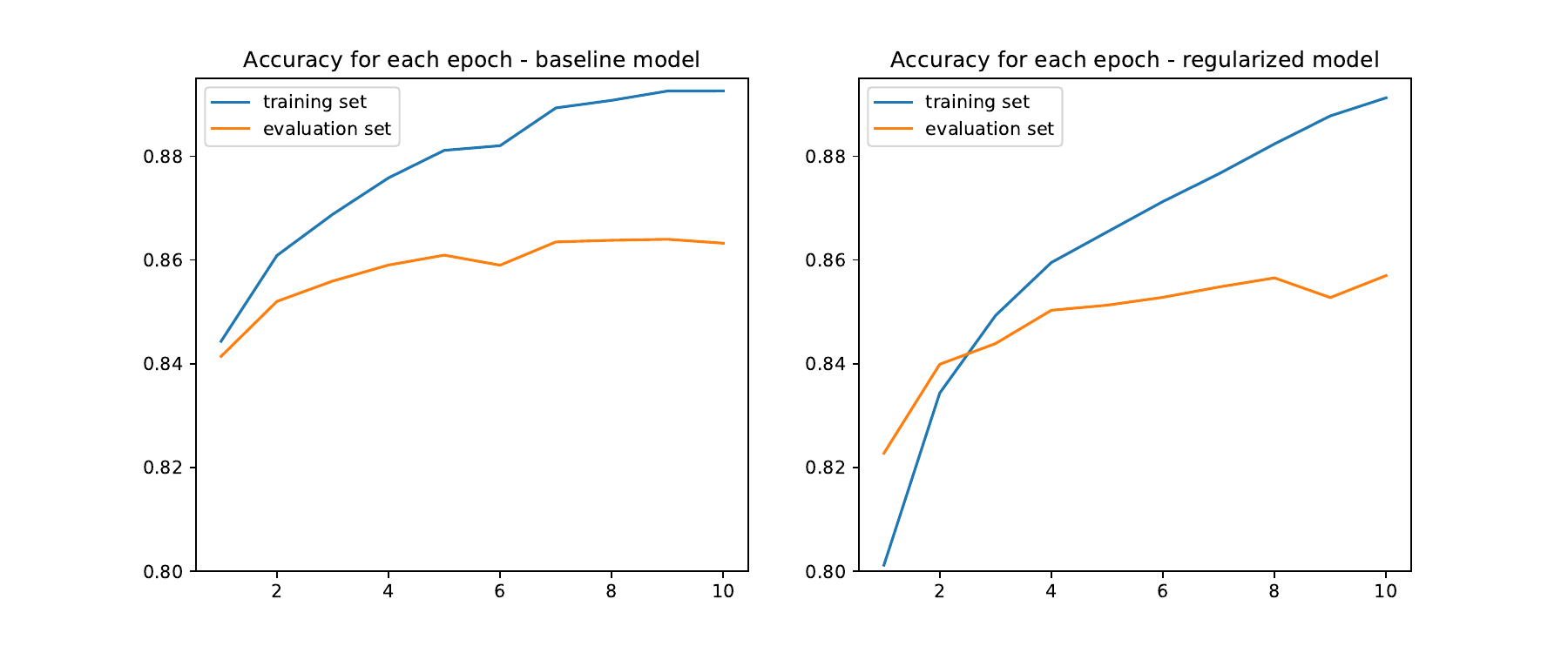}
    \caption{Learning curves with accuracy for each epoch for both models, on training and evaluation set.}
    \label{regularization:fig:convergencecurves}
\end{figure}

We present in Fig.~\ref{regularization:fig:convergencecurves}, the   accuracy convergences for the training and the evaluation sets obtained when training a baseline and a regularized model. We kept the same axes to efficiently compare the two models. The accuracy of the regularized model is lower on the trained and evaluation set than that of the baseline on the first epochs. This difference however fades quickly on the following epochs. It therefore takes longer for the regularized model to have the same accuracy as the baseline model, which is standard for regularized models. Note that we had previously selected the typical amount of epochs after which the computations stop when using early stopping (\textit{i.e.}  when the accuracy continues increasing on the training set, but starts decreasing on the test set). We used this amount of epochs here, and did not detected any significant overfitting on the convergence curves.

